\documentclass[preprint,12pt]{elsarticle}

\usepackage{amssymb}
\usepackage{amsmath}

\journal{}
\usepackage[hidelinks]{hyperref}
\usepackage{booktabs}
\usepackage{subcaption}
\usepackage{array} 
\usepackage{multirow}
\usepackage{siunitx} 
\usepackage{array}   
\usepackage{placeins}
\usepackage{float}
\usepackage{svg}
\usepackage{graphicx}
\usepackage[export]{adjustbox}
\usepackage{subcaption}
\usepackage{amsmath} 

\makeatletter
\def\ps@pprintTitle{%
   \let\@oddhead\@empty
   \let\@evenhead\@empty
   \let\@oddfoot\@empty
   \let\@evenfoot\@oddfoot
}
\makeatletter
\def\ps@pprintTitle{%
   \let\@oddhead\@empty
   \let\@evenhead\@empty
   \let\@oddfoot\@empty
   \let\@evenfoot\@oddfoot
}
\makeatother

\begin{document}

\begin{frontmatter}

\title{Neural Operator-Based Surrogate Model for CFD: Helical Coil Steam Generator in Small Modular Reactor}

\author[a]{Minseo Lee} 
\author[b]{Seongmin Oh}
\author[c]{Chaehyeon Song}
\author[c]{Bumjin Cho}
\author[a]{Shilaj Baral}
\author[d]{Sangam Khanal}
\author[c]{Minseop Song\corref{cor1}}
\author[a]{Joongoo Jeon\corref{cor2}}

\cortext[cor1]{Corresponding author. Email: hysms@hanyang.ac.kr}
\cortext[cor2]{Corresponding author. Email: jgjeon41@postech.ac.kr}

\affiliation[a]{organization={Division of Advanced Nuclear Engineering, POSTECH},
            addressline={77 Cheongam-Ro}, 
            city={Pohang-si},
            postcode={37673}, 
            state={Gyeongbuk-do},
            country={Republic of Korea}}
\affiliation[b]{organization={Department of Quantum System Engineering, Jeonbuk National University
},
            addressline={567 Baekje-daero}, 
            city={Jeonju-si},
            postcode={54896}, 
            state={Jeollabuk-do},
            country={Republic of Korea}}
\affiliation[c]{organization={Department of Nuclear Engineering, Hanyang University},
            addressline={Wangsimni-ro, Seongdong-gu}, 
            city={Seongdong-gu},
            postcode={04763}, 
            state={Seoul},
            country={Republic of Korea}} 
\affiliation[d]{organization={Graduate School of Integrated Energy-AI, Jeonbuk National University},
            addressline={567 Baekje-daero}, 
            city={Jeonju-si},
            postcode={54896}, 
            state={Jeollabuk-do},
            country={Republic of Korea}}  
\begin{abstract}
Real-time thermal-hydraulic simulation is essential for digital twin (DT) technology that supports the safe and efficient operation of small modular reactors (SMRs). Computational fluid dynamics (CFD) provides high-fidelity flow analysis, but its computational cost prevents direct use in DT applications. AI-based surrogate modeling has been actively investigated to address this limitation, yet neural operator--based surrogates for CFD-level transient analysis of SMR-specific geometries have not been reported. This study presents an integrated framework that combines a reduced-order model (ROM) with neural operators, applied to the helical coil steam generator (HCSG) of the System-integrated Modular Advanced Reactor (SMART). Two ROM strategies tailored to each CFD data type were compared, an MLP-based autoencoder (AE) for unstructured mesh data and a convolutional autoencoder (CAE) for structured mesh data, and each was coupled with the deep operator network (DeepONet) to construct the latent DeepONet (L-DeepONet). The Fourier neural operator (FNO) was additionally adopted for comparison. A multi-scale technique was incorporated into both frameworks to mitigate spectral bias and improve the prediction of K\'{a}rm\'{a}n vortex streets developing inside the HCSG. The multi-scale L-DeepONet captured the instantaneous periodic vortex dynamics in both velocity and pressure fields, while the FNO and its multi-scale variant predicted the time-averaged mean flow and provided reliable pressure drop estimates. These complementary characteristics provide a practical model-selection guideline that links each architecture to specific DT objectives based on CFD data type and the required level of flow resolution.
\end{abstract}


\end{frontmatter}

\section{Introduction}
\label{sec1}

The rapid advancement of artificial intelligence (AI) technologies and the accelerated construction of data centers are driving a significant increase in electricity demand. The proliferation of large language models (LLMs) and cloud-based services has further underscored the importance of a reliable and continuous power supply. Among available energy sources, nuclear power is expected to play a significant role in addressing these challenges. In particular, small modular reactors (SMRs) are regarded as a key technology for future power generation, as their passive cooling systems and compact modular design provide both enhanced safety and siting flexibility. To support the safe and efficient operation of SMRs, digital twin (DT) technology synchronized with the actual reactor has attracted significant attention. A DT enables real-time monitoring of internal reactor conditions and supports the optimization of control strategies in SMR operations. According to this trend, the U.S. Nuclear Regulatory Commission (NRC) has published reports highlighting the potential benefits of DT technology, including reactor diagnostics and prognostics, operational optimization, autonomous operation, and regulatory innovation~\cite{yadav2023digital,yadav2023state,yadav2023technical}.

To realize DT in SMRs, accurate modeling and simulation of thermal-hydraulic phenomena are essential. System-level codes have been actively applied to thermal-hydraulic analysis of helical coil steam generator (HCSG). For example, MARS-KS with helical-tube correlations has been used to predict density-wave oscillations in parallel helical-tube experiments and applied to a System-integrated Modular Advanced Reactor (SMART) HCSG configuration~\cite{oh2024prediction}. To complement such system-scale analysis with high-fidelity local flow phenomena, computational fluid dynamics (CFD) is widely used as a representative analytical tool, enabling precise calculations of fluid flow and detailed analysis of complex phenomena. However, the CFD has inherent limitations, including long computation times and the demand for substantial computational resources, which constrain its direct implementation in real-time DT frameworks. To overcome these limitations, deep learning–based surrogate modeling has emerged as a promising alternative. Several studies have applied deep learning to accelerate CFD simulations. Vinuesa and Brunton provided a broad perspective on how machine learning can accelerate computational fluid dynamics, outlining key opportunities and challenges across high-fidelity simulations, turbulence modeling, and reduced-order modeling~\cite{vinuesa2022enhancing}. Srinivasan et al. employed multilayer perceptrons (MLPs) and long short-term memory (LSTM) networks to learn and predict turbulent shear flows, demonstrating that LSTM can accurately reproduce turbulence statistics and dynamics~\cite{srinivasan2019predictions}. Khanal et al. utilized a natural convection CFD dataset to systematically compare various deep learning architectures based on convolutional neural networks (CNNs) under identical conditions, evaluating their performance in unsteady CFD prediction within small-data environments~\cite{khanal2025comparison}. These studies collectively demonstrate that data-driven neural network models such as CNNs and LSTMs can substantially reduce computational cost and time compared with conventional numerical methods. Despite their efficiency, deep learning–based approaches still face several intrinsic limitations. Most models perform predictions in an auto-regressive manner, which leads to error accumulation as the prediction horizon increases during sequential time-series forecasting. In addition, when the input conditions extend beyond the range of the training data, prediction accuracy degrades significantly, necessitating model retraining. Achieving high-fidelity predictions for complex geometries and turbulent flows likewise remains a challenging task.

To overcome these challenges, scientific machine learning (SciML) methodologies have been actively developed, integrating scientific data or domain knowledge into traditional machine learning techniques to construct physically consistent models~\cite{baker2019workshop}. Representative SciML methods include physics-informed neural networks (PINNs), neural operators, and residual-based physics-informed transfer learning (RePIT). PINNs incorporate initial and boundary conditions, as well as governing equations, into the loss function of standard deep learning architectures, thereby guiding the network to directly approximate physical laws~\cite{raissi2019physics}. In the nuclear field, studies have been conducted to approximate the severe accident analysis code MELCOR using PINNs~\cite{shin2025node}. Neural operators are operator learning methods that approximate PDEs by mappings between function spaces, with the deep operator network (DeepONet) and the Fourier neural operator (FNO) being the most prominent examples~\cite{lu2021learning, li2020fourier}. RePIT is a hybrid solver that couples CFD and ML models, in which the solver switches to CFD computations whenever error accumulation during ML prediction reaches a threshold, effectively mitigating the chronic error accumulation problem inherent in auto-regressive methods~\cite{jeon2024residual}.

Among SciML approaches, the neural operator is well suited to DT applications. As an operator-learning method that maps functions to functions, the neural operator predicts the entire temporal field in a single forward pass. This enables real-time inference without error accumulation. It also generalizes well across diverse operating conditions, predicting flow fields without retraining. These properties align with the DT requirement of predicting flow fields under varying operating conditions. Despite these advantages, to the best of the authors' knowledge, no prior study has reported a neural operator-based surrogate model for CFD-level transient analysis of an SMR component. This study addresses the gap by integrating dimensionality reduction with operator learning, applied to the HCSG of the SMART. To the authors' knowledge, this study presents the first such model. The study further compares the performance of various surrogate models to derive a model-selection guideline.

The implementation of the proposed surrogate models reflects the following technical considerations. Each high-fidelity simulation produces a large-scale spatiotemporal field. Such high-dimensional data contain many redundant features that induce the curse of dimensionality and hinder network optimization, making direct learning difficult. The limited number of high-fidelity samples affordable in practice further compounds this difficulty. However, due to physical constraints, such data typically reside on a lower-dimensional latent manifold, from which a reduced-order model (ROM) can efficiently extract essential features and enable effective learning. Building on this insight, the present study adopts the latent DeepONet (L-DeepONet) framework, which integrates a ROM with DeepONet to perform operator learning within a compact latent space. By removing redundant features prior to operator learning, L-DeepONet not only reduces training cost but also improves predictive accuracy, particularly for problems with strongly nonlinear dynamics~\cite{kontolati2024learning}. The Fourier neural operator (FNO) is also adopted for comparison. During training, both L-DeepONet and FNO struggle to predict K\'{a}rm\'{a}n vortex streets from spectral bias, in which neural networks preferentially learn low-frequency components while failing to reproduce high-frequency features~\cite{rahaman2019spectral}. To accurately capture the high-frequency flow features arising from K\'{a}rm\'{a}n vortex streets, a multi-scale technique is incorporated into both frameworks. Accordingly, the objectives of this study are as follows.
\begin{itemize}
    \item To apply a framework that integrates dimensionality reduction with operator learning to construct a CFD-level transient surrogate model for an SMR component.
    \item To compare ROM strategies tailored to each CFD data type, namely an MLP-based autoencoder (AE) for unstructured mesh data and a convolutional autoencoder (CAE) for structured mesh data obtained by interpolation.
    \item To compare, under identical training and evaluation conditions, the following architectures: the multi-scale L-DeepONet combined with the MLP-based AE and CAE, the standard FNO, and the multi-scale FNO.
    \item To derive from this comparison a model-selection guideline that links each architecture's predictive characteristics to specific digital twin objectives.
\end{itemize}

\begin{figure}[H]
    \centering
    \includegraphics[width=\textwidth]{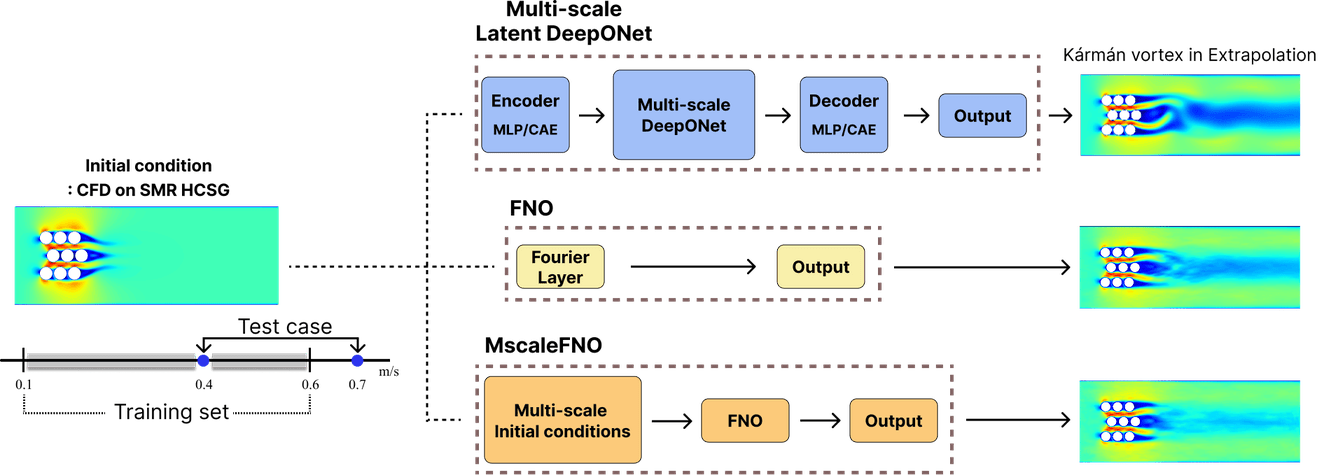}
    \caption{Overview of the neural operator surrogate framework for transient flow prediction in the HCSG of an SMR.}
    \label{fig:intro_overview}
\end{figure}

These objectives are addressed through the framework illustrated in Figure~\ref{fig:intro_overview}, which integrates a ROM and a multi-scale neural operator to enable CFD-level prediction of transient flow in the SMR HCSG.

The remainder of this paper is organized as follows. Section~\ref{sec_related} reviews related works on AI-driven DT frameworks that form the methodological foundation of this study. Section~\ref{sec4} describes the CFD dataset for the HCSG geometry. Section~\ref{sec5} presents the operator learning methodologies tailored to the HCSG flow, including the multi-scale techniques developed to mitigate spectral bias. Section~\ref{sec6} provides the results and discussion, comparing the four neural operator variants in their ability to predict the K\'{a}rm\'{a}n vortex dynamics and the pressure drop. Finally, Section~\ref{sec7} concludes the study by summarizing the key findings and discussing future directions.
 
\section{Related Works}
\label{sec_related}

\begin{table}[h]
    \centering
    \begin{adjustbox}{max width=\textwidth}
    \begin{tabular}{cccc}
        \toprule
        \textbf{Year} & \textbf{Author} & \textbf{Used AI} & \textbf{Application} \\
        \midrule
        2022 & He et al~\cite{he2022deep} & POD + Neural Network & Steam generator in PWR \\
        2023 & Li et al~\cite{li2023constructing} & Bayesian NN & Reactor core \\
        2024 & Kobayashi et al~\cite{kobayashi2024deep} & DeepONet & Neutron flux \\
        2024 & Kobayashi et al~\cite{kobayashi2024ai} & Polynomial chaos expansion & Accident-tolerant fuels \\
        2024 & Liu et al~\cite{liu2025development} & GNN & Whole system in advanced reactor \\
        2024 & Kajihara et al~\cite{kajihara2024machine} & Neural Network + FSM/DSE & Experimental capsule in ATR \\
        2025 & Hossain et al~\cite{hossain2025virtual} & DeepONet & Hot leg of PWR \\
        2025 & Chen et al~\cite{chen2025study} & DPNN and ResDPNN & Nuclear power plants \\
        2025 & Lim et al~\cite{lim2025ai} & GRU + LLM (GPT-4o) & Three-loop thermal-fluid testbed \\
        2025 & Cheng et al~\cite{cheng2025surrogate} & FB-DeepONet & Heat transfer problem in pipe \\
        2026 & Cho et al~\cite{cho2026efficient} & POD/DMD/AE + LSTM & Cross flow in HCSG of SMART \\
        \bottomrule
    \end{tabular}
    \end{adjustbox}
    \caption{Summary of related studies on AI-based DT in nuclear systems.}
    \label{tab:ai_table}
\end{table}

Recent studies have applied a wide variety of AI architectures to DT implementation in nuclear systems, as summarized in Table~\ref{tab:ai_table}. He et al. developed a deep-learning ROM that coupled proper orthogonal decomposition (POD) with neural network to rapidly estimate the void-fraction and temperature fields of a vertical U-tube natural-circulation steam generator in a pressurized water reactor (PWR)~\cite{he2022deep}. Their framework reduced the order of the void-fraction and temperature fields by 88.3\% and 96.7\%, respectively, while achieving approximately four orders of magnitude speedup over CFD with maximum absolute errors of 0.1 and 0.03 K. Li et al. proposed a probabilistic digital twin framework for a PWR core by integrating Bayesian neural networks (BNNs) with reduced-order modeling~\cite{li2023constructing}. Their approach addressed the inverse problem of core monitoring under uncertainty by replacing conventional data assimilation techniques with BNNs, thereby enabling probabilistic estimation of multi-physical fields with quantified confidence intervals. Kobayashi and Alam investigated the feasibility of DeepONet as a real-time surrogate modeling method for DT applications in nuclear energy systems~\cite{kobayashi2024deep}. They constructed a DeepONet-based surrogate model to predict the two-dimensional spatial distribution of neutron flux in a maze geometry, demonstrating that DeepONet achieves inference approximately 1500 times faster than conventional Monte Carlo particle transport simulations while outperforming traditional ML methods such as fully connected and CNNs. Kobayashi et al. further introduced an AI-driven non-intrusive uncertainty quantification method based on polynomial chaos expansion for DT-enabling technology in nuclear fuel performance evaluation~\cite{kobayashi2024ai}. Using the finite element–based fuel performance code BISON, they quantified the impact of uncertain input variables on accident-tolerant fuel models, demonstrating that fewer than 100 simulations were sufficient to achieve reliable uncertainty estimates. Liu et al. developed a whole-system DT for advanced nuclear reactors by leveraging graph neural networks (GNNs) combined with the System Analysis Module (SAM)~\cite{liu2025development}. Their approach modeled the reactor system as a heterogeneous graph in which components serve as nodes and physical interconnections as edges, enabling the DT to infer the entire system status from sparse sensor data and predict operational transients. The framework was validated on the Experimental Breeder Reactor II (EBR-II) and a generic fluoride-salt-cooled high-temperature reactor (gFHR). Kajihara et al. proposed a neural network based surrogate model that predicts the radial temperature and displacement distributions of an ATR experimental capsule, trained on Abaqus finite-element simulations parameterized by the outer-gas-gap thickness. They introduced feature similarity measurement (FSM) and data similarity enhancement (DSE) to improve predictive accuracy in extrapolation regions, and explicitly motivated the surrogate by the real-time inference requirements of digital twins~\cite{kajihara2024machine} .
 
More recently, Hossain et al. proposed a virtual sensing–enabled DT framework for real-time monitoring of nuclear systems using DeepONet, in which DeepONet was employed as a virtual sensor to predict key thermal-hydraulic parameters in the hot leg of a PWR, achieving predictions approximately 1400 times faster than conventional CFD simulations~\cite{hossain2025virtual}. Chen et al. proposed a hybrid physics- and data-driven approach for constructing the digital model of a nuclear power plant DT, in which a data-driven and physical-model combined neural network (DPNN) and its residual variant (ResDPNN) were developed by embedding the law of energy conservation into a neural network with residual connections~\cite{chen2025study}. Their model demonstrated improved generalization and predictive accuracy for coolant average temperature using real operational data. Lim et al. presented an AI-driven thermal-fluid testbed for advanced SMR technologies that integrates a three-loop experimental facility, a SAM-based DT accelerated by a gated recurrent unit (GRU) neural network, and an intelligent operator assistance system powered by a large language model~\cite{lim2025ai}. The GRU-accelerated DT enables real-time prediction of complex thermal-fluid dynamics, while the LLM-based operator assistant provides actionable operational guidance in natural language. Cheng et al. developed a Fourier basis–deep operator network (FB-DeepONet) as a surrogate model for predicting heat transfer under fluctuating flow conditions in nuclear reactor systems, in which the input function space was constructed using Fourier basis functions to effectively capture recurring oscillatory flow patterns while incorporating Bayesian uncertainty quantification to assess prediction reliability~\cite{cheng2025surrogate}. Cho et al. developed ROM-based LSTM frameworks for predicting turbulent cross flow in the HCSG of an SMR by coupling POD, dynamic mode decomposition (DMD), and an AE with a long short-term memory (LSTM) network~\cite{cho2026efficient}. Their framework predict the temporal evolution of high-fidelity LES data and achieved substantial computational speed-up over LES while comparing the predictive accuracy of linear and nonlinear ROM frameworks for non-periodic turbulent flows.

As summarized above, AI-based DT research in nuclear systems has explored a wide spectrum of architectures and applications, including transient analyses of actual reactor components such as the PWR hot leg~\cite{hossain2025virtual} and the SMART HCSG~\cite{cho2026efficient}. While neural operators such as DeepONet and FB-DeepONet have been adopted for nuclear DT applications including neutron flux prediction~\cite{kobayashi2024deep}, virtual sensing in a PWR hot leg~\cite{hossain2025virtual}, and heat transfer in pipes~\cite{cheng2025surrogate}, these studies have been confined to relatively simple geometries or single-component flows and have each relied on a single neural operator architecture. To the best of the authors' knowledge, neural operator–based surrogate modeling has not yet been applied to CFD-level thermal-hydraulic transient analysis of an SMR-specific geometry, and a systematic comparison of multiple neural operator architectures within such a configuration has not been reported. The present study fills these gaps by developing L-DeepONet and FNO surrogate models for the SMART HCSG, by incorporating a multi-scale technique into both frameworks to mitigate spectral bias in periodic vortex flows, and by systematically comparing four neural operator variants under identical training and evaluation conditions across multiple inlet velocity conditions.


\section{Neural Operators Methods}
\label{sec3}
\subsection{Deep Operator Network (DeepONet)}
\label{subsec3.1}

DeepONet is an operator learning model designed to map a function to another function, based on the universal approximation theorem for operators~\cite{lu2021learning,chen1995universal}. An operator $\mathcal{G}$ takes an input function $u_0$ and returns the corresponding output function $s = \mathcal{G}(u_0)$, as expressed in Eq.~(\ref{eq:deeponet_map}). In this study, $u_0$ denotes the initial flow field and $s$ denotes the corresponding output function. When $\mathcal{G}(u_0)$ is evaluated at a coordinate $y$, the universal approximation theorem for nonlinear operators takes the form given in Eq.~(\ref{eq:universal}).

\begin{equation}
\label{eq:deeponet_map}
\begin{aligned}
\mathcal{G} &: u_0 \mapsto s,\\
\end{aligned}
\end{equation}

\begin{equation}
    \left| \mathcal{G}(u_0)(y) - 
    \sum_{k=1}^{p} \sum_{i=1}^{n} 
    c_{ki} \, \sigma \!\left( \sum_{j=1}^{m} \xi_{kij} \, u_0(x_j) + \theta_{ki} \right) 
    \sigma \!\left( \mathbf{w}_k \cdot y + \zeta_k \right) \right|
    < \epsilon
    \label{eq:universal}
\end{equation}

In Eq.~(\ref{eq:universal}), $\sigma$ denotes a non-linear activation function, $\{x_j\}_{j=1}^{m}$ are the sensor locations at which the input function $u_0$ is evaluated, and $u_0(x_j)$ are the corresponding sensor values. The integers $p$, $n$, and $m$ represent the number of basis functions, hidden units, and sensors, respectively. The trainable network parameters consist of the branch-net weights $\xi_{kij}$ and biases $\theta_{ki}$, the trunk-net weights $\mathbf{w}_k$ and biases $\zeta_k$, and the output coefficients $c_{ki}$. The bound $\epsilon > 0$ denotes the approximation error, indicating that the operator $\mathcal{G}(u_0)(y)$ can be approximated by the network output to within an arbitrarily small error.

The DeepONet architecture consists of two neural networks, namely the branch net and the trunk net. The branch net takes the input function $u_0$ evaluated at the sensor locations as input, and learns the features in function space. The trunk net takes the spatial or temporal coordinates $y$ as input, at which $\mathcal{G}(u_0)(y)$ is predicted. The outputs of the branch net and the trunk net are then combined through a dot product to produce the output function, as summarized in Eq.~(\ref{eq:dotproduct}).

\begin{equation}
    \hat{s}(y) = \mathcal{G}(u_0)(y) = \sum_{i=1}^{p} b_i \, t_i
    \label{eq:dotproduct}
\end{equation}

In Eq.~(\ref{eq:dotproduct}), $\hat{s}(y)$ denotes the predicted output function evaluated at the coordinate $y$, and $\{b_i\}_{i=1}^{p}$ and $\{t_i\}_{i=1}^{p}$ are the output vectors of the branch net and the trunk net, respectively. The hyperparameter $p$ controls the number of basis functions and determines the size of the final hidden layer of both sub-networks.

\subsection{Latent DeepONet (L-DeepONet)}
\label{subsec3.2}

In this study, we adopt L-DeepONet to efficiently learn high-dimensional physical systems. Accurate prediction of complex phenomena typically requires sufficient labeled data. However, in the case of intricate geometries such as SMRs, performing high-fidelity CFD simulations becomes computationally prohibitive. Data generated by numerical solvers, such as the finite-volume method (FVM) or finite-element method (FEM), exhibit rapidly increasing dimensionality with higher spatiotemporal grid resolutions, while the attainable dataset size remains limited, thereby impeding effective training. To overcome these limitations, L-DeepONet first compresses input fields into a latent space, trains DeepONet within that space, and then decodes predictions back to the original space. This procedure shortens training time and removes redundant features, thereby facilitating optimization. Consequently, L-DeepONet has been reported to maintain or improve accuracy relative to baseline DeepONet while substantially reducing training time and memory~\cite{kontolati2024learning}.

The L-DeepONet architecture integrates a ROM with a DeepONet. The ROM can be developed through methods such as POD and AE. In this study, an AE is adopted as a non-linear ROM to map the high-dimensional CFD data into a compact latent space, where the DeepONet is trained on the latent representation.

\begin{figure}[H]
    \centering
    \makebox[\textwidth]{%
        \includegraphics[width=1.2\textwidth]{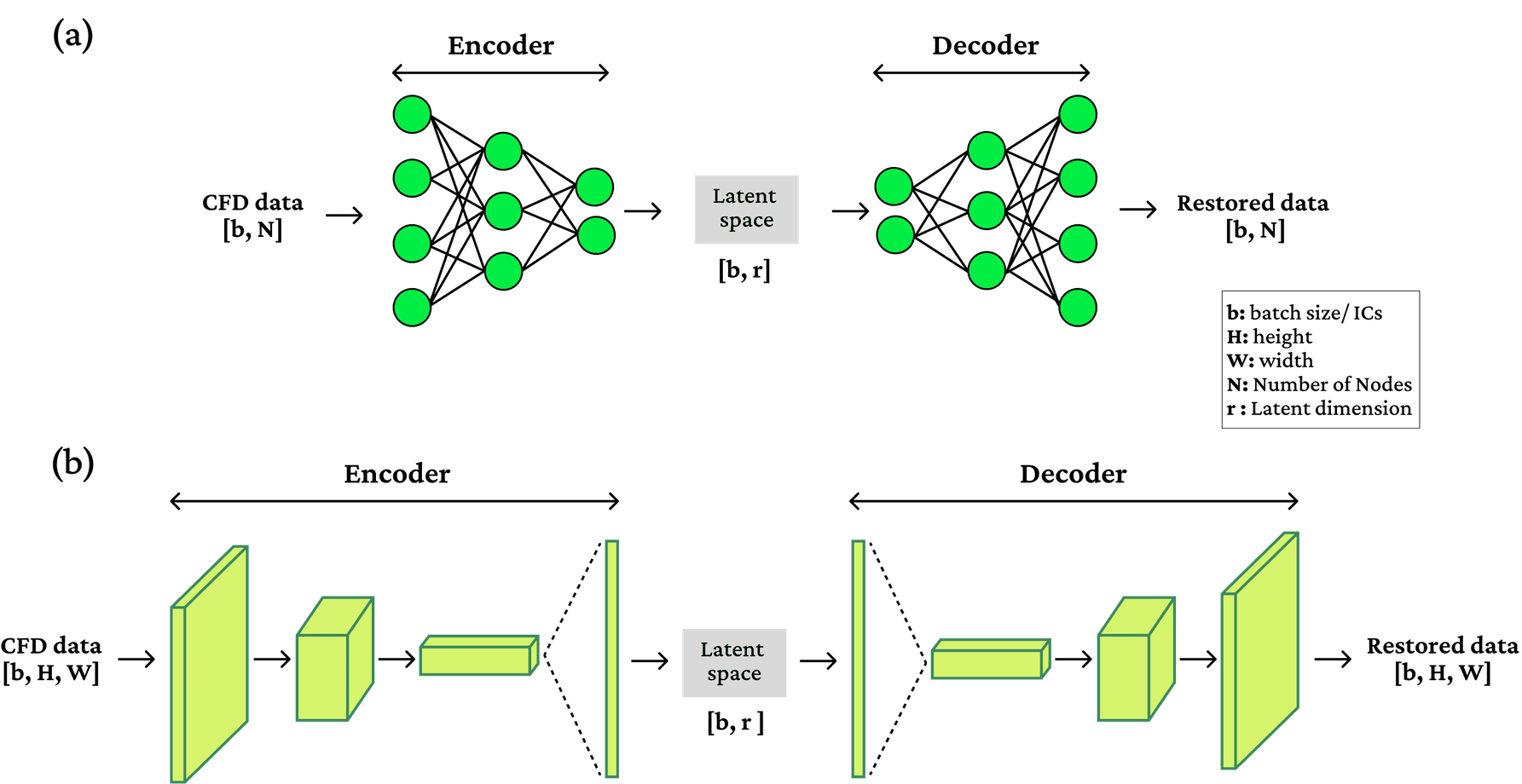}%
    }
    \caption{AE architecture: (a) MLP-based AE, (b) convolutional AE.}
    \label{fig:AE architecture}
\end{figure}

The AE is an unsupervised learning–based neural network composed of an encoder and a decoder. The encoder compresses the input data into a latent space, while the decoder reconstructs the original data from the latent representation. As shown in Eq.~(\ref{eq:ldeop_flow}), the encoder $\mathcal{E}$ maps the original dataset $\mathbf{z}$ to a low-dimensional latent representation $\mathbf{z}^{r}$, where the superscript $r$ denotes a latent space quantity, and the decoder $\mathcal{D}$ maps $\mathbf{z}^{r}$ back to the reconstructed dataset $\tilde{\mathbf{z}}$. Here, the original dataset is defined as $\mathbf{z} \equiv [u_0, s]$, in which $u_0$ is the input function representing the initial flow field at $t = 0$, and $s$ is the output function representing the temporal difference field (described in detail in Section~\ref{sec4}), with the full flow field recovered by adding $u_0$ to the predicted output. The model is trained to minimize the mean-squared reconstruction error 
between $\mathbf{z}_{i}$ and $\tilde{\mathbf{z}}_{i}$ over $n$ training samples, as defined in Eq.~(\ref{eq:lae_mse}). The CFD data used for training consist of time-discretized snapshots, each of which serves as an input to the AE. As illustrated in Figure~\ref{fig:AE architecture}, a multi-layer AE architecture with several hidden layers is employed in this study.

\begin{equation}
\label{eq:ldeop_flow}
\begin{aligned}
\mathcal{E} &: \mathbf{z} \mapsto \mathbf{z}^{r},\\
\mathcal{D} &: \mathbf{z}^{r} \mapsto \tilde{\mathbf{z}}.
\end{aligned}
\end{equation}

\begin{equation}
\mathcal{L}_{\mathrm{AE}}
= \frac{1}{n} \sum_{i=1}^{n} \left( \mathbf{z}_{i} - \tilde{\mathbf{z}}_{i} \right)^{2}
\label{eq:lae_mse}
\end{equation}

\begin{figure}[htbp]
    \centering
    \makebox[\textwidth][c]{
        \includegraphics[width=1.2\textwidth]{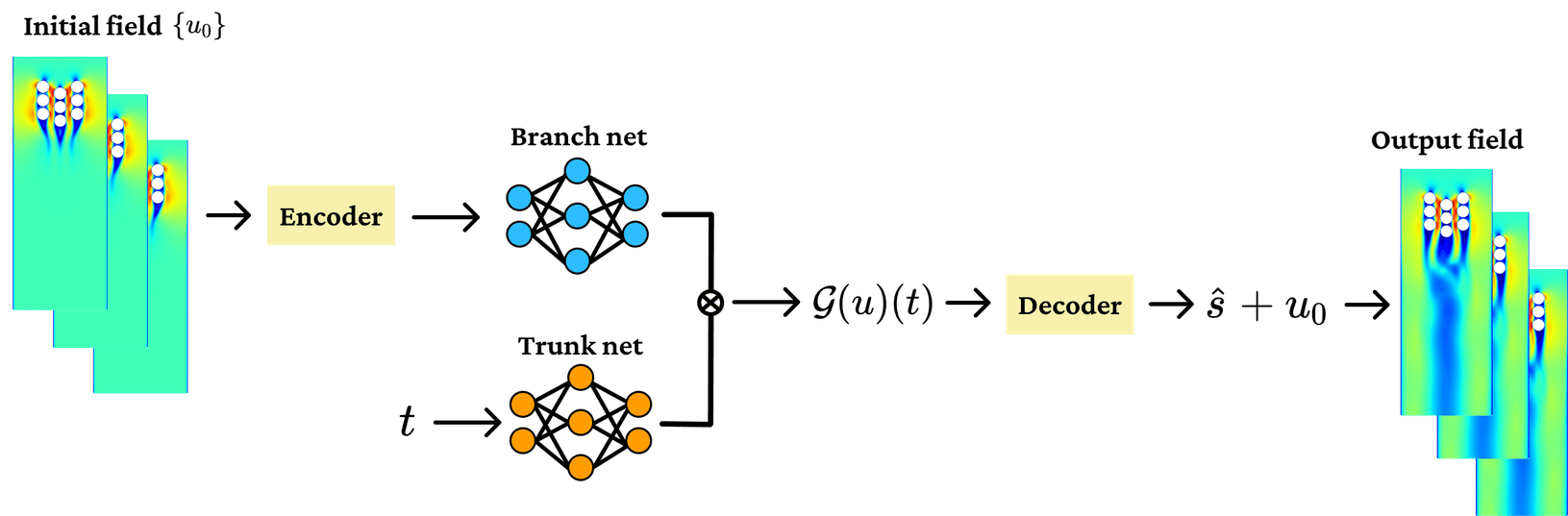}
    }
    \caption{L-DeepONet architecture. The latent data compressed by the encoder is used to train the DeepONet, and the DeepONet outputs are reconstructed in the original space by the decoder.}
    \label{fig:L-DeepONet architecture}
\end{figure}

After the AE has been trained, the DeepONet is further trained in the latent space. First, the combined data $\mathbf{z} = [u_0, s]$ is compressed into the latent representation $\mathbf{z}^{r}$ by the encoder $\mathcal{E}$, from which the latent input $u_0^{r}$ and the latent output $s^{r}$ are extracted as the latent-space counterparts of $u_0$ and $s$. Next, the latent-space DeepONet, denoted by $\mathcal{G}_{r}$, is trained to map $u_0^{r}$ to the predicted latent output $\hat{s}^{r}$, where the hat symbol $\hat{(\cdot)}$ indicates a network prediction. In the prediction stage, $\hat{s}^{r}$ is passed through the decoder $\mathcal{D}$ and reconstructed as $\hat{s}$ in the original space. This process is summarized in Eq.~(\ref{eq:sss}) and Figure~\ref{fig:L-DeepONet architecture}. During training, $\mathcal{G}_{r}$ is optimized to minimize the mean-squared error between the reference latent output $s^{r}$ and the predicted latent output $\hat{s}^{r}$, as defined in Eq.~(\ref{eq:mse_latent}).

\begin{equation}
\label{eq:sss}
\begin{alignedat}{3}
\mathcal{E}     &: & \quad \mathbf{z}   &\mapsto \mathbf{z}^{r},\\
\mathcal{G}_{r} &: & \quad u_0^{r}        &\mapsto \hat{s}^{r},\\
\mathcal{D}     &: & \quad \hat{s}^{r}  &\mapsto \hat{s}.
\end{alignedat}
\end{equation}

\begin{equation}
\mathcal{L}_{\mathcal{G}_{r}}
= \frac{1}{n} \sum_{i=1}^{n} \left( s_{i}^{r} - \hat{s}_{i}^{r} \right)^{2}
\label{eq:mse_latent}
\end{equation}

\subsection{Fourier neural operator (FNO)}
\label{subsec3.3}

\begin{figure}[h]
    \centering
    \includegraphics[width=\textwidth]{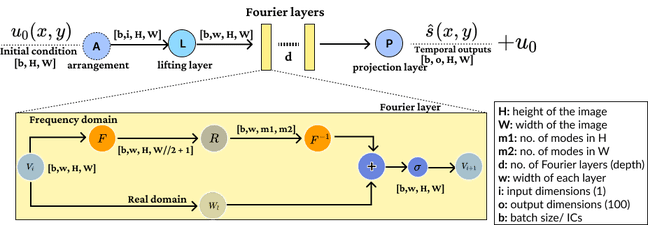}
    \caption{Overall architecture of the Fourier Neural Operator.}
    \label{fig:FNO2D_simple}
\end{figure}

In this study, the FNO is also adopted as an operator learning method that maps the input function to the output function. Following the notation introduced in Section~\ref{subsec3.1}, the input function $u_0(x, y)$ denotes the initial flow field at $t = 0$, and the output function $s(x, y)$ denotes the temporal difference field, where the temporal evolution is encoded in the output channel dimension. The full flow field is recovered by adding $u_0$ to the predicted output. As illustrated in Figure~\ref{fig:FNO2D_simple}, the overall architecture of the FNO consists of three primary stages, namely the lifting layer ($L$), iterative Fourier layers, and the projection layer ($P$).

The input function $u_0(x, y)$, representing the initial flow field, first undergoes an arrangement step ($A$) followed by the lifting layer ($L$). As depicted on the left side of Figure~\ref{fig:FNO2D_simple}, the input data with batch size $b$, height $H$, and width $W$ is first reshaped into a tensor of shape $[b, i, H, W]$, where $i$ denotes the input channel dimension and is set to $i = 1$ in this study. The lifting layer then maps this input to a high-dimensional latent space by expanding the channel dimension from $i$ to the layer width $w$, while preserving the physical grid structure. The output of the lifting layer, denoted by $V_0$ with dimensions $[b, w, H, W]$, then serves as the input to the iterative Fourier layers.
 
The feature tensor $V_0$ passes through a block of $d$ Fourier layers, where $d$ denotes the depth of the Fourier-layer block. Each Fourier layer updates the features by combining a local linear transformation $W_l$ in the spatial domain and a kernel integral operator $\mathcal{K}$ that captures global features, followed by a non-linear activation function $\sigma$. The update rule is defined as follows:
 
\begin{equation}
    V_{j+1}(x, y) = \sigma \left( W_{l} V_{j}(x, y) + (\mathcal{K} V_{j})(x, y) \right),
    \label{eq:fno_update}
\end{equation}
 
\noindent where $V_{j}$ and $V_{j+1}$ denote the feature tensors at the input and output of the $j$-th Fourier layer ($j = 0, 1, \dots, d-1$).
 
The operator $\mathcal{K}$ is efficiently implemented via operations in the Fourier space. Since the input data is real-valued, its Fourier transform exhibits Hermitian symmetry, meaning that only approximately half of the frequency components carry independent information. By applying the fast Fourier transform (FFT) and retaining only $W/2 + 1$ frequency modes along the width direction, redundant components are discarded, thereby maximizing computational and memory efficiency. In the frequency domain, the transformed data is linearly transformed by multiplying it with a learnable weight tensor $R$. At this stage, $m_1$ and $m_2$ determine the range of low-frequency modes to be preserved along the $H$ and $W$ directions, respectively, while high-frequency components exceeding this range are truncated. This truncation acts as a smoothing process, filtering out high-frequency noise and facilitating the efficient learning of global governing equations. The processed signal is then restored to the physical space with dimensions $[b, w, H, W]$ via the inverse Fourier transform ($\mathcal{F}^{-1}$), and the kernel operation is expressed as:
 
\begin{equation}
    (\mathcal{K} V_{j})(x, y) = \mathcal{F}^{-1} \left( R \cdot \mathcal{F}(V_{j}) \right)(x, y).
    \label{eq:fno_kernel}
\end{equation}
 
Consequently, the FNO learns global information ($\mathcal{K}$) in the Fourier domain and local features ($W_l$) in the real domain. The latent representation $V_{d}$, generated after $d$ iterations of the Fourier layer, finally passes through the projection layer ($P$). In this final stage, the layer width $w$ is mapped to the output channel dimension $o$, which corresponds to the number of predicted time steps, ultimately generating the full spatiotemporal output $s(x, y)$ with dimensions $[b, o, H, W]$.

\section{CFD Dataset}
\label{sec4}
\subsection{Unstructured dataset}
\label{subsec4.1}

\begin{figure}[H]
    \centering
    \includegraphics[width=0.8\linewidth]{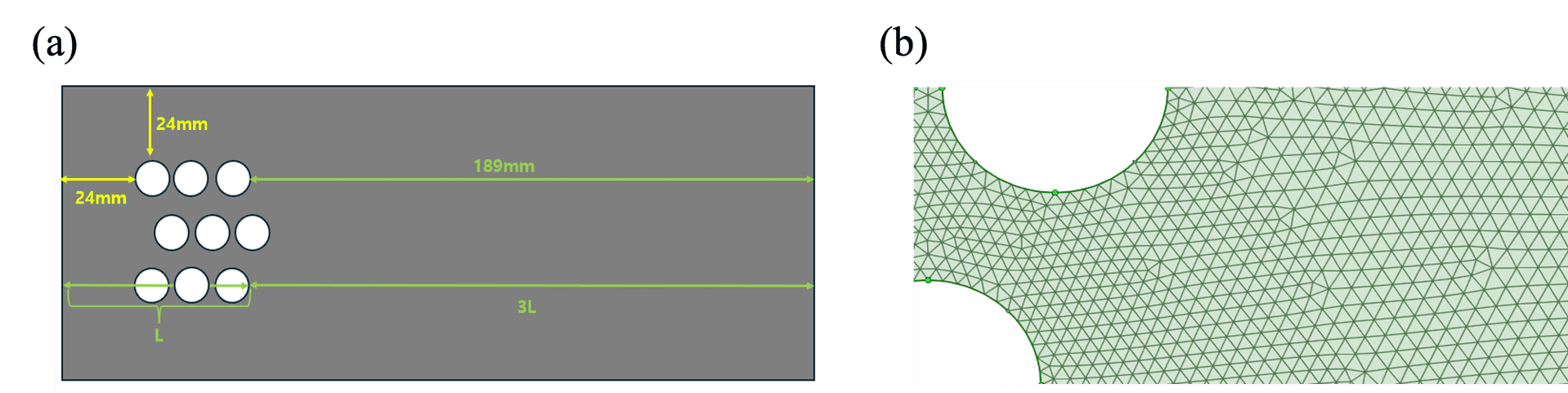}
    \caption{(a) Geometry information of HCSG. (b) Medium mesh visualization of HCSG.}
    \label{fig:Geometry}
\end{figure}

In this study, the HCSG of the SMART is adopted as the target geometry for constructing a high-fidelity CFD dataset. The cross-flow inside an HCSG governs not only the pressure drop and the convective heat transfer but also the flow-induced vibration (FIV) of the helical tubes, which can drive fretting wear and, in extreme cases, tube rupture through fluidelastic instability~\cite{jo2008flow}. Therefore, CFD-level fidelity is required to analyze these phenomena, which cannot be captured by system-level codes. Accelerating such analyses with AI-based surrogates enables rapid evaluation of diverse SMR HCSG geometries and operating conditions, supporting both design optimization and enhanced safety margins. A two-dimensional transient dataset representing the primary-side cross-flow was constructed using ANSYS SpaceClaim 24.2.0 for the geometry and mesh generation and ANSYS Fluent 24.2.0 for transient flow simulations. The rod-bundle geometry adopted a staggered-array configuration with offset rows to investigate wake deflection and vortex interactions. The computational domain was defined by setting the cylinder located in the first row of the first column as the reference point, from which the vertical and horizontal length ratios were established. The downstream length was set to three times the upstream length with respect to the last-column cylinder. The specific design parameters included a radial length of 252 mm, an axial length of 94 mm, a tube diameter of 12 mm, and pitches of 13.5 mm and 18.29 mm. The corresponding geometry and design parameters are shown in Figure~\ref{fig:Geometry} and Table~\ref{tab:geometry_parameter}, respectively.

\begin{table}[htbp]
\centering
\caption{Geometry Parameters}
\label{tab:geometry_parameter}
\renewcommand{\arraystretch}{1.15}
\begin{tabular}{>{\bfseries}p{0.45\linewidth} p{0.4\linewidth}}
\toprule
Parameter              & Length \\
\midrule
Radial Length          & \SI{252}{\milli\meter} \\
Axial Length           & \SI{94}{\milli\meter} \\
Tube Diameter          & \SI{12}{\milli\meter} \\
Pitch                  & \SI{13.5}{\milli\meter} / \SI{18.29}{\milli\meter} \\
Upstream Length        & \SI{24}{\milli\meter} \\
Downstream Length      & \SI{189}{\milli\meter} \\
\bottomrule
\end{tabular}
\end{table}

A two-dimensional unstructured mesh composed of triangular elements was employed. Triangular meshes offer superior flexibility for complex geometries and accurately represent curved surfaces, making them well-suited for capturing the K\'{a}rm\'{a}n vortex street in the cylinder wake and the intricate cross-flow patterns between adjacent cylinders. To assess the influence of spatial resolution on the simulation results, a mesh sensitivity study was performed with three different mesh sizes: coarse (1.6 mm), medium (1.0 mm), and fine (0.6 mm). For accurate resolution of the complex flow near the cylinders, the local mesh sizes were refined to 0.94 mm, 0.79 mm, and 0.31 mm, respectively. The corresponding total number of mesh elements was 5,066 for the coarse mesh, 19,458 for the medium mesh, and 58,912 for the fine mesh. The results of the mesh sensitivity analysis confirmed that the medium mesh (1.0 mm) effectively captures the K\'{a}rm\'{a}n vortex street comparably to the fine mesh, while the coarse mesh failed to sufficiently resolve the vortex structures. Considering the balance between computational cost, data storage efficiency, and prediction accuracy, the medium mesh was selected for data generation in this study. The mesh configuration is illustrated in Figure~\ref{fig:Geometry}.

To cover a range of HCSG operating conditions, simulations were conducted for 50 inlet velocity conditions ranging from 0.10 m/s to 0.60 m/s (excluding 0.4 m/s). Among the 50 generated datasets, 45 were used for training, while five datasets corresponding to inlet velocities of 0.1 m/s, 0.2 m/s, 0.3 m/s, 0.5 m/s, and 0.6 m/s were used for validation. To evaluate the model's flow-field prediction performance, the datasets with inlet velocities of 0.4 and 0.7 m/s were used as test data. The outlet boundary condition was specified as a gauge pressure of 0 Pa, and a no-slip condition was applied to all solid walls. Water was used as the working fluid, with a density of 998.2 kg/m\textsuperscript{3} and a viscosity of 0.001 Pa$\cdot$s. To accurately capture the turbulent boundary-layer behavior, the shear stress transport (SST) $k$--$\omega$ turbulence model was adopted. The specific solver settings are summarized in Table~\ref{tab:solver_settings}.

\begin{table}[H]
\centering
\caption{Solver Settings}
\label{tab:solver_settings}
\renewcommand{\arraystretch}{1.3}
\begin{tabular}{>{\bfseries}p{0.2\linewidth} p{0.22\linewidth}
                >{\bfseries}p{0.22\linewidth} p{0.22\linewidth}}
\toprule
CFD Tool            & Ansys FLUENT          & Fluid Material    & Water \\
Solver              & Pressure-based        & Density           & 998.2 [\si{\kilogram/\meter\cubed}] \\
Turbulence model    & SST $k$--$\omega$     & Viscosity         & 0.001 [\si{\pascal\cdot\second}] \\
Tube wall condition & No-slip condition     & \# of Time Steps  & 1{,}000 \\
Heat                & Not considered        & Time step size    & 0.01 s \\
\bottomrule
\end{tabular}
\end{table}

The CFD solver computed 1{,}000 time steps with a step size of 0.01 s, and every 10th snapshot was extracted to construct a dataset of 100 time steps per simulation, resulting in a total of 5{,}000 instantaneous snapshots across the 50 CFD simulations. The resulting dataset comprehensively represents the unsteady flow characteristics within the HCSG, with particular emphasis on the formation of the K\'{a}rm\'{a}n vortex street in the wake region downstream of the tube bundle. The staggered-array configuration with a smaller radial spacing (252 mm) and a larger axial spacing (94 mm) intensified wake interactions, sustaining vortex streets under all flow conditions and producing enhanced turbulence and mixing effects, which are critical for characterizing cross-flow behavior in HCSG designs.

To improve the training efficiency and prediction accuracy of the surrogate model, the dataset was reformulated in terms of temporal difference fields. Specifically, the initial field $u_0(\mathbf{x})$ was subtracted from the field at each subsequent time step, yielding the difference field:

\begin{equation}
    \Delta u(\mathbf{x}, t) = u(\mathbf{x}, t) - u_0(\mathbf{x})
    \label{eq:diff_field}
\end{equation}
This formulation is motivated by the scale separation principle demonstrated in the finite volume method network (FVMN) proposed by Jeon et al.~\cite{jeon2022finite}, where the derivative system employs the temporal change between two consecutive time steps as the network output instead of the absolute field values. In fluid analysis, the scale of the actual variable value is substantially larger than the scale of its temporal change, and training the network to predict the full field directly can impair performance due to this scale imbalance. By subtracting the initial field, the difference representation reduces the dynamic range of the target data and allows the network to focus on capturing the transient flow variations, such as vortex shedding and wake development. During inference, the full field is recovered by adding the initial condition back to the predicted difference field:
\begin{equation}
    \hat{u}(\mathbf{x}, t) = \Delta\hat{u}(\mathbf{x}, t) + u_0(\mathbf{x})
    \label{eq:recover_field}
\end{equation}

\begin{figure}[H]
    \centering
    \captionsetup[subfigure]{labelformat=parens, labelsep=space, singlelinecheck=false, justification=raggedright, position=top}
    
    \begin{subfigure}[t]{\textwidth}
        \caption{}
        \label{fig:reference_full}
        \centering
        \setlength{\tabcolsep}{1pt}
        \makebox[\textwidth][c]{%
        \begin{tabular}{ccc}
            \textbf{$t = 2$} & \textbf{$t = 50$} & \textbf{$t = 100$} \\
            \includegraphics[width=0.4\textwidth]{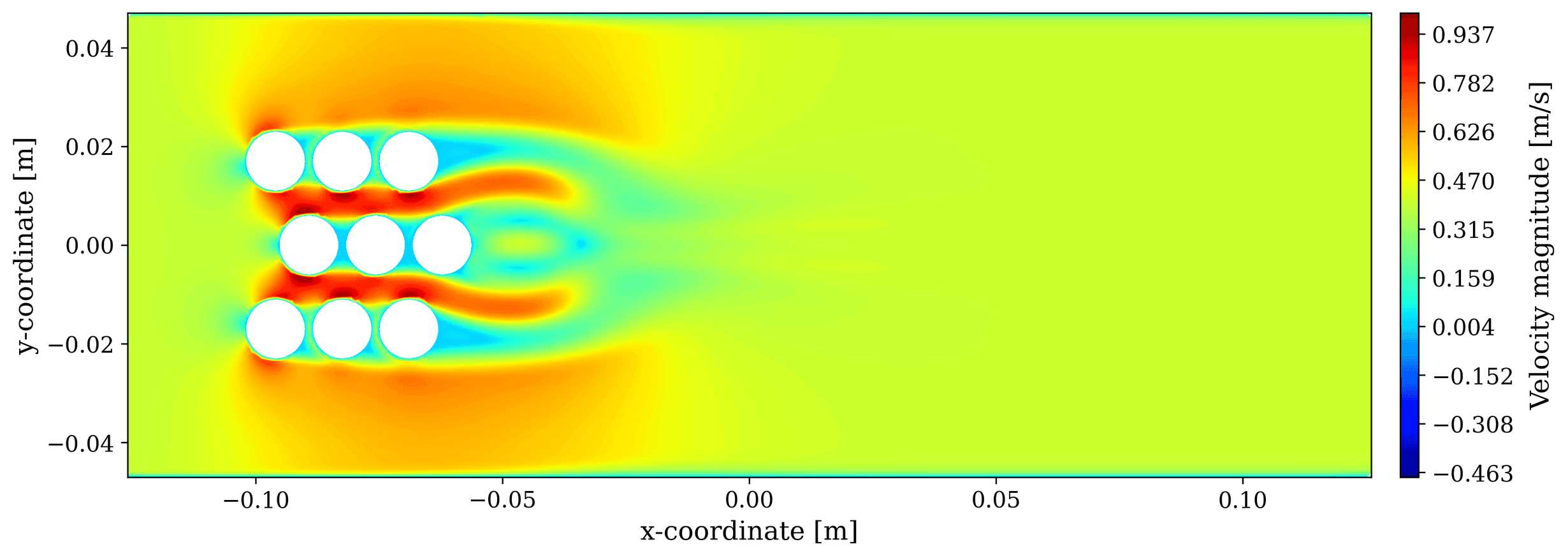} &
            \includegraphics[width=0.4\textwidth]{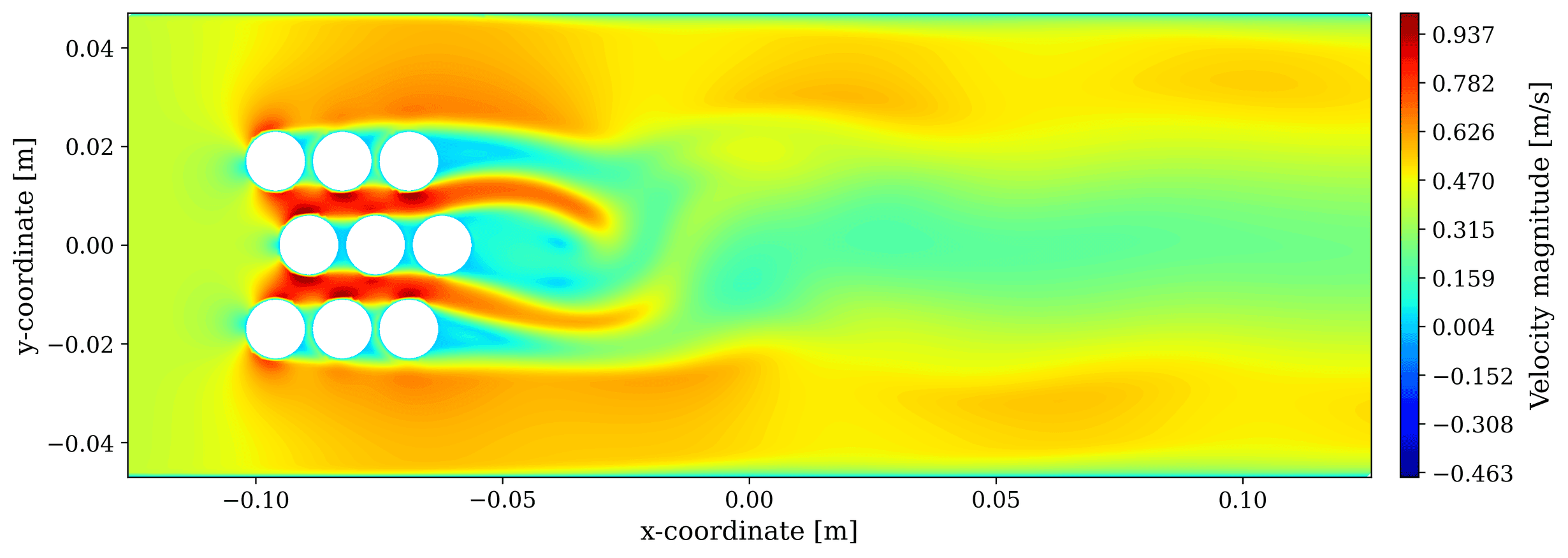} &
            \includegraphics[width=0.4\textwidth]{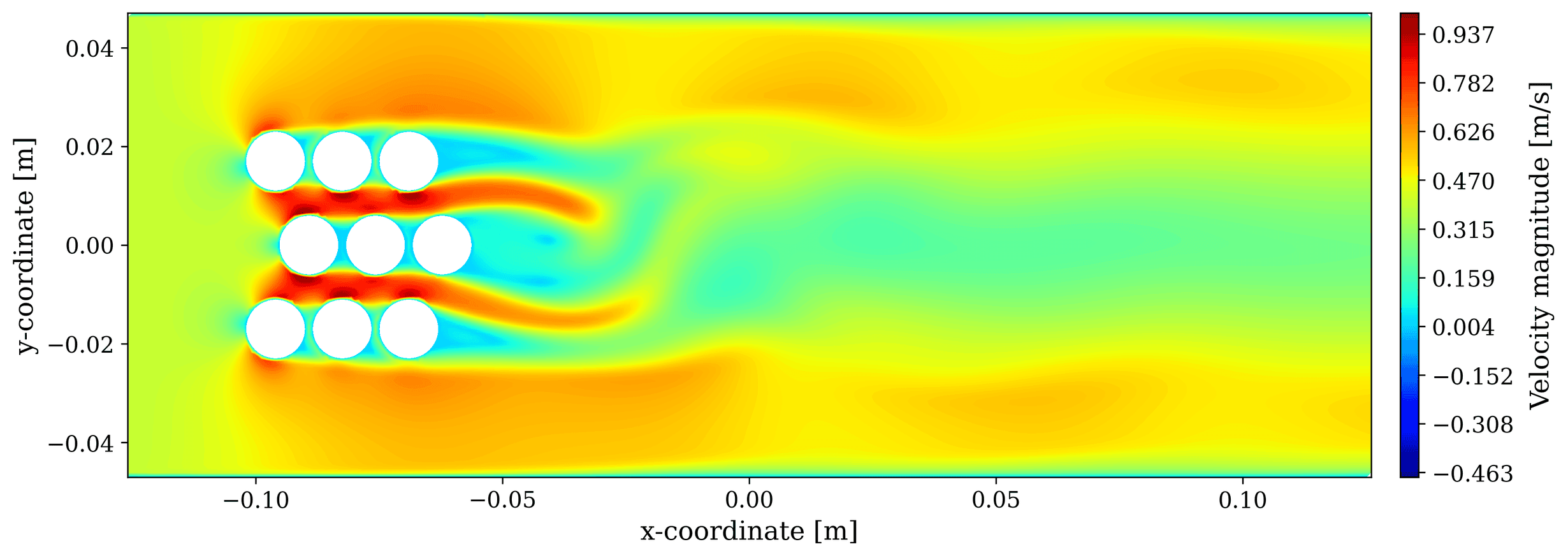} \\
        \end{tabular}
        }%
    \end{subfigure}
    
    \vspace{4pt}
    
    \begin{subfigure}[t]{\textwidth}
        \caption{}
        \label{fig:reference_diff}
        \centering
        \setlength{\tabcolsep}{1pt}
        \makebox[\textwidth][c]{%
        \begin{tabular}{ccc}
            \includegraphics[width=0.4\textwidth]{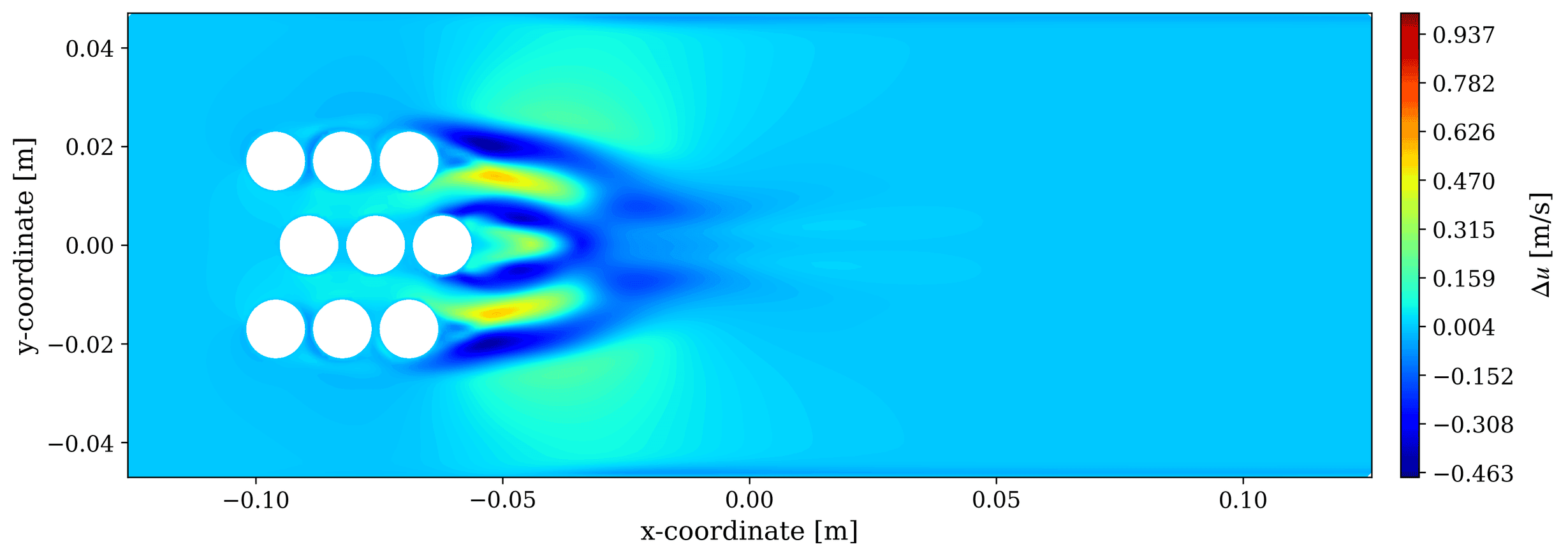} &
            \includegraphics[width=0.4\textwidth]{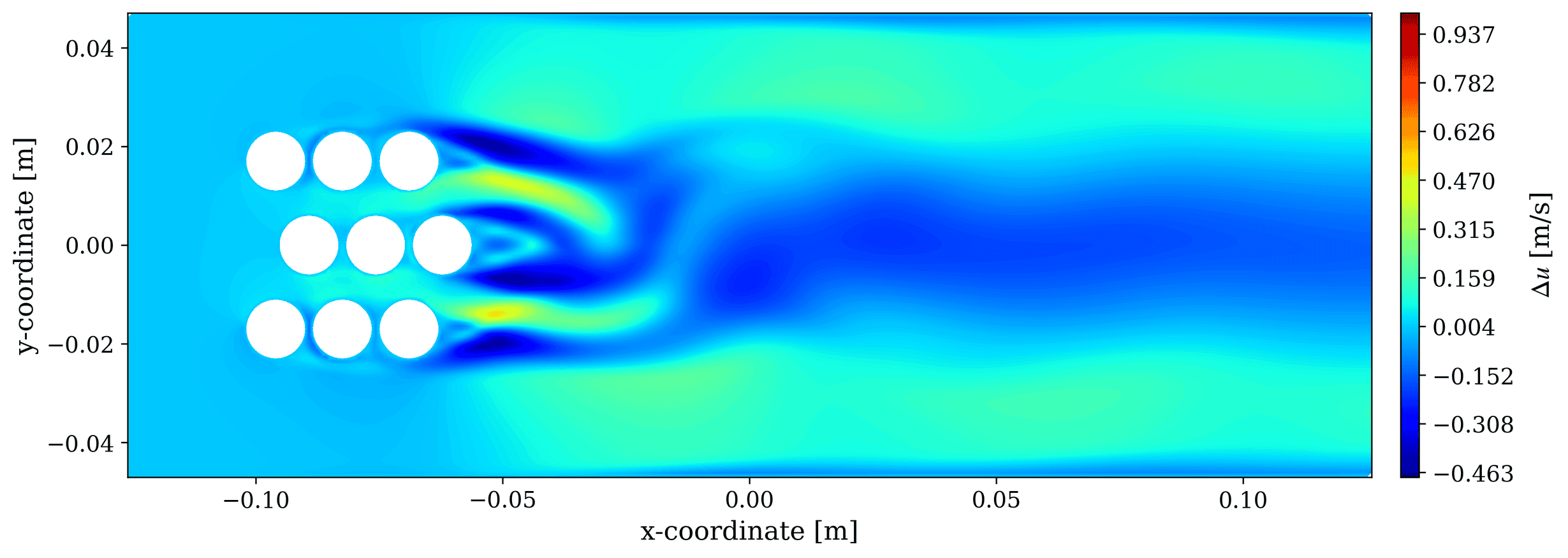} &
            \includegraphics[width=0.4\textwidth]{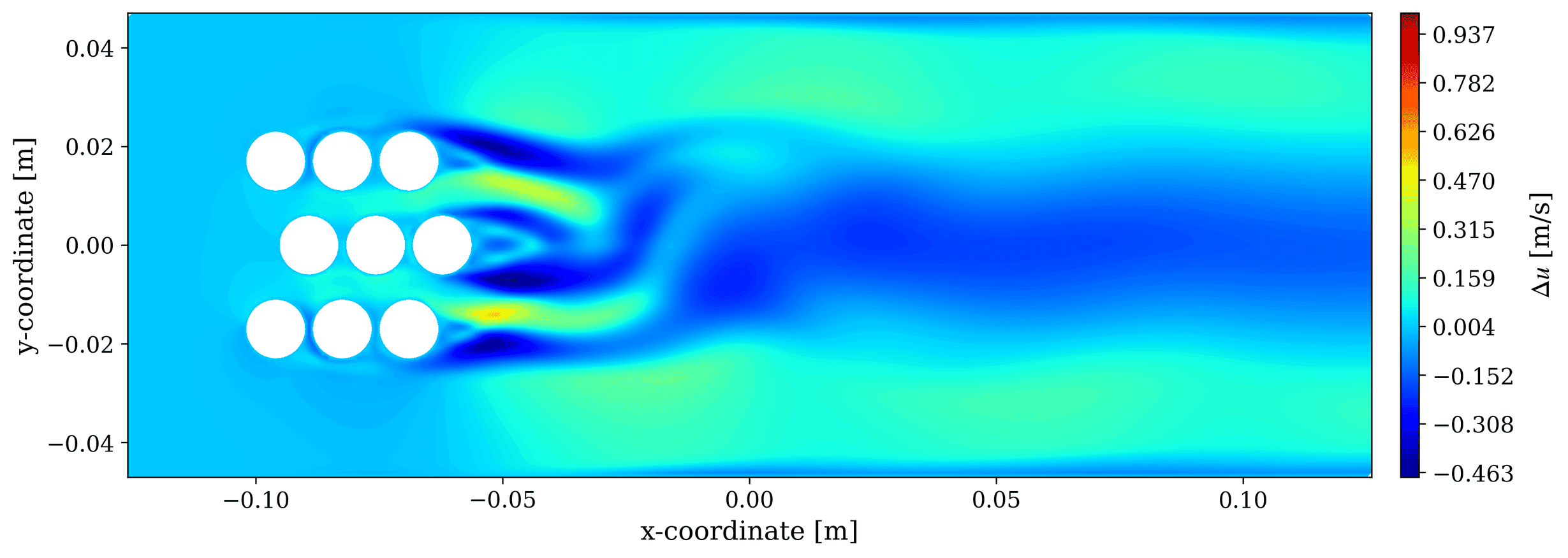} \\
        \end{tabular}
        }%
    \end{subfigure}
    
    \caption{Reference CFD solution at the inlet velocity of 0.4 m/s. (a) Full velocity field $u(\mathbf{x}, t)$, illustrating the temporal development of the K\'{a}rm\'{a}n vortex street downstream of the cylinder bundle. (b) Temporal difference field $\Delta u(\mathbf{x}, t) = u(\mathbf{x}, t) - u_0(\mathbf{x})$, used as the surrogate-model training input. The flow remains relatively uniform at the early time step ($t = 2$) and progressively develops into a fully periodic vortex streets regime at $t = 50$ and $t = 100$.}
    \label{fig:reference_velocity_field}
\end{figure}

Figure~\ref{fig:reference_velocity_field} compares the full velocity field, in which the K\'{a}rm\'{a}n vortex street progressively develops downstream of the cylinder bundle, with the corresponding temporal difference field used as the surrogate-model training input, at the inlet velocity of 0.4 m/s.

\subsection{Structured dataset using inverse distance weighting}
\label{subsec4.2}

In the current two-dimensional framework, an MLP-based AE operating directly on unstructured data can achieve sufficient computational efficiency. However, extending the analysis to three-dimensional CFD simulations of complex SMR geometries would expose a fundamental limitation of fully connected architectures: the number of trainable parameters grows exponentially with input dimensionality, making such models computationally prohibitive. To address this, the present study also evaluates a CAE, which significantly reduces the number of trainable parameters while leveraging spatially structured feature maps for efficient hierarchical feature extraction. Since not only CAE but also FNO requires uniformly spaced Cartesian inputs, inverse distance weighting (IDW) interpolation~\cite{Cheng2025} was applied to project the unstructured CFD data onto structured meshes prior to model training.
 
IDW estimates the value at an unsampled location as a distance-weighted average of the $k$ nearest source nodes, with weights inversely proportional to the $p$-th power of the distance:
 
\begin{equation}
u^{\ast}(\mathbf{x})
    = \frac{\displaystyle\sum_{i=1}^{k} w_i(\mathbf{x})\,u_i}
           {\displaystyle\sum_{i=1}^{k} w_i(\mathbf{x})},
\qquad
w_i(\mathbf{x}) = \frac{1}{d(\mathbf{x},\,\mathbf{x}_i)^{p}},
\label{eq:idw}
\end{equation}
 
\noindent

where $\mathbf{x}$ is the target mesh point, $\mathbf{x}_i$ and $u_i$ denote the coordinate and field value of the $i$-th neighboring source node, $d(\cdot)$ is the Euclidean distance, $k$ is the number of nearest neighbors, and $p$ is the distance-decay exponent. The parameters $k{=}6$ and $p{=}1$ were determined empirically based on interpolation accuracy tests conducted on a prior dataset of a different geometry. Nearest-neighbor queries were accelerated using a  k-dimensional tree, enabling efficient processing across all cases and time steps.
 
To examine the effect of spatial resolution on the geometric representation of the cylindrical obstacles, interpolation was performed at three scales derived from the physical domain of the preceding section, with the mesh spacing progressively reduced from a baseline of approximately 1\,mm to one-third of that interval. The resulting configurations are listed in Table~\ref{tab:mesh_scales}.
 
\begin{table}[htbp]
\centering
\caption{Structured mesh configurations for the three interpolation scales.}
\label{tab:mesh_scales}
\begin{tabular}{cccc}
\hline
Scale & $N_x \times N_y$ & $\Delta x$ (mm) & $\Delta y$ (mm) \\
\hline
1$\times$ (a) & $252 \times 94$  & 1.004 & 1.011 \\
2$\times$ (b) & $504 \times 188$ & 0.501 & 0.503 \\
3$\times$ (c) & $756 \times 282$ & 0.334 & 0.335 \\
\hline
\end{tabular}
\end{table}
 
To identify solid interior regions, a distance-based masking criterion was applied. Points at which the distance to the nearest source node exceeded a prescribed threshold were set to zero:
 
\begin{equation}
u(\mathbf{x}) = \mathrm{0}
\quad\text{if}\quad
d_{\min}(\mathbf{x}) > 2.5\cdot\max(\Delta x,\,\Delta y).
\label{eq:mask}
\end{equation}
 
\noindent
When a threshold factor smaller than 2.5 was tested during initial validation, valid flow regions near the domain boundaries were incorrectly masked. This artifact arises in areas where the structured mesh spacing is finer than the local node spacing of the original unstructured mesh, causing the distance to the nearest source node to spuriously exceed the
threshold. Setting the factor to 2.5 eliminated this over-masking while accurately delineating the solid boundaries.

\begin{figure}[H]
    \centering
    \includegraphics[width=1\linewidth]{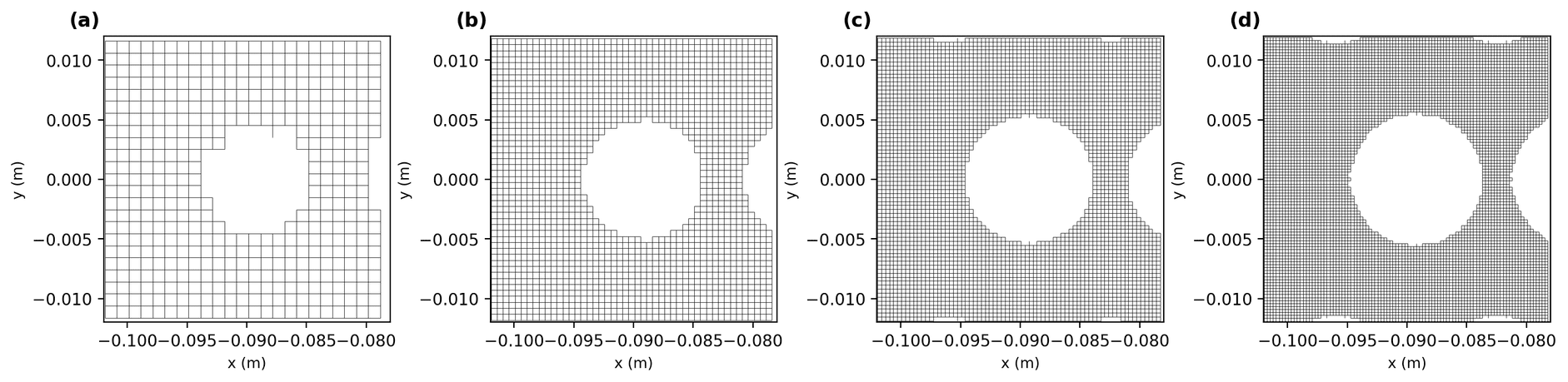}
    \caption{Mesh resolution comparison near a cylindrical obstacle.
(a) 1×, (b) 2×, (c) 3×, (d) 4x.}
    \label{fig:2d_structure_data_zoom}
\end{figure}

Figure~\ref{fig:2d_structure_data_zoom} presents a magnified comparison of a single cylinder at each scale. At scale (a), the circular boundary is approximated in a staircase fashion due to the coarse mesh resolution. The geometric fidelity improves progressively at scales (b) and (c), with scale (c) providing an accurate representation of the curved surface. At scale (d), the improvement in the curved boundary is marginal compared to (c), whereas the grid count increases by approximately 1.78 times. On this basis, considering both shape preservation and data size, the $3\times$ scale was selected as the final interpolation resolution for model training.
 
The two interpolated quantities pressure and velocity magnitude were stored separately as structured arrays of shape $[b, o, H, W]$. The inlet velocity conditions and dataset partitioning follow those established in the preceding section. To validate the interpolation quality, the pressure drop was evaluated as a quantitative metric. Under an initial flow velocity of 0.6~m/s, the pressure drop was averaged over 100 time steps and compared between the original unstructured mesh data and the preprocessed structured mesh data. The pressure drop error between the two datasets was found to be 0.063\%, confirming that the interpolation process preserves the physical characteristics of the flow field with negligible loss of accuracy.

\section{Helical Coil Steam Generator-Specific Operator Methods}
\label{sec5}

\subsection{Spectral Bias in L-DeepONet}
\label{subsec5.1}
When L-DeepONet was trained on the HCSG dataset described in Section~\ref{sec3}, we encountered a critical challenge known as spectral bias. Spectral bias refers to the tendency of deep networks, particularly ReLU-based MLPs, to approximate low-frequency (global, smooth) components more rapidly than high-frequency (local, rapidly varying) components during the training process.

This phenomenon arises from the combination of three factors. (i) ReLU networks operate as continuous piecewise linear functions, whose structure inherently causes spectral attenuation at higher frequencies, making high-frequency representation intrinsically disadvantageous. (ii) Gradient descent-based optimization exhibits a frequency-dependent convergence rate, where low-frequency components converge first while high-frequency components converge much later. (iii) High-frequency representations are less robust, as they depend more sensitively on precise parameter coordination and collapse more easily under parameter perturbation~\cite{rahaman2019spectral}.

\begin{figure}[H]
    \centering
    \includegraphics[width=\textwidth]{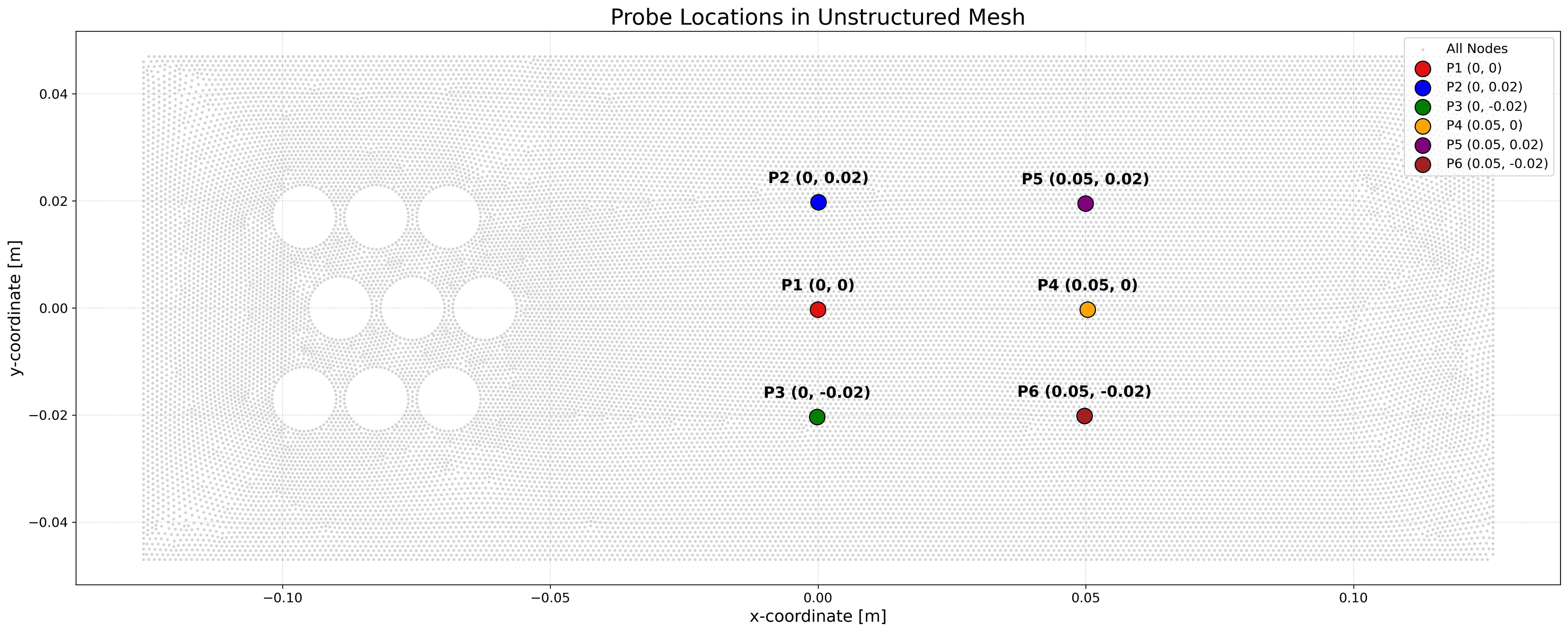}
    \caption{Probe locations on the unstructured mesh for evaluating the velocity at each node.}
    \label{fig:unstructured_probe_locations}
\end{figure}

This limitation is particularly damaging in the present problem because the K\'{a}rm\'{a}n vortex generated within the HCSG exhibits fully periodic flow patterns across the training data range of inlet velocities. Therefore, the surrogate model must reproduce these intrinsically high-frequency oscillatory structures. To evaluate this capability, we selected several representative probe locations on the unstructured mesh, as illustrated in Figure~\ref{fig:unstructured_probe_locations}, and compared the predicted velocity time histories against the reference CFD solutions. These probe nodes were chosen to evaluate the model's ability to predict regions with high velocity fluctuations.

\begin{figure}[H]
    \centering
    \begin{subfigure}[b]{\textwidth}
        \centering
        \includegraphics[width=0.85\textwidth]{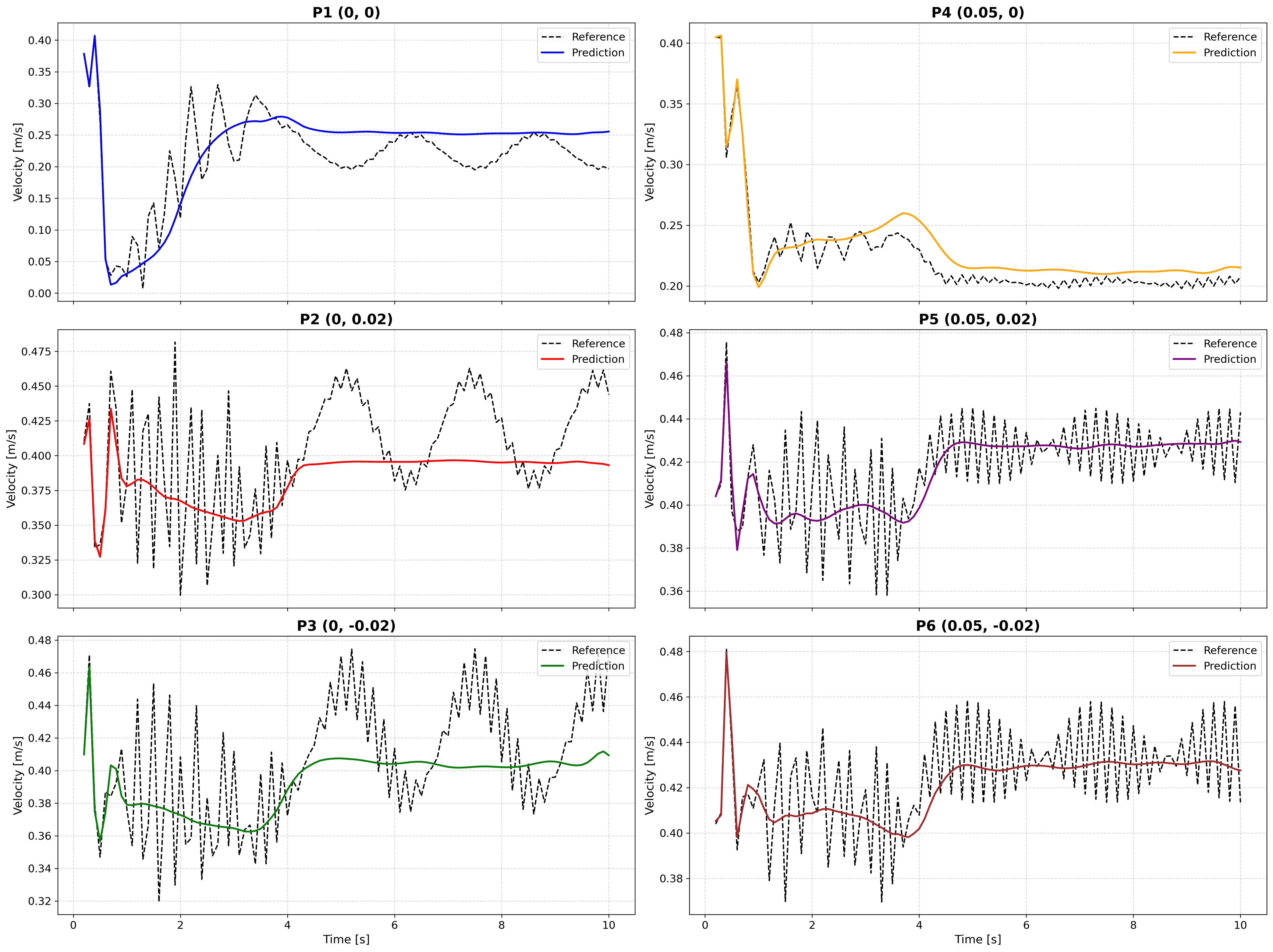}
        \caption{Inlet velocity 0.4 m/s}
        \label{fig:without_multiscale_040}
    \end{subfigure}
    \vspace{0.2cm}
    \begin{subfigure}[b]{\textwidth}
        \centering
        \includegraphics[width=0.85\textwidth]{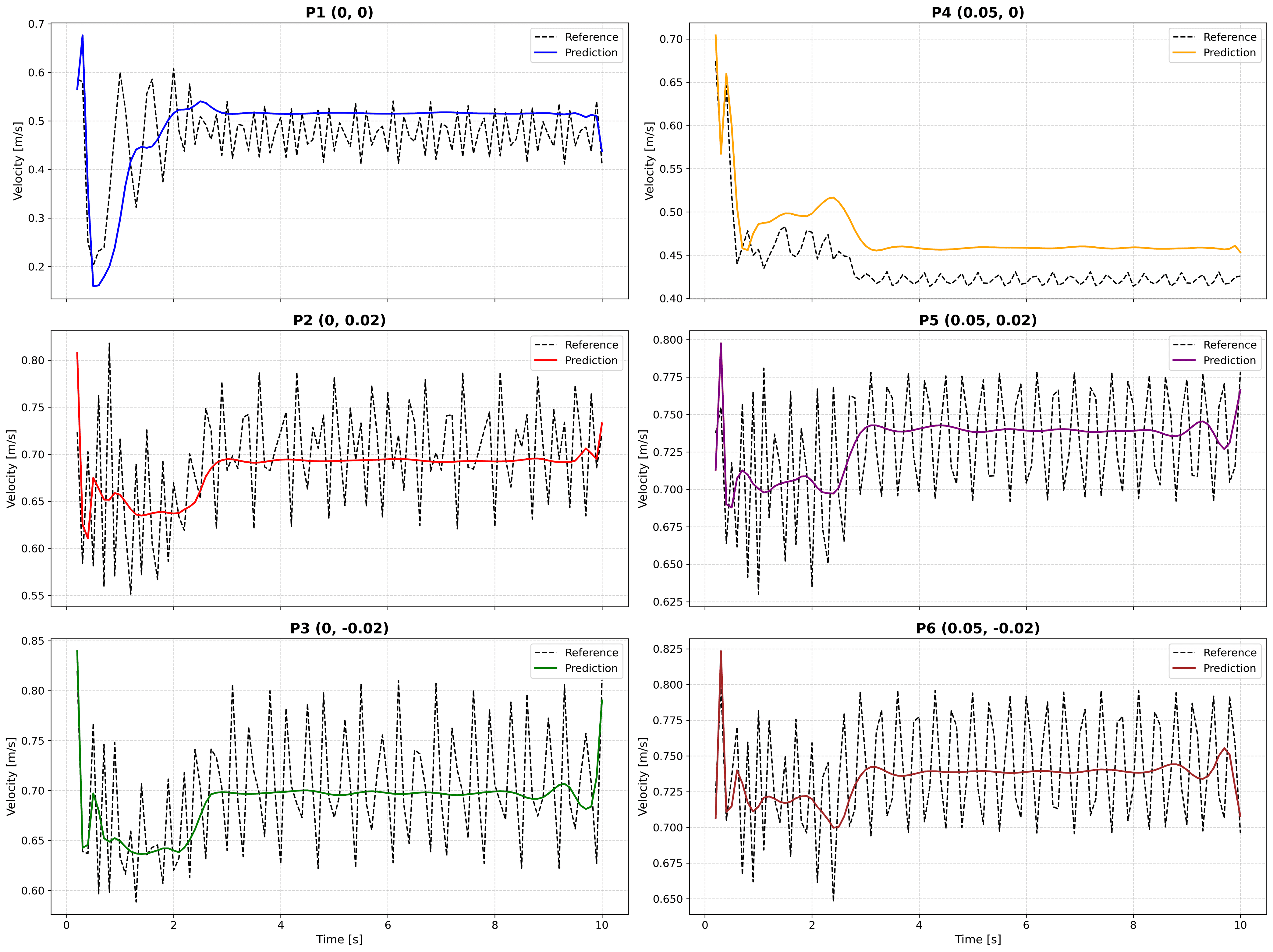}
        \caption{Inlet velocity 0.7 m/s}
        \label{fig:without_multiscale_070}
    \end{subfigure}
    \caption{Velocity time histories predicted by the vanilla L-DeepONet at probe locations for inlet velocities of (a) 0.4 m/s and (b) 0.7 m/s. Dashed lines indicate reference CFD and solid lines indicate predictions.}
    \label{fig:without_multiscale}
\end{figure}

However, as shown in Figure~\ref{fig:without_multiscale}, the vanilla L-DeepONet failed to predict the periodic flow variations at these probe locations, even though the training loss had converged sufficiently low. Instead of capturing the physically meaningful oscillatory patterns, the model predicted only the time-averaged mean flow field, which is a characteristic manifestation of spectral bias.

In this study, three approaches were employed to address the spectral bias problem: (1) a comparison of FNO results with varying Fourier mode configurations, (2) the application of a multi-scale technique to the L-DeepONet, and (3) the application of a multi-scale technique to the FNO (MscaleFNO).

\subsection{Fourier Mode Analysis in FNO}
\label{subsec5.2}
While the FNO operates directly in the Fourier domain by retaining a finite number of frequency modes, the relationship between mode truncation and spectral bias has not been systematically investigated. Therefore, we examine whether adjusting the number of retained Fourier modes in the FNO can mitigate spectral bias when learning periodic K\'{a}rm\'{a}n vortex flows.

\begin{figure}[H]
    \centering
    \begin{subfigure}[b]{\textwidth}
        \centering
        \includegraphics[width=\textwidth]{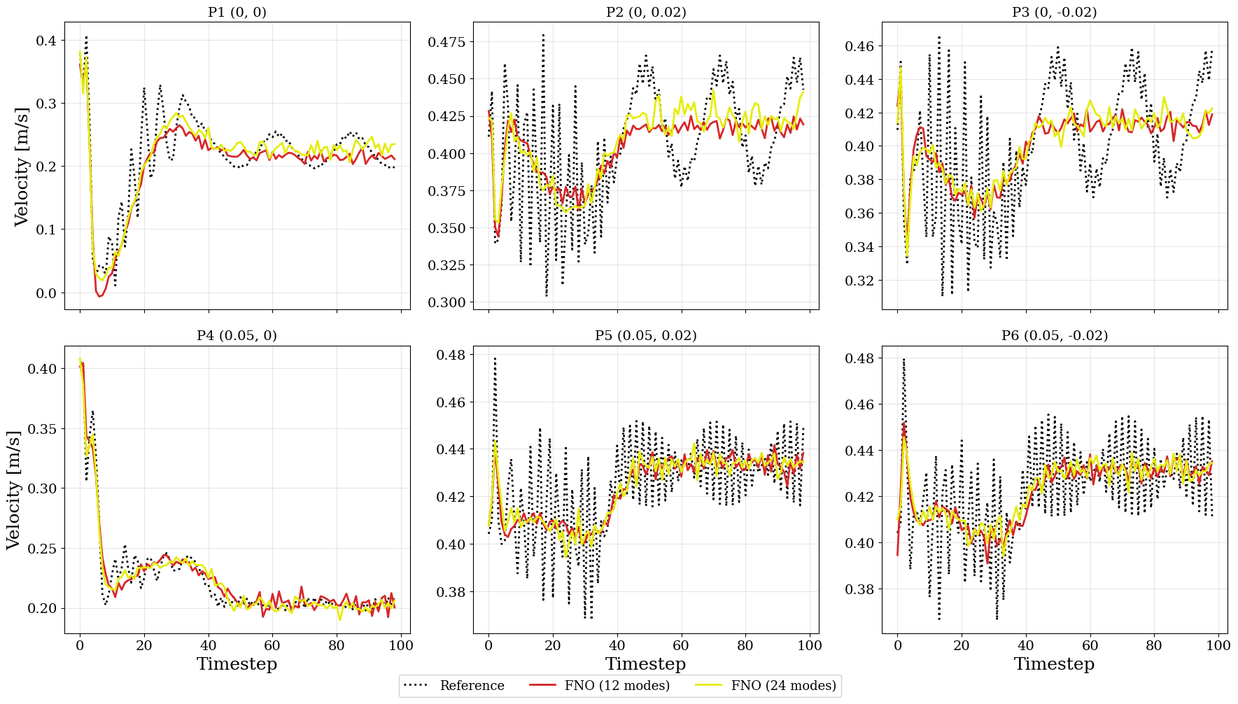}
        \caption{Inlet velocity 0.4 m/s}
        \label{fig:fno_mode_040}
    \end{subfigure}
    \vspace{0.5cm}
    \begin{subfigure}[b]{\textwidth}
        \centering
        \includegraphics[width=\textwidth]{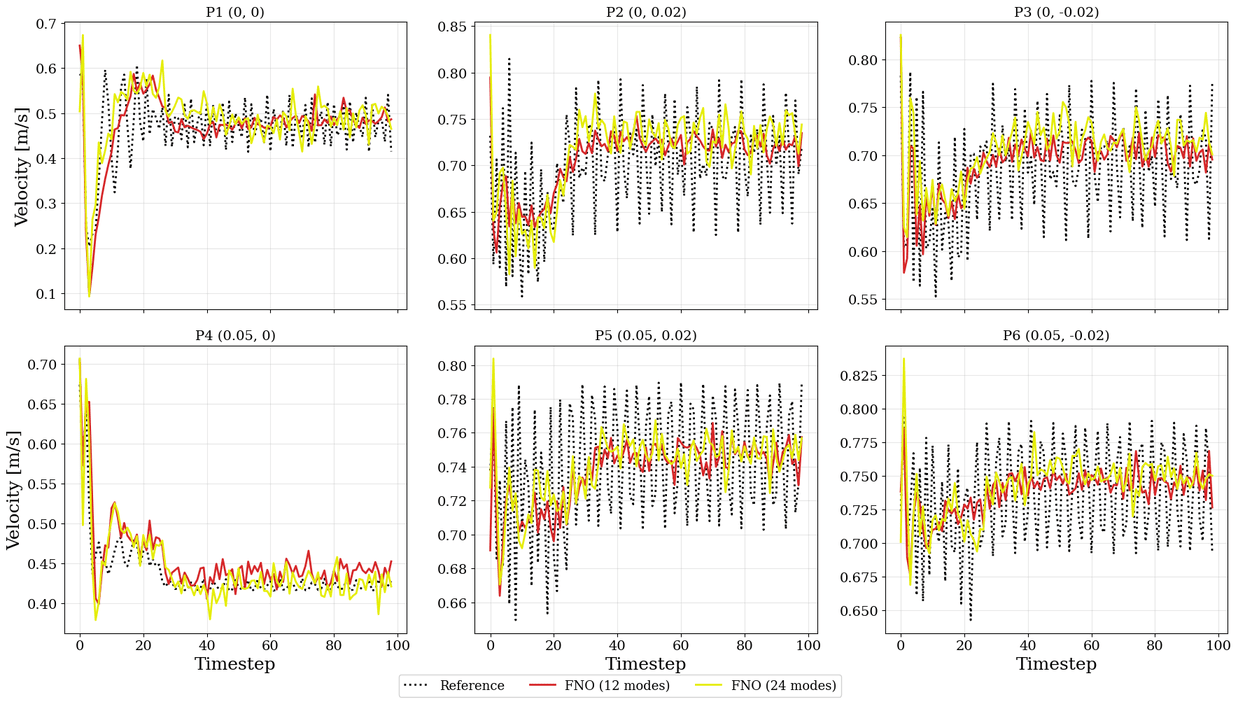}
        \caption{Inlet velocity 0.7 m/s}
        \label{fig:fno_mode_070}
    \end{subfigure}
    \caption{Comparison of FNO predictions with different Fourier mode configurations at six probe locations for two inlet velocity conditions: (a) 0.4 m/s and (b) 0.7 m/s. The black dashed lines represent the reference CFD solutions, the red solid lines represent the FNO with 12 Fourier modes, and the yellow solid lines represent the FNO with 24 Fourier modes. Doubling the Fourier modes does not significantly improve the prediction of high-frequency oscillatory patterns.}
    \label{fig:fno_mode_comparison}
\end{figure}

As shown in Figure~\ref{fig:fno_mode_comparison}, the FNO with 12 Fourier modes and 24 Fourier modes produced similar prediction results at all six probe locations. Although the FNO with 24 modes captures slightly more fluctuation compared to the 12-mode configuration, both models fail to reproduce the high-frequency oscillatory patterns observed in the reference CFD solutions. This result indicates that simply increasing the number of retained Fourier modes is insufficient to overcome the spectral bias in predicting the periodic K\'{a}rm\'{a}n vortex flow within the HCSG.

\subsection{Multi-Scale L-DeepONet}
\label{subsec5.3}

Unlike both FNO and vanilla DeepONet, L-DeepONet operates in a compressed low-dimensional latent space rather than directly in the Fourier domain or in infinite-dimensional function spaces. The AE employed for dimensionality reduction maps the input function into a low-dimensional space that retains only the most dominant features. During this compression, high-frequency features are suppressed and only low-frequency features are preserved; this applies equally to both MLP-based AE and CAE~\cite{asanjan2026improving}. Consequently, spectral bias arises in both the AE and DeepONet training stages of L-DeepONet, making the learning and prediction of K\'{a}rm\'{a}n vortex flow particularly challenging.

Liu et al. and Wang et al. both pointed out that DeepONet exhibits spectral bias characteristics, where low-frequency components are preferentially learned, leading to performance degradation in operator learning involving strongly oscillatory high-frequency functions ~\cite{liu2021multiscale, wang2025multi}. To mitigate this, both studies proposed Multi-scale DeepONet by combining multi-scale neural network techniques with the DeepONet architecture to simultaneously capture components across diverse frequency bands. Wang et al. demonstrated significant improvement over baseline DeepONet in high-frequency wave scattering problems, while Liu and Cai showed similar improvements in seismic wave response prediction for multi-story buildings. Notably, both studies demonstrated that applying the multi-scale technique to the trunk net, which receives spatial or temporal coordinates as input, is more effective than applying it to the branch net, yielding meaningful performance improvements. However, these studies validated the multi-scale technique primarily on problems governed by the Helmholtz equation or linear structural dynamics, where the oscillatory behavior originates from the mathematical structure of the governing equation itself. In contrast, the applicability of multi-scale neural operators to intrinsically nonlinear unsteady flows, such as K\'{a}rm\'{a}n vortex where periodicity emerges spontaneously from flow instability rather than being explicitly prescribed by the governing equation or boundary conditions, remains unexplored.

\begin{figure}[H]
    \centering
    \includegraphics[width=\textwidth]{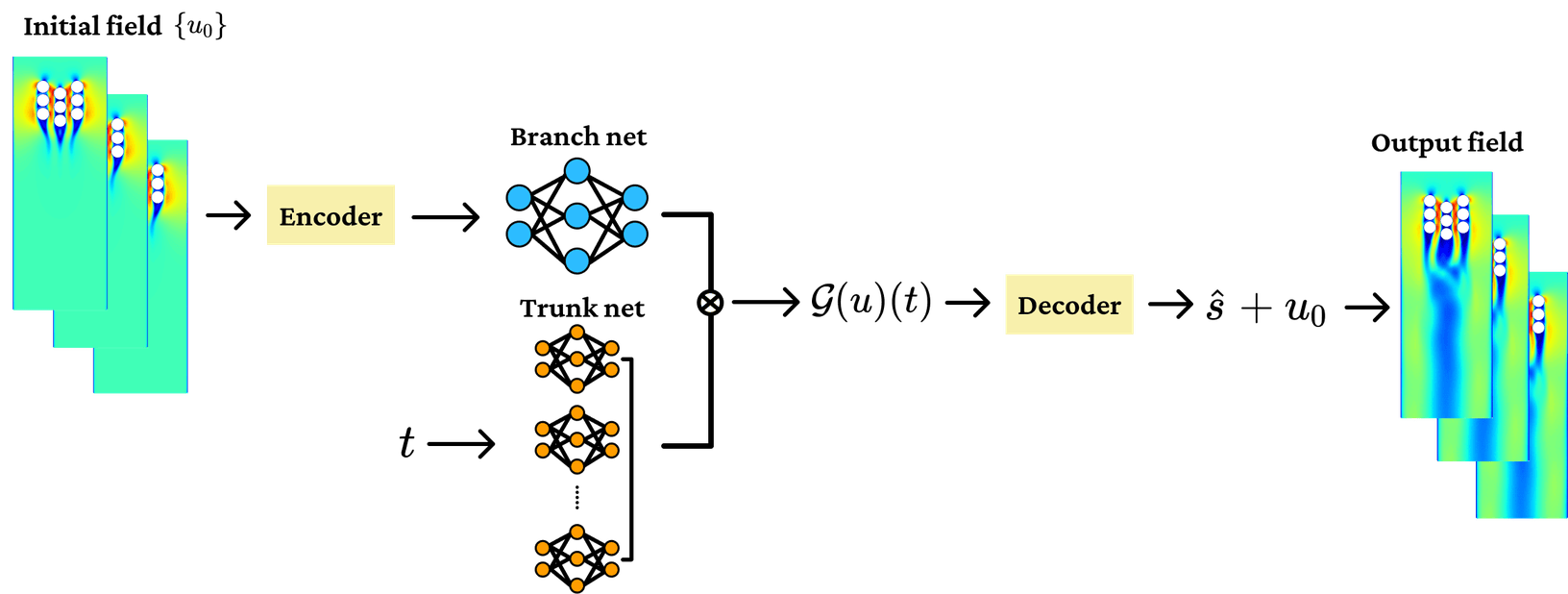}
    \caption{Architecture of the proposed Multi-scale L-DeepONet.}
    \label{fig:ms_ldon_architecture}
\end{figure}

The internal flow of the HCSG is characterized by unsteady flow dynamics dominated by periodic K\'{a}rm\'{a}n vortices. Drawing on the insights from these previous studies, we introduce the multi-scale technique into the L-DeepONet architecture to enhance its ability to capture high-frequency components. Specifically, parallel sub-networks that process inputs transformed at various scales are incorporated into the trunk network, which receives the temporal coordinate as input. The original temporal coordinate $t$ is scaled by predefined coefficients ($t$, $10t$, $20t$, \ldots, $500t$), and each scaled coordinate is fed into a separate sub-trunk network. This multi-scale trunk network structure effectively learns diverse basis representations spanning a wide range of frequency bands, thereby significantly improving the reconstruction accuracy of periodic and rapidly varying unsteady flow structures such as the K\'{a}rm\'{a}n vortex. The overall architecture of the proposed model is illustrated in Figure~\ref{fig:ms_ldon_architecture}.

\subsection{Multi-Scale FNO}
\label{subsec5.4}

In addition to the multi-scale L-DeepONet, we also investigate the multi-scale Fourier neural operator (MscaleFNO) proposed by You et al.~\cite{you2025mscalefno} as an alternative approach to mitigate the spectral bias in operator learning. While the multi-scale L-DeepONet applies scaling to the temporal coordinate in the trunk network, the MscaleFNO extends the multi-scale paradigm to the FNO framework through parallel sub-networks that process scaled inputs.

The key idea of MscaleFNO is inspired by the multi-scale deep neural network ~\cite{liu2020multi}, which uses input scaling to convert high-frequency components into lower-frequency ones that neural networks can learn more efficiently. For a general nonlinear operator with wide-band frequency content, a single scaling is insufficient to ensure uniform training across all frequency ranges. Therefore, the MscaleFNO employs a set of $N$ parallel FNO sub-networks, each operating at a distinct scale. In this study, the scaling is applied to the input flow field. The initial condition tensor with dimensions $[b, H, W]$ is multiplied by the scaling coefficient $c_n$ before being fed into each sub-network. The output of the MscaleFNO is formulated as:
\begin{equation}
    s(x,y) = \sum_{n=1}^{N} \gamma_n \, \text{FNO}_{\theta_n} \left[ c_n \cdot u_0(x, y) \right]
    \label{eq:mscalefno}
\end{equation}
where $\text{FNO}_{\theta_n}$ denotes the $n$-th FNO sub-network parameterized by $\theta_n$, $c_n$ is the scaling coefficient applied to the input flow field $u_0(x, y)$, and $\gamma_n$ is the combination weight. Both the scaling factors $\{c_n\}_{n=1}^{N}$ and the combination weights $\{\gamma_n\}_{n=1}^{N}$ are implemented as trainable parameters. Each sub-network maintains the same internal architecture as the standard FNO described in Section~\ref{subsec3.3}, including the arrangement and lifting layer, iterative Fourier layers, and the projection layer.

\begin{figure}[H]
    \centering
    \includegraphics[width=\textwidth]{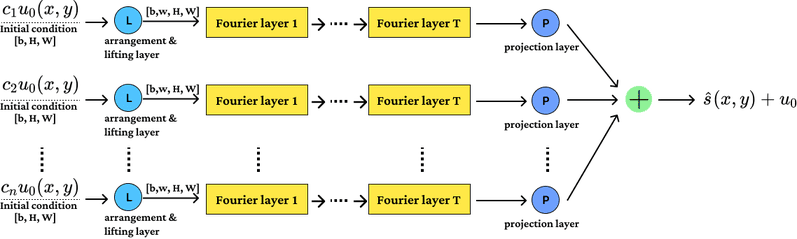}
    \caption{Architecture of the Multi-scale FNO (MscaleFNO). Each parallel sub-network receives the scaled initial condition $c_n \cdot u_0(x, y)$ as input, and their outputs are combined via a weighted summation.}
    \label{fig:mscalefno_architecture}
\end{figure}

This multi-scale design enables each sub-network to capture frequency components at its corresponding scale: sub-networks with smaller $c_n$ capture low-frequency patterns, while those with larger $c_n$ extract high-frequency details. The weighted summation of all sub-network outputs collectively achieves a comprehensive representation across the entire frequency spectrum. Furthermore, the sine activation function is employed throughout all Fourier layers, as activation functions with localized frequency spectra encourage frequency separation among different sub-networks~\cite{liu2020multi}. The overall architecture of the MscaleFNO is illustrated in Figure~\ref{fig:mscalefno_architecture}.

\section{Results and Discussion}
\label{sec6}
\subsection{Autoencoder Reconstruction Accuracy}
\label{subsec6.1}
 
Before evaluating the performance of the neural operator models, the reconstruction accuracy of the AEs was first assessed, as the fidelity of the dimensionality reduction directly affects the prediction quality of L-DeepONet. Both the MLP-based AE and the CAE were trained on the temporal difference fields of velocity magnitude, as described in Section~\ref{sec3}. The reconstruction performance was evaluated on two test conditions with inlet velocities of 0.4 m/s and 0.7 m/s, which were excluded from the training dataset.
 
\begin{figure}[H]
    \centering
    \setlength{\tabcolsep}{1pt}
    \makebox[\textwidth][c]{%
    \begin{tabular}{c@{\hspace{4pt}}ccc}
        & \textbf{$t = 2$} & \textbf{$t = 50$} & \textbf{$t = 100$} \\
        \adjustbox{valign=c}{\rotatebox[origin=c]{90}{\small\textbf{Reference}}} &
        \includegraphics[width=0.45\textwidth,valign=c]{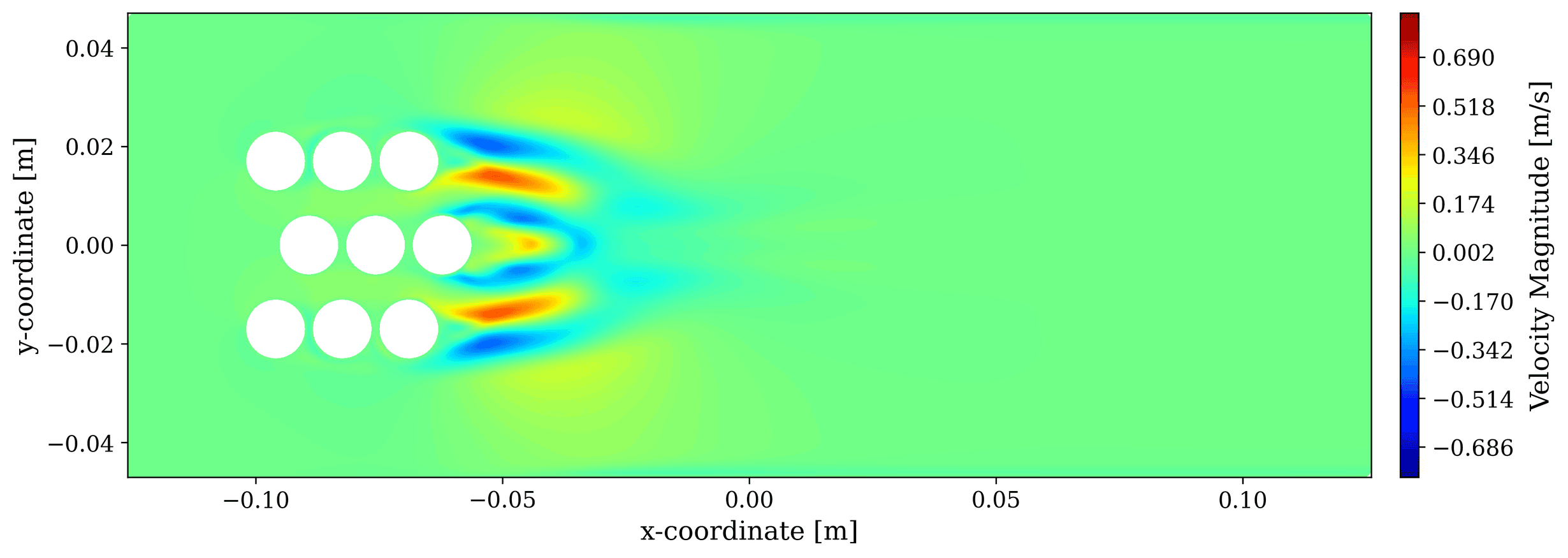} &
        \includegraphics[width=0.45\textwidth,valign=c]{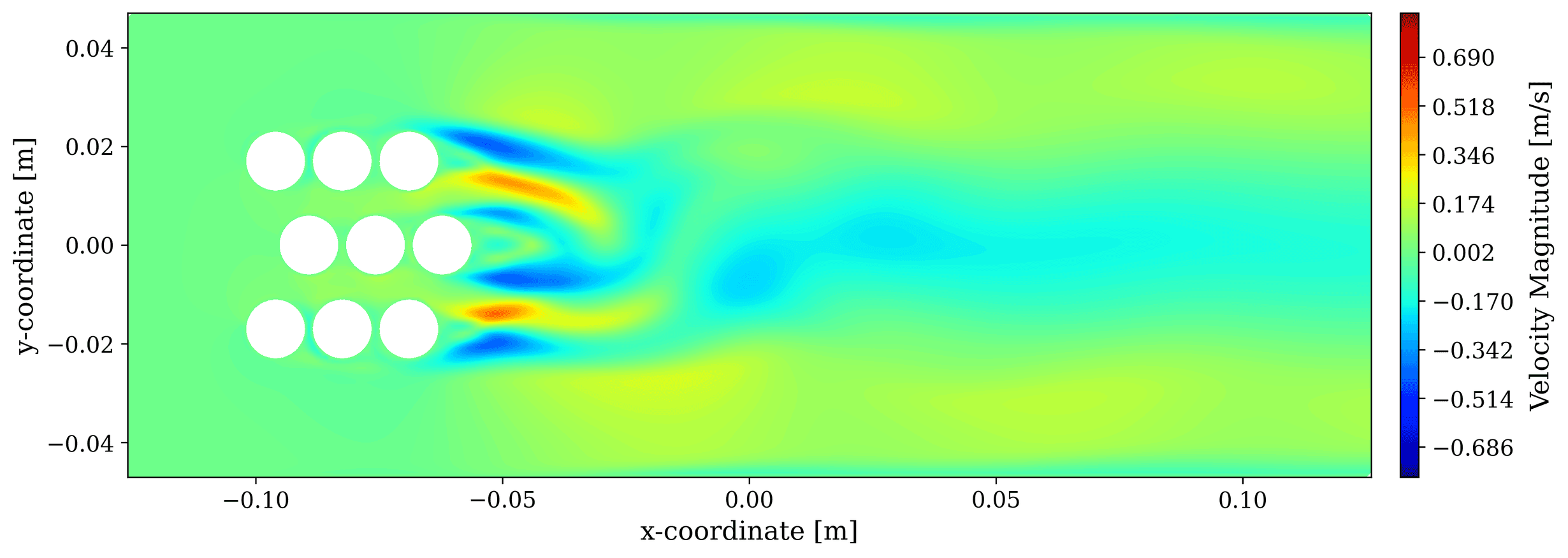} &
        \includegraphics[width=0.45\textwidth,valign=c]{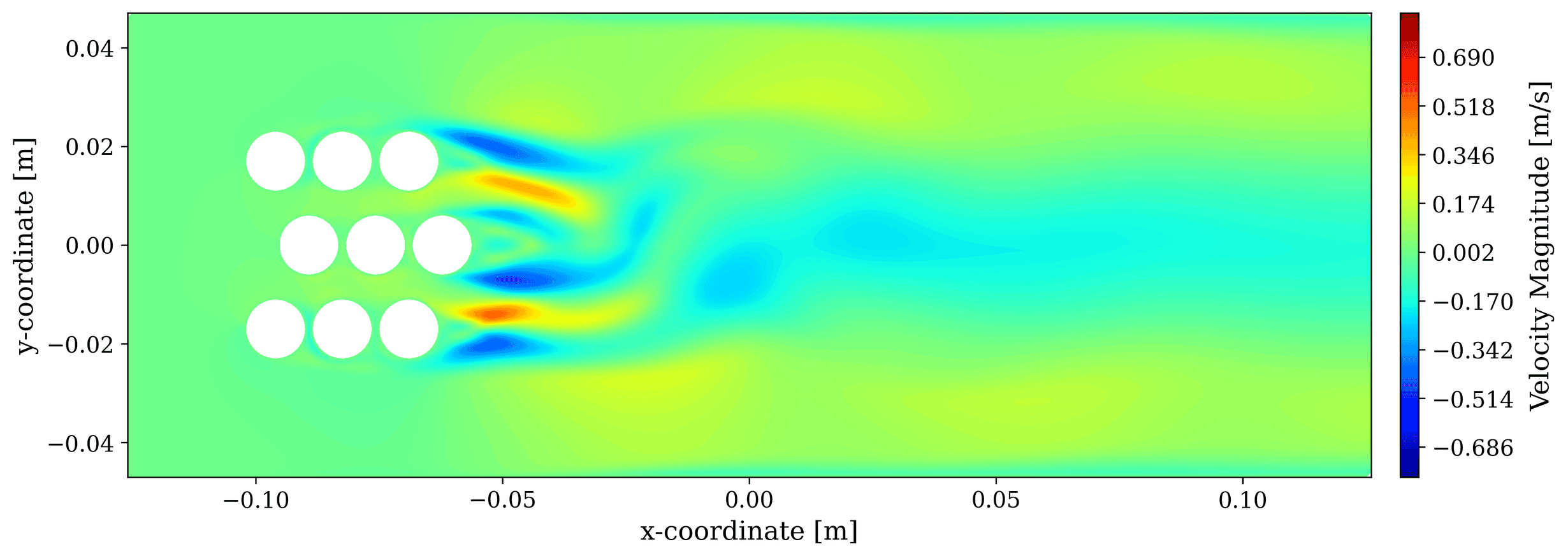} \\[2pt]
        \adjustbox{valign=c}{\rotatebox[origin=c]{90}{\small\textbf{Restored}}} &
        \includegraphics[width=0.45\textwidth,valign=c]{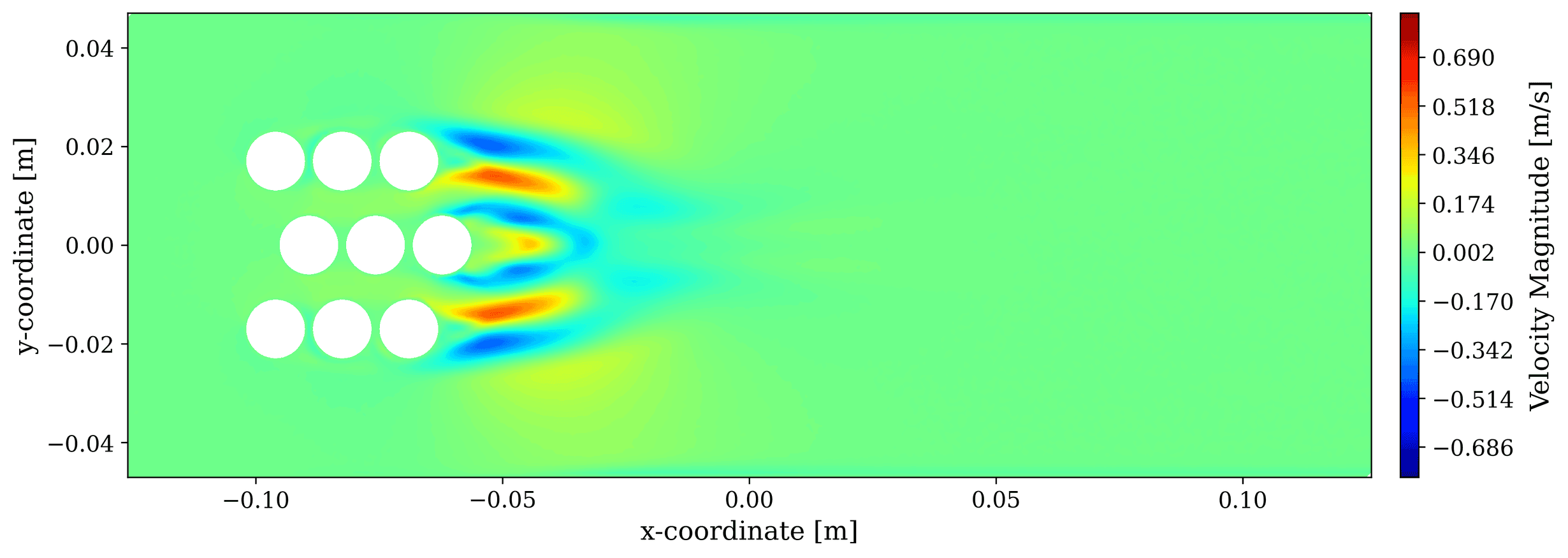} &
        \includegraphics[width=0.45\textwidth,valign=c]{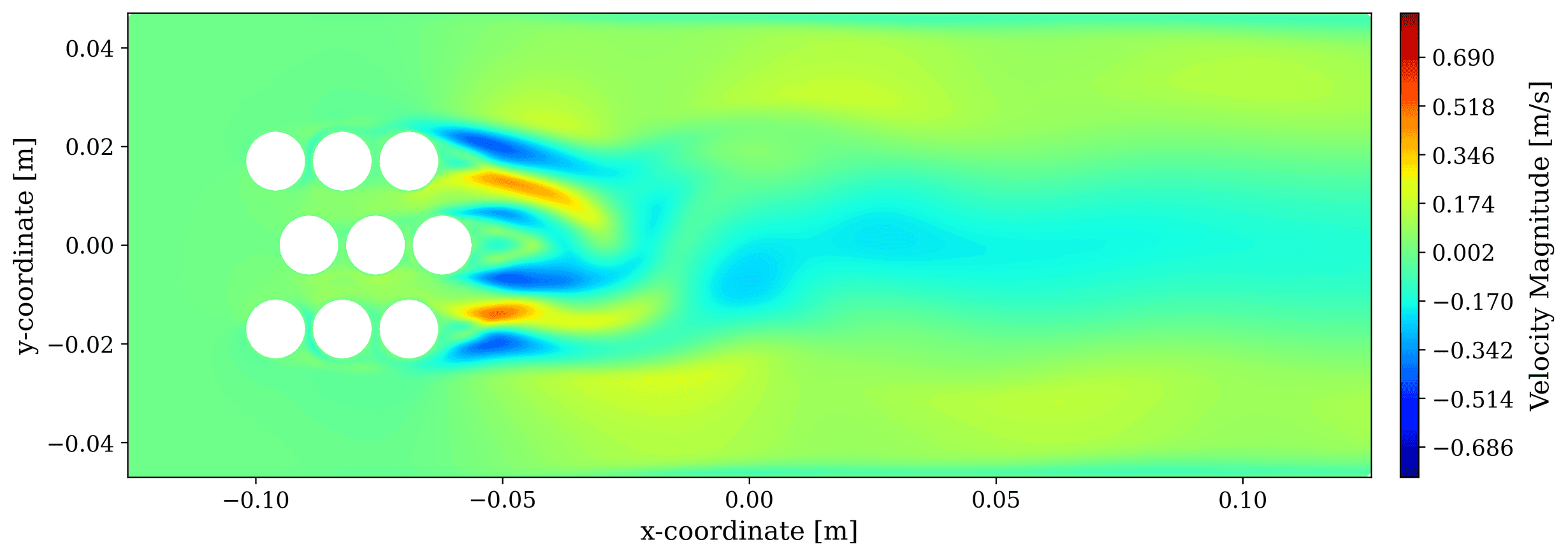} &
        \includegraphics[width=0.45\textwidth,valign=c]{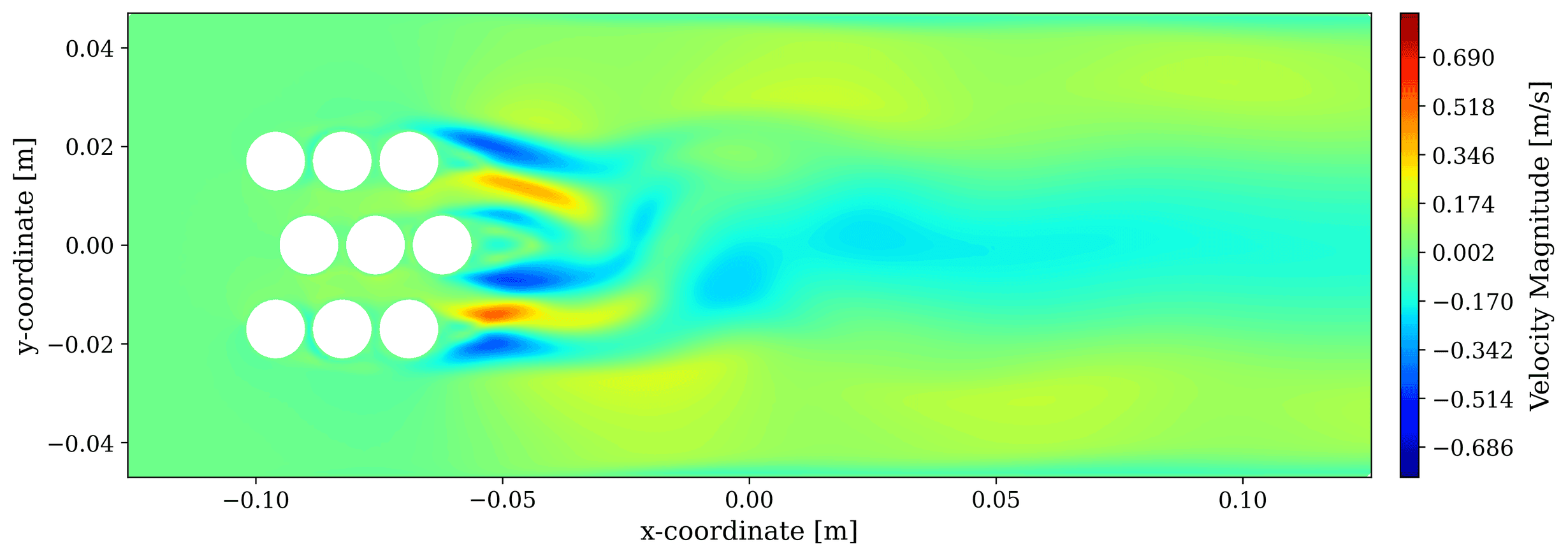} \\[2pt]
        \adjustbox{valign=c}{\rotatebox[origin=c]{90}{\small\textbf{Error}}} &
        \includegraphics[width=0.45\textwidth,valign=c]{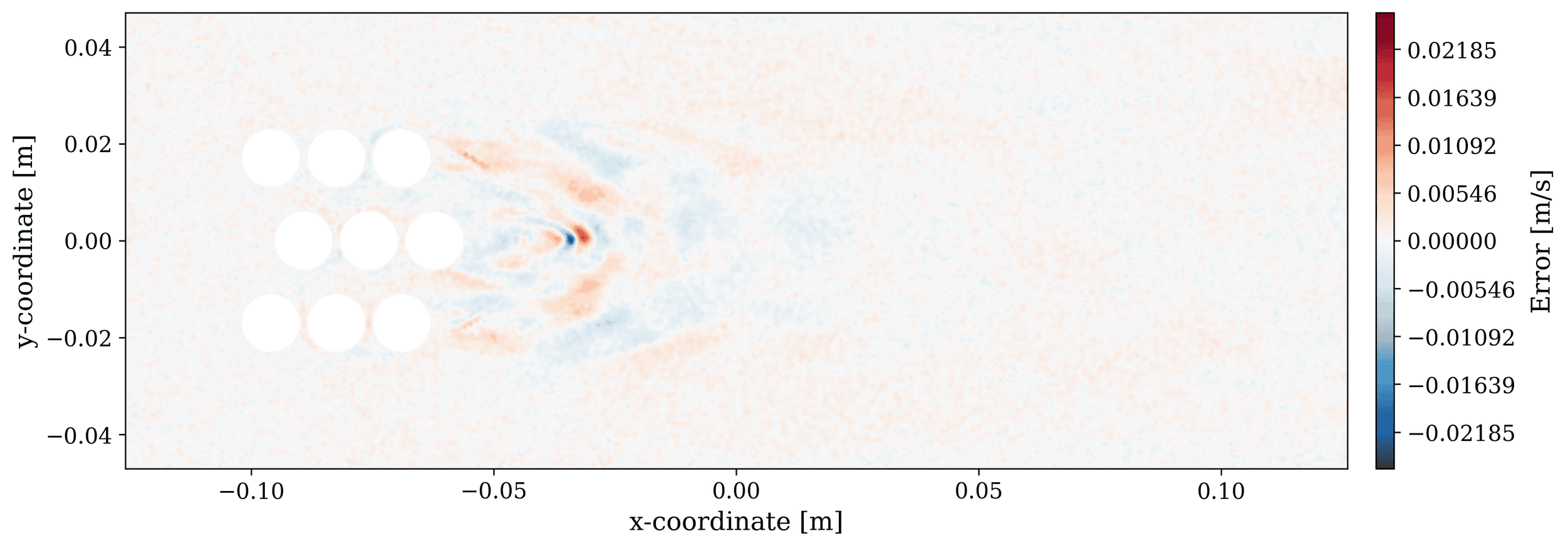} &
        \includegraphics[width=0.45\textwidth,valign=c]{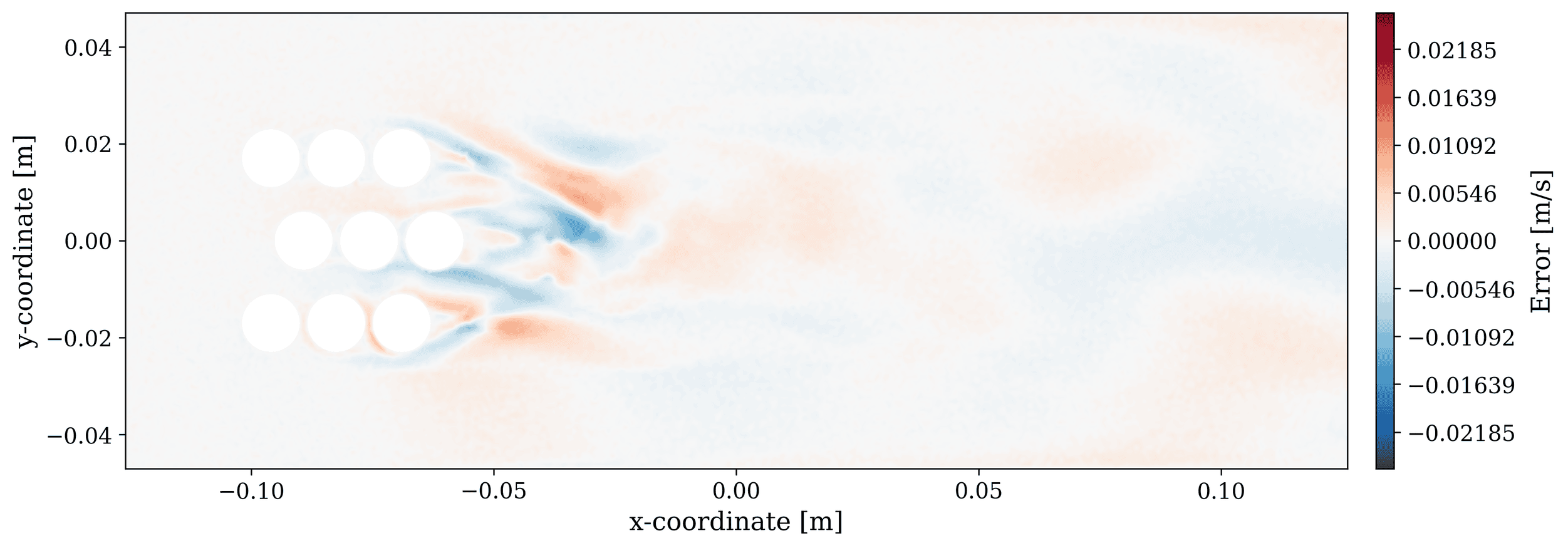} &
        \includegraphics[width=0.45\textwidth,valign=c]{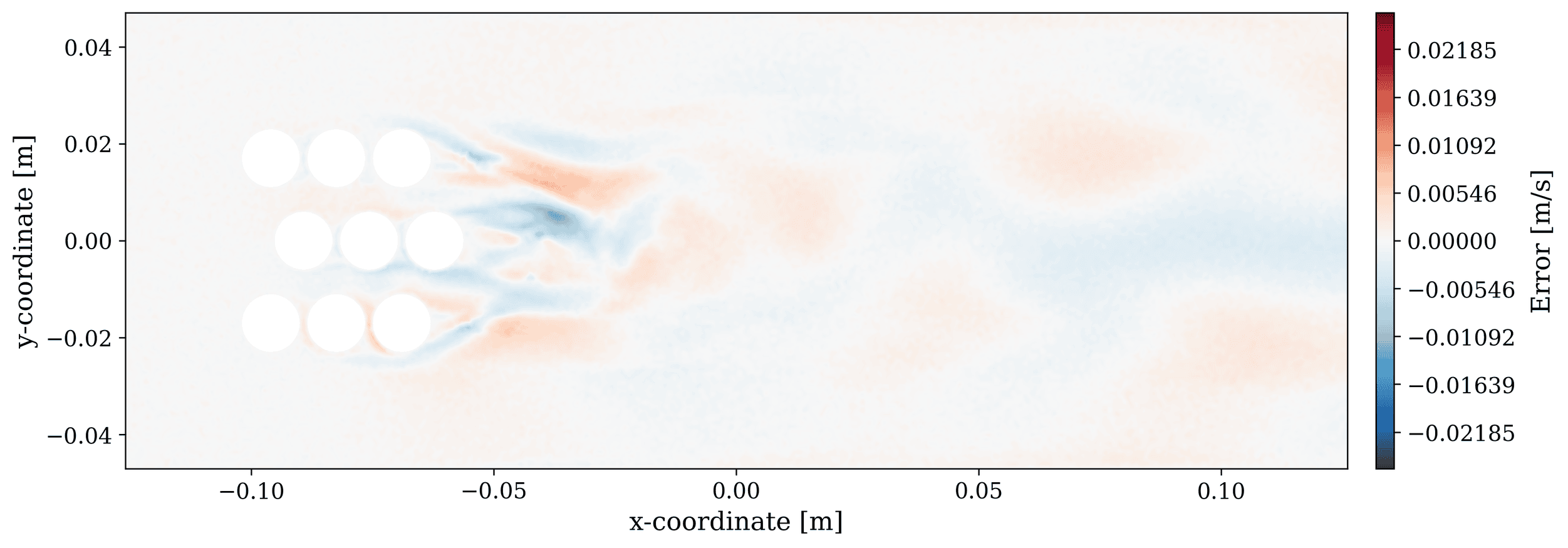} \\
    \end{tabular}
    }%
    \caption{MLP-based AE results for velocity field with inlet velocity 0.4 $m/s$. Reference, restored, and error at different timesteps.}
    \label{fig:ae_velocity_inlet040}
\end{figure}

\begin{figure}[H]
    \centering
    \setlength{\tabcolsep}{1pt}
    \makebox[\textwidth][c]{%
    \begin{tabular}{c@{\hspace{4pt}}ccc}
        & \textbf{$t = 2$} & \textbf{$t = 50$} & \textbf{$t = 100$} \\
        \adjustbox{valign=c}{\rotatebox[origin=c]{90}{\small\textbf{Reference}}} &
        \includegraphics[width=0.45\textwidth,valign=c]{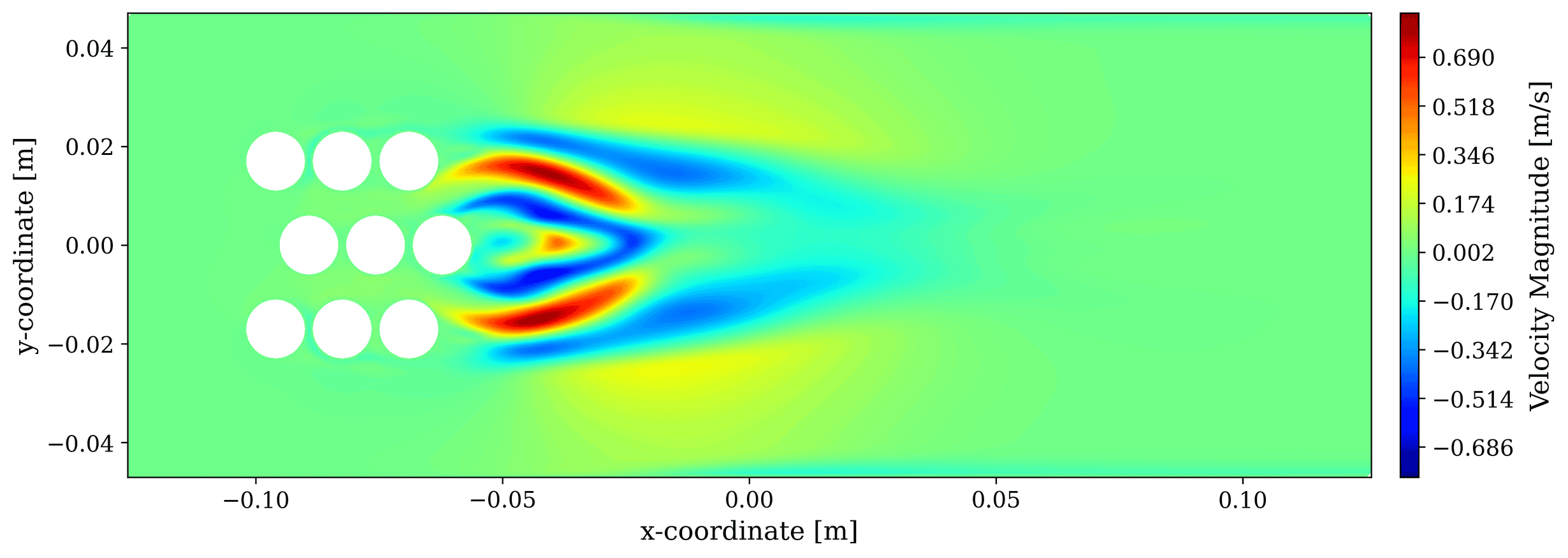} &
        \includegraphics[width=0.45\textwidth,valign=c]{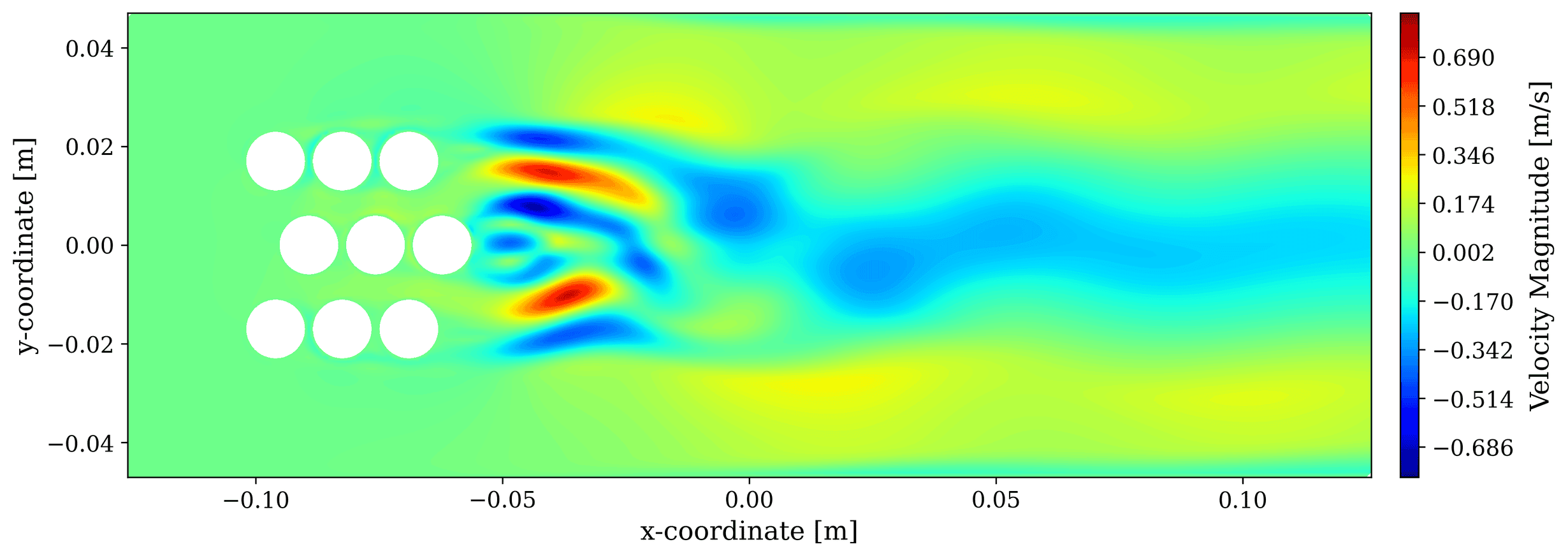} &
        \includegraphics[width=0.45\textwidth,valign=c]{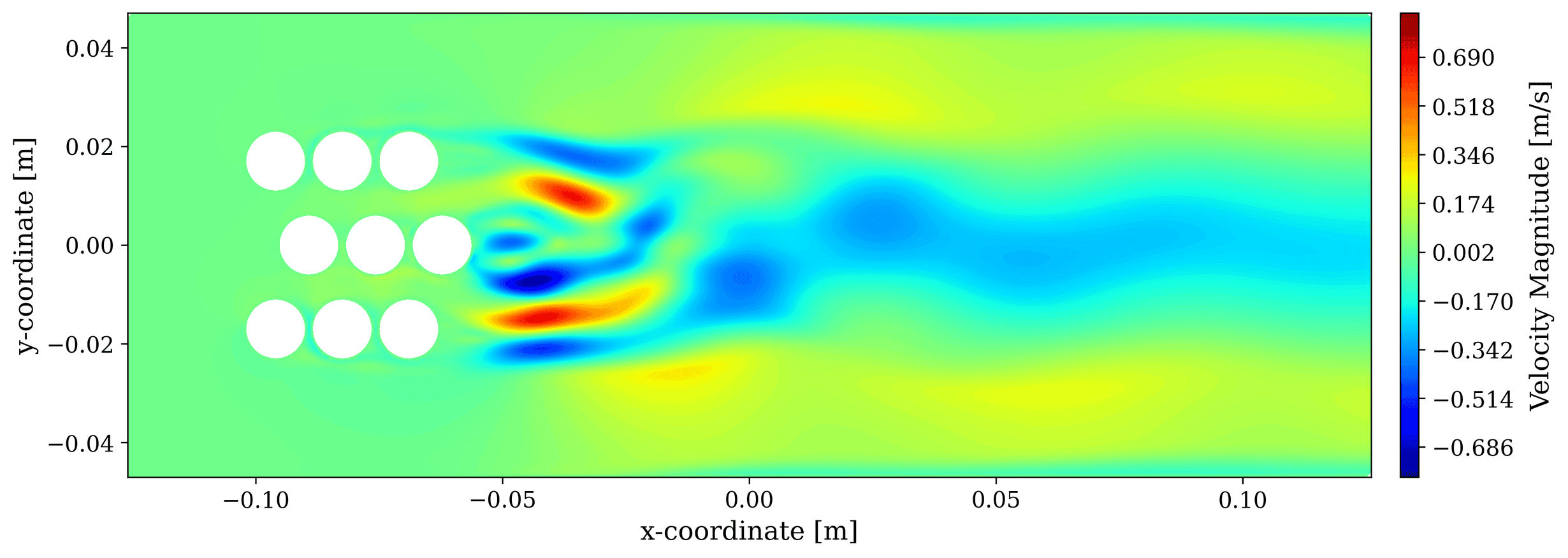} \\[2pt]
        \adjustbox{valign=c}{\rotatebox[origin=c]{90}{\small\textbf{Restored}}} &
        \includegraphics[width=0.45\textwidth,valign=c]{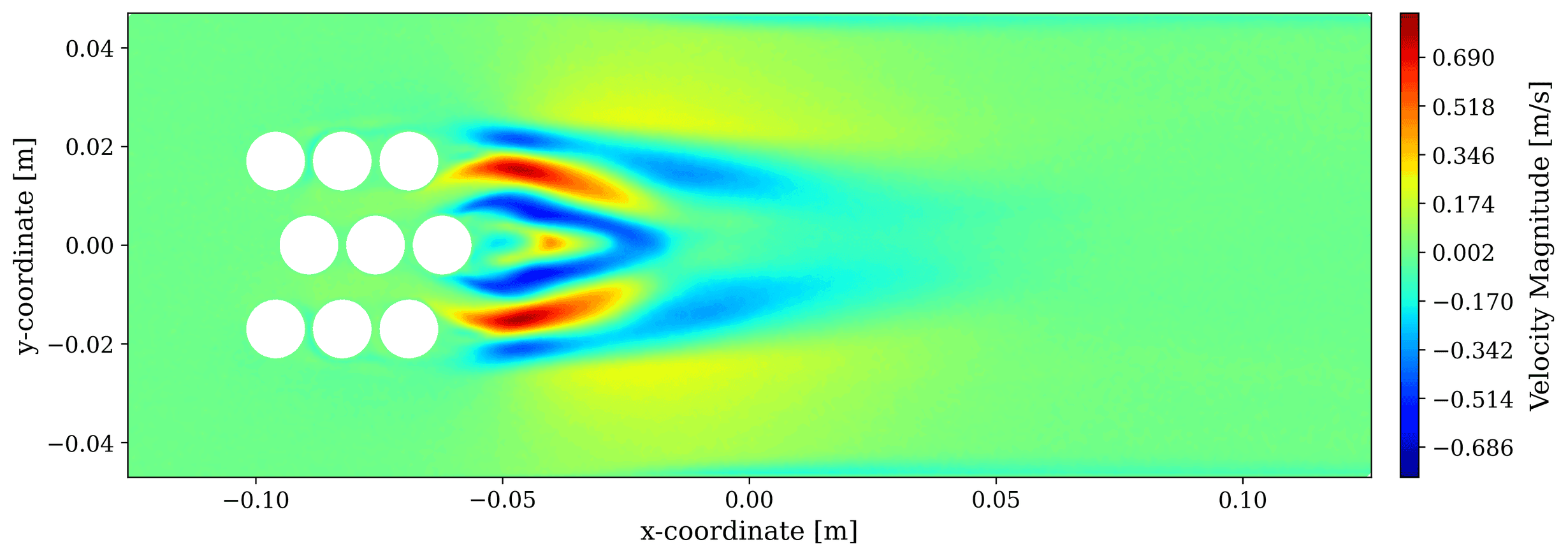} &
        \includegraphics[width=0.45\textwidth,valign=c]{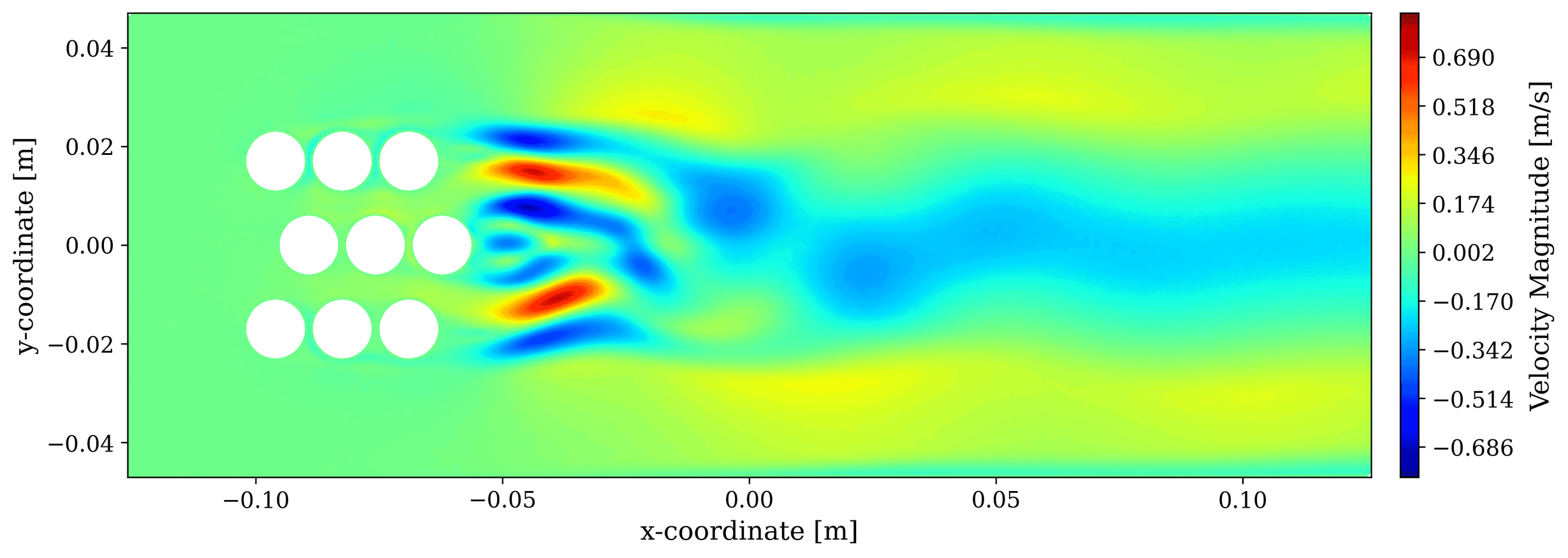} &
        \includegraphics[width=0.45\textwidth,valign=c]{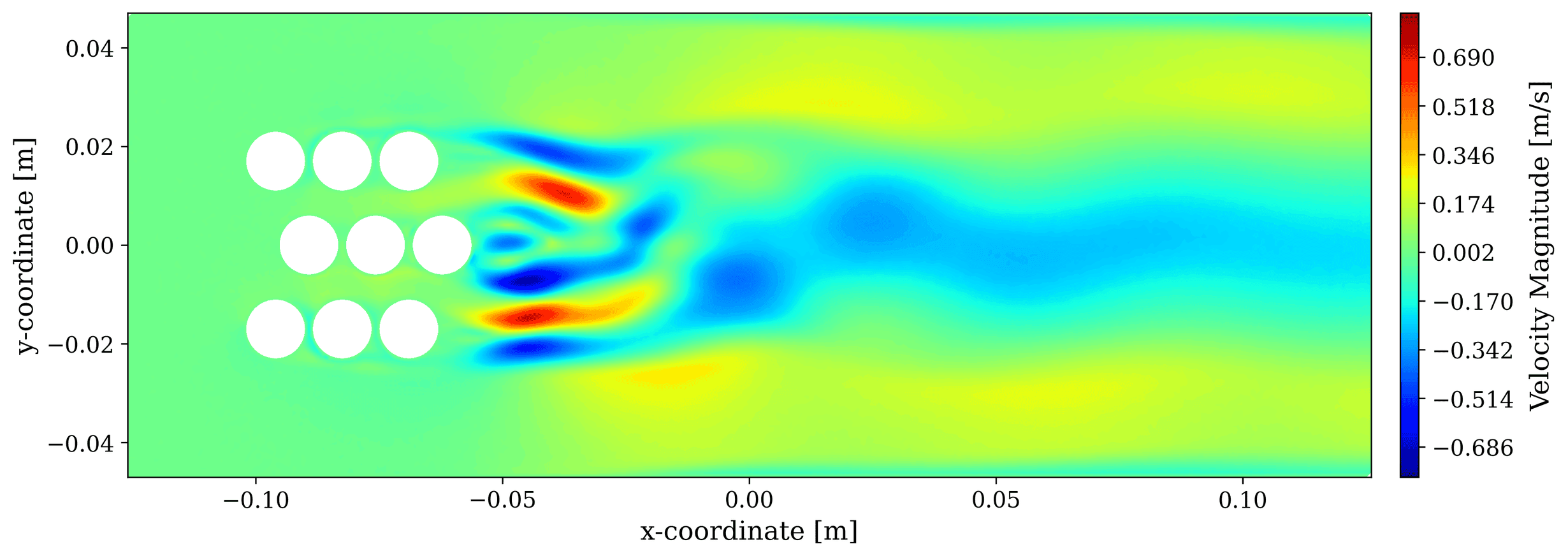} \\[2pt]
        \adjustbox{valign=c}{\rotatebox[origin=c]{90}{\small\textbf{Error}}} &
        \includegraphics[width=0.45\textwidth,valign=c]{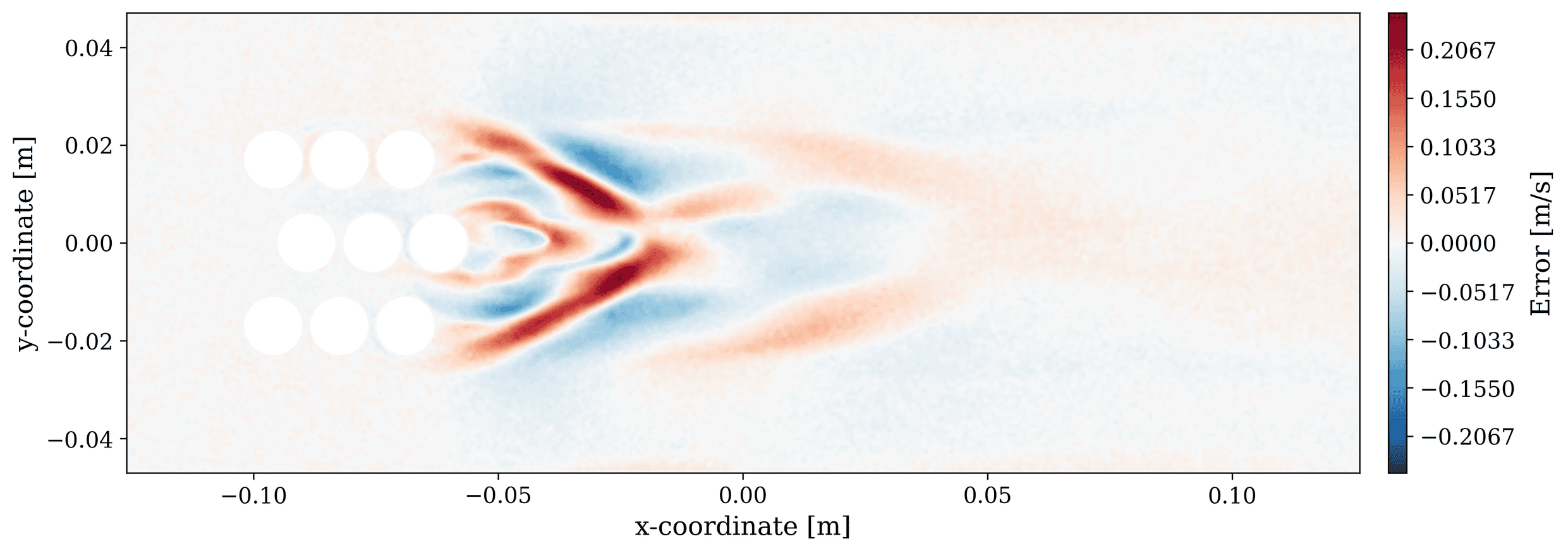} &
        \includegraphics[width=0.45\textwidth,valign=c]{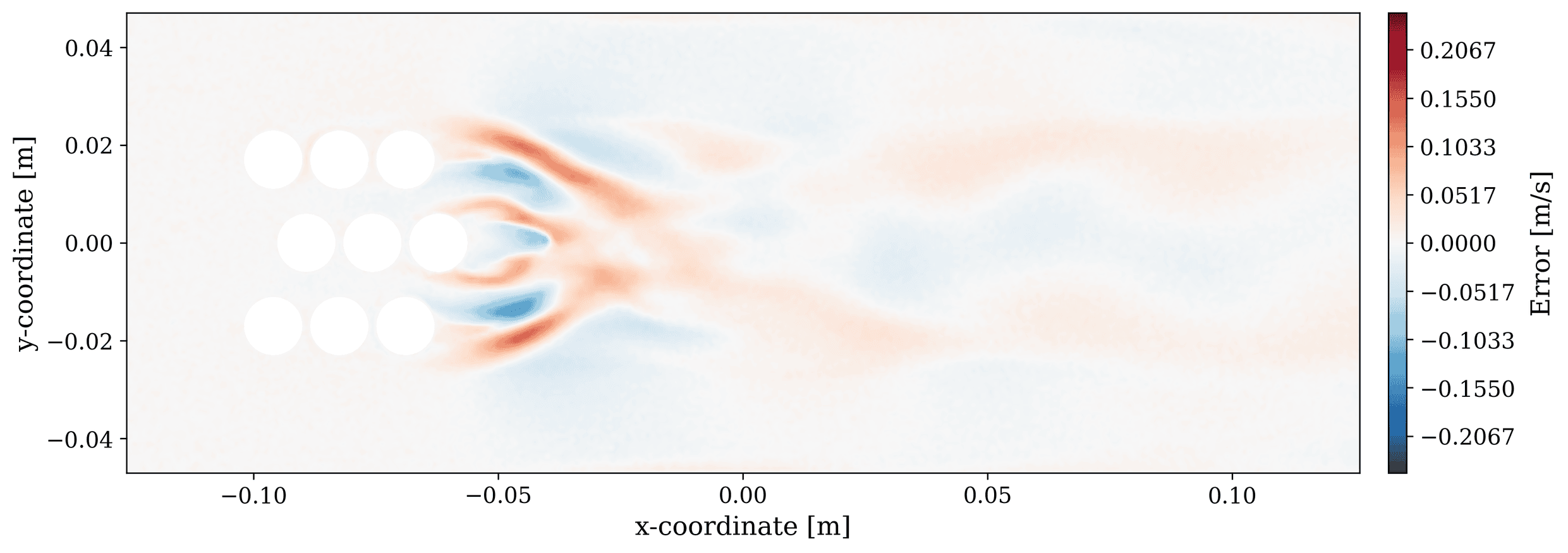} &
        \includegraphics[width=0.45\textwidth,valign=c]{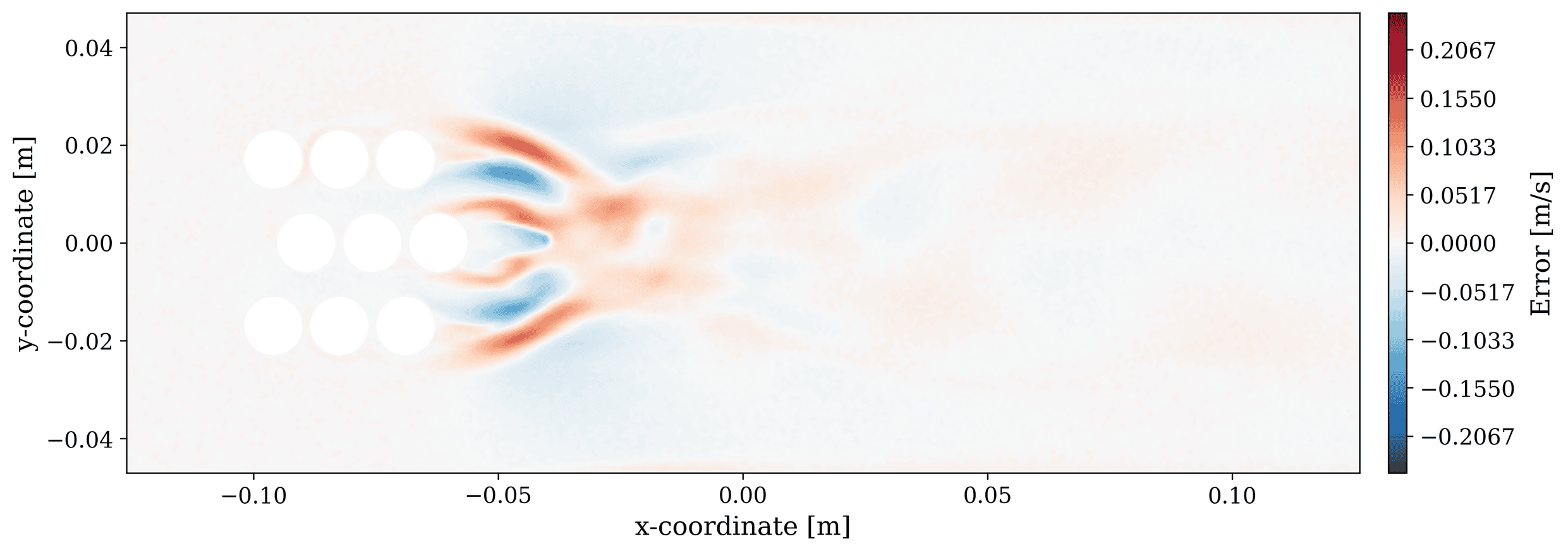} \\
    \end{tabular}
    }%
    \caption{MLP-based AE results for velocity field with inlet velocity 0.7 $m/s$. Reference, restored, and error at different timesteps.}
    \label{fig:ae_velocity_inlet070}
\end{figure}

Figure~\ref{fig:ae_velocity_inlet040} and Figure~\ref{fig:ae_velocity_inlet070} present the reconstruction results of the MLP-based AE for inlet velocities of 0.4 m/s and 0.7 m/s, respectively. Each figure shows the reference CFD solution (top row), the reconstructed field (middle row), and the error (bottom row) at three representative time steps ($t = 2$, $50$, and $100$). For the 0.4 m/s case, the MLP-based AE accurately reconstructs the overall flow structure, including the progressive development of the K\'{a}rm\'{a}n vortex street in the wake region downstream of the cylinder bundle. At the early time step ($t = 2$), the flow field remains relatively uniform and the reconstruction error is minimal. As the flow develops toward the fully periodic vortex streets regime at $t = 50$ and $t = 100$, the reconstructed fields faithfully preserve the spatial patterns of the alternating vortices, with errors concentrated primarily near the cylinder surfaces and the wake region where velocity gradients are steepest. For the 0.7 m/s case shown in Figure~\ref{fig:ae_velocity_inlet070}, despite the higher inlet velocity producing more intense vortex interactions and stronger velocity fluctuations, the MLP-based AE maintains a comparable level of reconstruction accuracy. The vortex structures are clearly captured, and the error magnitudes remain at a similar level to the 0.4 m/s case, confirming that the MLP-based AE generalizes well across different operating conditions.

\begin{figure}[H]
    \centering
    \setlength{\tabcolsep}{1pt}
    \makebox[\textwidth][c]{%
    \begin{tabular}{c@{\hspace{4pt}}ccc}
        & \textbf{$t = 2$} & \textbf{$t = 50$} & \textbf{$t = 100$} \\
        \adjustbox{valign=c}{\rotatebox[origin=c]{90}{\small\textbf{Reference}}} &
        \includegraphics[width=0.45\textwidth,valign=c]{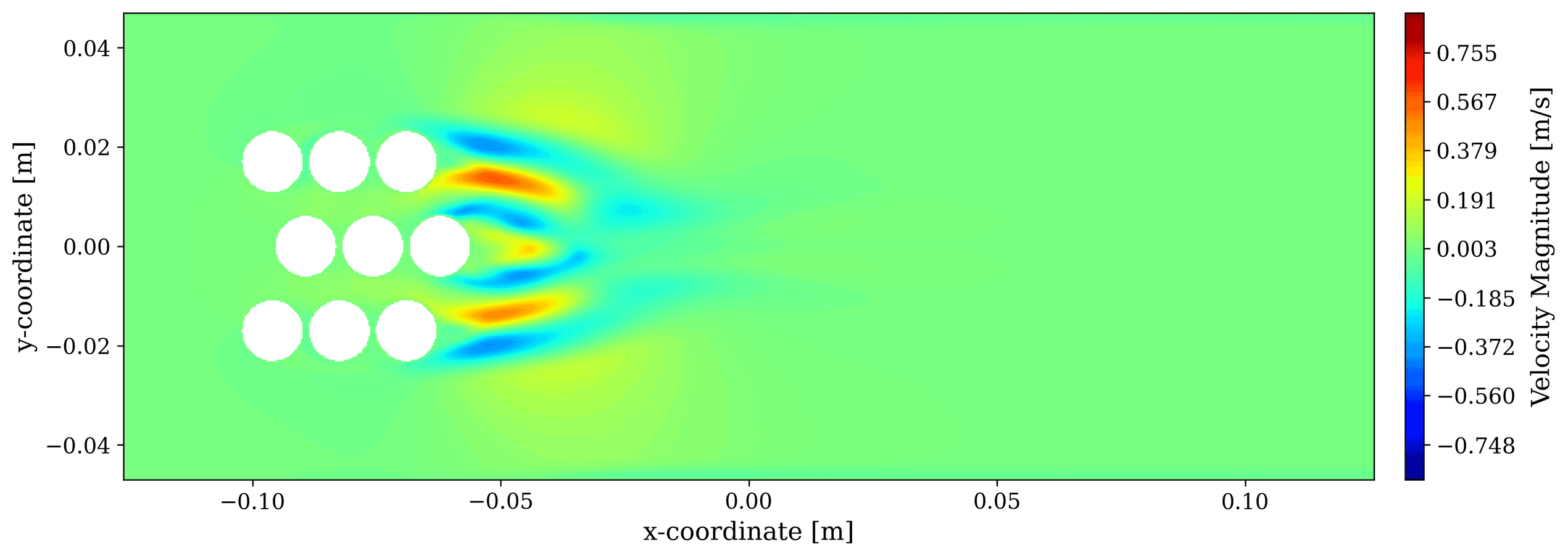} &
        \includegraphics[width=0.45\textwidth,valign=c]{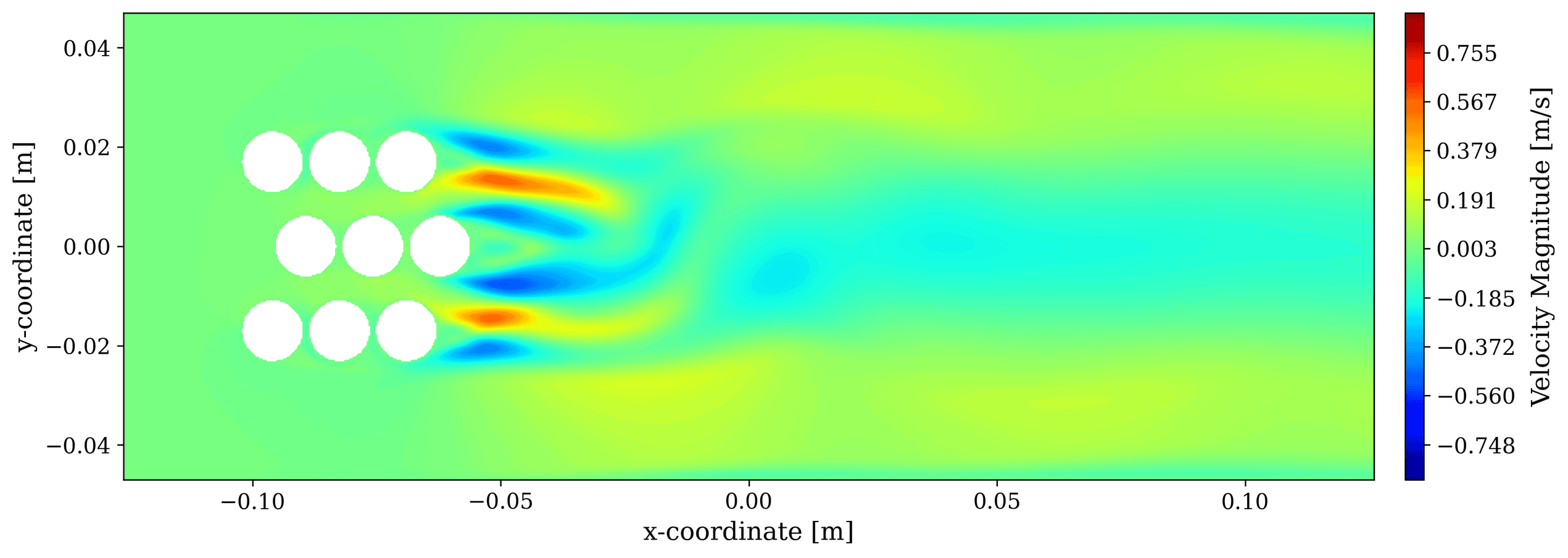} &
        \includegraphics[width=0.45\textwidth,valign=c]{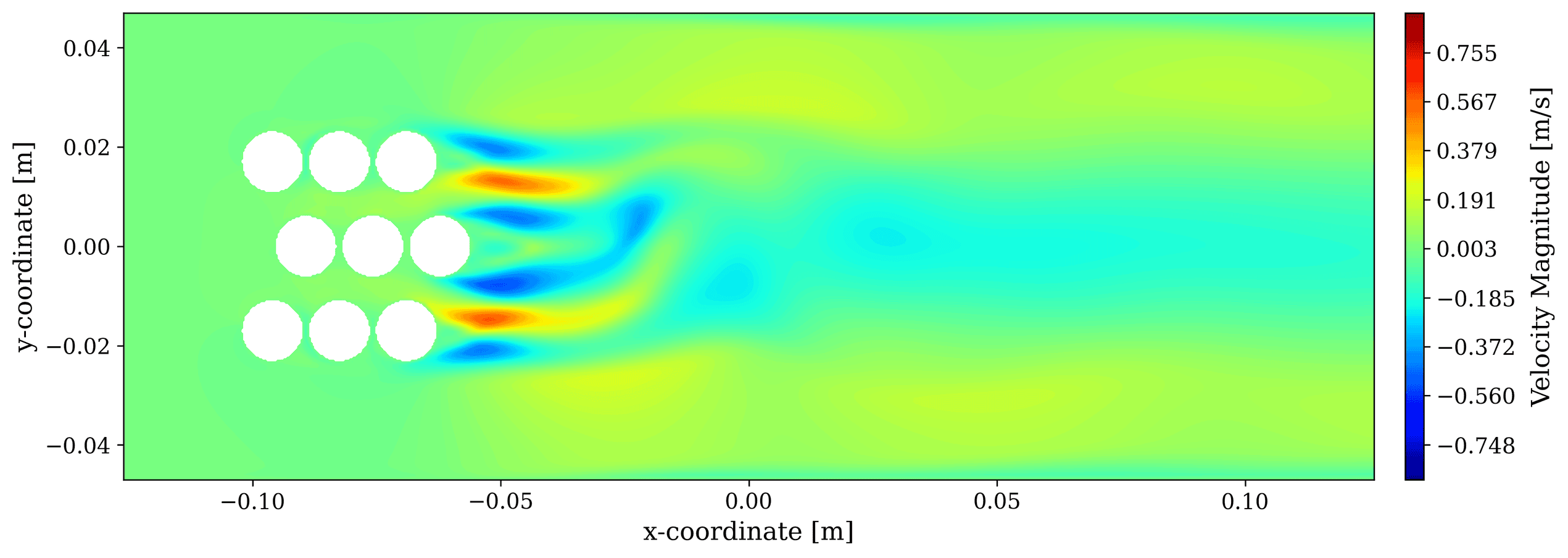} \\[2pt]
        \adjustbox{valign=c}{\rotatebox[origin=c]{90}{\small\textbf{Restored}}} &
        \includegraphics[width=0.45\textwidth,valign=c]{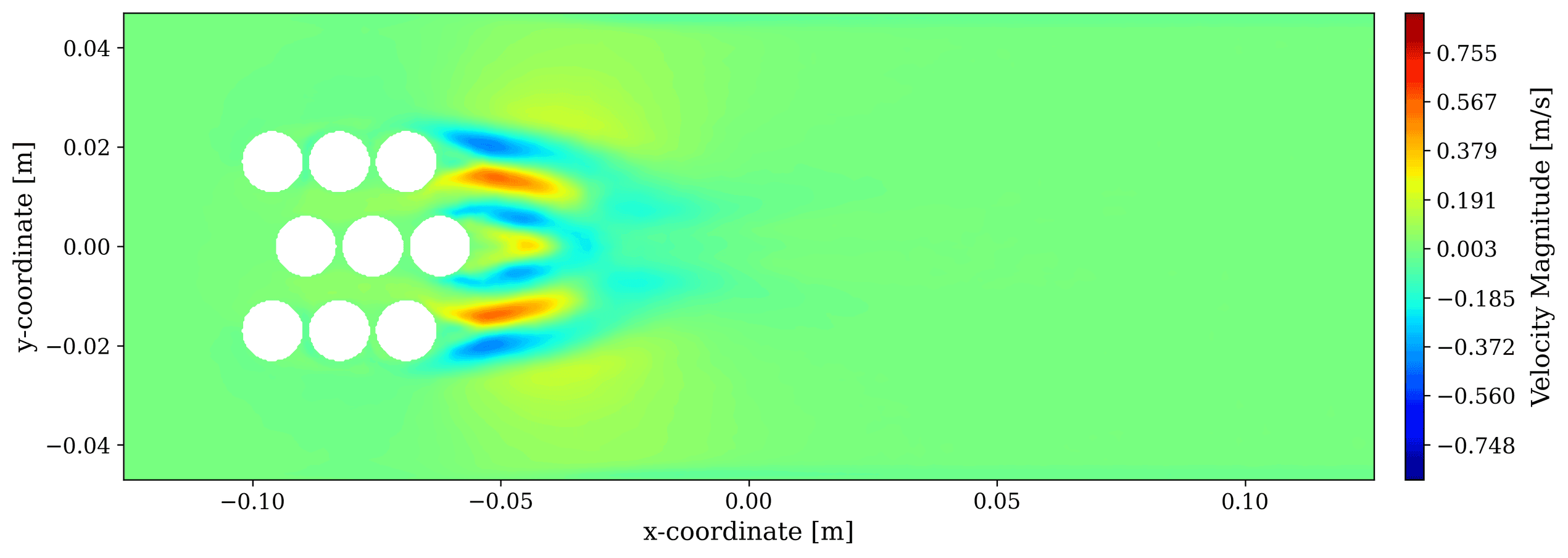} &
        \includegraphics[width=0.45\textwidth,valign=c]{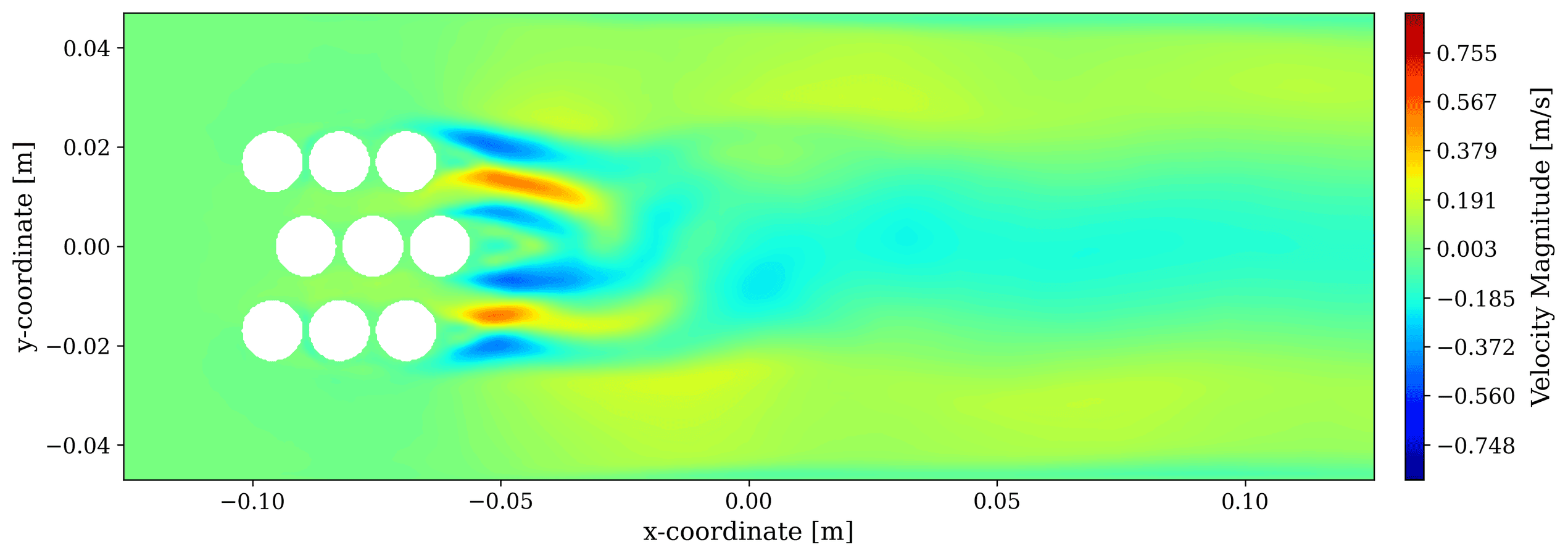} &
        \includegraphics[width=0.45\textwidth,valign=c]{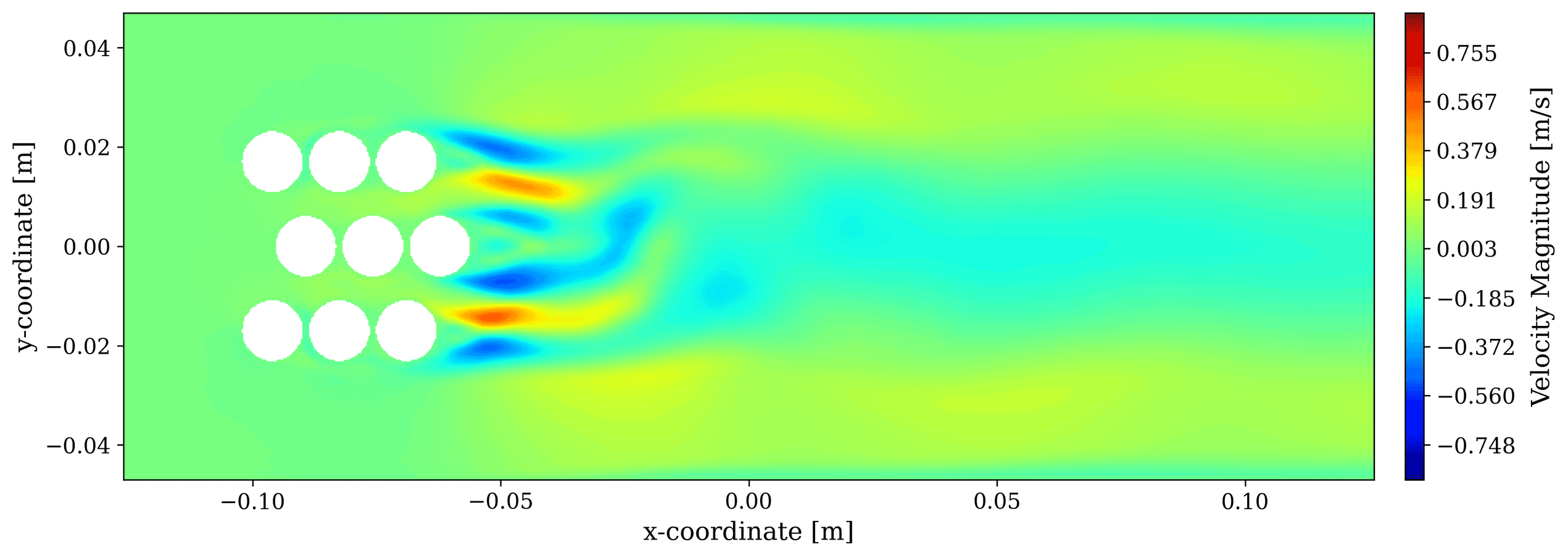} \\[2pt]
        \adjustbox{valign=c}{\rotatebox[origin=c]{90}{\small\textbf{Error}}} &
        \includegraphics[width=0.45\textwidth,valign=c]{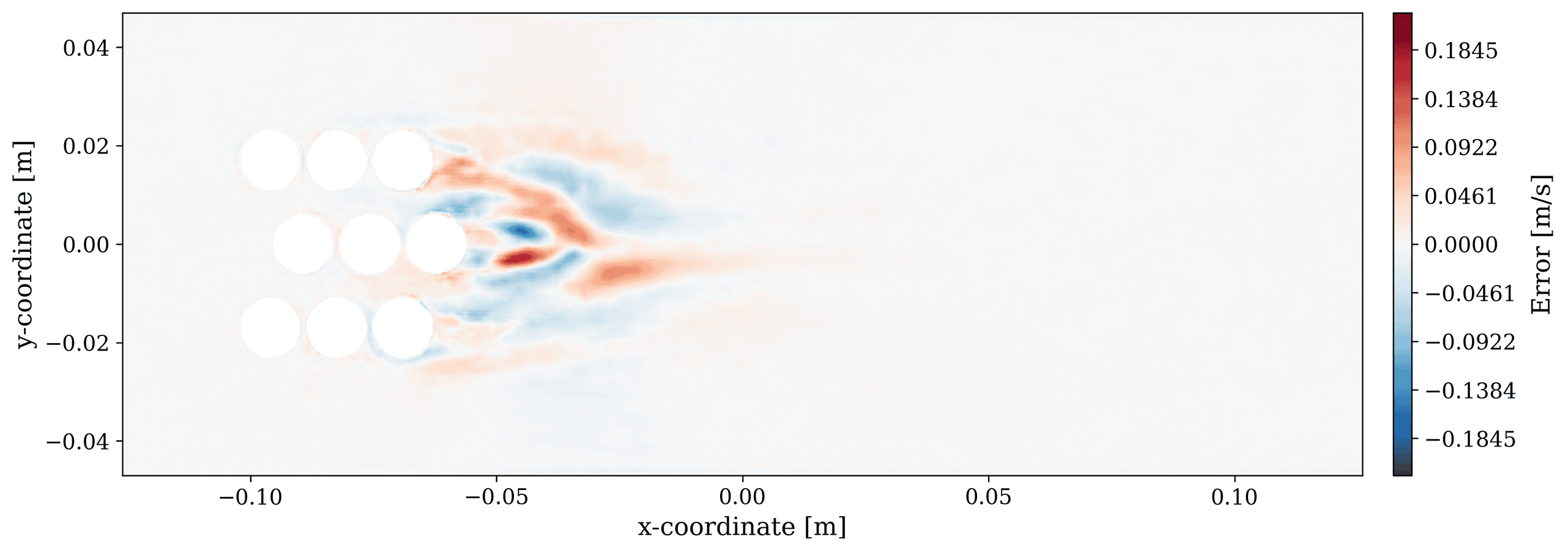} &
        \includegraphics[width=0.45\textwidth,valign=c]{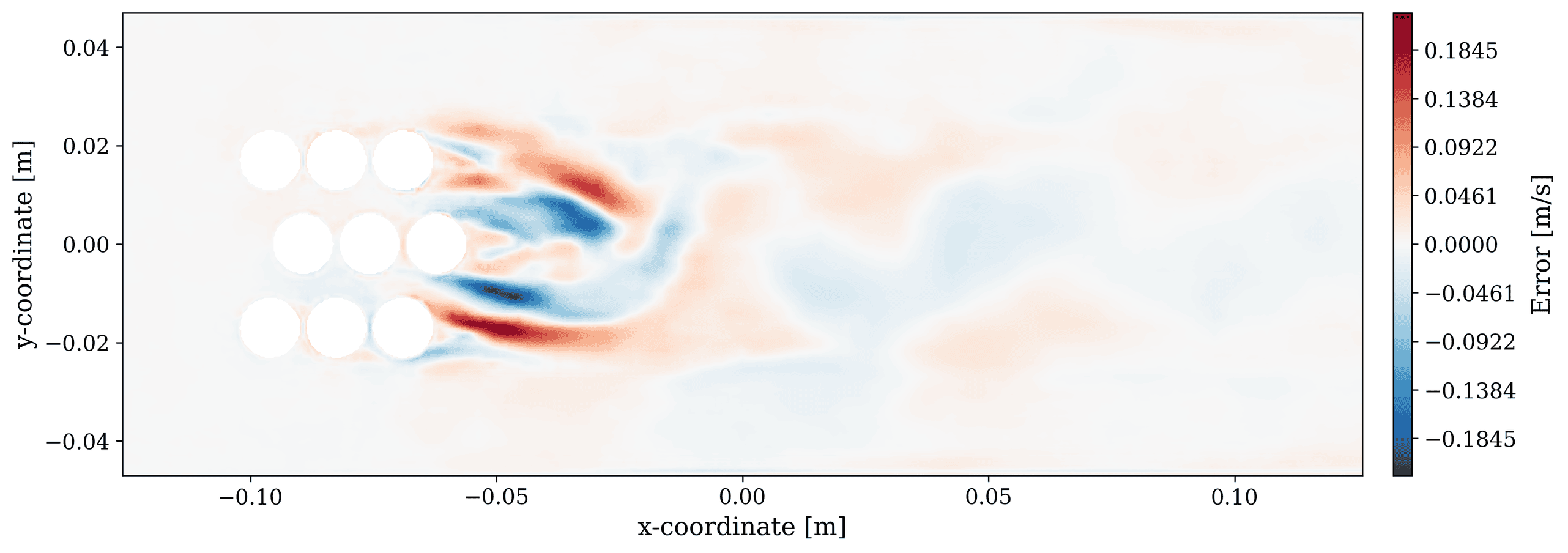} &
        \includegraphics[width=0.45\textwidth,valign=c]{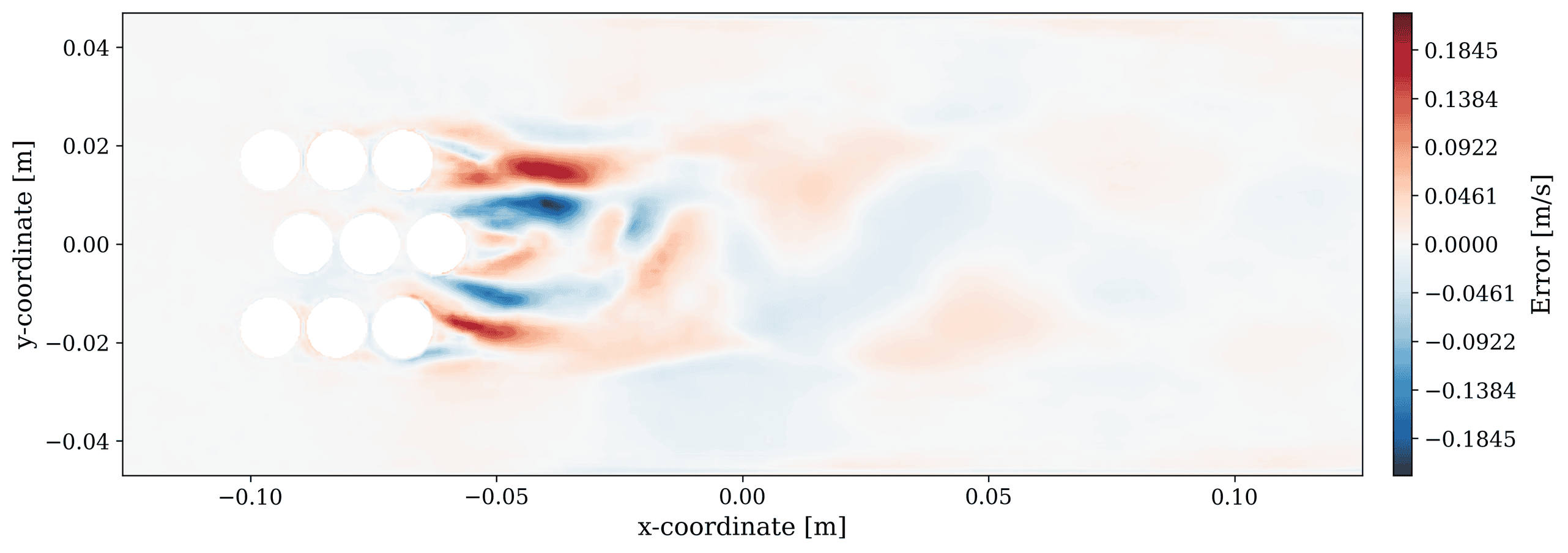} \\
    \end{tabular}
    }%
    \caption{CAE results for velocity field with inlet velocity 0.4 $m/s$. Reference, restored, and error at different timesteps.}
    \label{fig:cae_velocity_inlet040}
\end{figure}

\begin{figure}[H]
    \centering
    \setlength{\tabcolsep}{1pt}
    \makebox[\textwidth][c]{%
    \begin{tabular}{c@{\hspace{4pt}}ccc}
        & \textbf{$t = 2$} & \textbf{$t = 50$} & \textbf{$t = 100$} \\
        \adjustbox{valign=c}{\rotatebox[origin=c]{90}{\small\textbf{Reference}}} &
        \includegraphics[width=0.45\textwidth,valign=c]{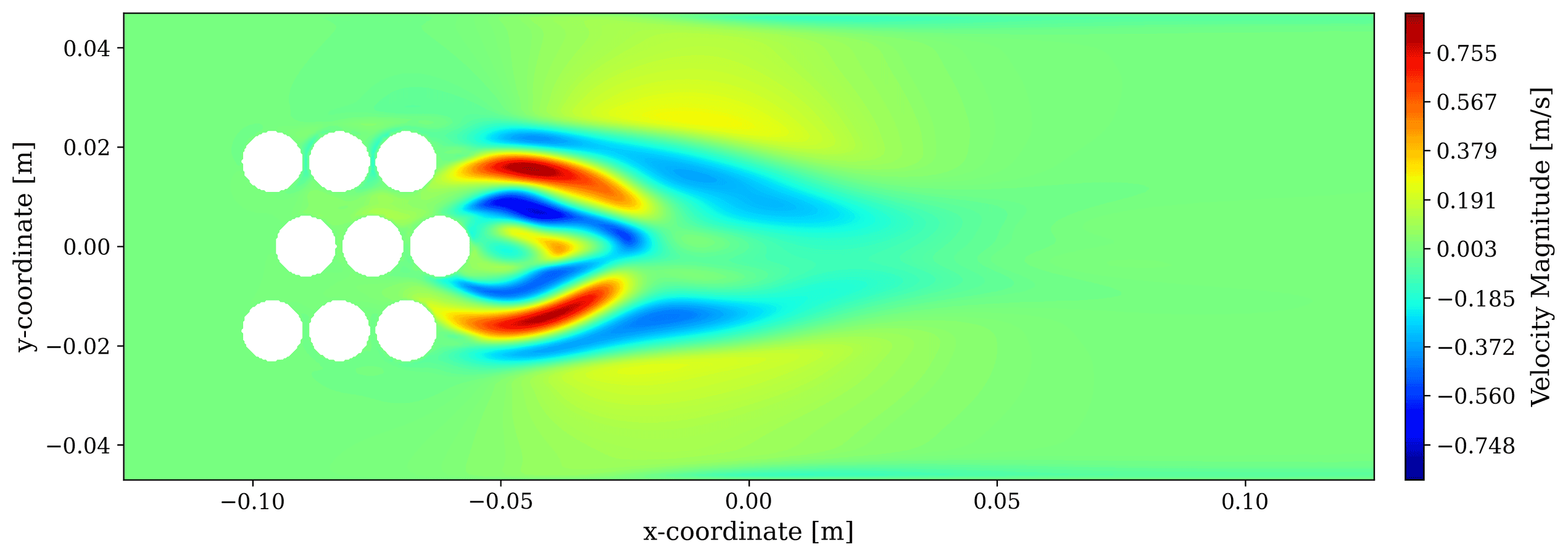} &
        \includegraphics[width=0.45\textwidth,valign=c]{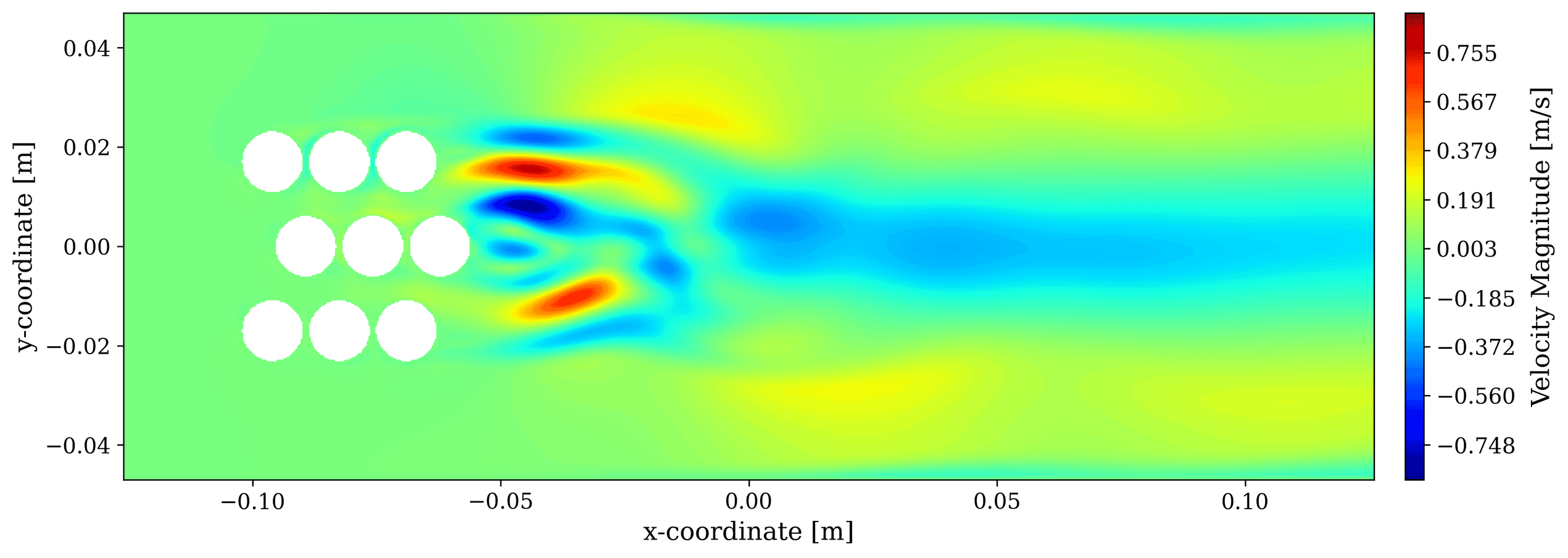} &
        \includegraphics[width=0.45\textwidth,valign=c]{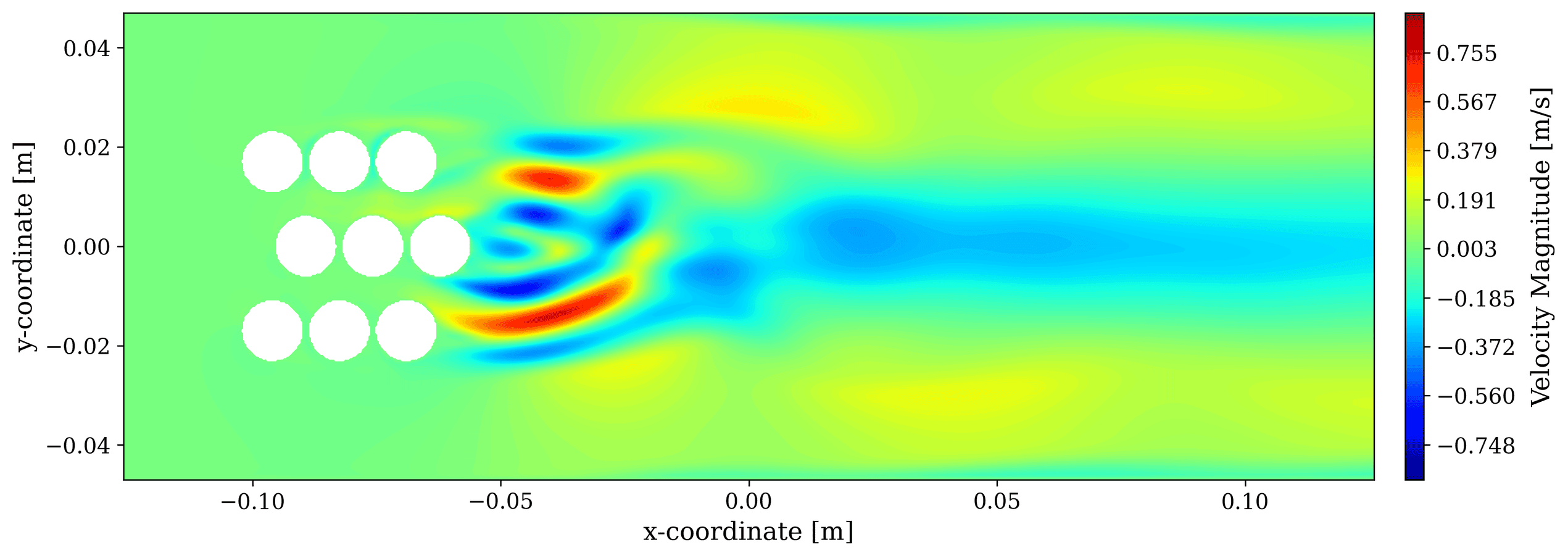} \\[2pt]
        \adjustbox{valign=c}{\rotatebox[origin=c]{90}{\small\textbf{Restored}}} &
        \includegraphics[width=0.45\textwidth,valign=c]{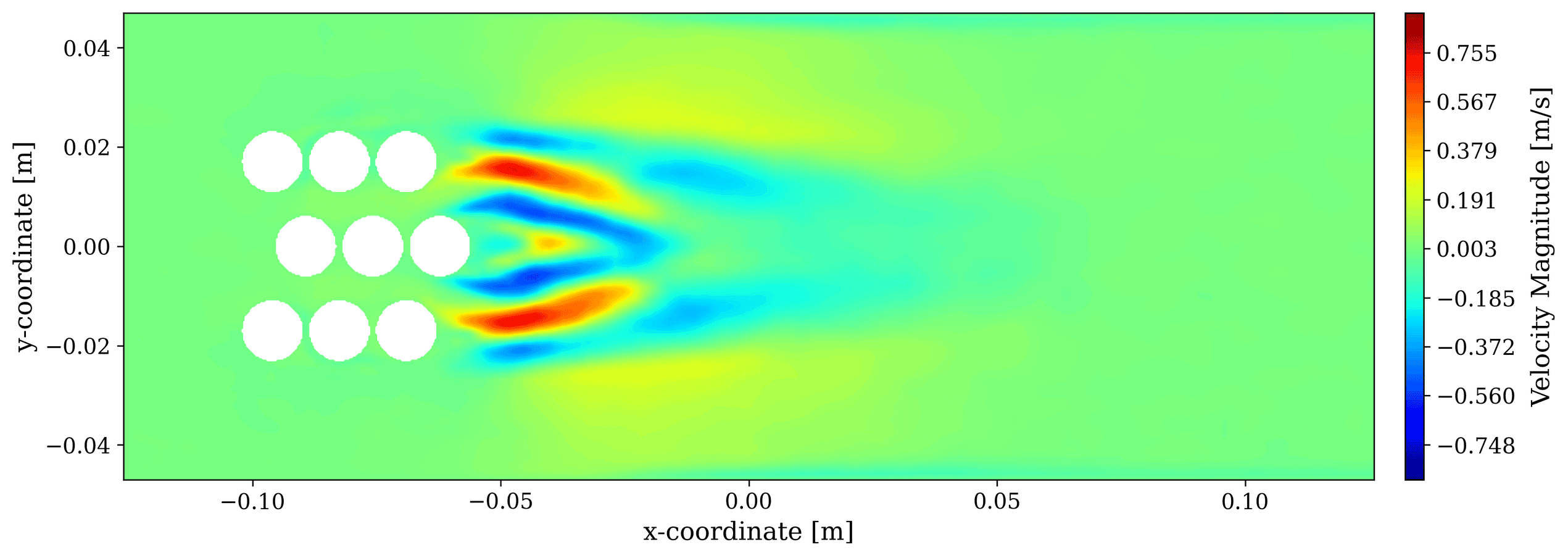} &
        \includegraphics[width=0.45\textwidth,valign=c]{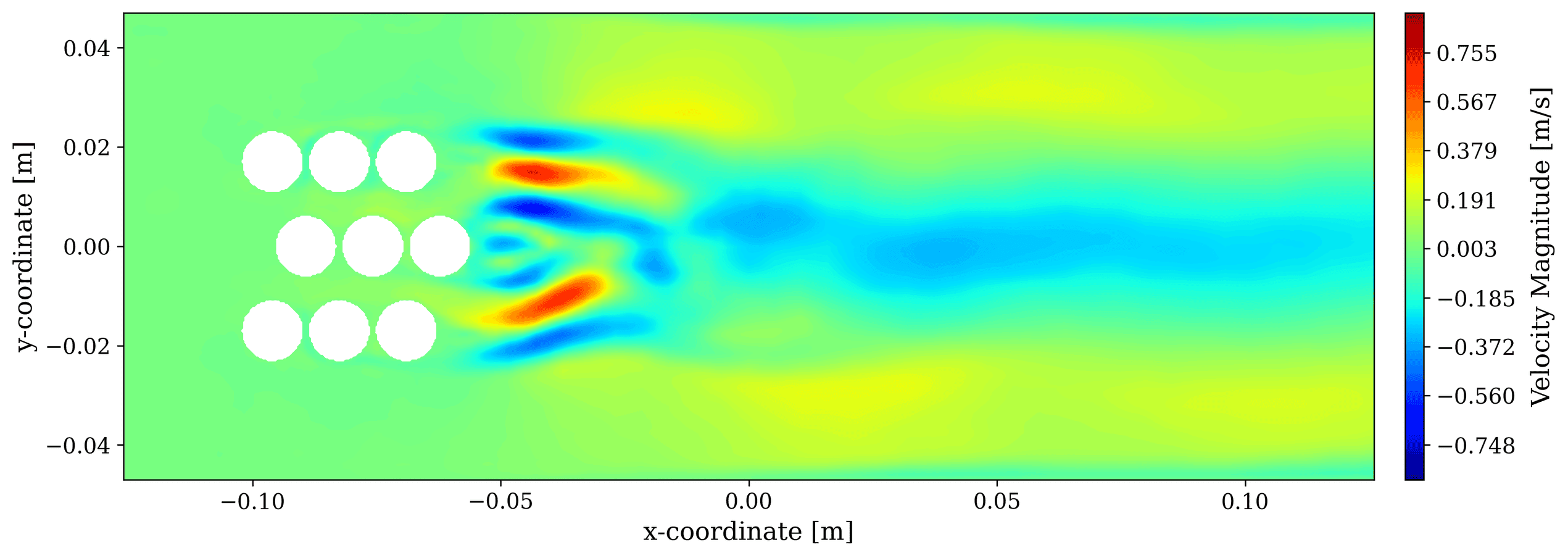} &
        \includegraphics[width=0.45\textwidth,valign=c]{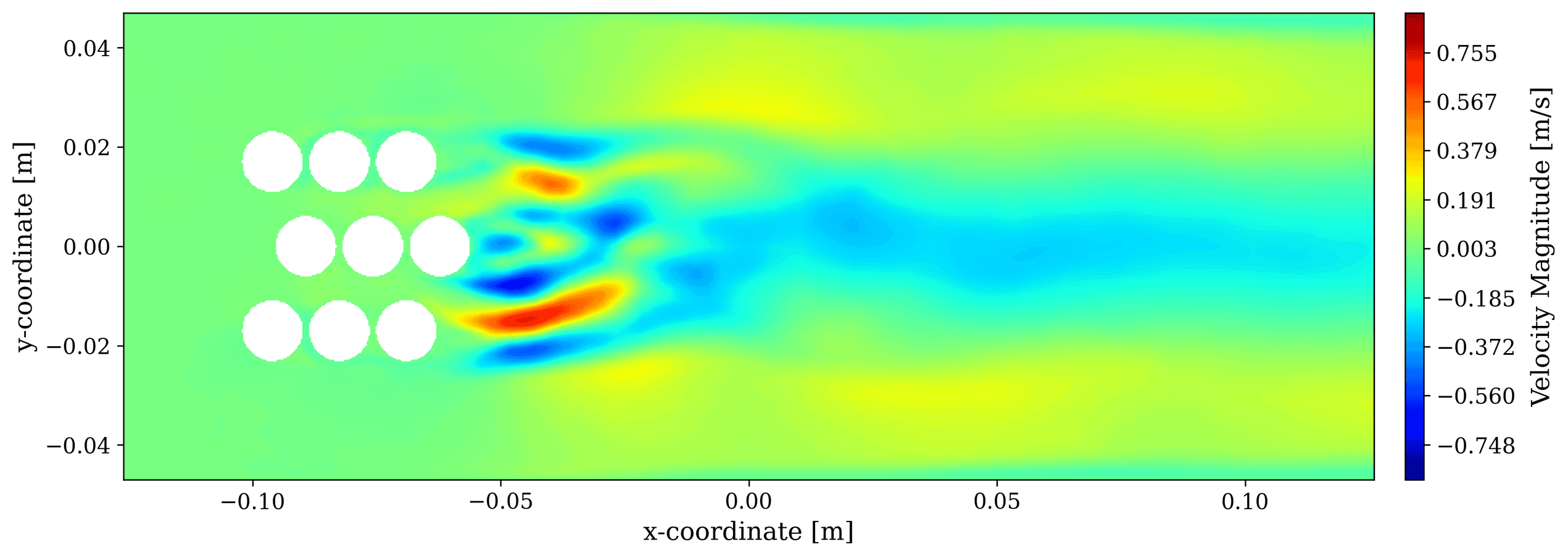} \\[2pt]
        \adjustbox{valign=c}{\rotatebox[origin=c]{90}{\small\textbf{Error}}} &
        \includegraphics[width=0.45\textwidth,valign=c]{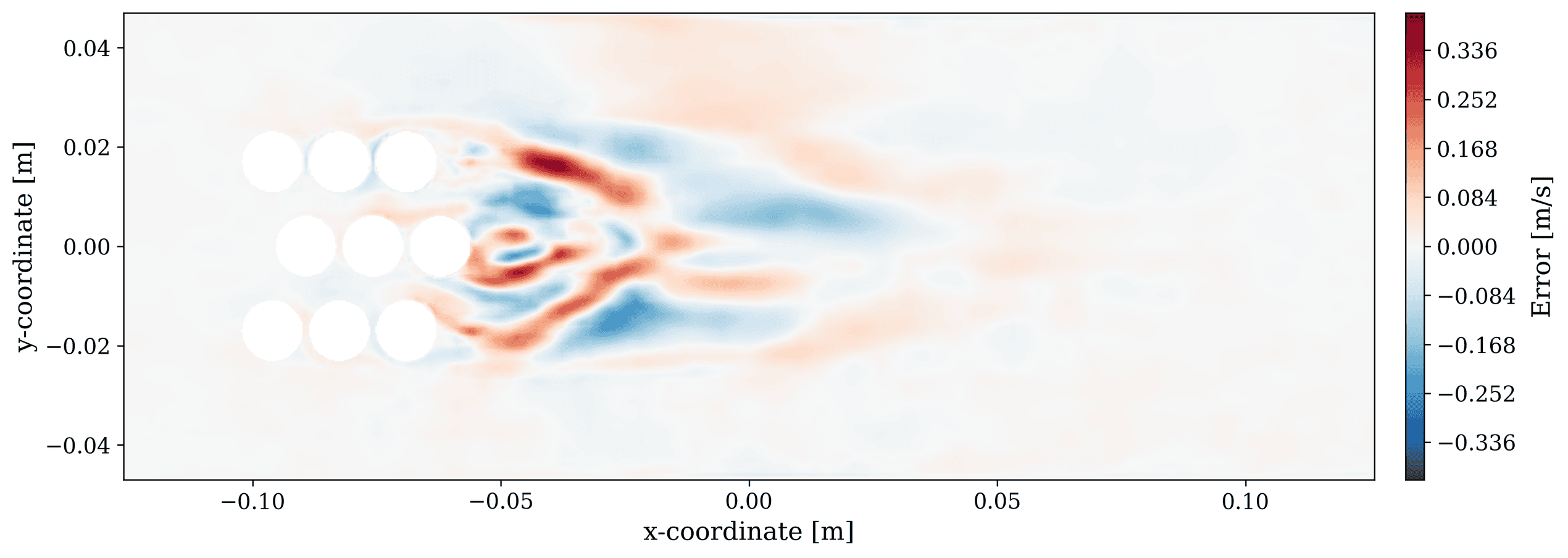} &
        \includegraphics[width=0.45\textwidth,valign=c]{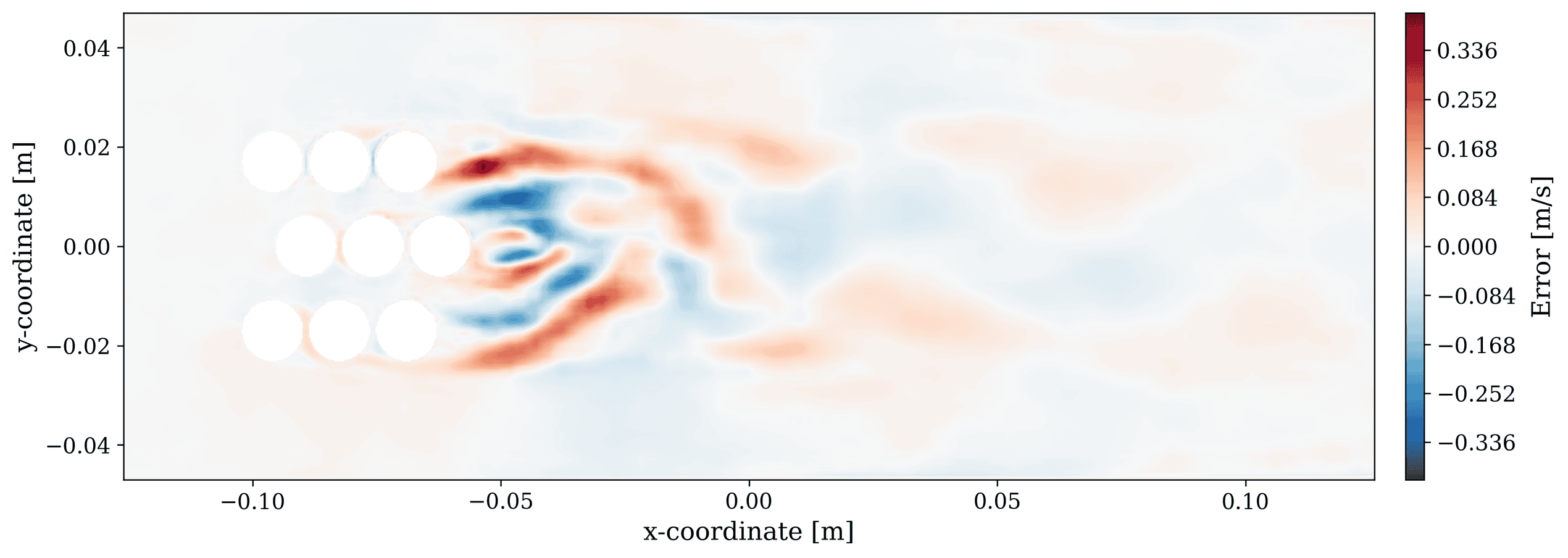} &
        \includegraphics[width=0.45\textwidth,valign=c]{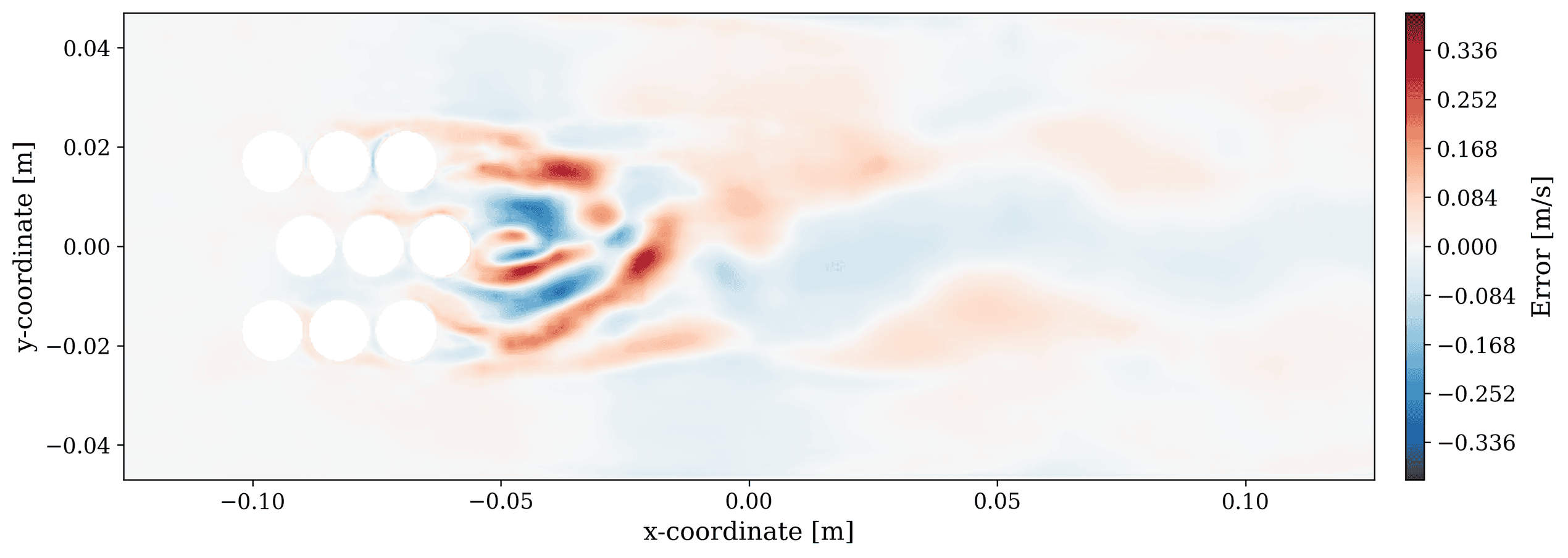} \\
    \end{tabular}
    }%
    \caption{CAE results for velocity field with inlet velocity 0.7 $m/s$. Reference, restored, and error at different timesteps.}
    \label{fig:cae_velocity_inlet070}
\end{figure}

Figure~\ref{fig:cae_velocity_inlet040} and Figure~\ref{fig:cae_velocity_inlet070} show the corresponding reconstruction results of the CAE for inlet velocities of 0.4 m/s and 0.7 m/s, respectively. The CAE also demonstrates accurate reconstruction of the velocity difference fields, effectively capturing the spatial structure of the vortex patterns at all time steps. The error distributions for the CAE exhibit similar patterns to those of the MLP-based AE, with the largest deviations occurring in the vicinity of the cylinders.

Overall, both the MLP-based AE and CAE demonstrate sufficient reconstruction accuracy for the temporal difference fields across the tested inlet velocity conditions. These results confirm that the essential flow features, including the periodic K\'{a}rm\'{a}n vortex structures, are well preserved through the encoding and decoding process, providing a reliable results for the subsequent training of L-DeepONet in the latent space.

\subsection{Velocity Field Prediction}
\label{subsec6.2}

The velocity field prediction performance of four neural operator models was evaluated: (1) the multi-scale L-DeepONet with MLP-based AE, (2) the multi-scale L-DeepONet with CAE, (3) the standard FNO, and (4) the MscaleFNO. All models were tested on two unseen inlet velocity conditions of 0.4 m/s and 0.7 m/s. For each model, the reference CFD solution, the predicted velocity field, and the error are presented at three representative time steps ($t = 2$, $50$, and $100$). The corresponding pressure field predictions are provided in Appendix~\ref{app:pressure_field}.

\begin{figure}[H]
    \centering
    \setlength{\tabcolsep}{1pt}
    \makebox[\textwidth][c]{%
    \begin{tabular}{c@{\hspace{4pt}}ccc}
        & \textbf{$t = 2$} & \textbf{$t = 50$} & \textbf{$t = 100$} \\
        \adjustbox{valign=c}{\rotatebox[origin=c]{90}{\small\textbf{Reference}}} &
        \includegraphics[width=0.45\textwidth,valign=c]{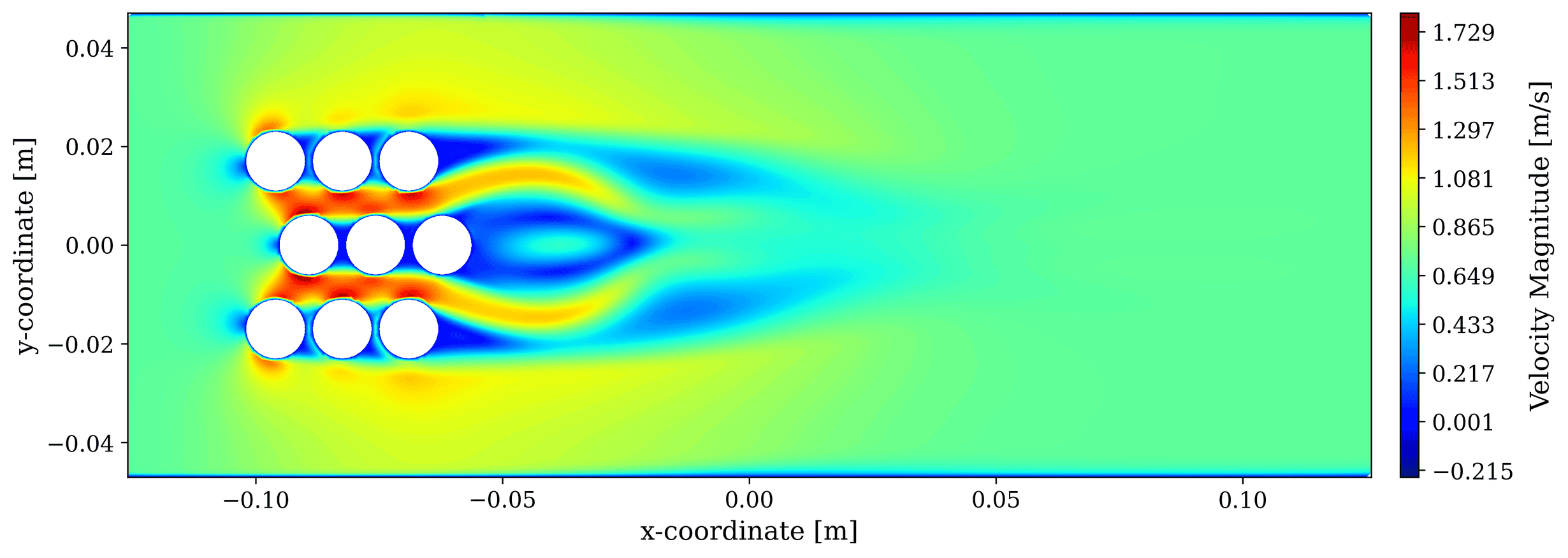} &
        \includegraphics[width=0.45\textwidth,valign=c]{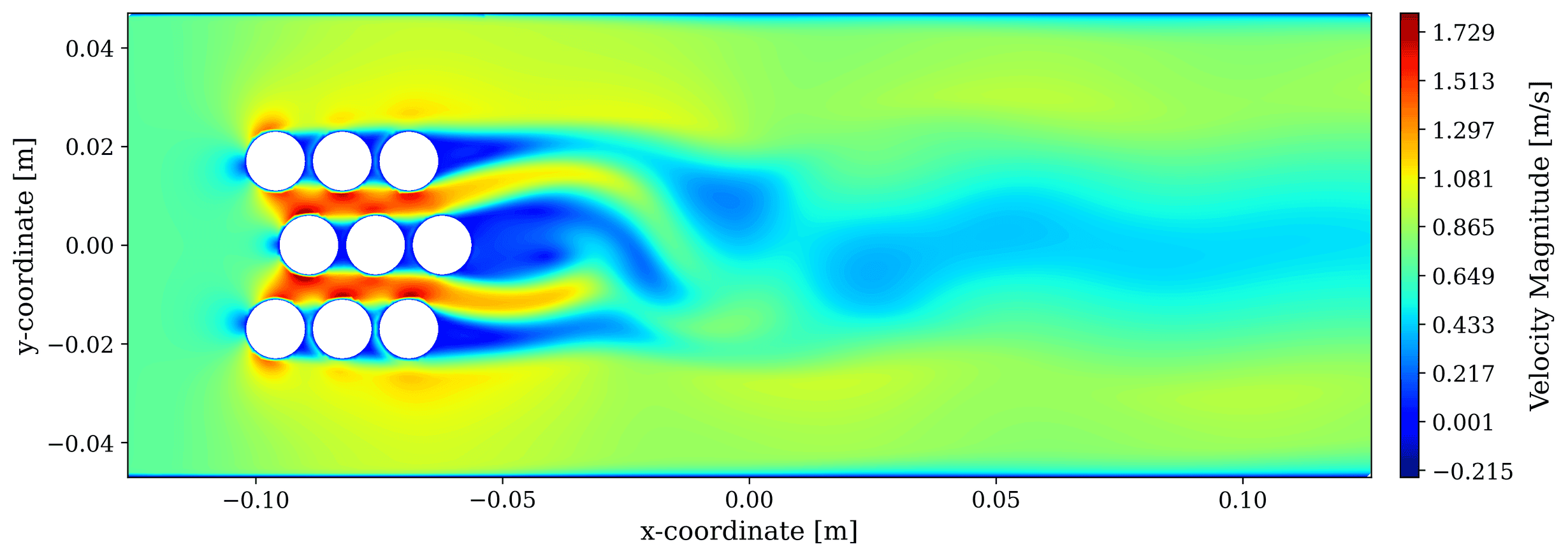} &
        \includegraphics[width=0.45\textwidth,valign=c]{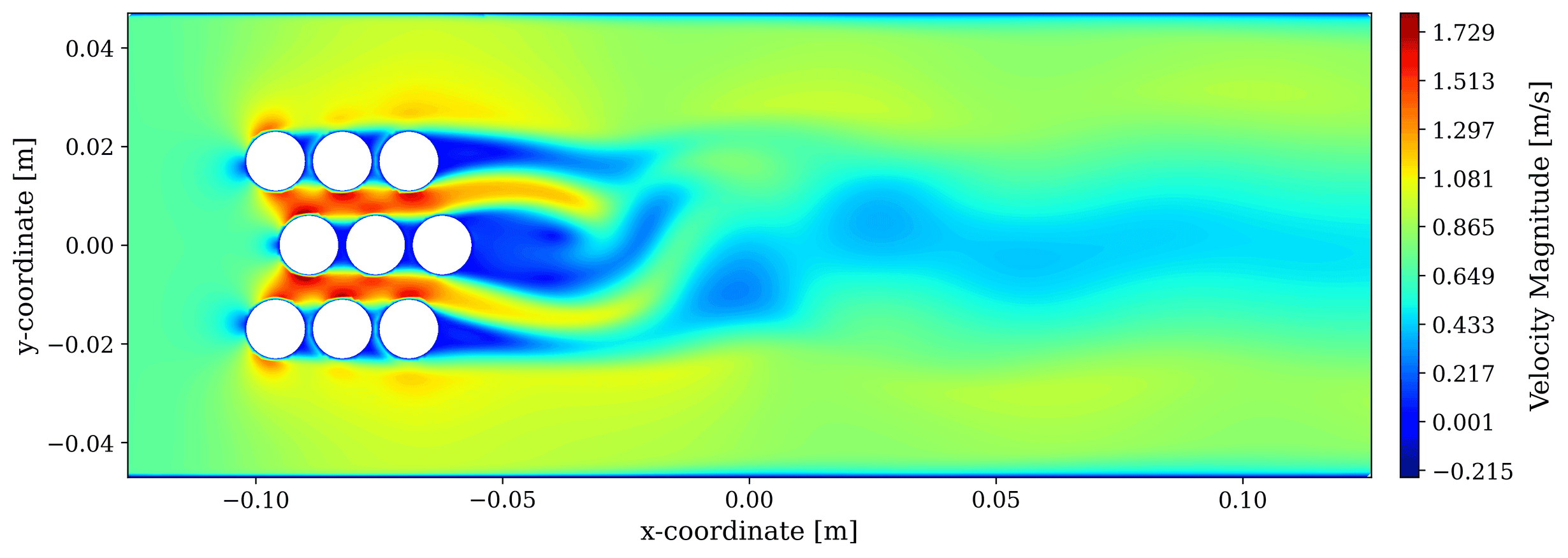} \\[2pt]
        \adjustbox{valign=c}{\rotatebox[origin=c]{90}{\small\textbf{Predicted}}} &
        \includegraphics[width=0.45\textwidth,valign=c]{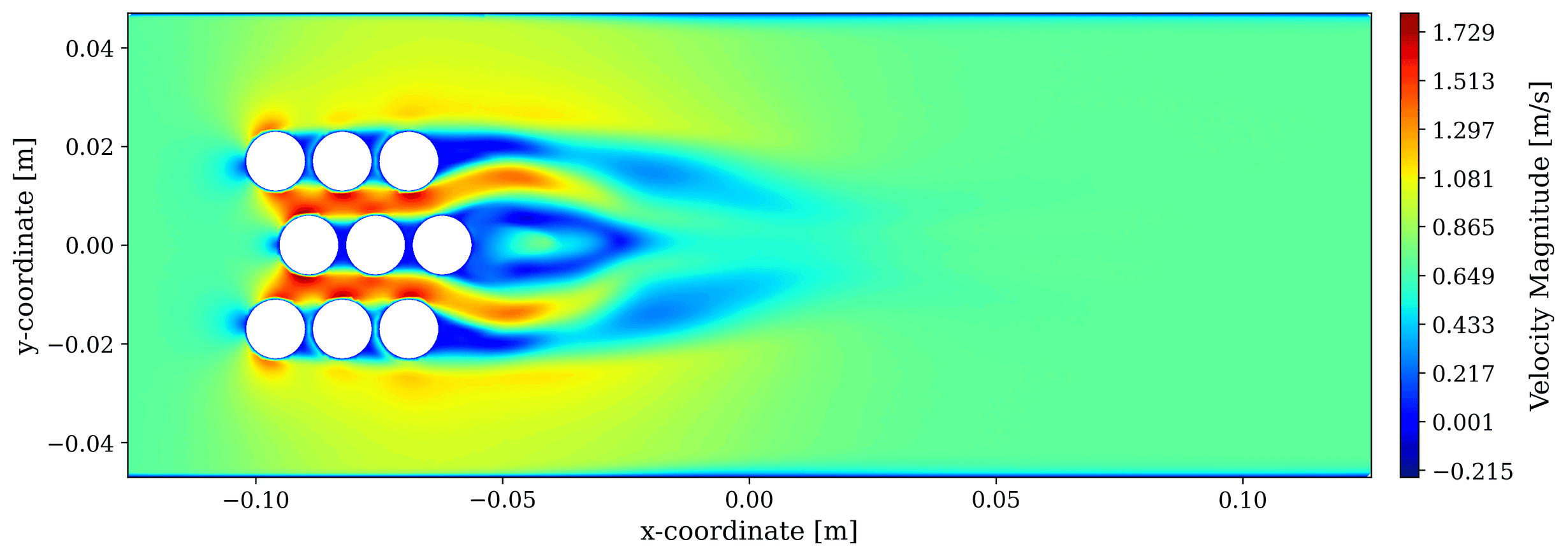} &
        \includegraphics[width=0.45\textwidth,valign=c]{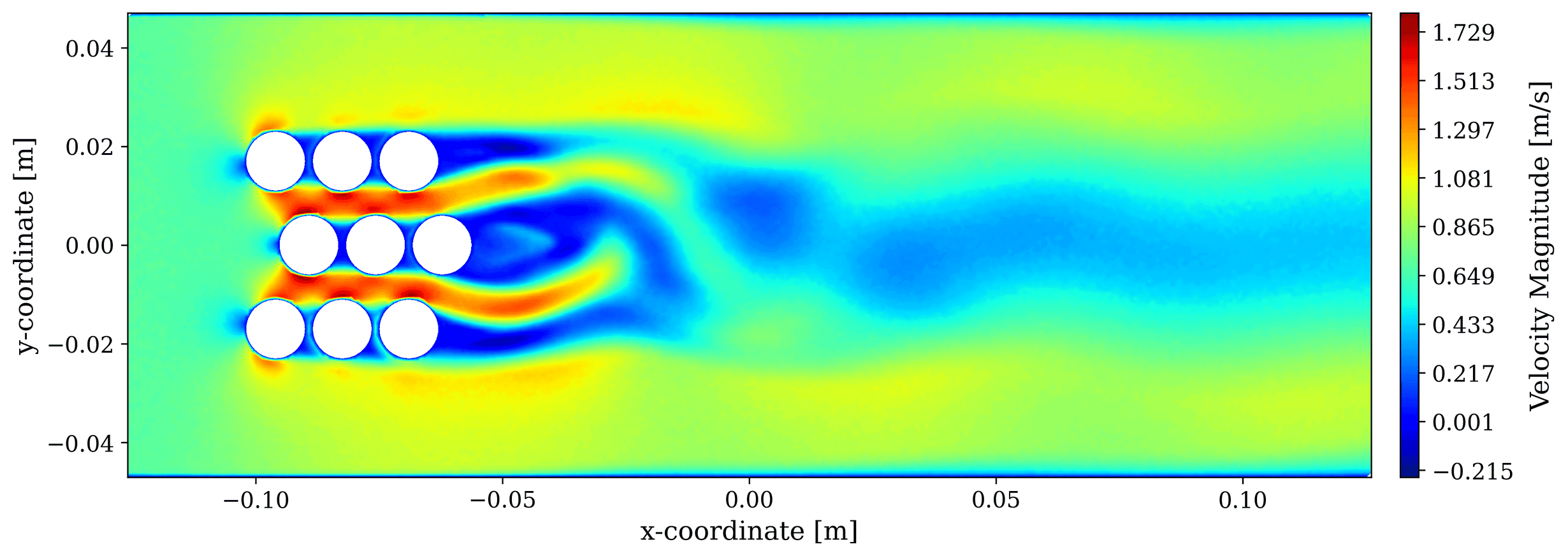} &
        \includegraphics[width=0.45\textwidth,valign=c]{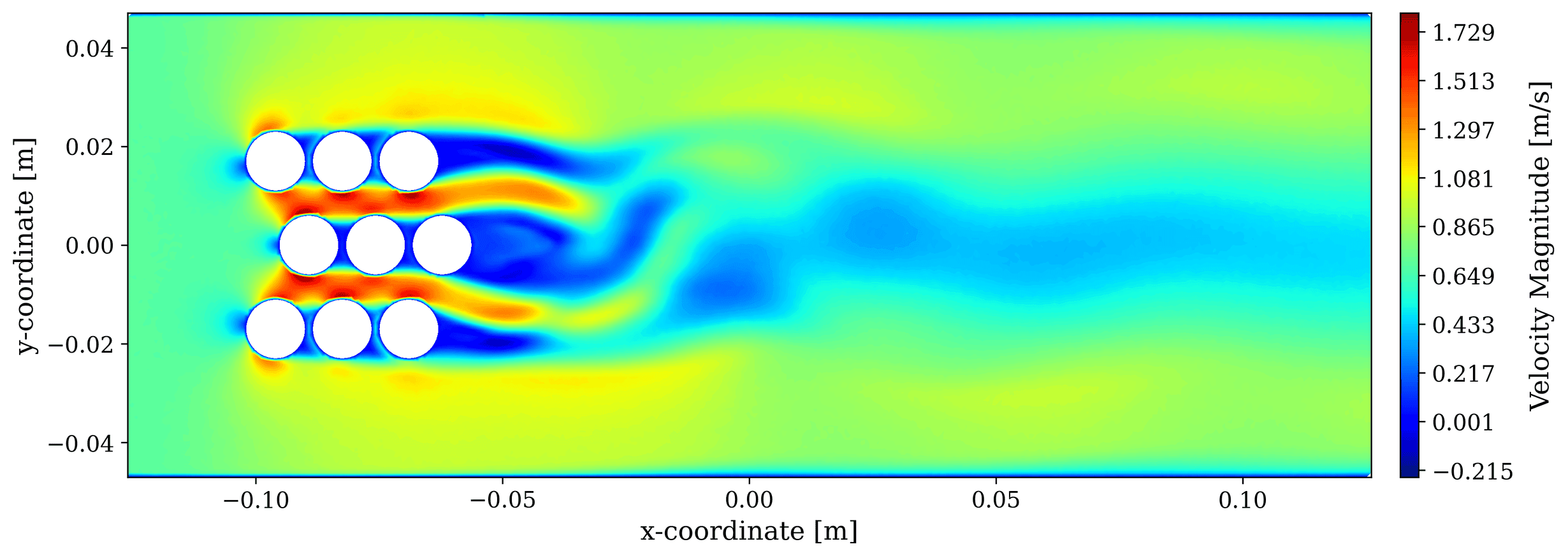} \\[2pt]
        \adjustbox{valign=c}{\rotatebox[origin=c]{90}{\small\textbf{Error}}} &
        \includegraphics[width=0.45\textwidth,valign=c]{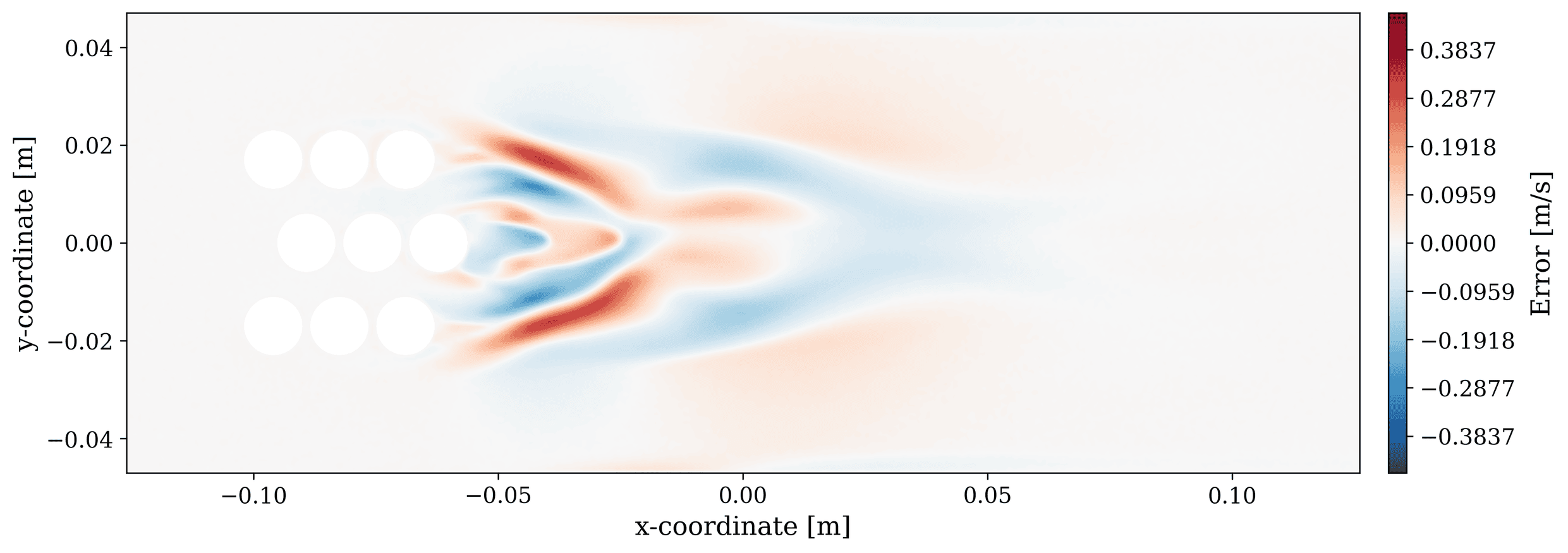} &
        \includegraphics[width=0.45\textwidth,valign=c]{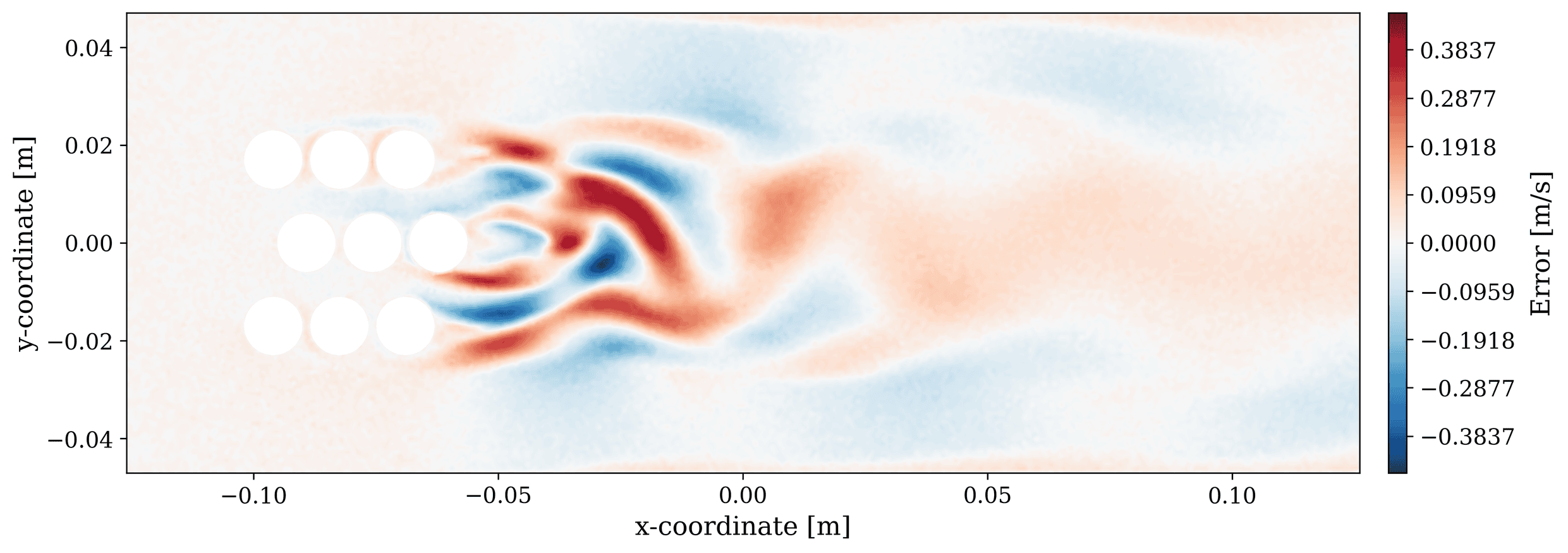} &
        \includegraphics[width=0.45\textwidth,valign=c]{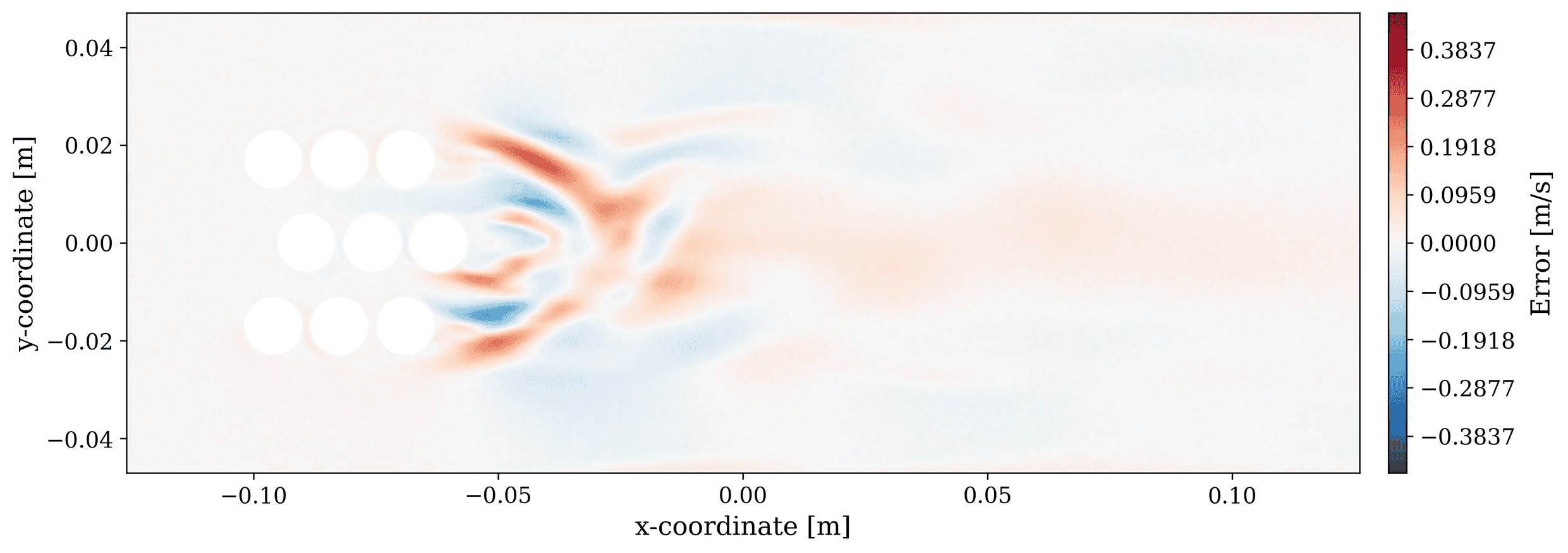} \\
    \end{tabular}
    }%
    \caption{Predicted velocity field by the multi-scale L-DeepONet with MLP-based AE at inlet velocity 0.7 m/s. Top: reference CFD solution, middle: predicted field, bottom: absolute error at $t = 2$, $50$, and $100$.}
    \label{fig:ldon_unstructured_velocity_inlet070}
\end{figure}

Figure~\ref{fig:ldon_unstructured_velocity_inlet040} and Figure~\ref{fig:ldon_unstructured_velocity_inlet070} show the prediction results of the multi-scale L-DeepONet with MLP-based AE for inlet velocities of 0.4 m/s and 0.7 m/s, respectively. The model successfully captures the periodic K\'{a}rm\'{a}n vortex streets that develop in the wake region downstream of the cylinder bundle. At the early time step ($t = 2$), the predicted flow field closely matches the reference solution during the initial transient development. At later time steps ($t = 50$ and $t = 100$), where the flow exhibits fully developed periodic vortex streets, the model faithfully reproduces the alternating vortex patterns with accurate spatial positioning and intensity. The error fields remain small across all time steps, confirming that the multi-scale technique effectively mitigates the spectral bias identified in Section~\ref{subsec5.1}.

\begin{figure}[H]
    \centering
    \setlength{\tabcolsep}{1pt}
    \makebox[\textwidth][c]{%
    \begin{tabular}{c@{\hspace{4pt}}ccc}
        & \textbf{$t = 2$} & \textbf{$t = 50$} & \textbf{$t = 100$} \\
        \adjustbox{valign=c}{\rotatebox[origin=c]{90}{\small\textbf{Reference}}} &
        \includegraphics[width=0.45\textwidth,valign=c]{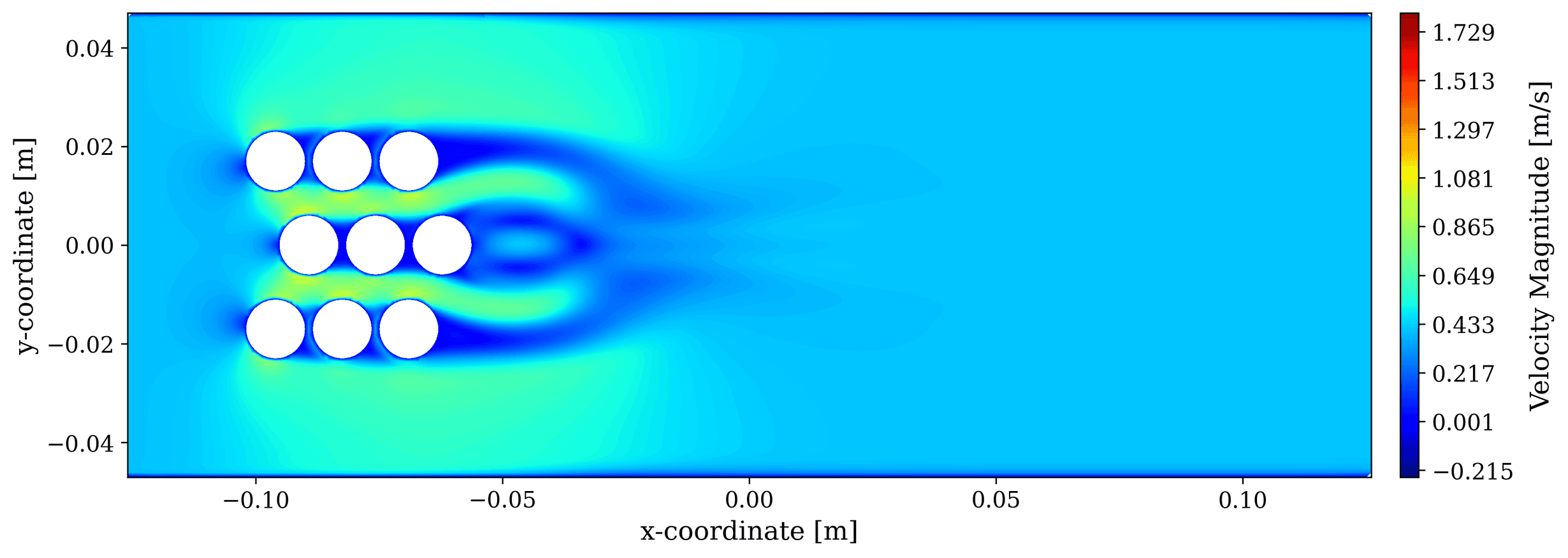} &
        \includegraphics[width=0.45\textwidth,valign=c]{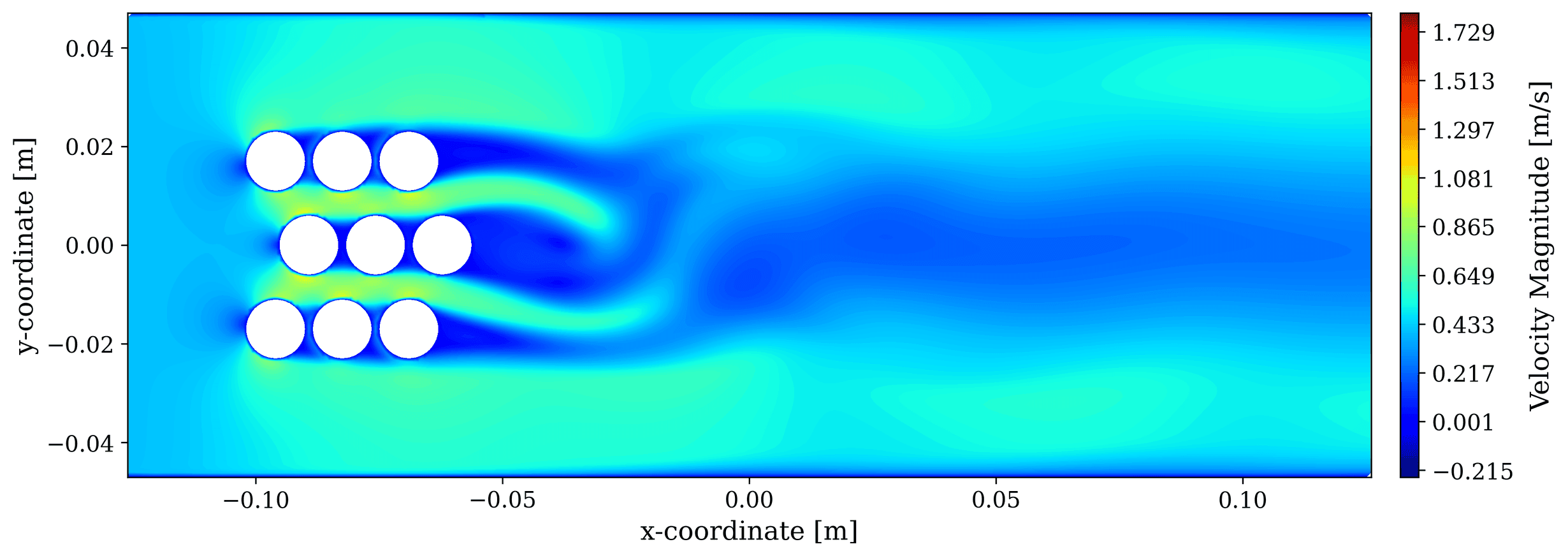} &
        \includegraphics[width=0.45\textwidth,valign=c]{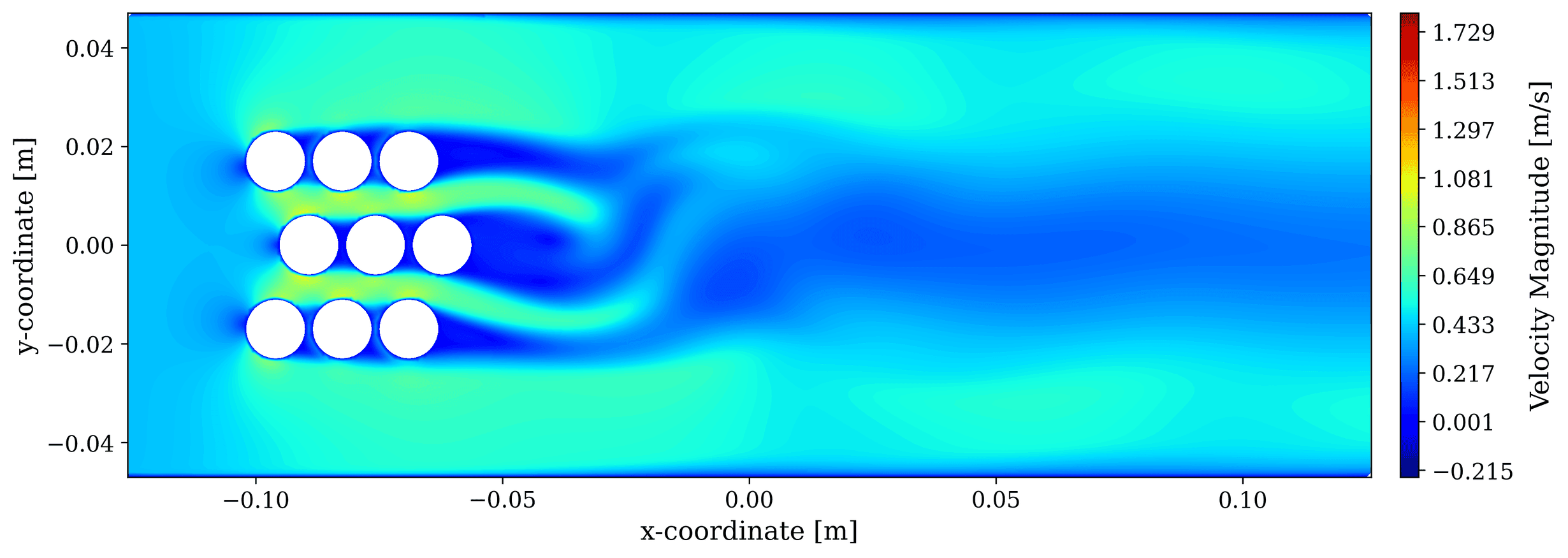} \\[2pt]
        \adjustbox{valign=c}{\rotatebox[origin=c]{90}{\small\textbf{Predicted}}} &
        \includegraphics[width=0.45\textwidth,valign=c]{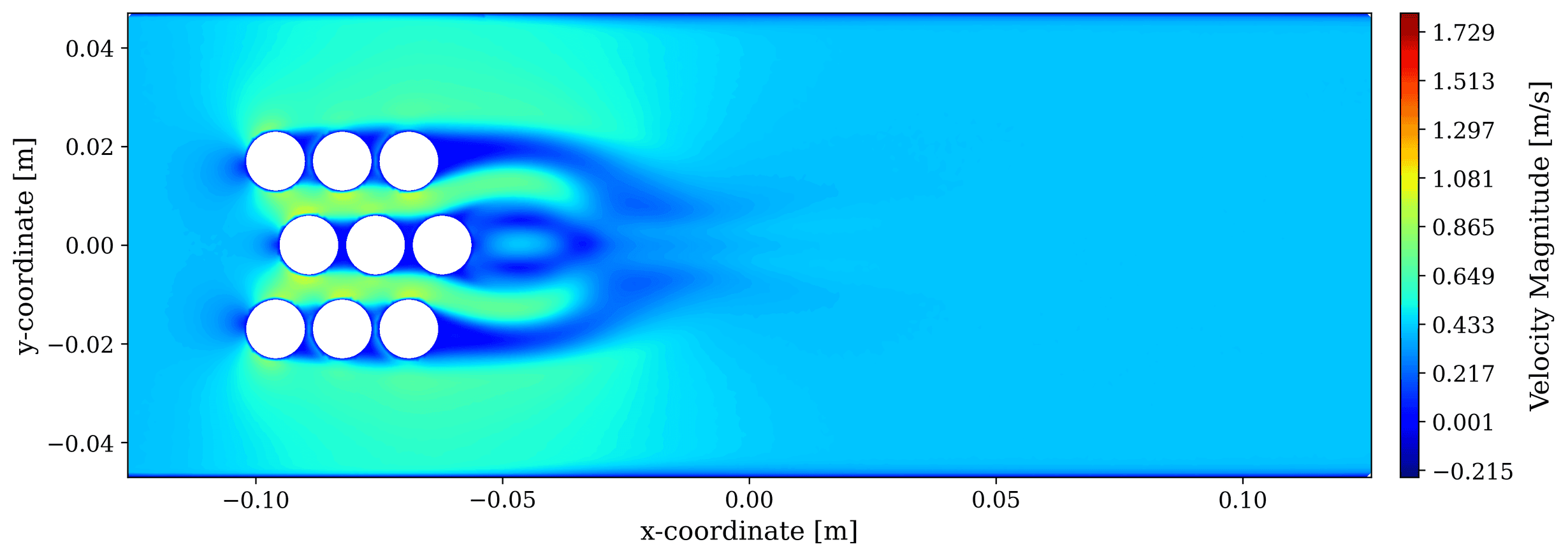} &
        \includegraphics[width=0.45\textwidth,valign=c]{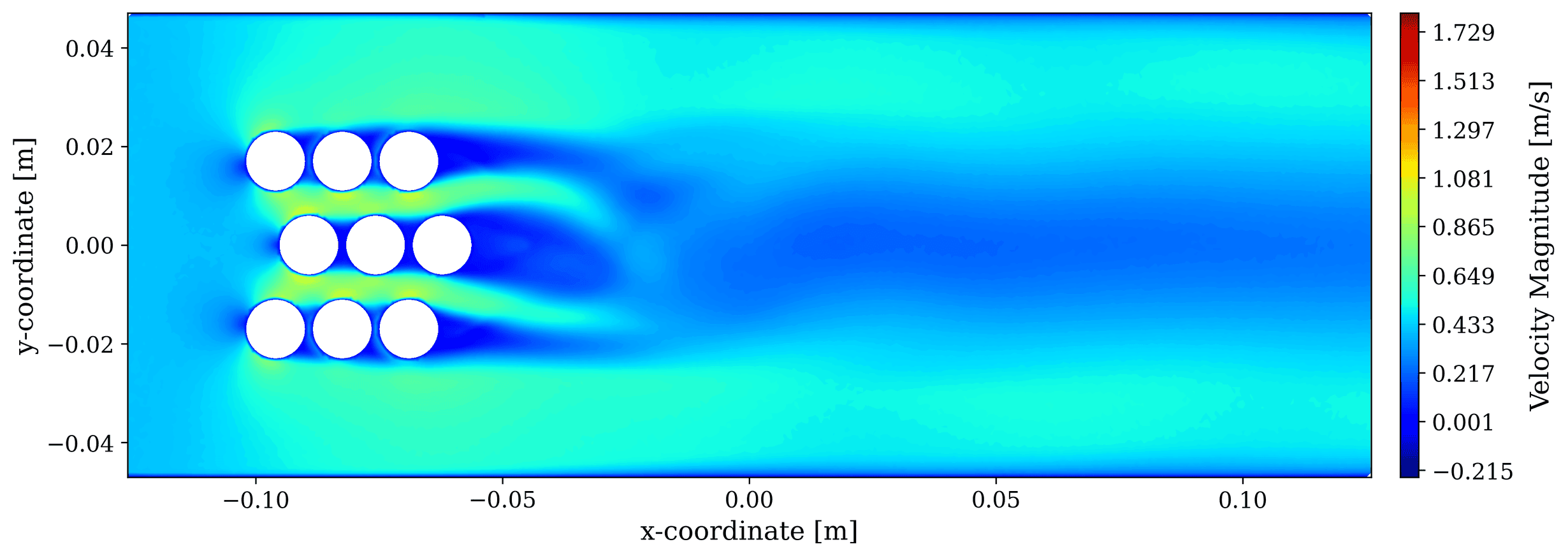} &
        \includegraphics[width=0.45\textwidth,valign=c]{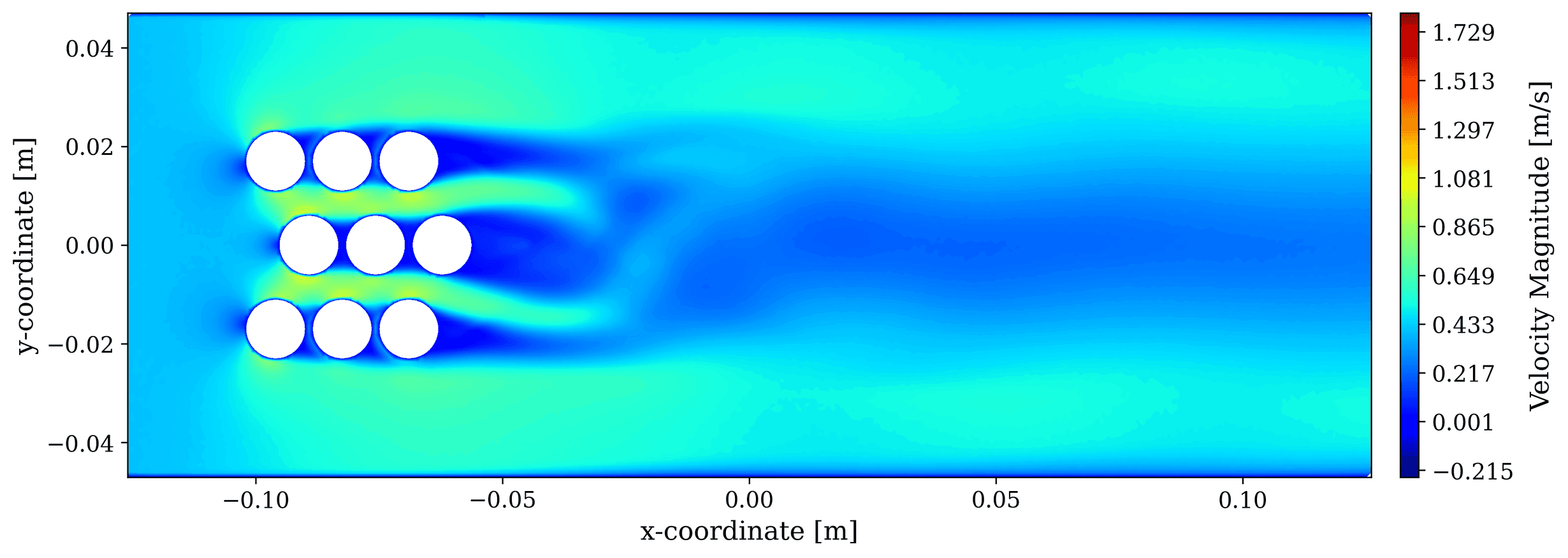} \\[2pt]
        \adjustbox{valign=c}{\rotatebox[origin=c]{90}{\small\textbf{Error}}} &
        \includegraphics[width=0.45\textwidth,valign=c]{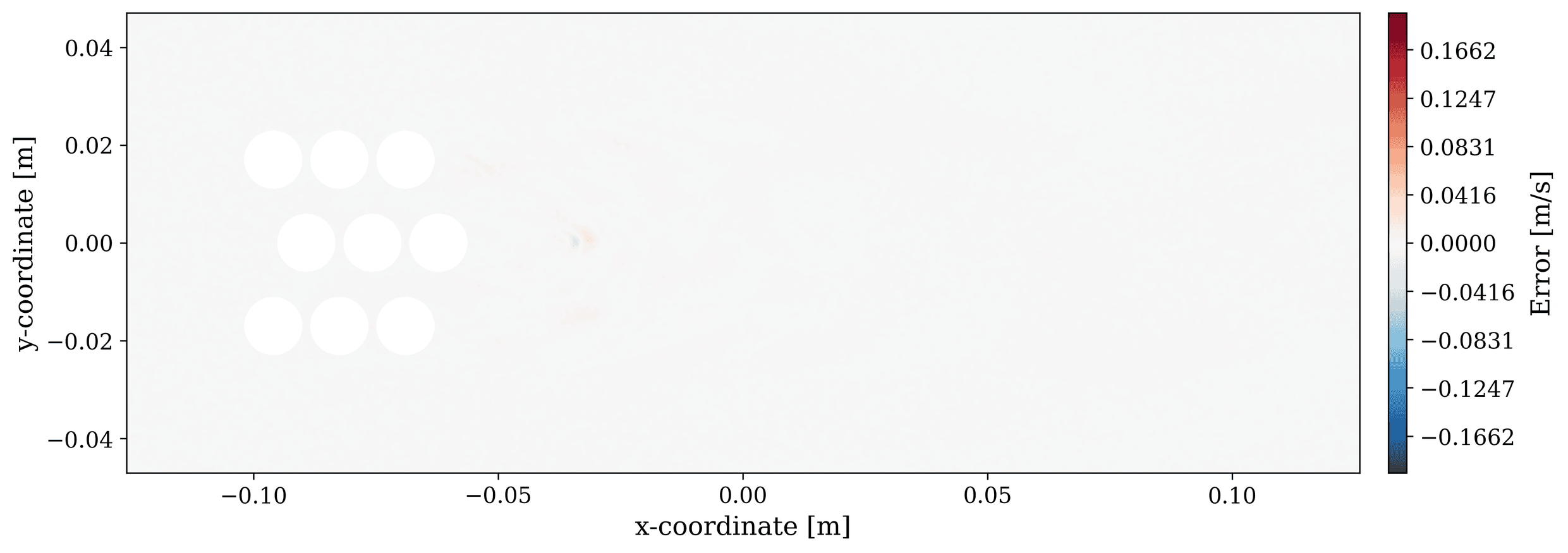} &
        \includegraphics[width=0.45\textwidth,valign=c]{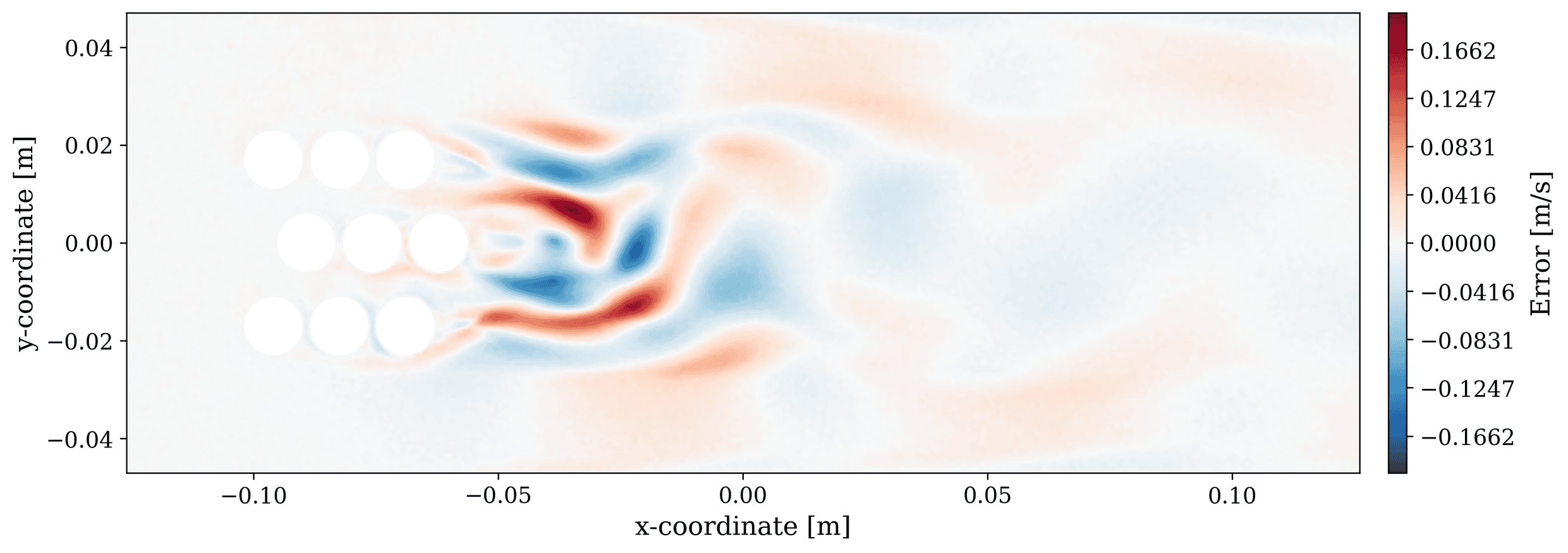} &
        \includegraphics[width=0.45\textwidth,valign=c]{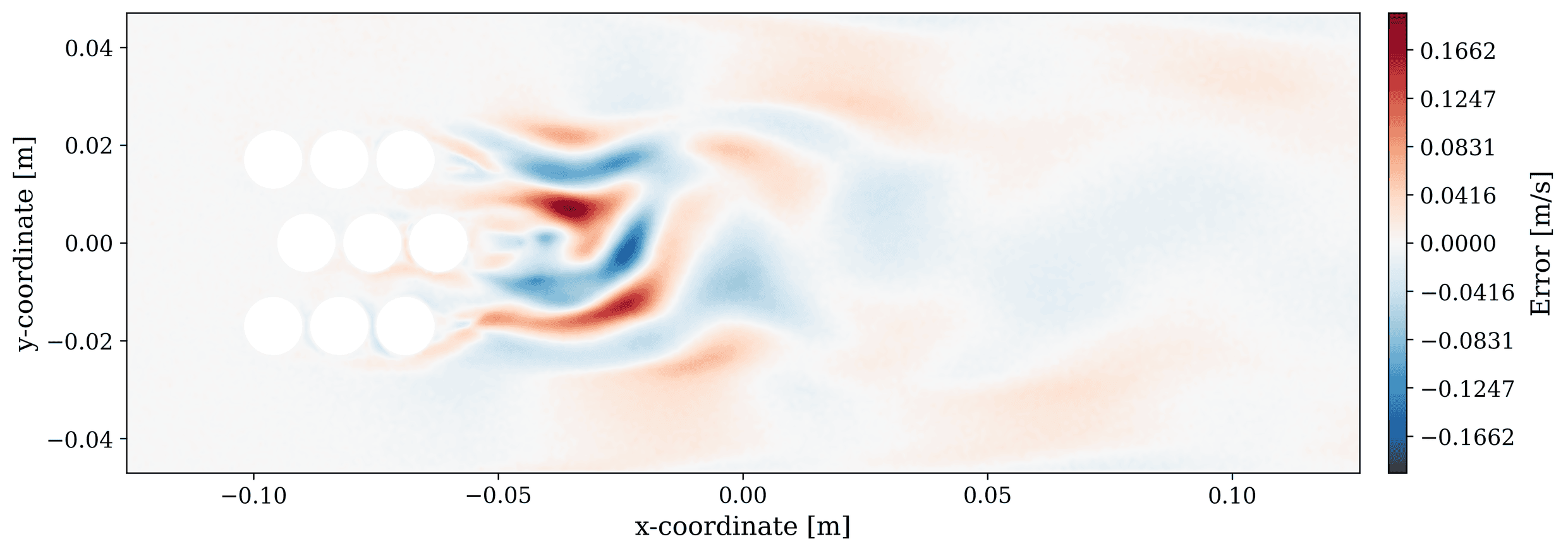} \\
    \end{tabular}
    }%
    \caption{Predicted velocity field by the multi-scale L-DeepONet with MLP-based AE at inlet velocity 0.4 m/s. Top: reference CFD solution, middle: predicted field, bottom: absolute error at $t = 2$, $50$, and $100$.}
    \label{fig:ldon_unstructured_velocity_inlet040}
\end{figure}

\begin{figure}[H]
    \centering
    \setlength{\tabcolsep}{1pt}
    \makebox[\textwidth][c]{%
    \begin{tabular}{c@{\hspace{4pt}}ccc}
        & \textbf{$t = 2$} & \textbf{$t = 50$} & \textbf{$t = 100$} \\
        \adjustbox{valign=c}{\rotatebox[origin=c]{90}{\small\textbf{Reference}}} &
        \includegraphics[width=0.45\textwidth,valign=c]{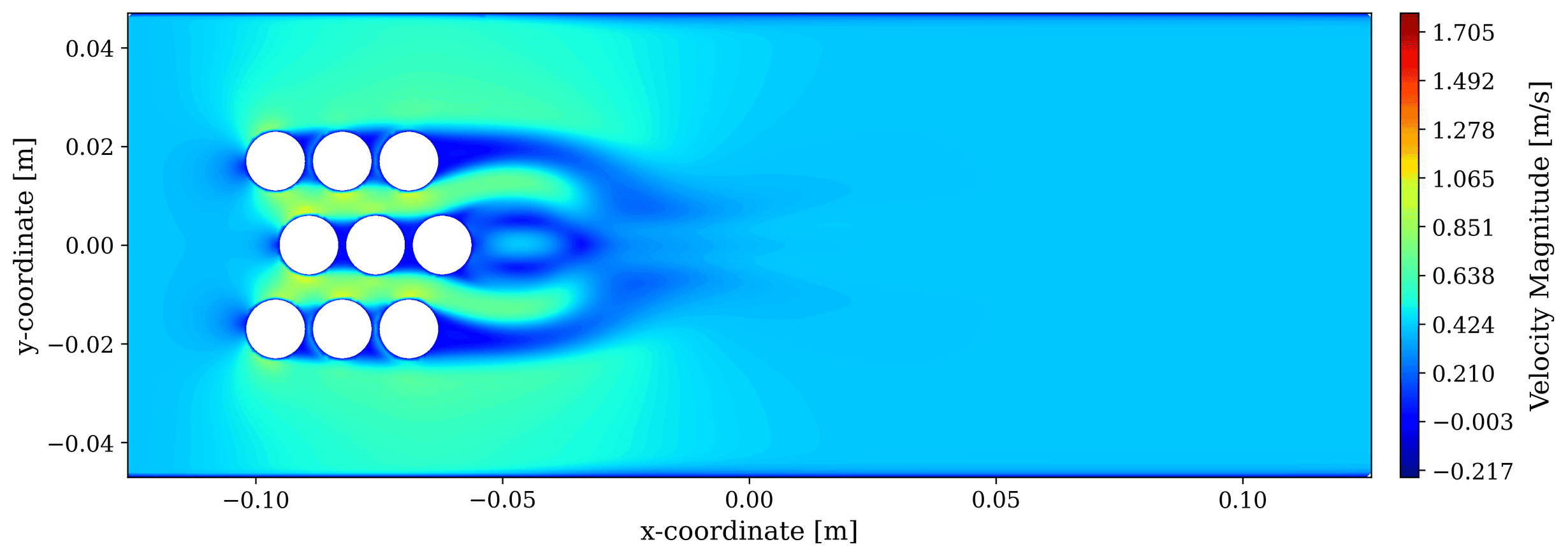} &
        \includegraphics[width=0.45\textwidth,valign=c]{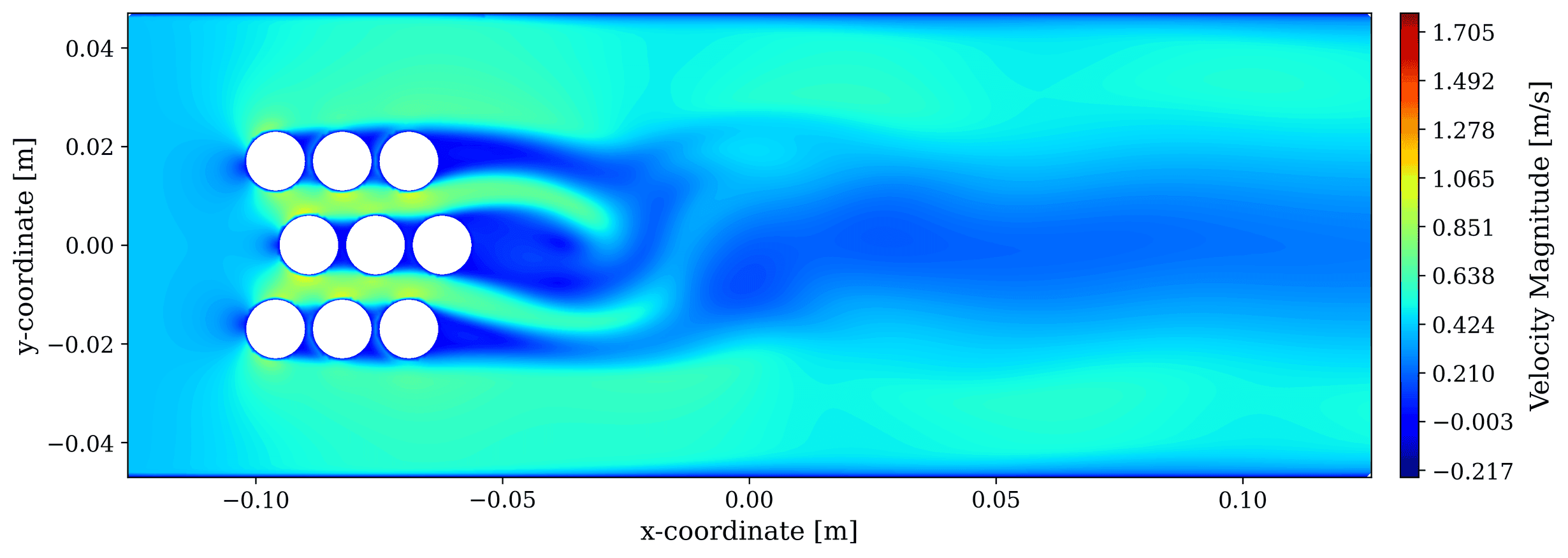} &
        \includegraphics[width=0.45\textwidth,valign=c]{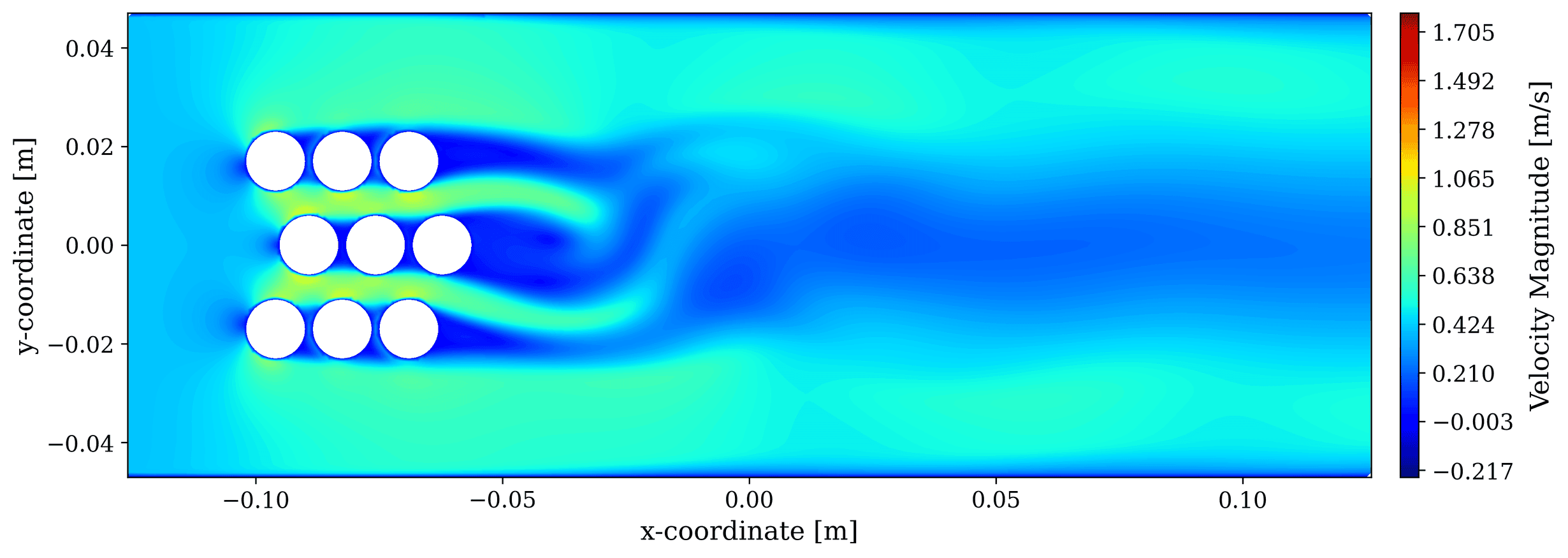} \\[2pt]
        \adjustbox{valign=c}{\rotatebox[origin=c]{90}{\small\textbf{Predicted}}} &
        \includegraphics[width=0.45\textwidth,valign=c]{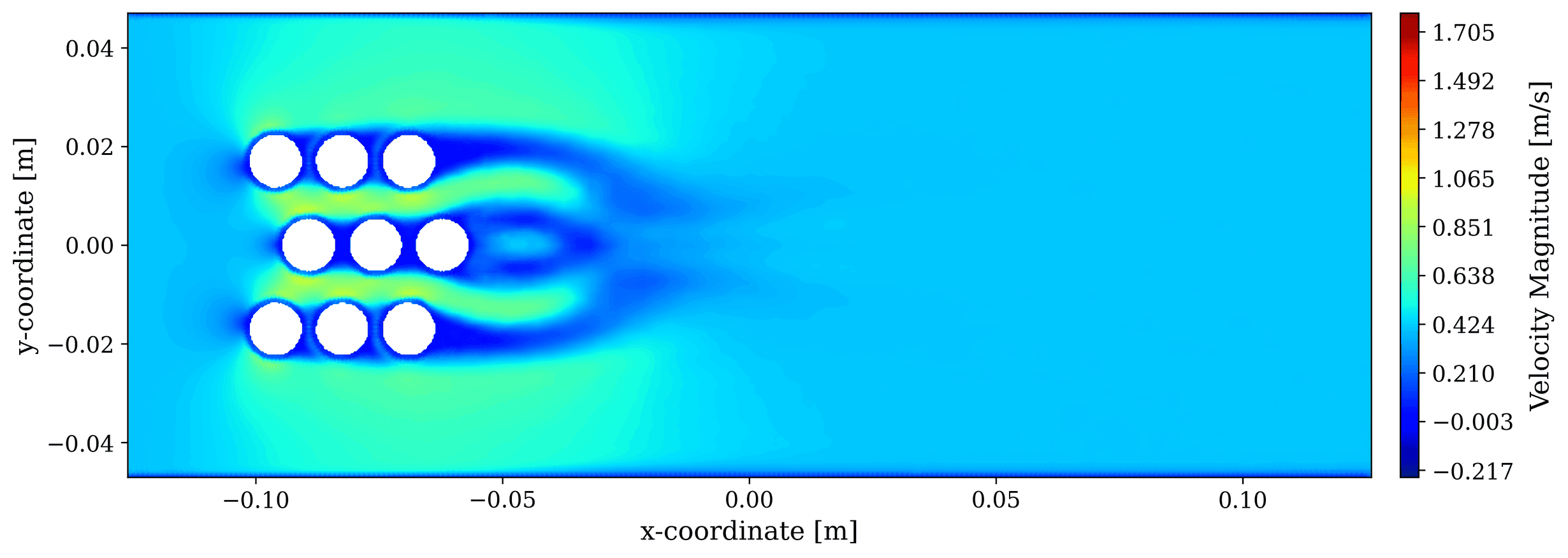} &
        \includegraphics[width=0.45\textwidth,valign=c]{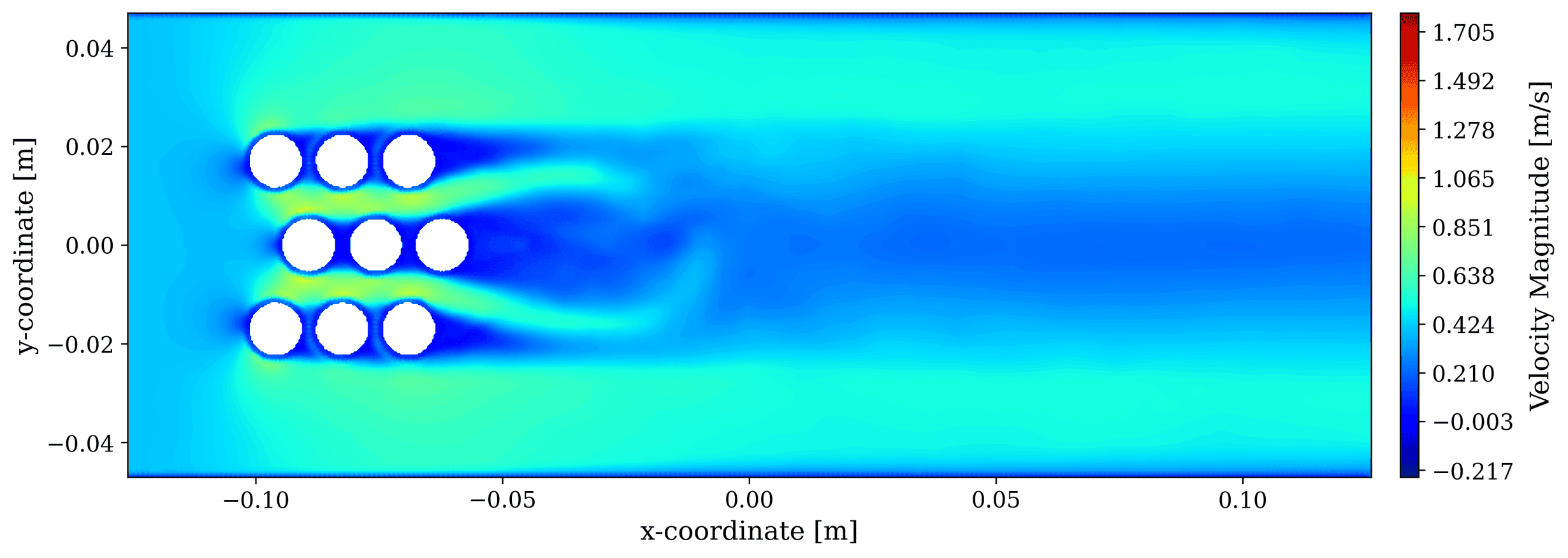} &
        \includegraphics[width=0.45\textwidth,valign=c]{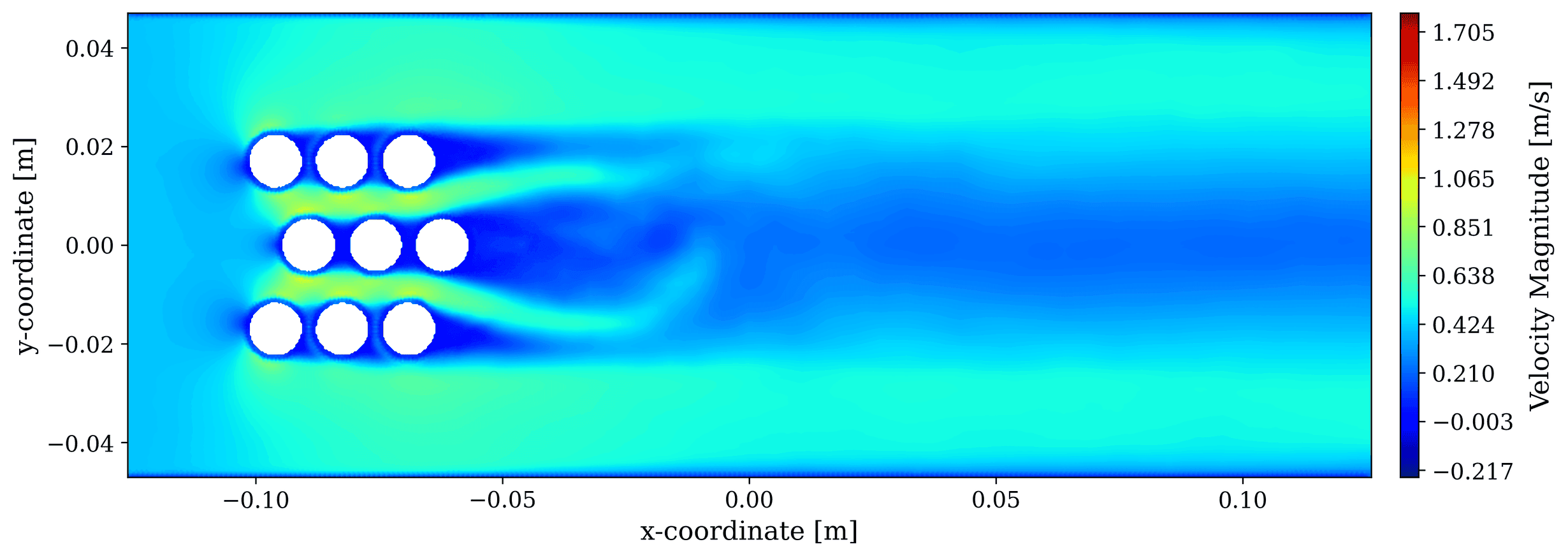} \\[2pt]
        \adjustbox{valign=c}{\rotatebox[origin=c]{90}{\small\textbf{Error}}} &
        \includegraphics[width=0.45\textwidth,valign=c]{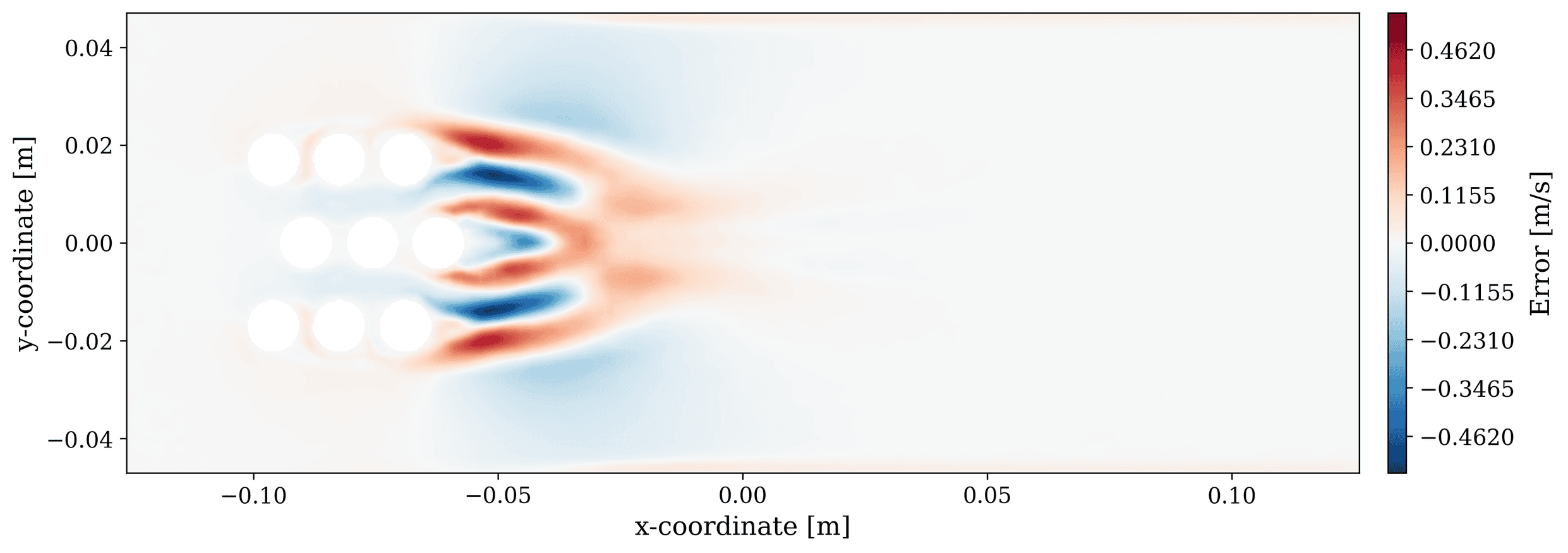} &
        \includegraphics[width=0.45\textwidth,valign=c]{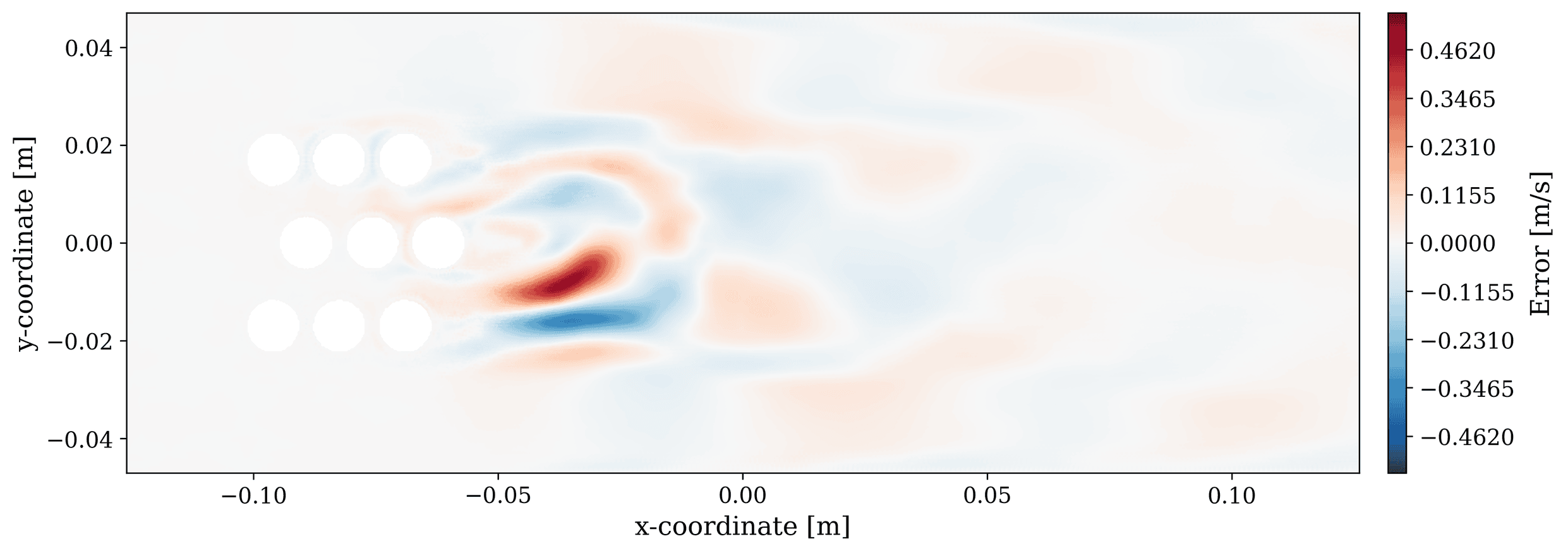} &
        \includegraphics[width=0.45\textwidth,valign=c]{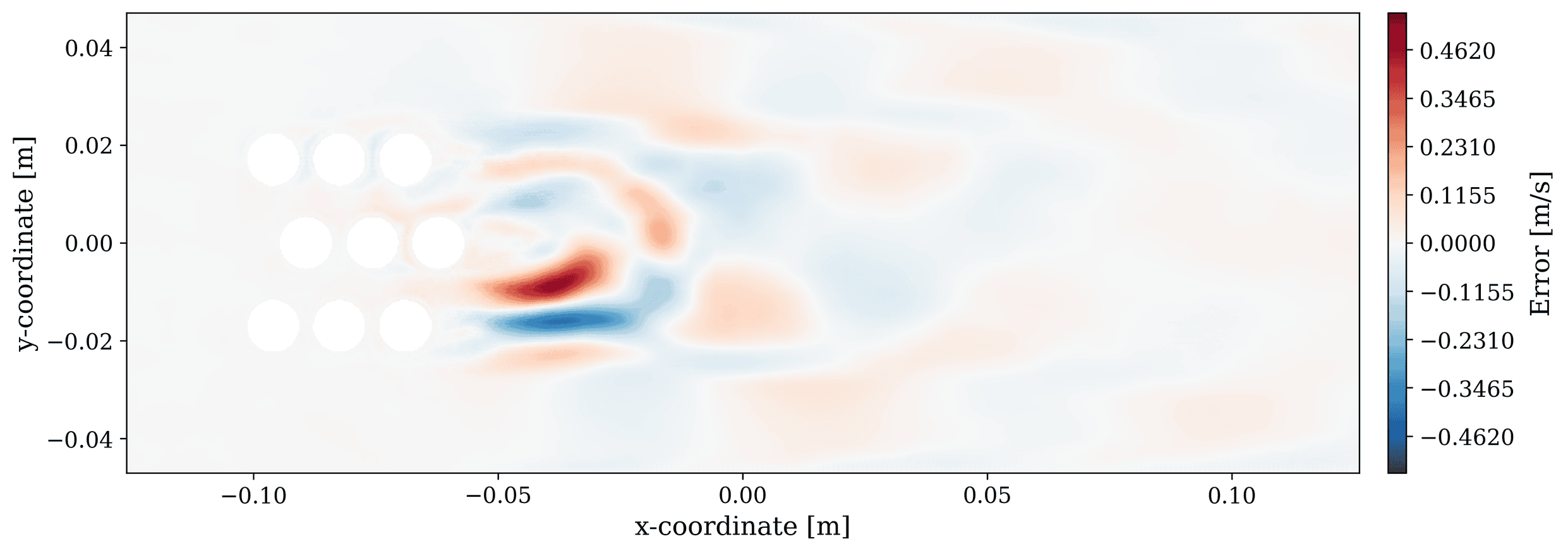} \\
    \end{tabular}
    }%
    \caption{Predicted velocity field by the multi-scale L-DeepONet with CAE at inlet velocity 0.4 m/s. Top: reference CFD solution, middle: predicted field, bottom: absolute error at $t = 2$, $50$, and $100$.}
    \label{fig:cae_ldon_velocity_inlet040}
\end{figure}

\begin{figure}[H]
    \centering
    \setlength{\tabcolsep}{1pt}
    \makebox[\textwidth][c]{%
    \begin{tabular}{c@{\hspace{4pt}}ccc}
        & \textbf{$t = 2$} & \textbf{$t = 50$} & \textbf{$t = 100$} \\
        \adjustbox{valign=c}{\rotatebox[origin=c]{90}{\small\textbf{Reference}}} &
        \includegraphics[width=0.45\textwidth,valign=c]{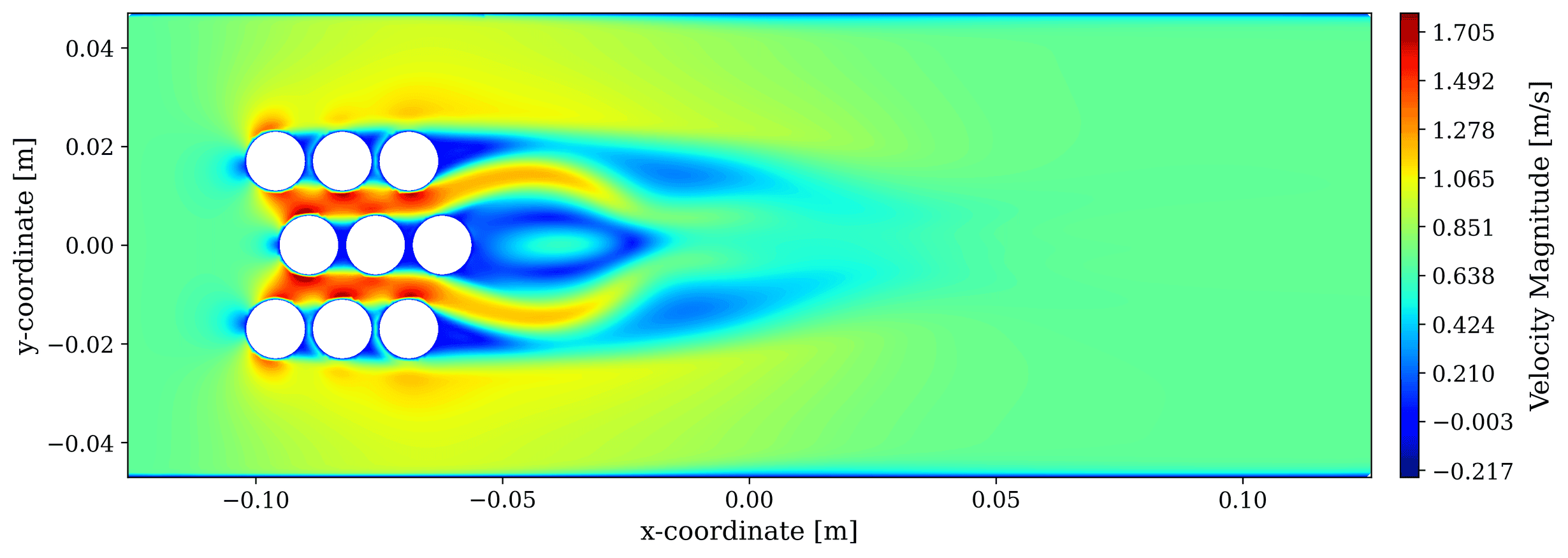} &
        \includegraphics[width=0.45\textwidth,valign=c]{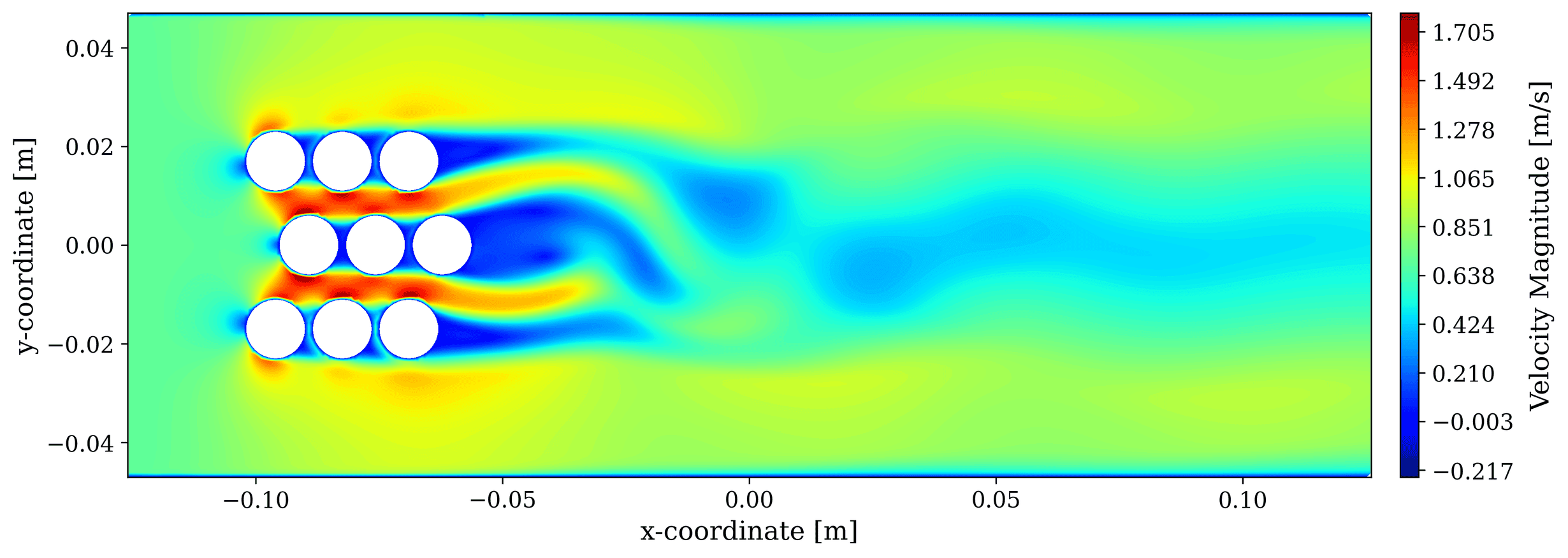} &
        \includegraphics[width=0.45\textwidth,valign=c]{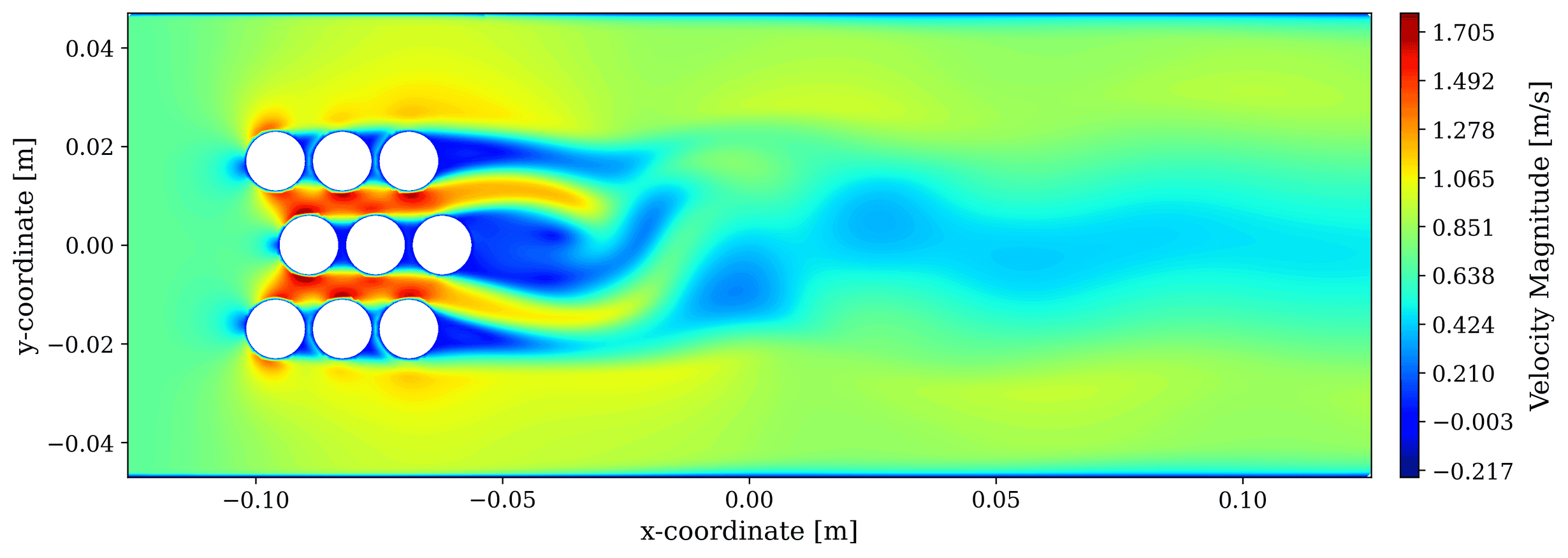} \\[2pt]
        \adjustbox{valign=c}{\rotatebox[origin=c]{90}{\small\textbf{Predicted}}} &
        \includegraphics[width=0.45\textwidth,valign=c]{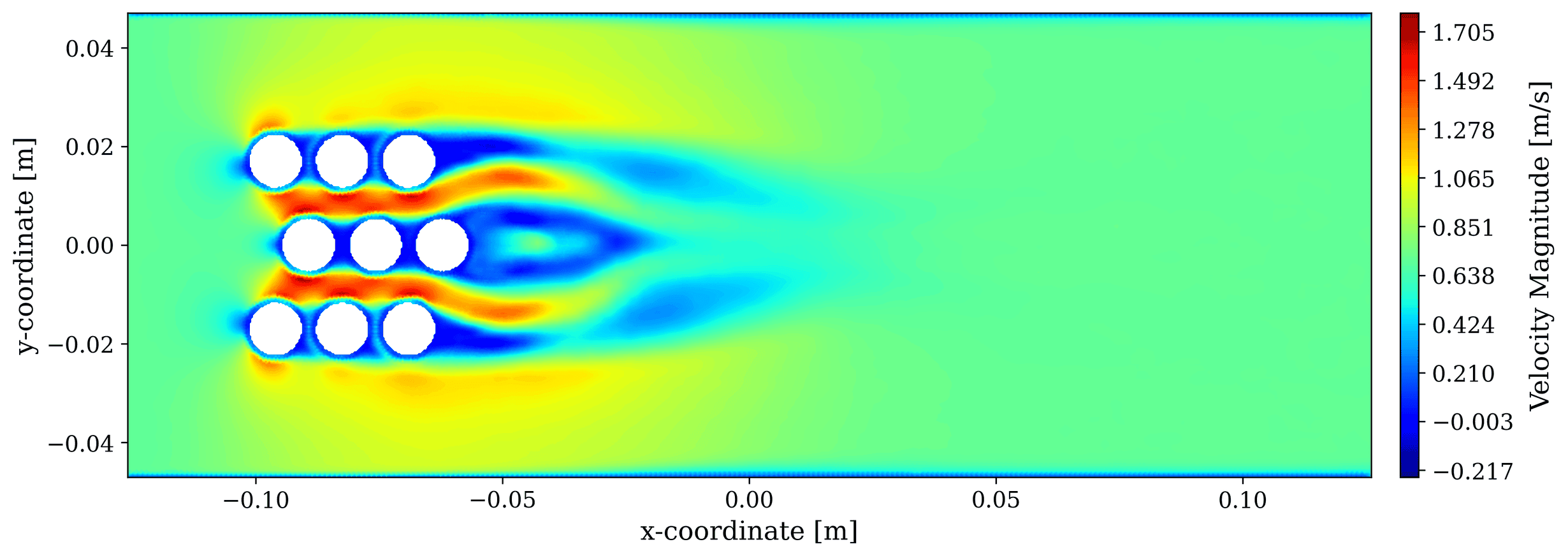} &
        \includegraphics[width=0.45\textwidth,valign=c]{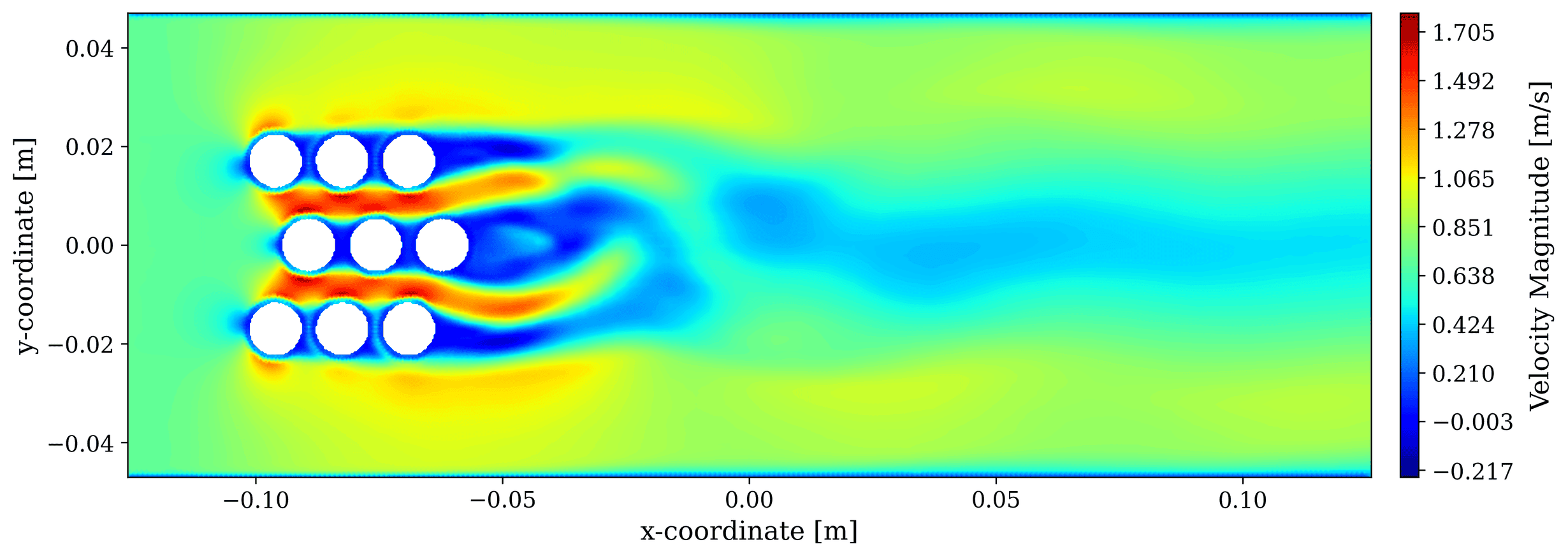} &
        \includegraphics[width=0.45\textwidth,valign=c]{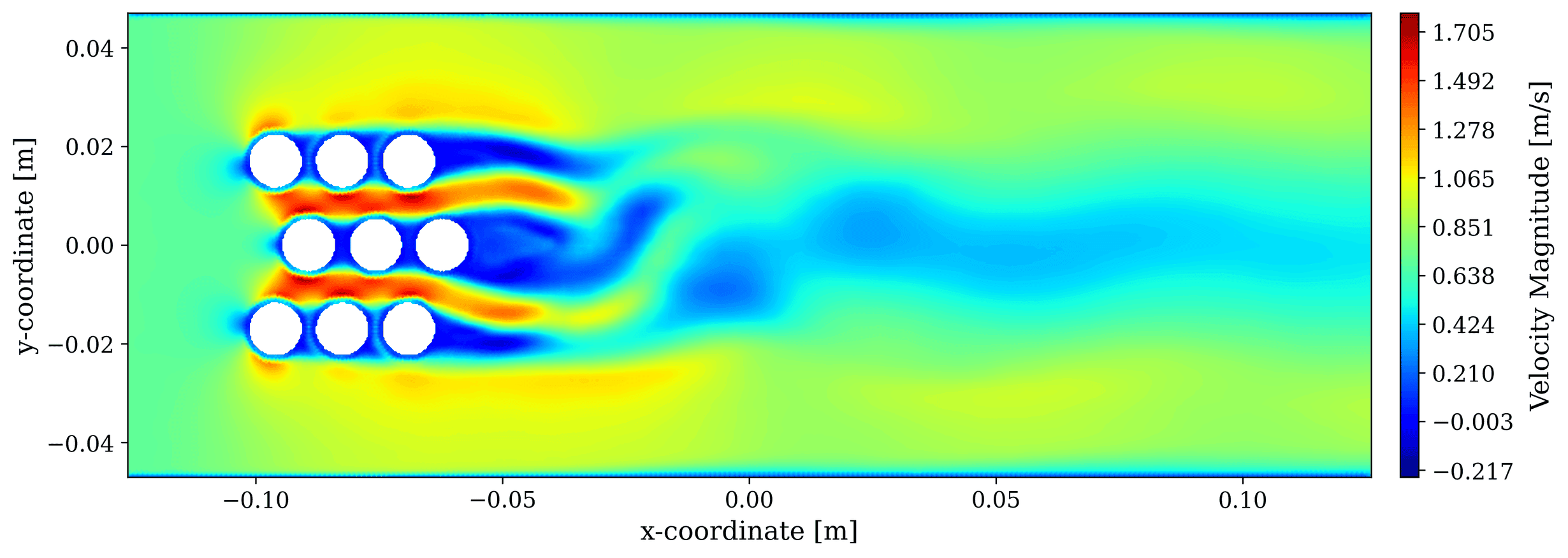} \\[2pt]
        \adjustbox{valign=c}{\rotatebox[origin=c]{90}{\small\textbf{Error}}} &
        \includegraphics[width=0.45\textwidth,valign=c]{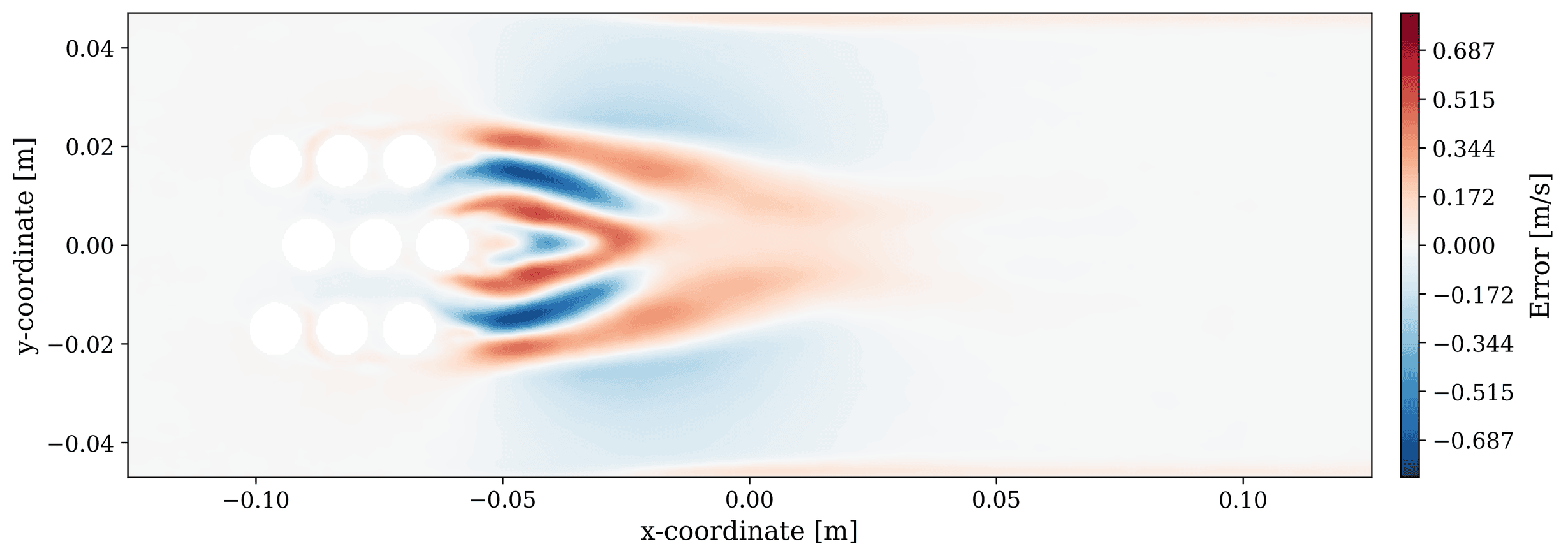} &
        \includegraphics[width=0.45\textwidth,valign=c]{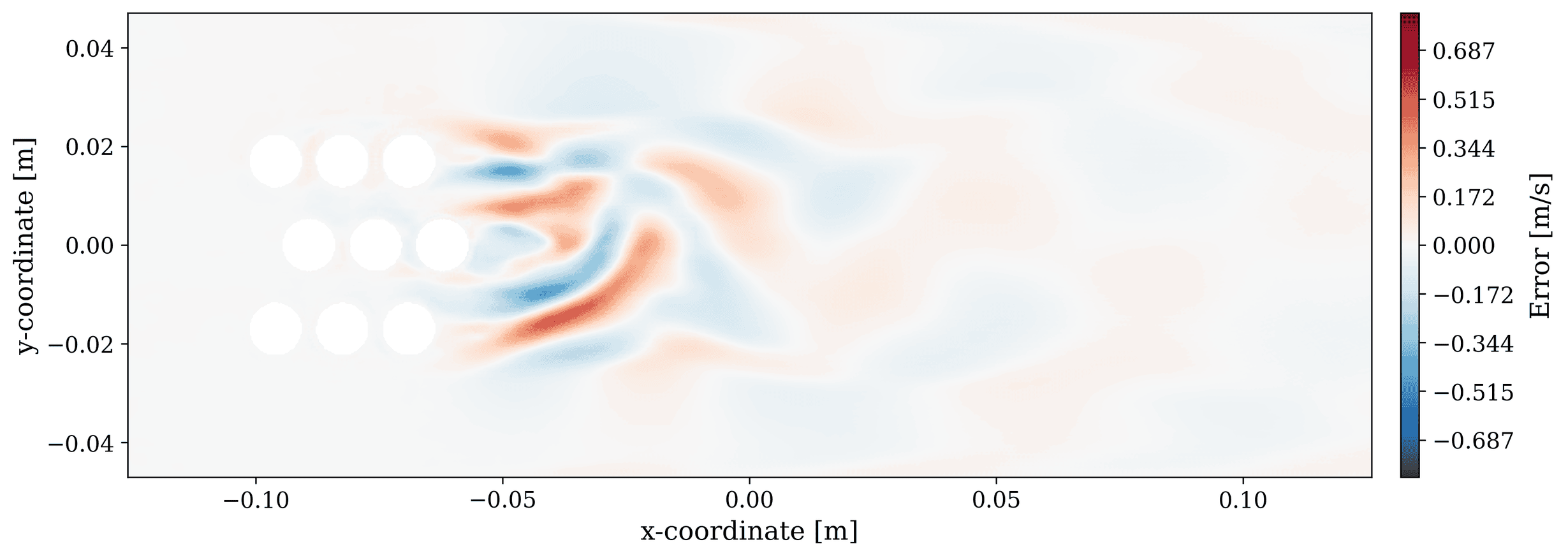} &
        \includegraphics[width=0.45\textwidth,valign=c]{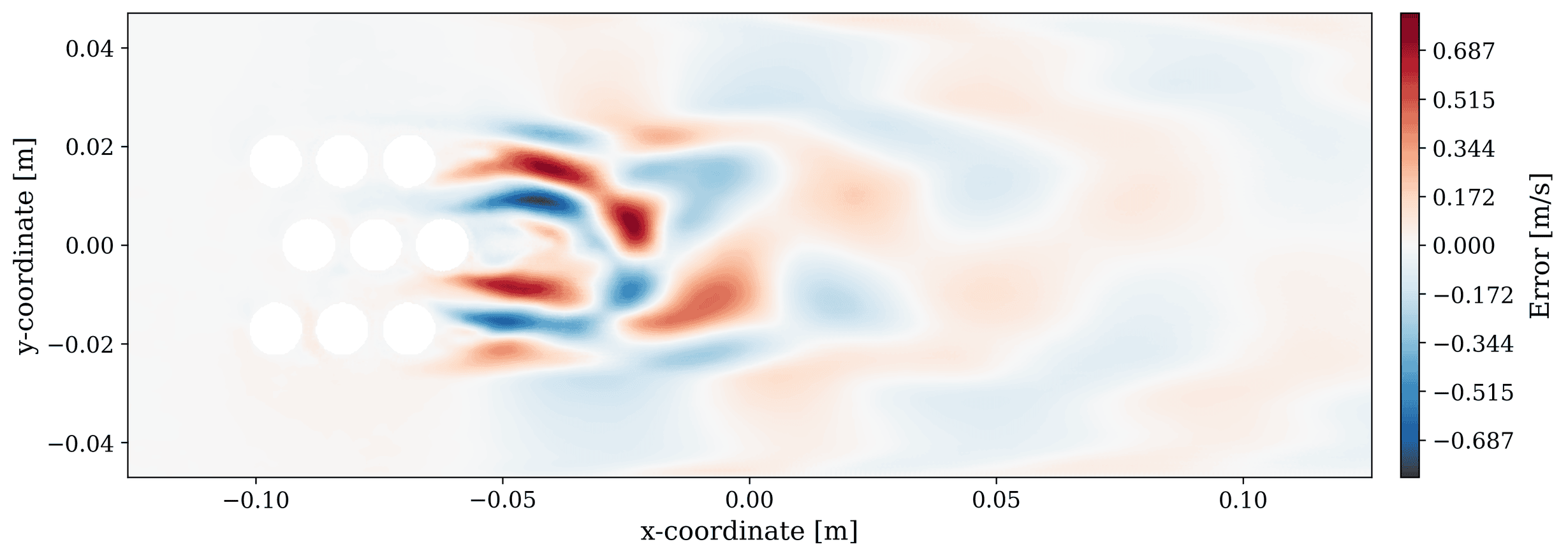} \\
    \end{tabular}
    }%
    \caption{Predicted velocity field by the multi-scale L-DeepONet with CAE at inlet velocity 0.7 m/s. Top: reference CFD solution, middle: predicted field, bottom: absolute error at $t = 2$, $50$, and $100$.}
    \label{fig:cae_ldon_velocity_inlet070}
\end{figure}

Similarly, Figure~\ref{fig:cae_ldon_velocity_inlet040} and Figure~\ref{fig:cae_ldon_velocity_inlet070} present the results of the multi-scale L-DeepONet with CAE. This model also demonstrates the ability to predict the periodic flow variations, capturing the vortex streets patterns at all time steps for both inlet velocity conditions. The overall prediction quality is comparable to that of the MLP-based AE variant, with errors concentrated near the cylinder surfaces and the downstream wake region. These results indicate that the multi-scale technique is effective regardless of the choice of dimensionality reduction model, successfully enabling L-DeepONet to learn the oscillatory flow dynamics of the HCSG.

\begin{figure}[H]
    \centering
    \setlength{\tabcolsep}{1pt}
    \makebox[\textwidth][c]{%
    \begin{tabular}{c@{\hspace{4pt}}ccc}
        & \textbf{$t = 2$} & \textbf{$t = 50$} & \textbf{$t = 100$} \\
        \adjustbox{valign=c}{\rotatebox[origin=c]{90}{\small\textbf{Reference}}} &
        \includegraphics[width=0.45\textwidth,valign=c]{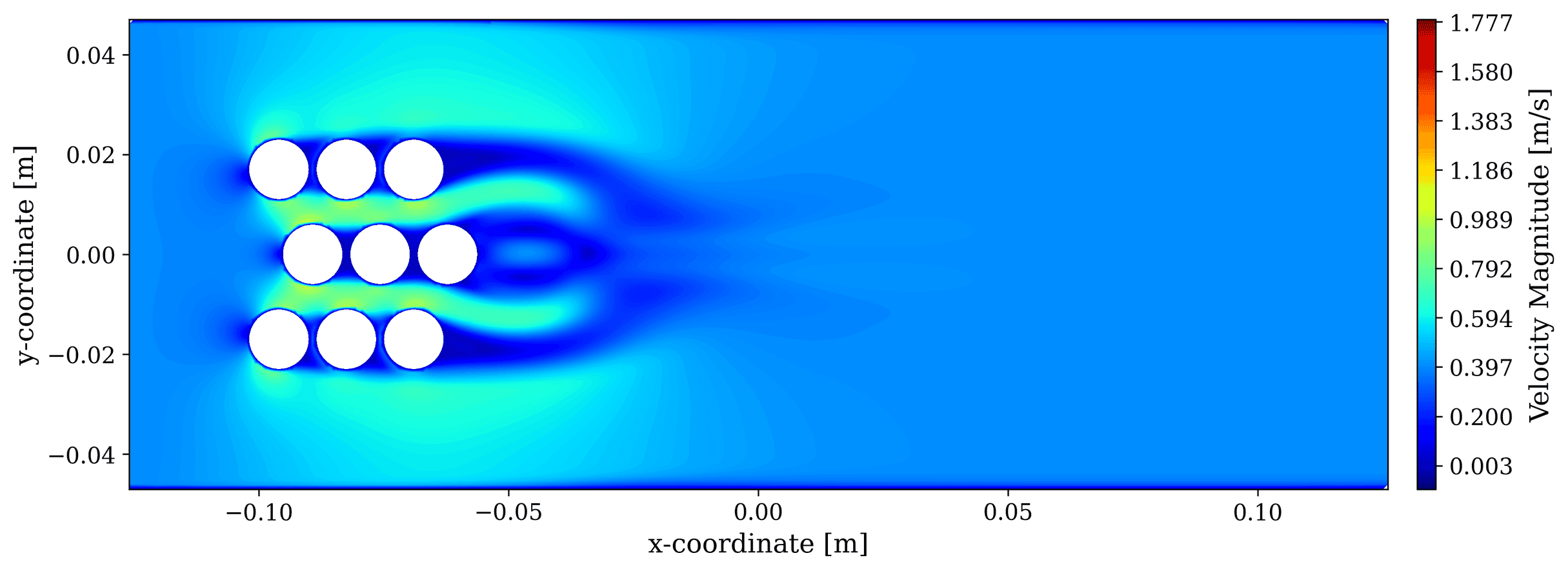} &
        \includegraphics[width=0.45\textwidth,valign=c]{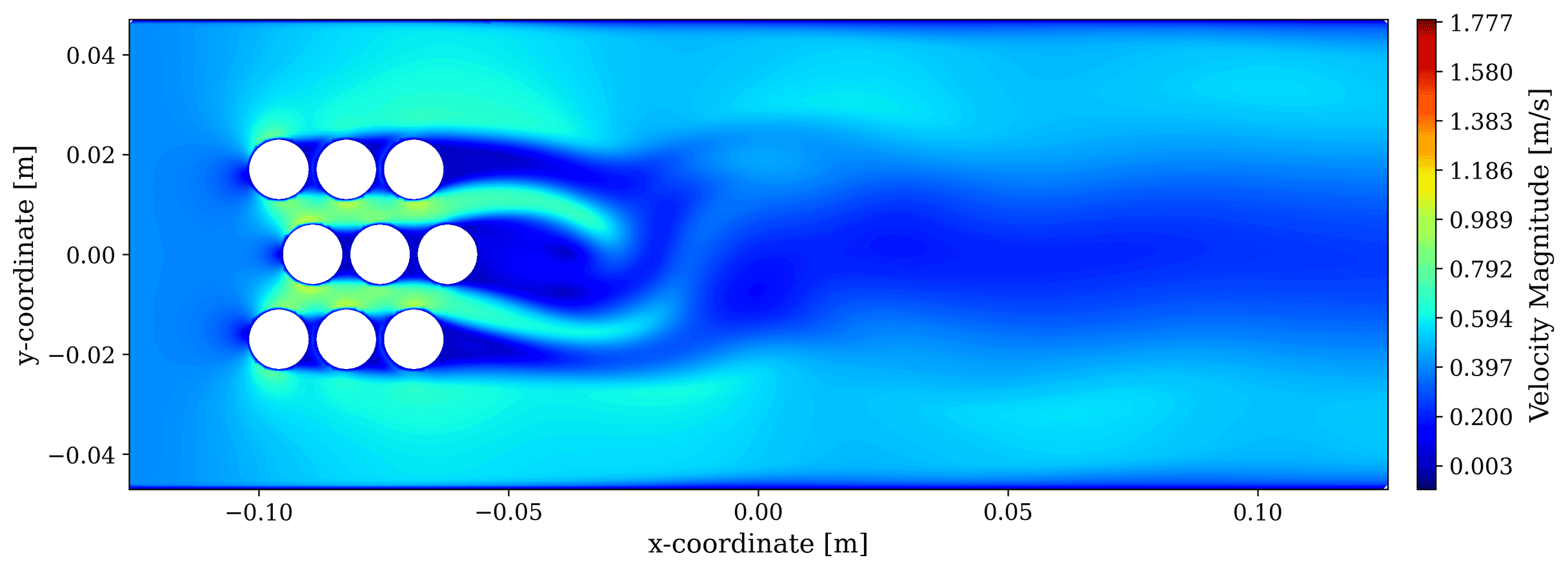} &
        \includegraphics[width=0.45\textwidth,valign=c]{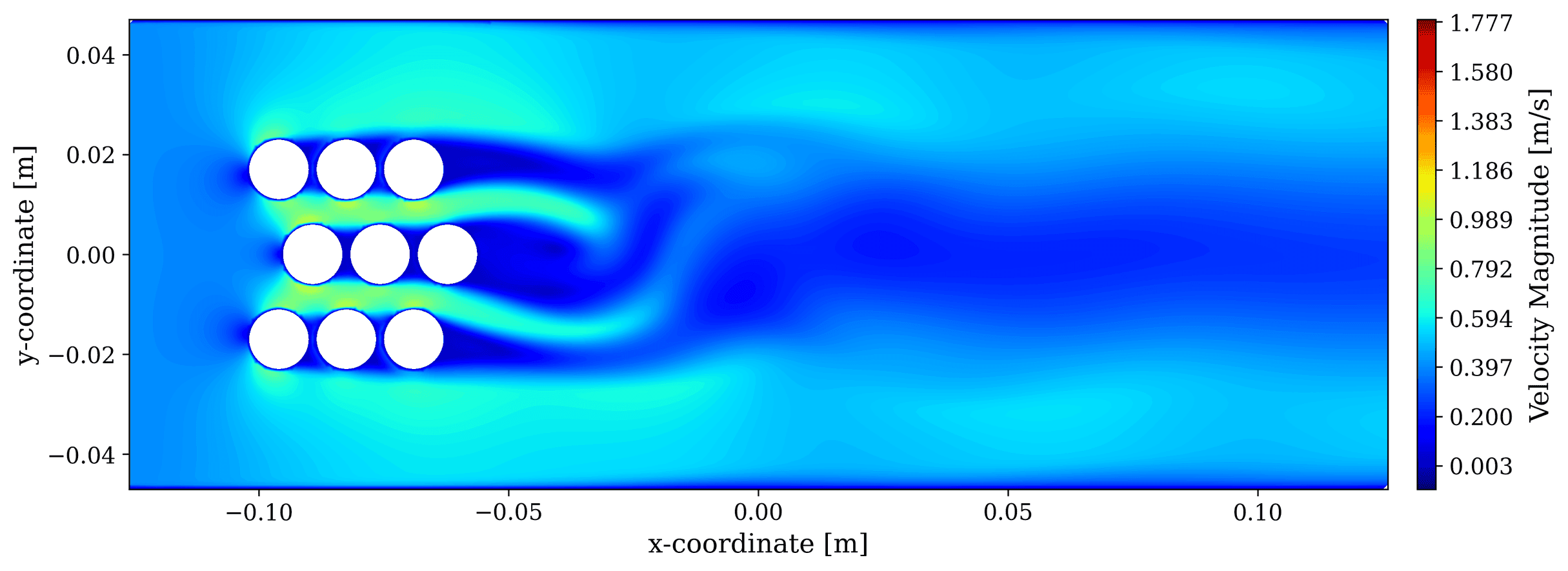} \\[2pt]
        \adjustbox{valign=c}{\rotatebox[origin=c]{90}{\small\textbf{Predicted}}} &
        \includegraphics[width=0.45\textwidth,valign=c]{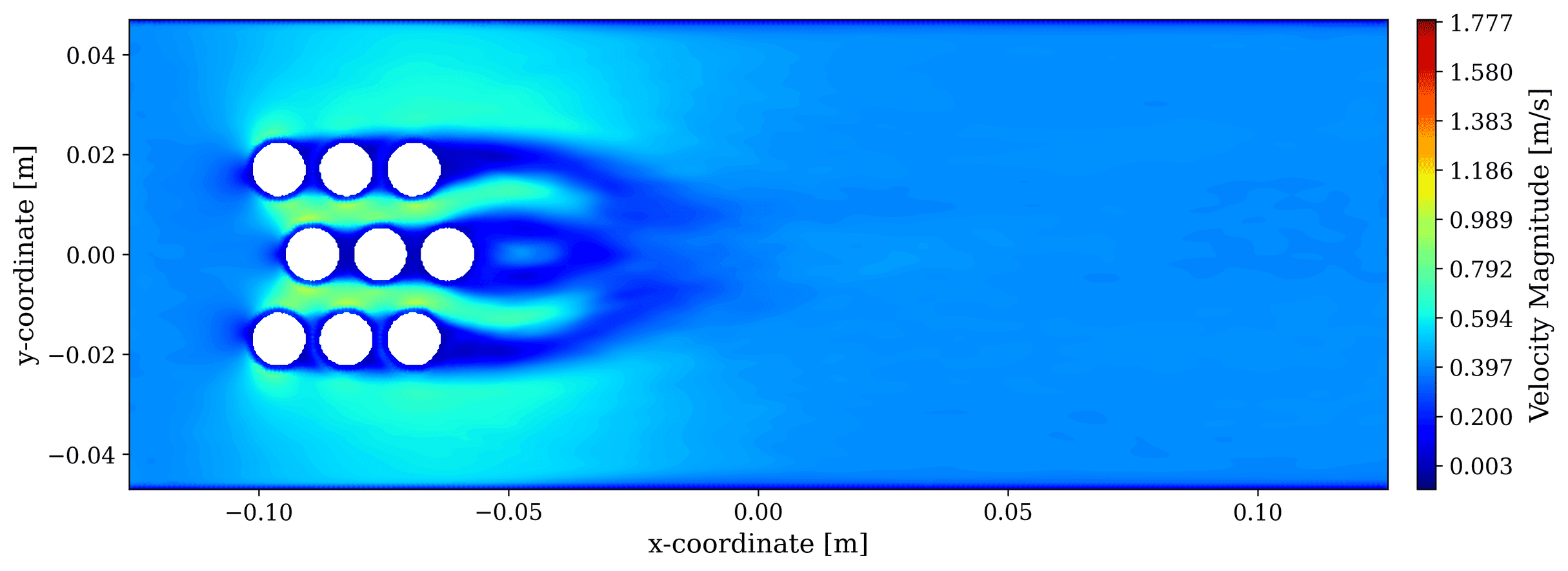} &
        \includegraphics[width=0.45\textwidth,valign=c]{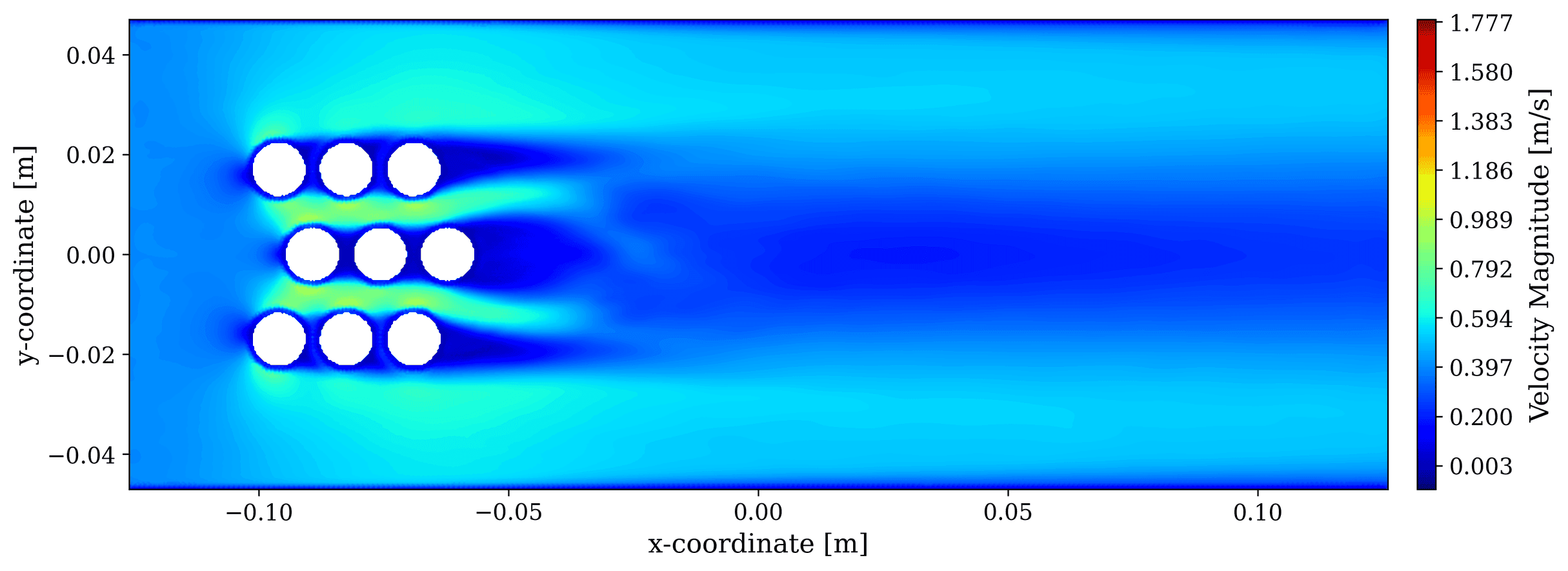} &
        \includegraphics[width=0.45\textwidth,valign=c]{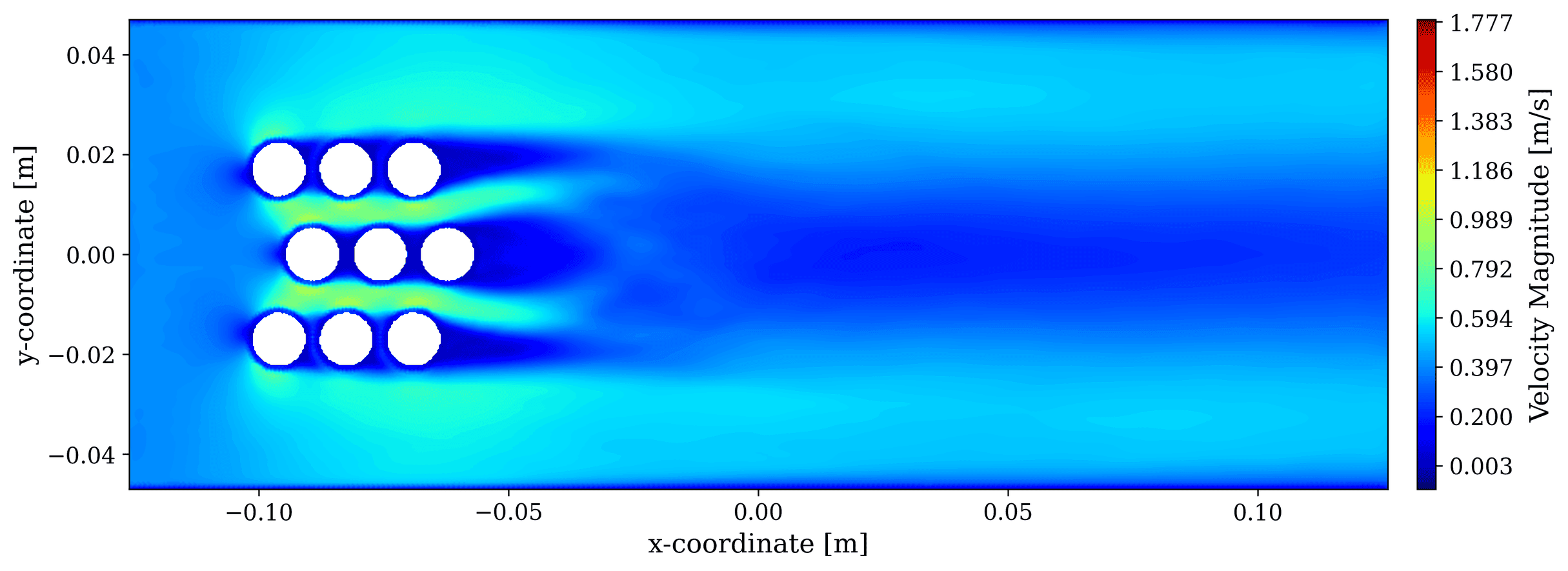} \\[2pt]
        \adjustbox{valign=c}{\rotatebox[origin=c]{90}{\small\textbf{Error}}} &
        \includegraphics[width=0.45\textwidth,valign=c]{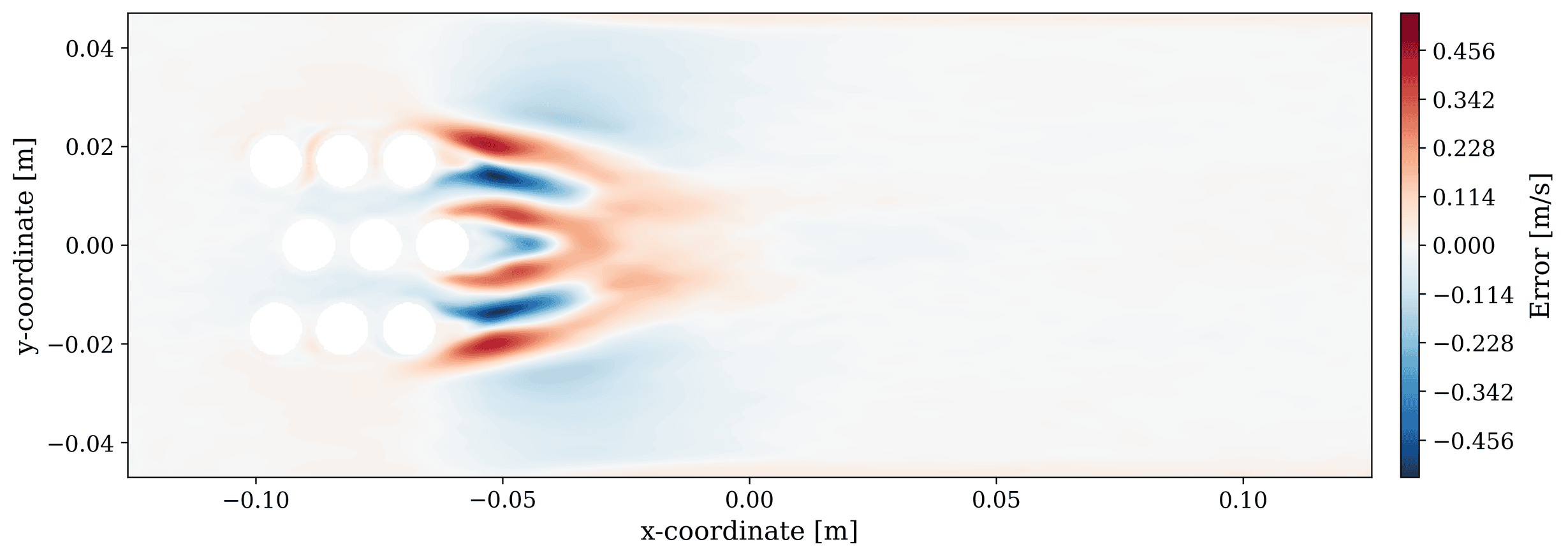} &
        \includegraphics[width=0.45\textwidth,valign=c]{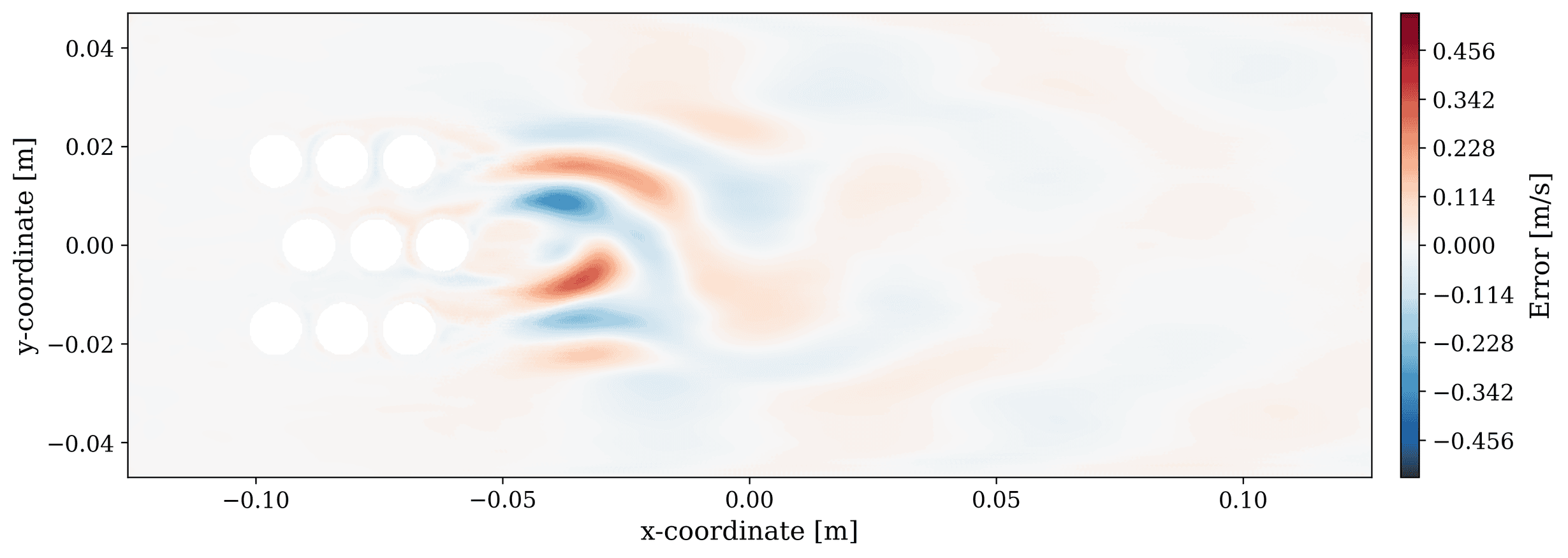} &
        \includegraphics[width=0.45\textwidth,valign=c]{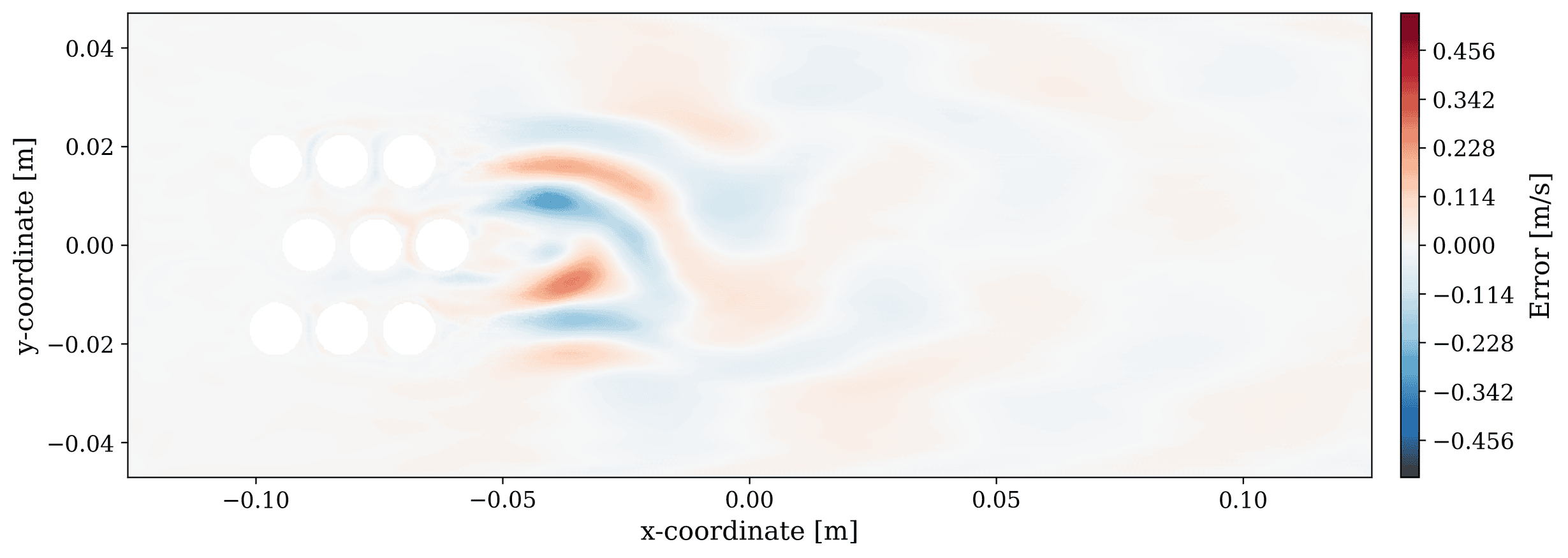} \\
    \end{tabular}
    }%
    \caption{Predicted velocity field by the FNO (24 Fourier modes) at inlet velocity 0.4 m/s. Top: reference CFD solution, middle: predicted field, bottom: absolute error at $t = 2$, $50$, and $100$.}
    \label{fig:fno_velocity_inlet040}
\end{figure}

\begin{figure}[H]
    \centering
    \setlength{\tabcolsep}{1pt}
    \makebox[\textwidth][c]{%
    \begin{tabular}{c@{\hspace{4pt}}ccc}
        & \textbf{$t = 2$} & \textbf{$t = 50$} & \textbf{$t = 100$} \\
        \adjustbox{valign=c}{\rotatebox[origin=c]{90}{\small\textbf{Reference}}} &
        \includegraphics[width=0.45\textwidth,valign=c]{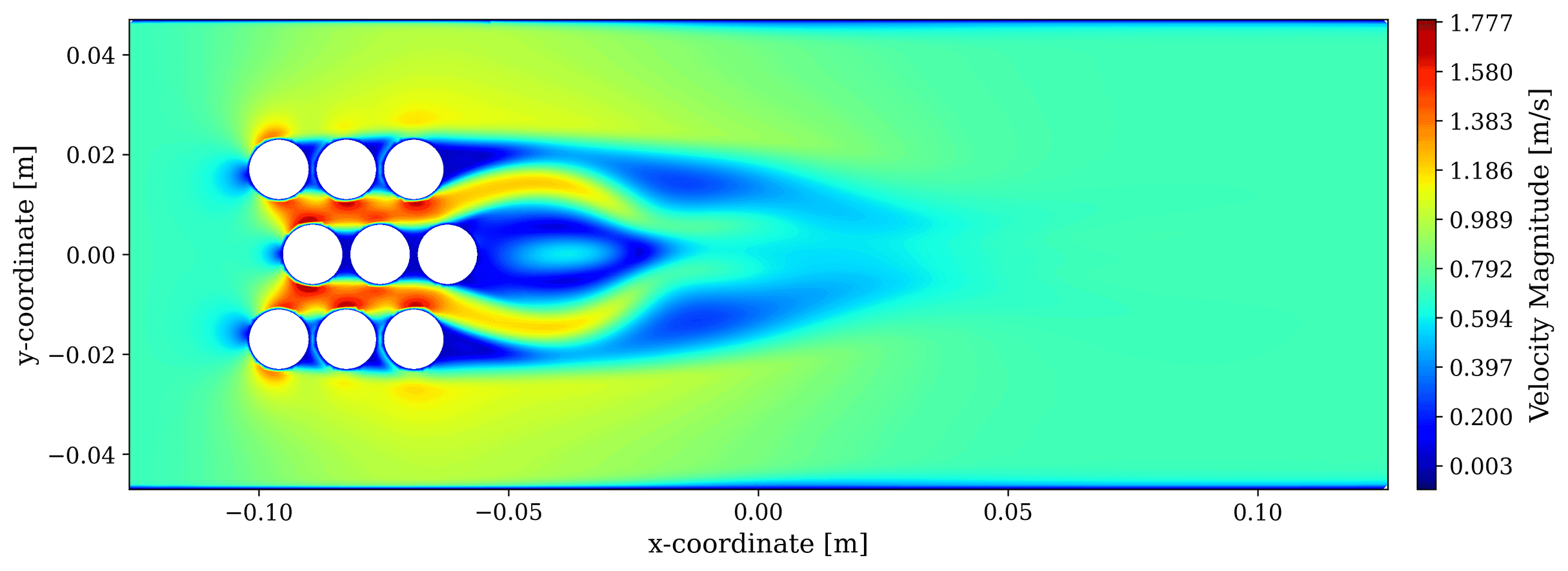} &
        \includegraphics[width=0.45\textwidth,valign=c]{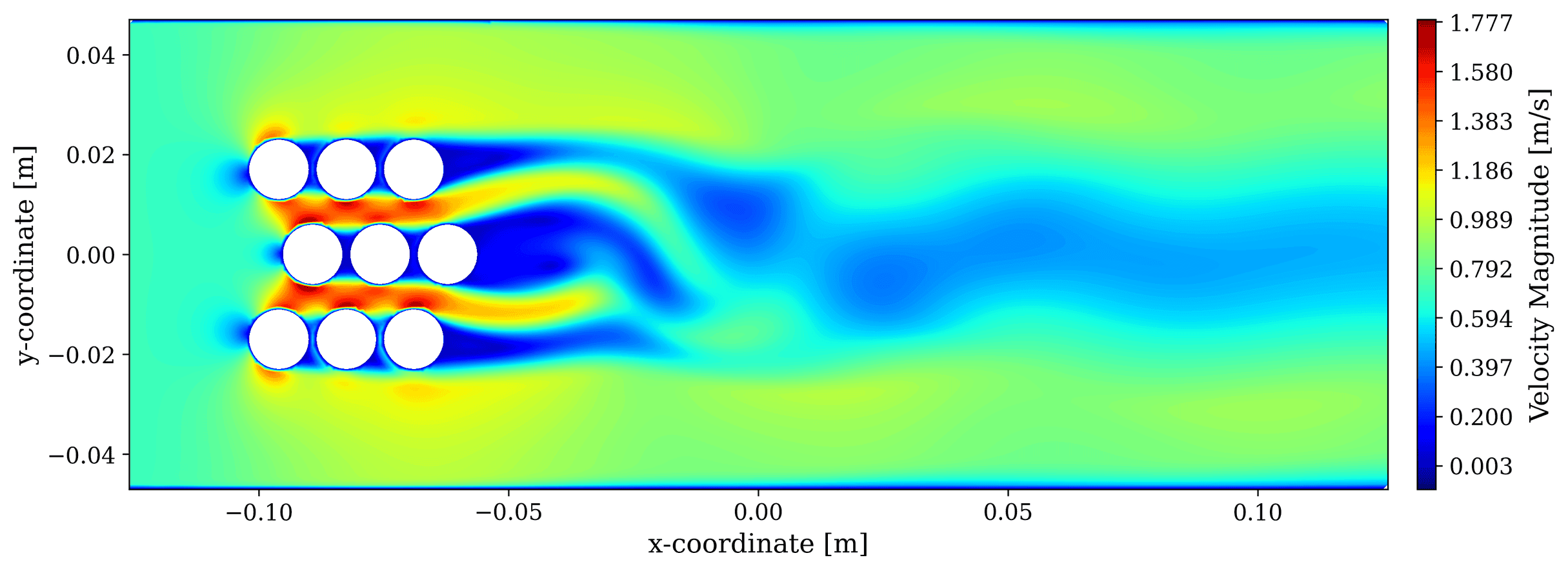} &
        \includegraphics[width=0.45\textwidth,valign=c]{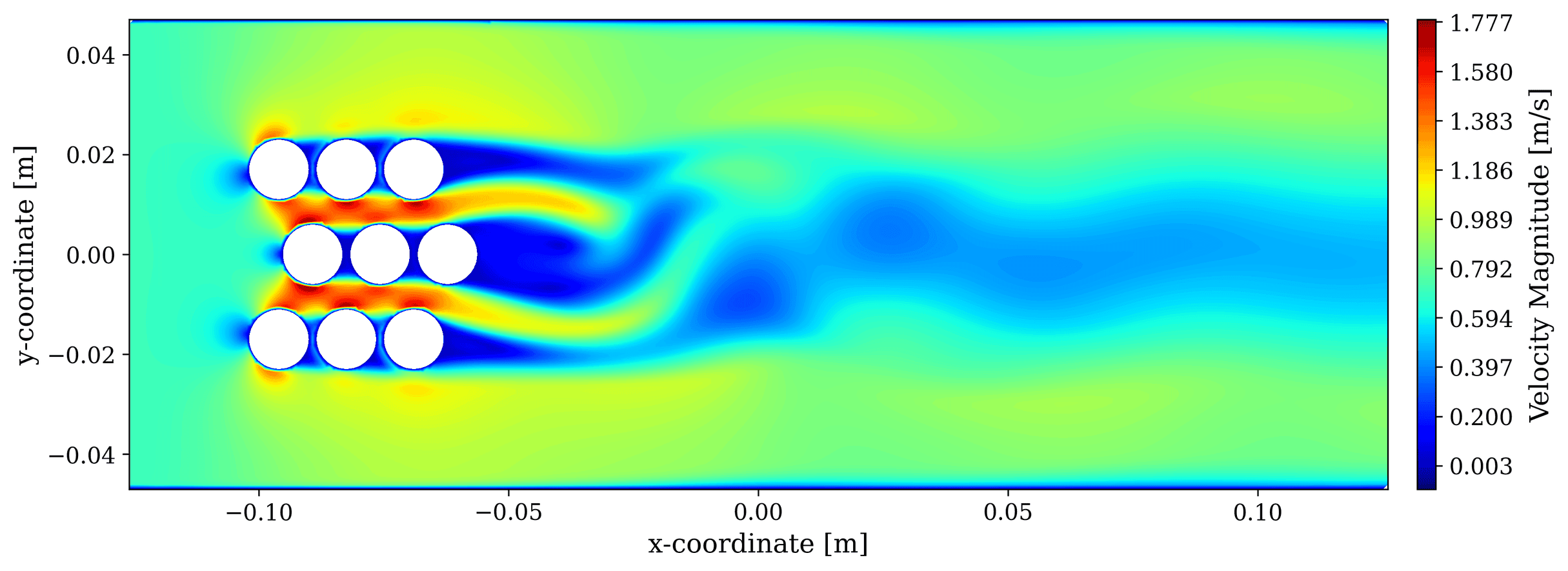} \\[2pt]
        \adjustbox{valign=c}{\rotatebox[origin=c]{90}{\small\textbf{Predicted}}} &
        \includegraphics[width=0.45\textwidth,valign=c]{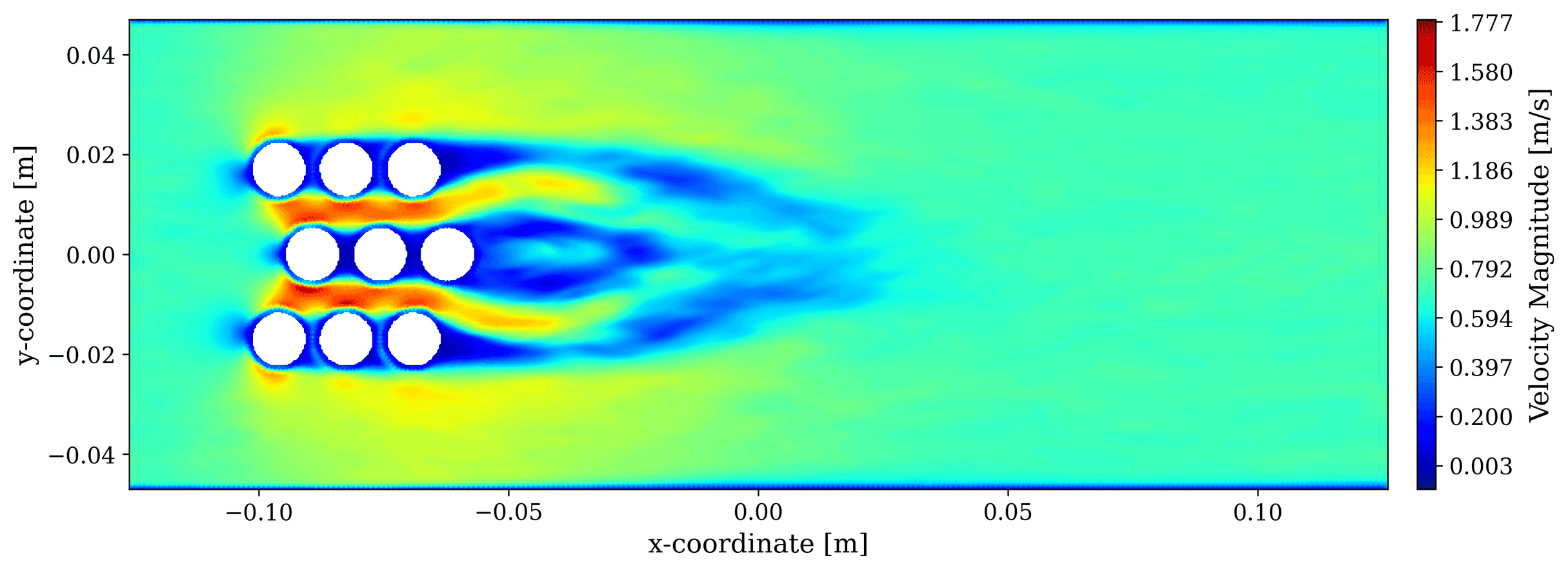} &
        \includegraphics[width=0.45\textwidth,valign=c]{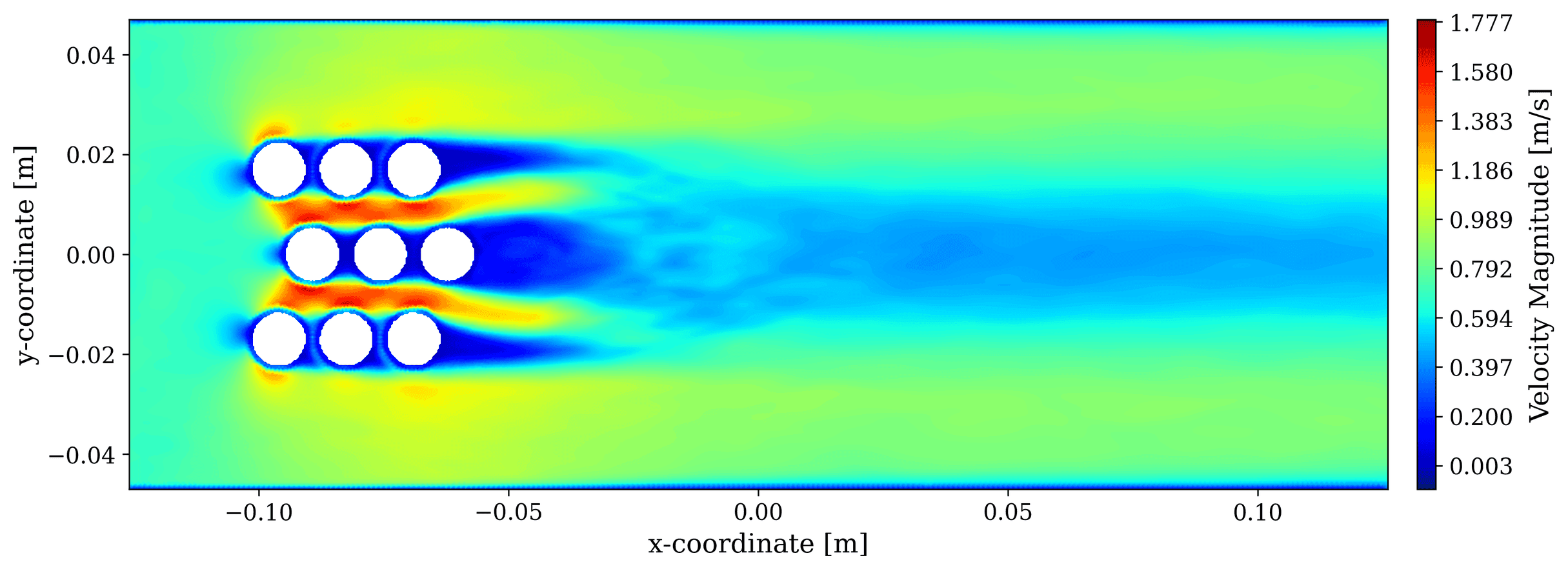} &
        \includegraphics[width=0.45\textwidth,valign=c]{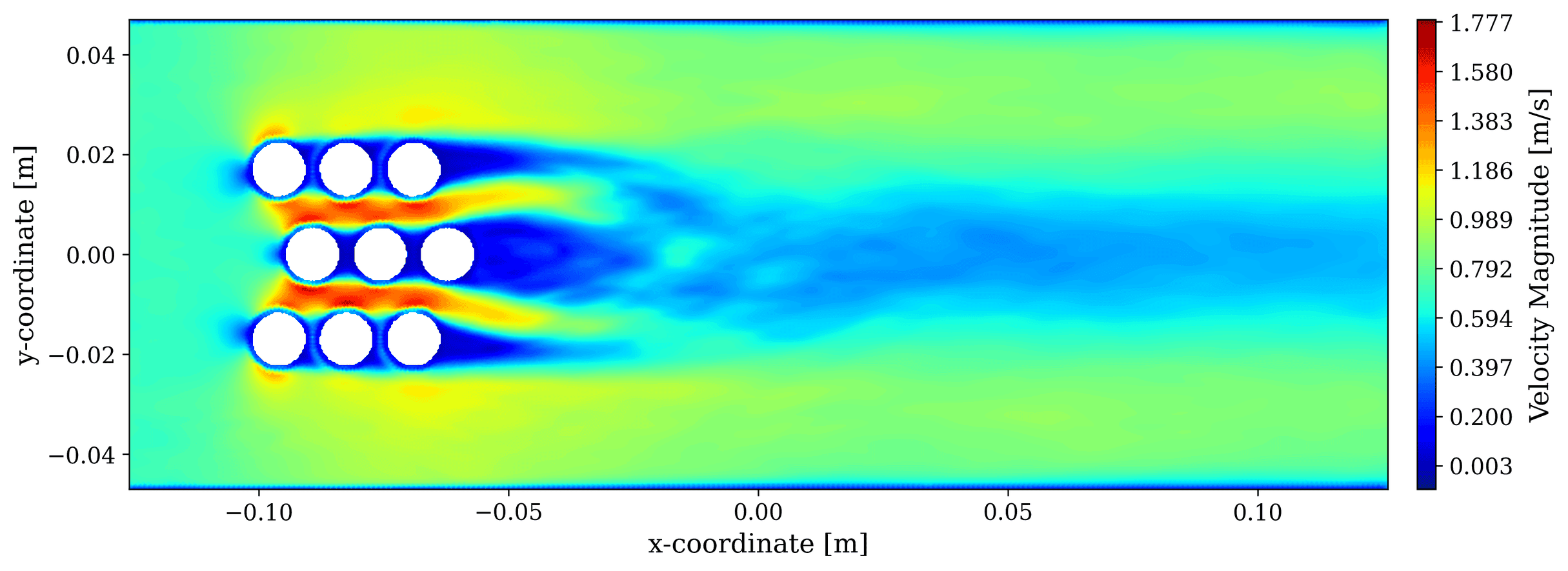} \\[2pt]
        \adjustbox{valign=c}{\rotatebox[origin=c]{90}{\small\textbf{Error}}} &
        \includegraphics[width=0.45\textwidth,valign=c]{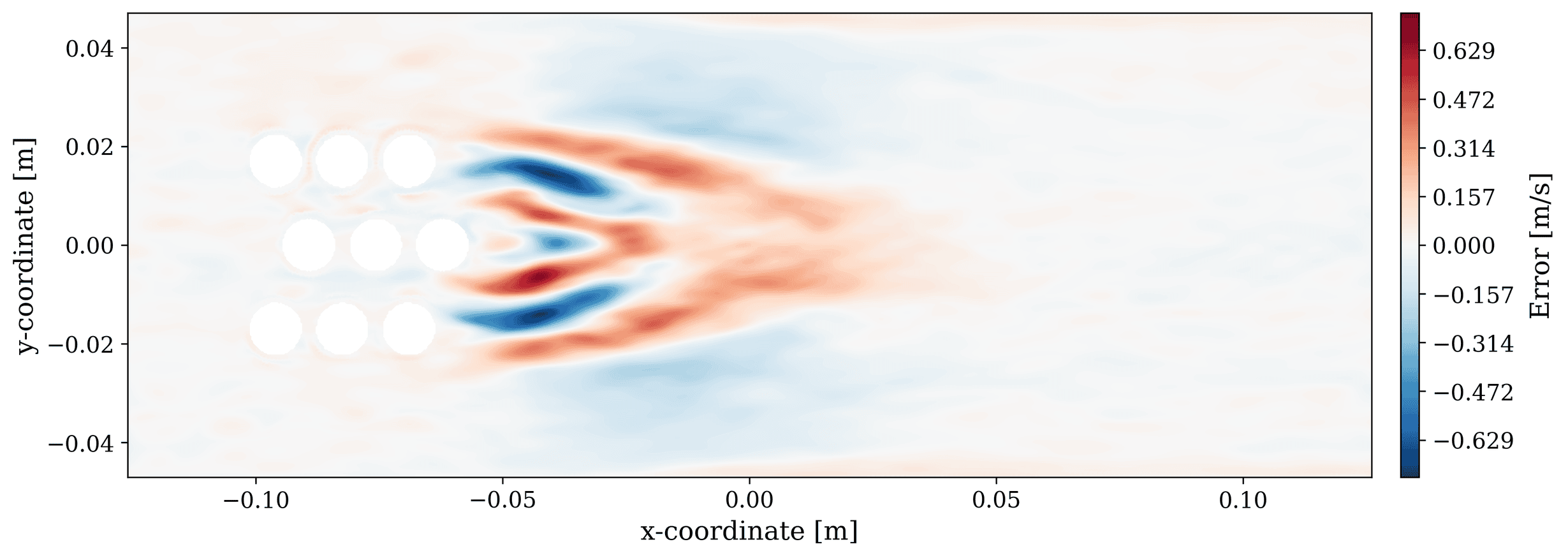} &
        \includegraphics[width=0.45\textwidth,valign=c]{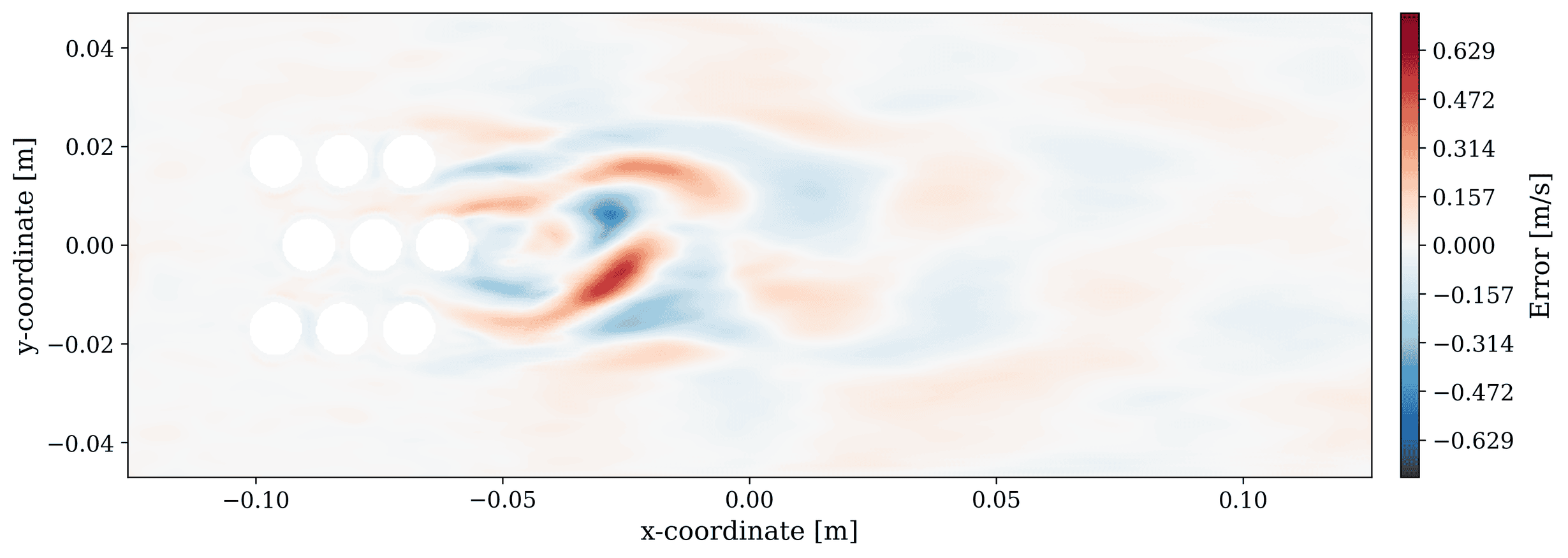} &
        \includegraphics[width=0.45\textwidth,valign=c]{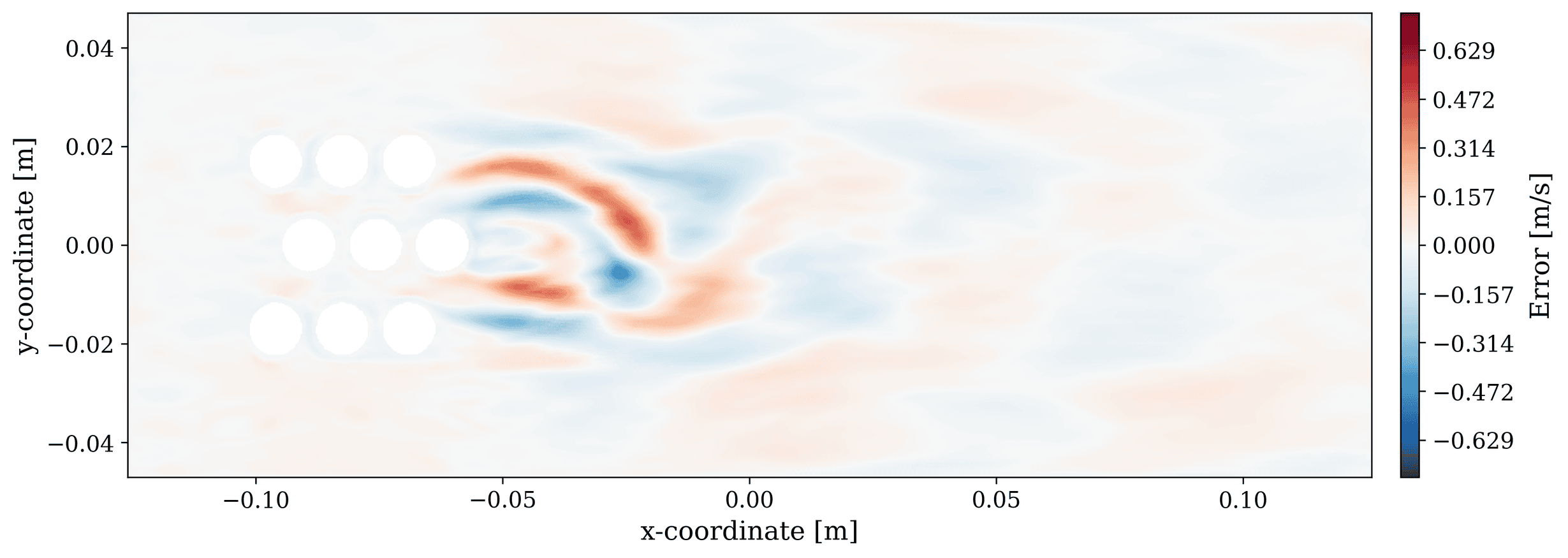} \\
    \end{tabular}
    }%
   \caption{Predicted velocity field by the FNO (24 Fourier modes) at inlet velocity 0.7 m/s. Top: reference CFD solution, middle: predicted field, bottom: absolute error at $t = 2$, $50$, and $100$.}
    \label{fig:fno_velocity_inlet070}
\end{figure}

In contrast, the standard FNO exhibits markedly different prediction characteristics. As shown in Figure~\ref{fig:fno_velocity_inlet040} and Figure~\ref{fig:fno_velocity_inlet070}, the FNO predictions capture the overall magnitude scale and spatial distribution of the velocity field reasonably well, matching the global scale of the flow. However, the predicted fields lack the periodic vortex structures observed in the reference solutions. Instead of reproducing the time-varying K\'{a}rm\'{a}n vortex streets, the FNO predicts a smoothed, time-averaged mean flow field. This behavior is consistent with the spectral bias analysis presented in Section~\ref{subsec5.2}, where increasing the number of Fourier modes alone was shown to be insufficient for capturing high-frequency oscillatory patterns.

\begin{figure}[H]
    \centering
    \setlength{\tabcolsep}{1pt}
    \makebox[\textwidth][c]{%
    \begin{tabular}{c@{\hspace{4pt}}ccc}
        & \textbf{$t = 2$} & \textbf{$t = 50$} & \textbf{$t = 100$} \\
        \adjustbox{valign=c}{\rotatebox[origin=c]{90}{\small\textbf{Reference}}} &
        \includegraphics[width=0.45\textwidth,valign=c]{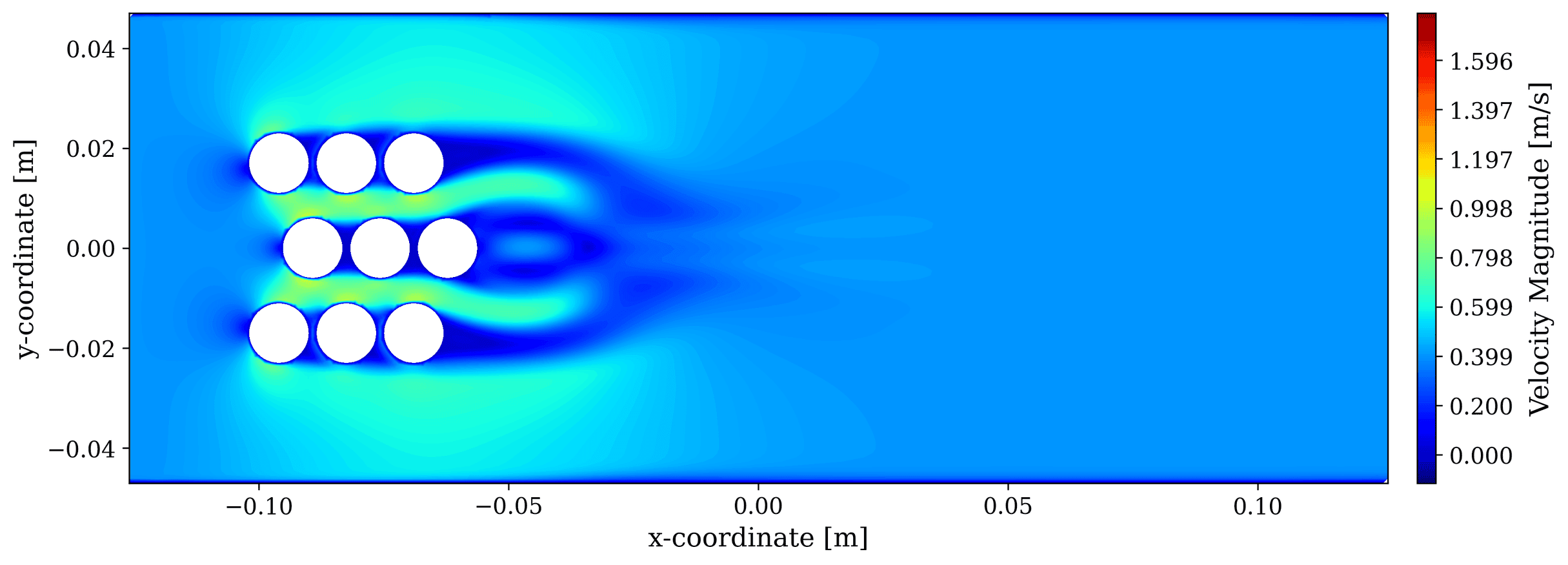} &
        \includegraphics[width=0.45\textwidth,valign=c]{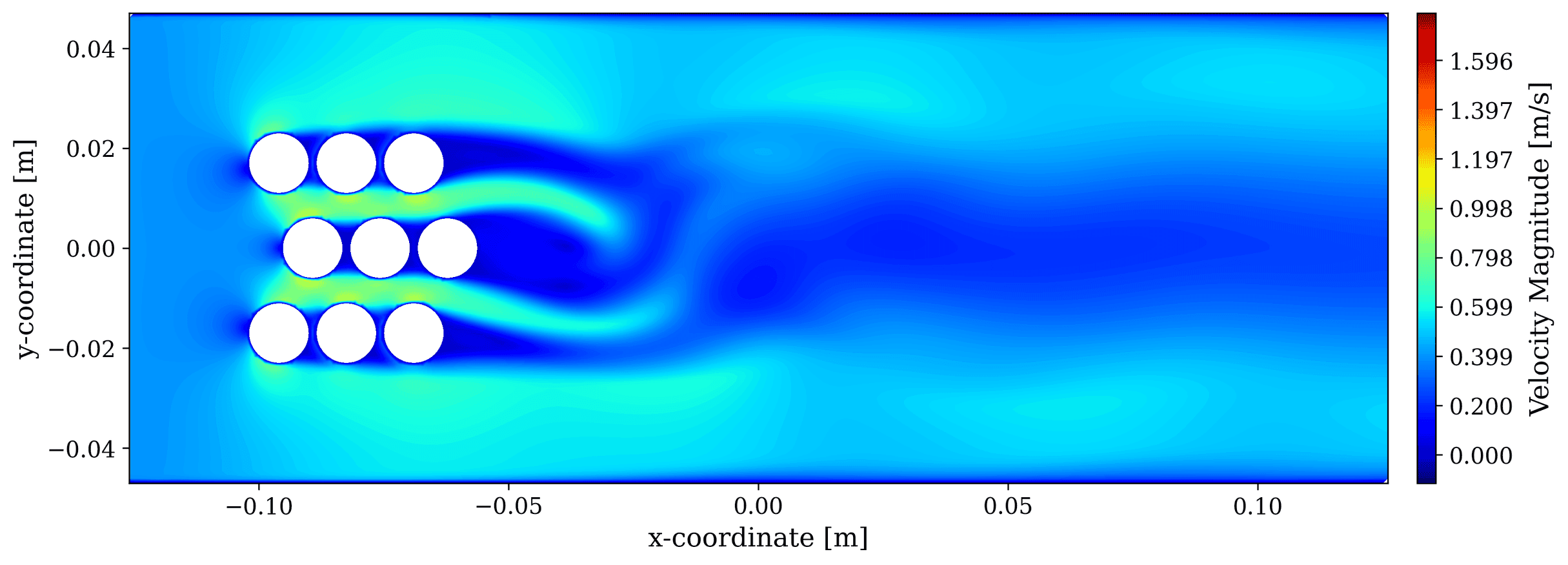} &
        \includegraphics[width=0.45\textwidth,valign=c]{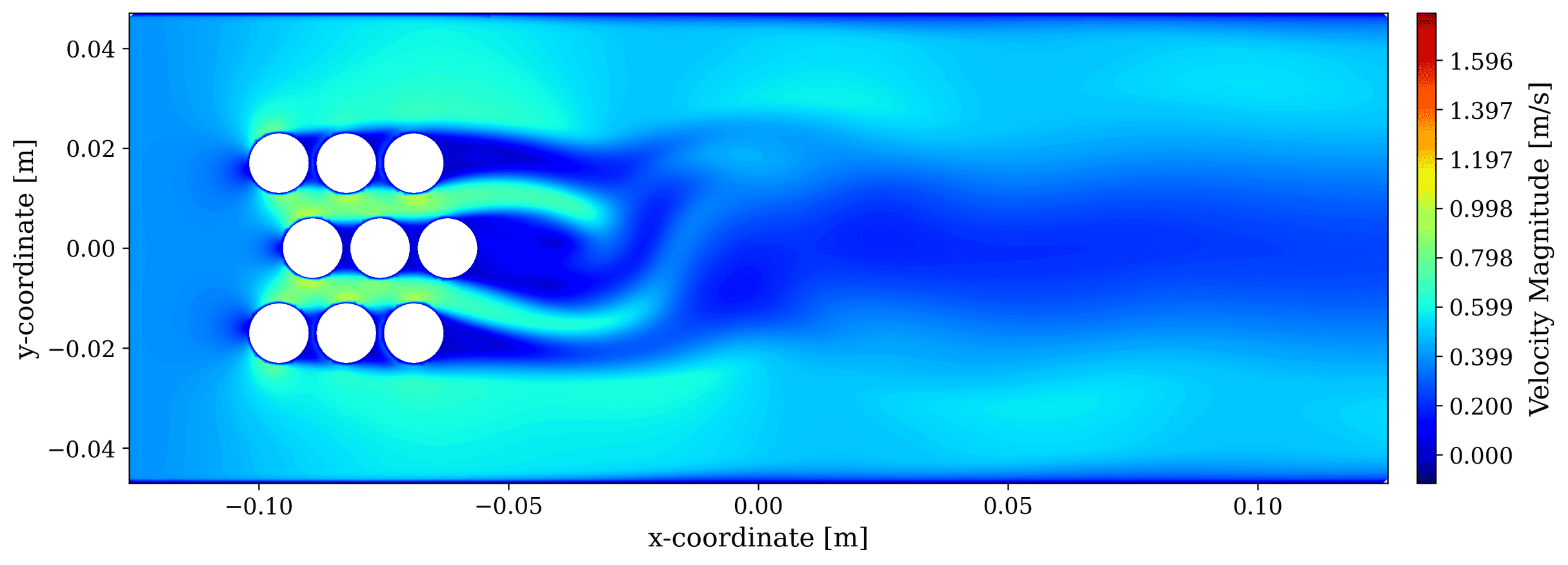} \\[2pt]
        \adjustbox{valign=c}{\rotatebox[origin=c]{90}{\small\textbf{Predicted}}} &
        \includegraphics[width=0.45\textwidth,valign=c]{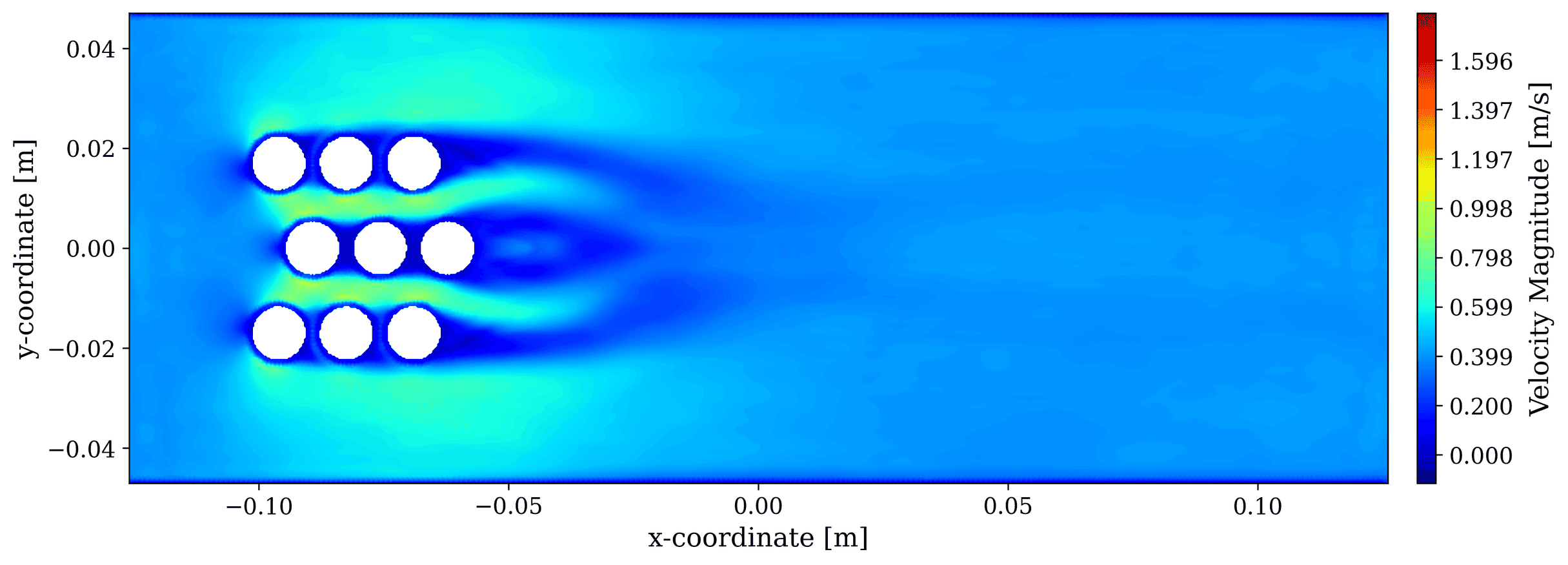} &
        \includegraphics[width=0.45\textwidth,valign=c]{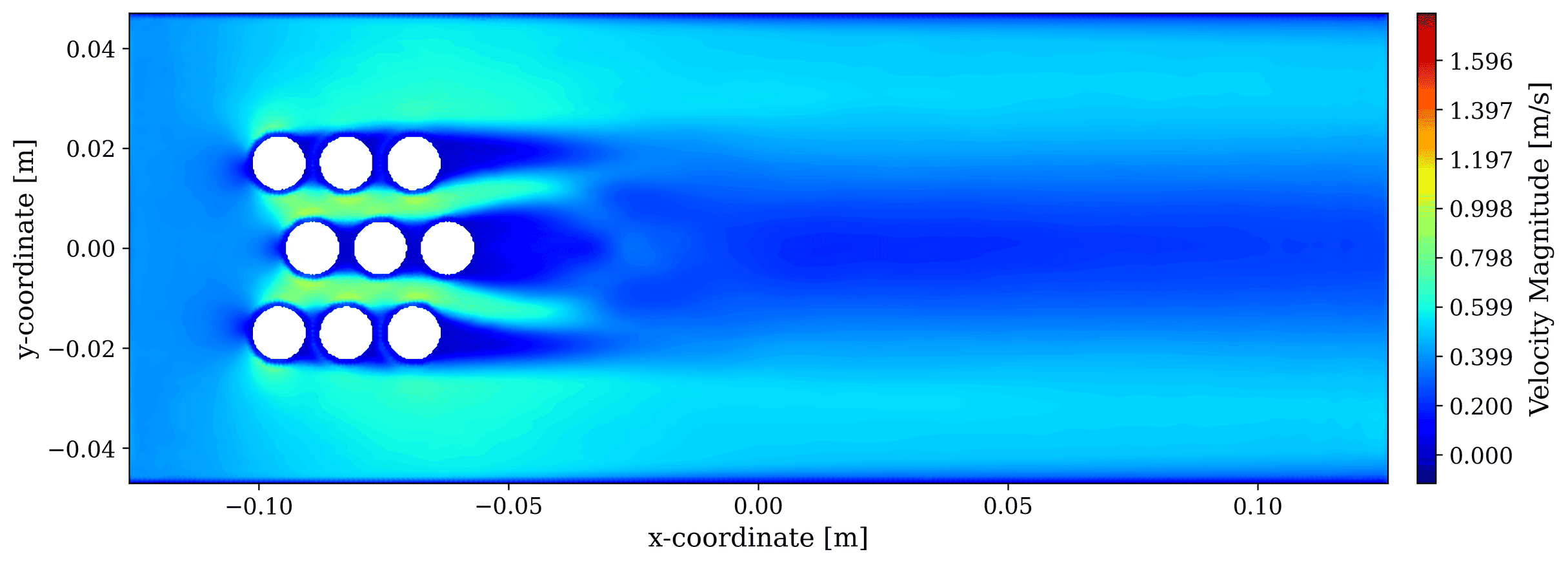} &
        \includegraphics[width=0.45\textwidth,valign=c]{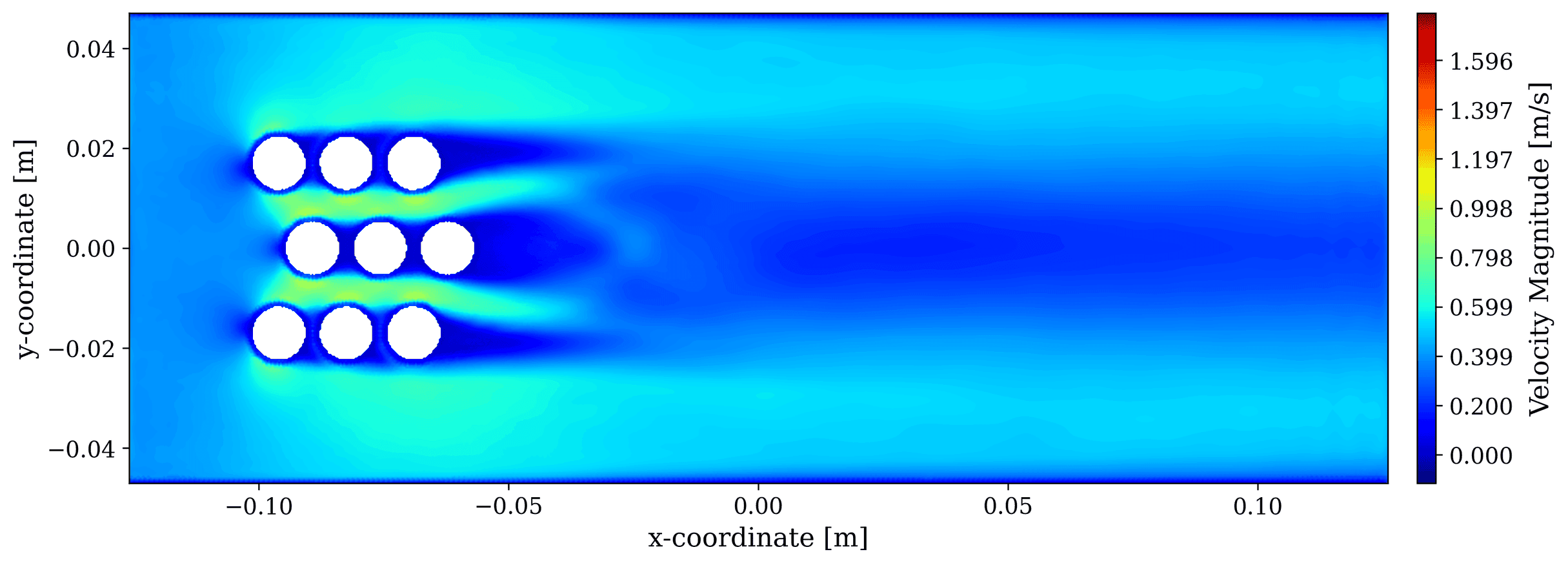} \\[2pt]
        \adjustbox{valign=c}{\rotatebox[origin=c]{90}{\small\textbf{Error}}} &
        \includegraphics[width=0.45\textwidth,valign=c]{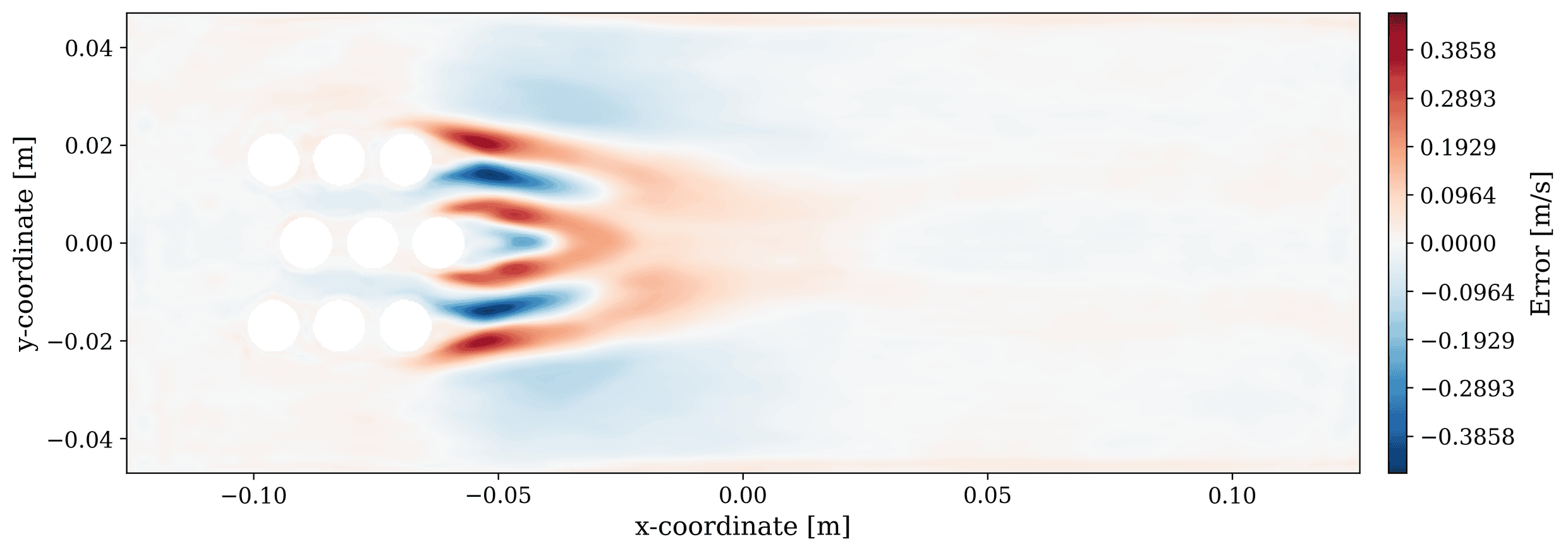} &
        \includegraphics[width=0.45\textwidth,valign=c]{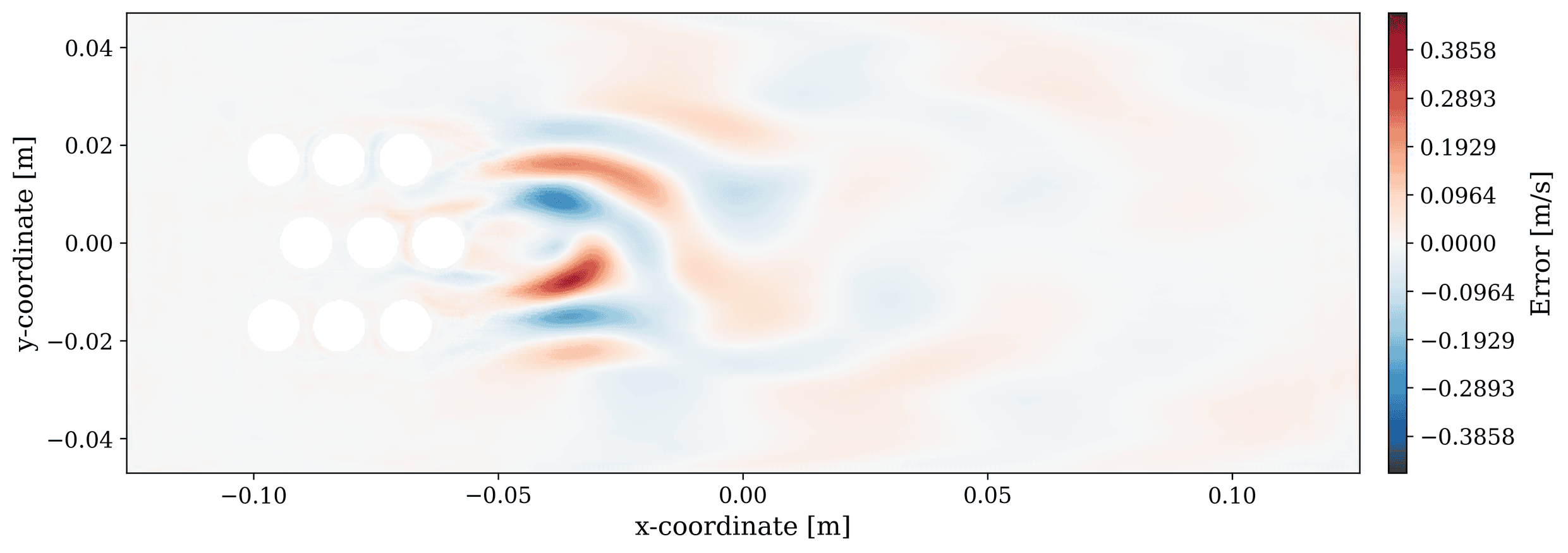} &
        \includegraphics[width=0.45\textwidth,valign=c]{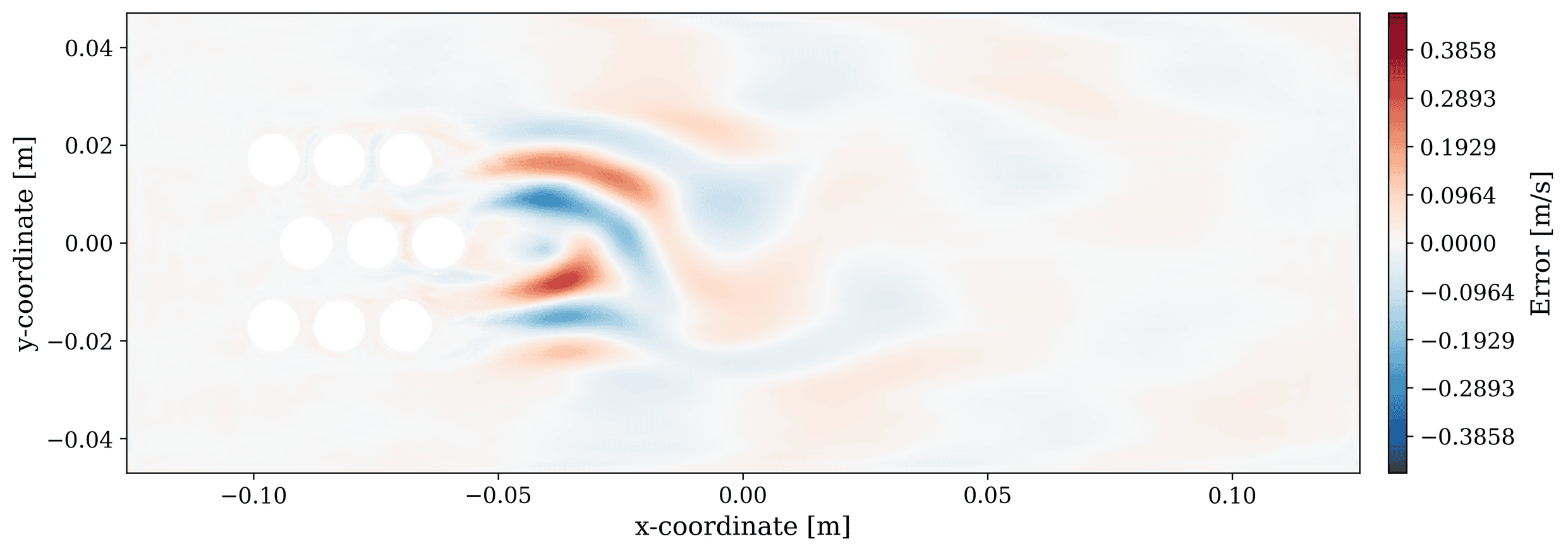} \\
    \end{tabular}
    }%
   \caption{Predicted velocity field by the MscaleFNO at inlet velocity 0.4 m/s. Top: reference CFD solution, middle: predicted field, bottom: absolute error at $t = 2$, $50$, and $100$.}
    \label{fig:mscalefno_velocity_inlet040}
\end{figure}

\begin{figure}[H]
    \centering
    \setlength{\tabcolsep}{1pt}
    \makebox[\textwidth][c]{%
    \begin{tabular}{c@{\hspace{4pt}}ccc}
        & \textbf{$t = 2$} & \textbf{$t = 50$} & \textbf{$t = 100$} \\
        \adjustbox{valign=c}{\rotatebox[origin=c]{90}{\small\textbf{Reference}}} &
        \includegraphics[width=0.45\textwidth,valign=c]{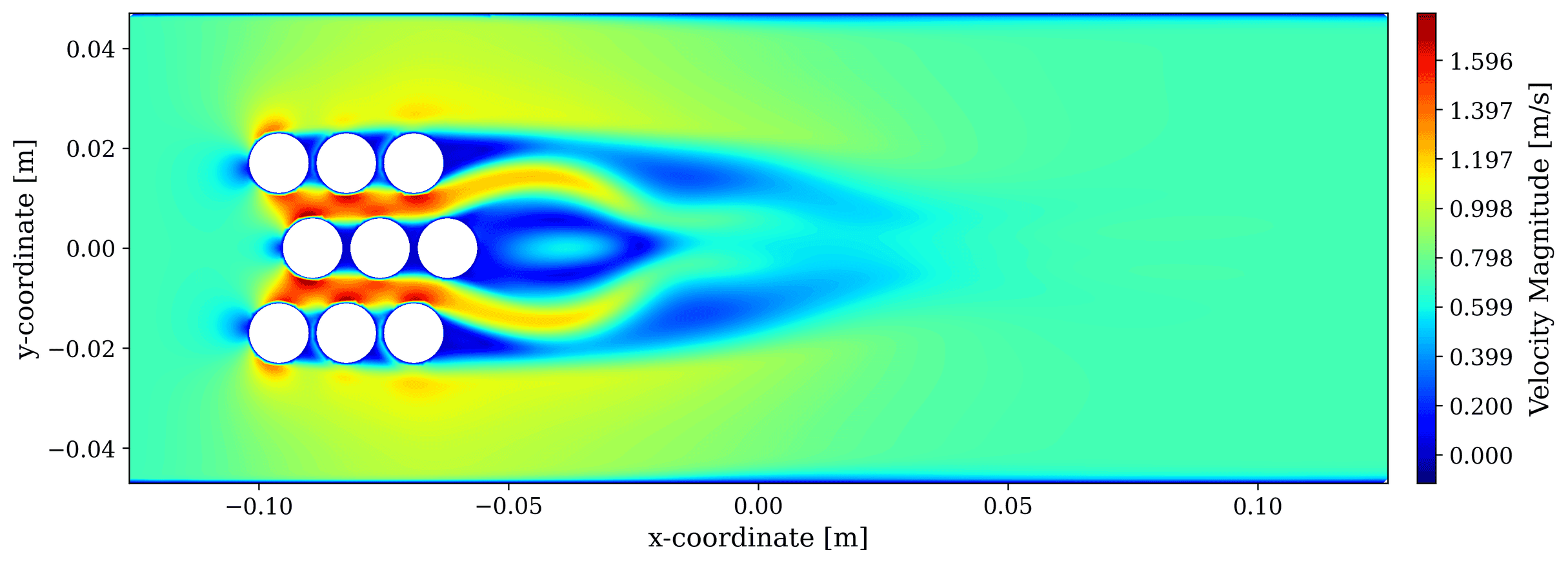} &
        \includegraphics[width=0.45\textwidth,valign=c]{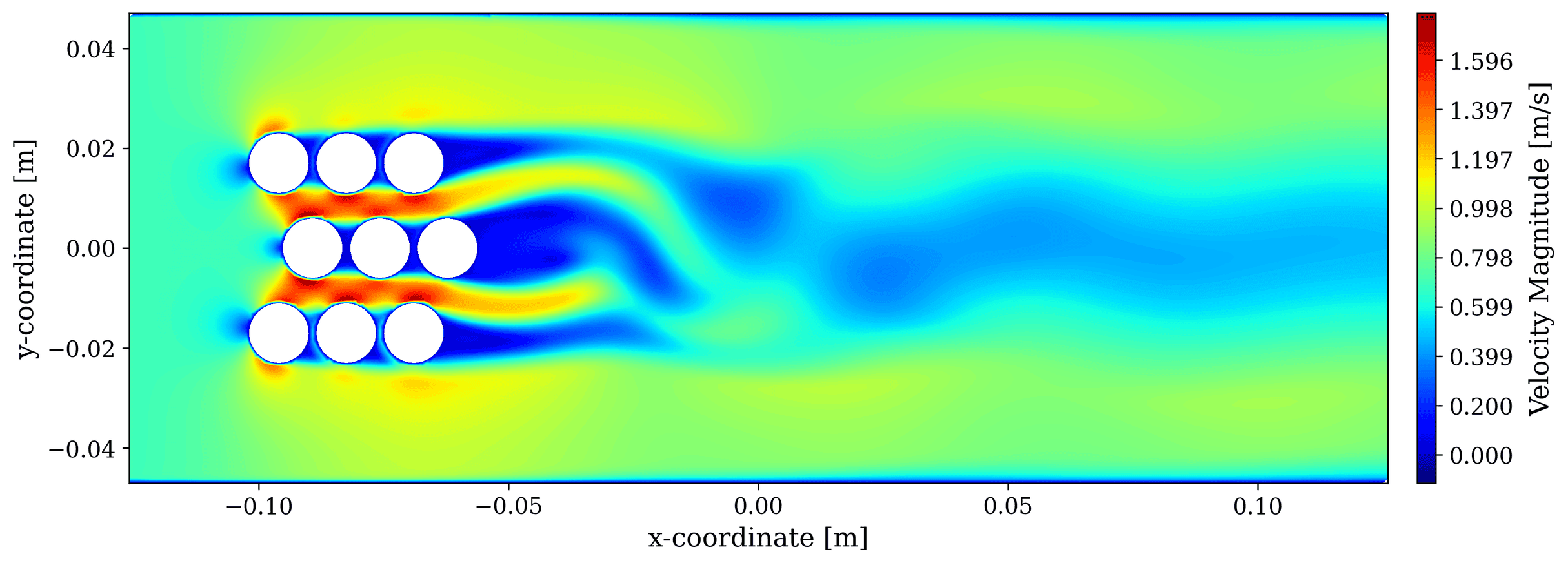} &
        \includegraphics[width=0.45\textwidth,valign=c]{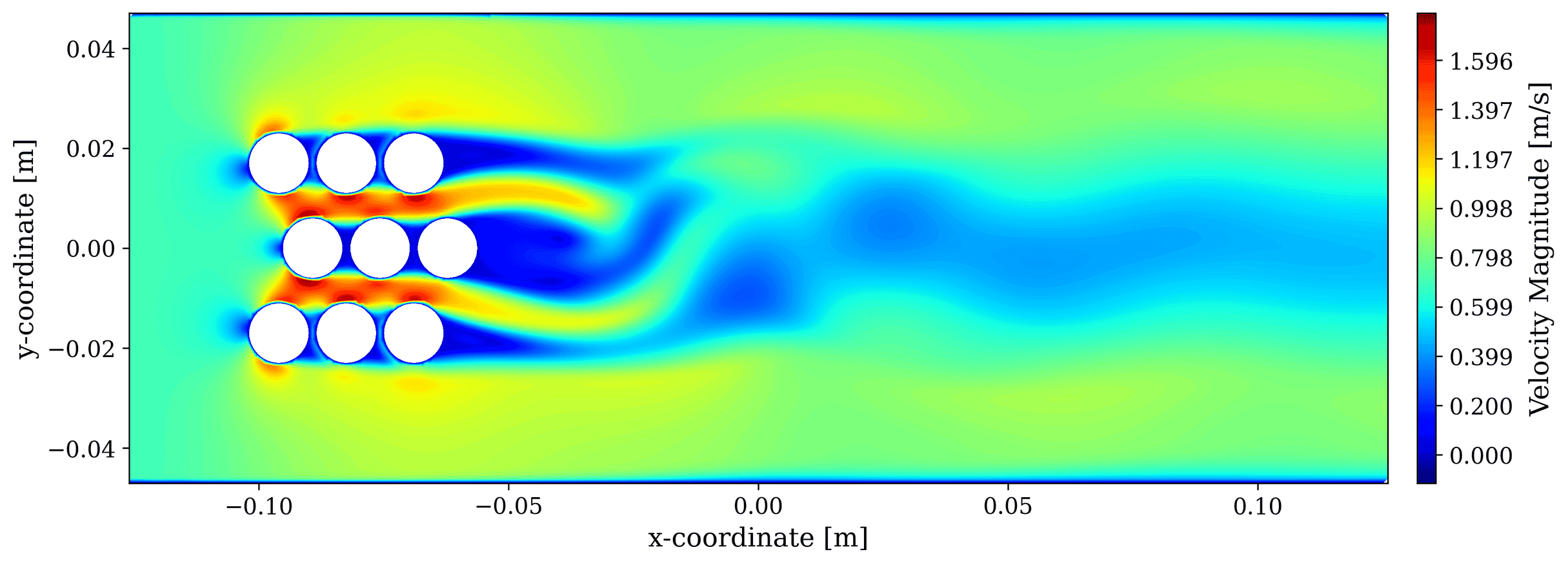} \\[2pt]
        \adjustbox{valign=c}{\rotatebox[origin=c]{90}{\small\textbf{Predicted}}} &
        \includegraphics[width=0.45\textwidth,valign=c]{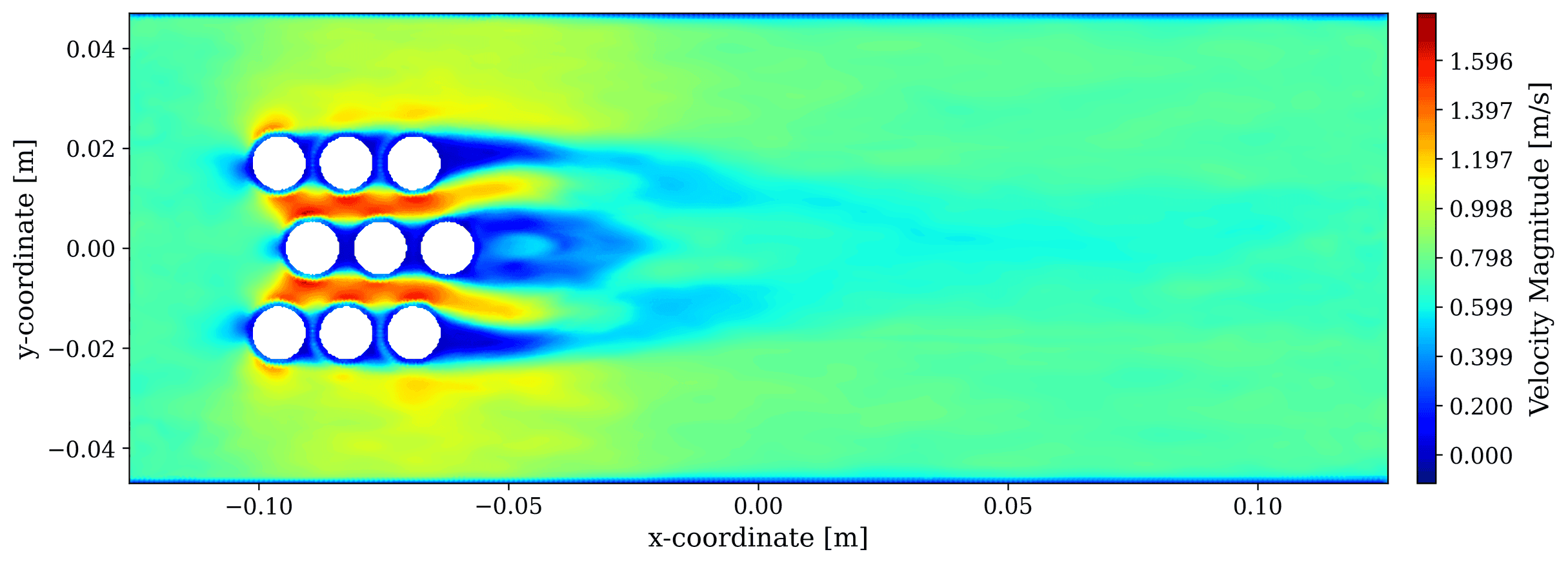} &
        \includegraphics[width=0.45\textwidth,valign=c]{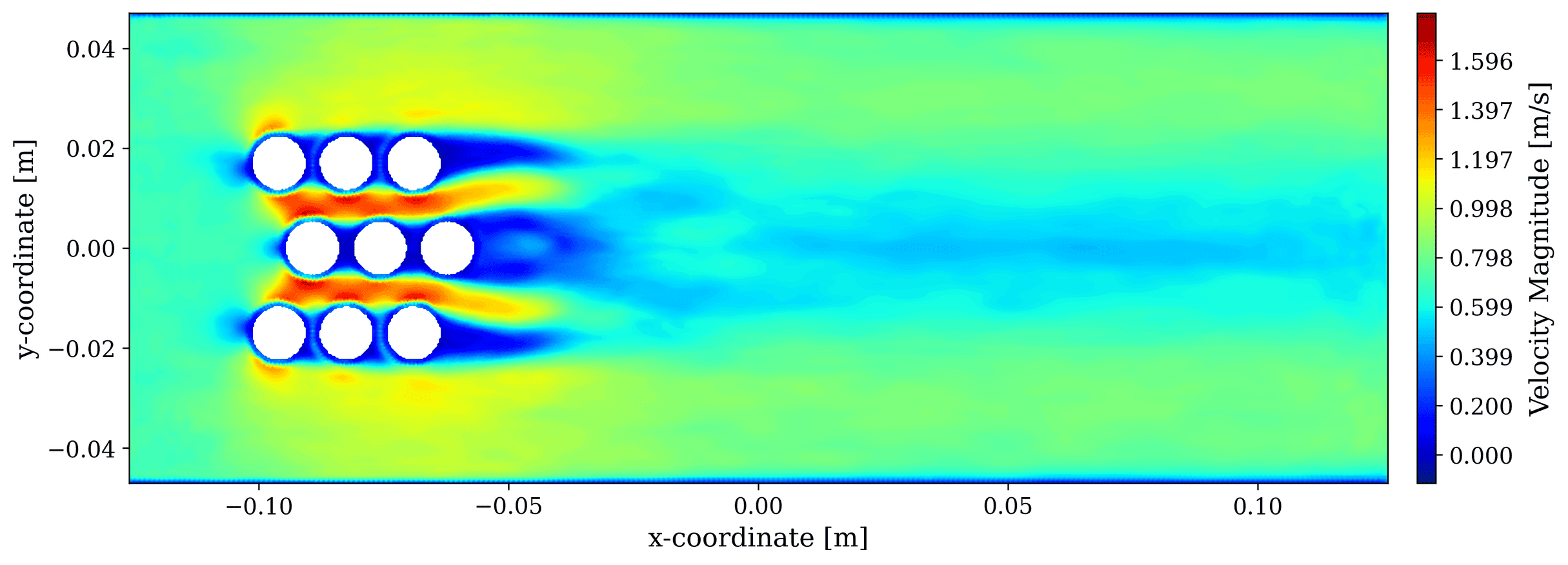} &
        \includegraphics[width=0.45\textwidth,valign=c]{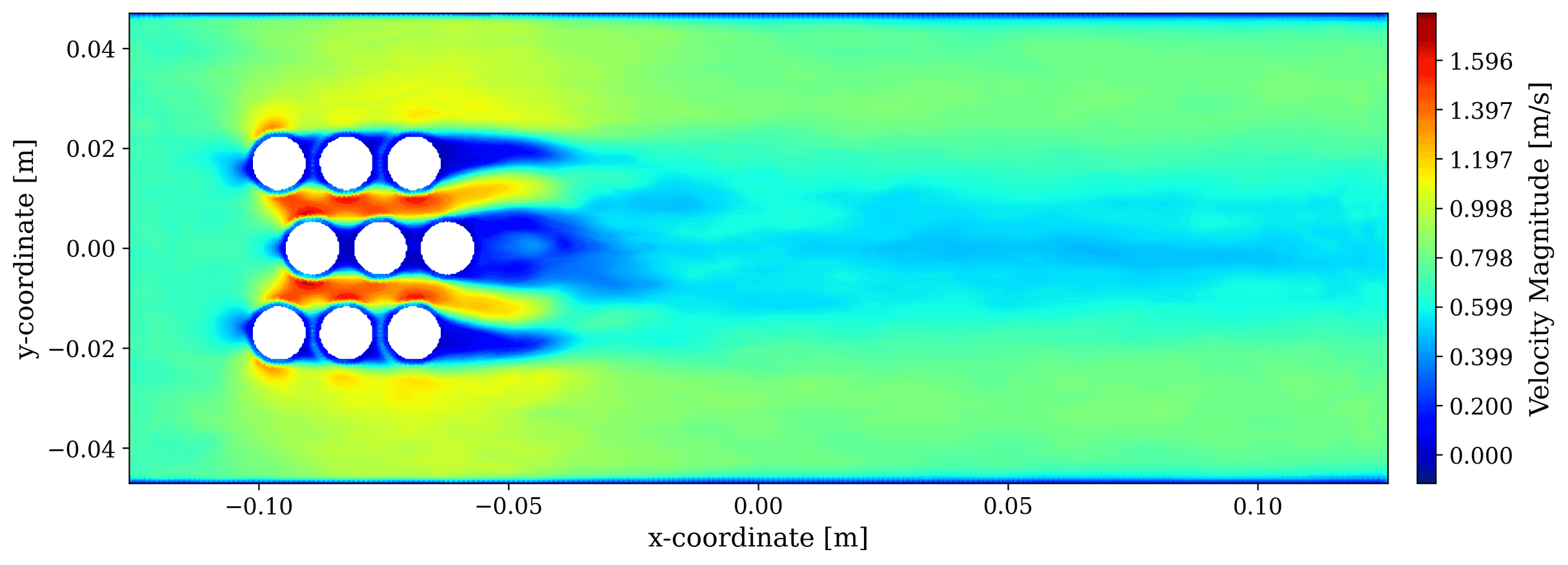} \\[2pt]
        \adjustbox{valign=c}{\rotatebox[origin=c]{90}{\small\textbf{Error}}} &
        \includegraphics[width=0.45\textwidth,valign=c]{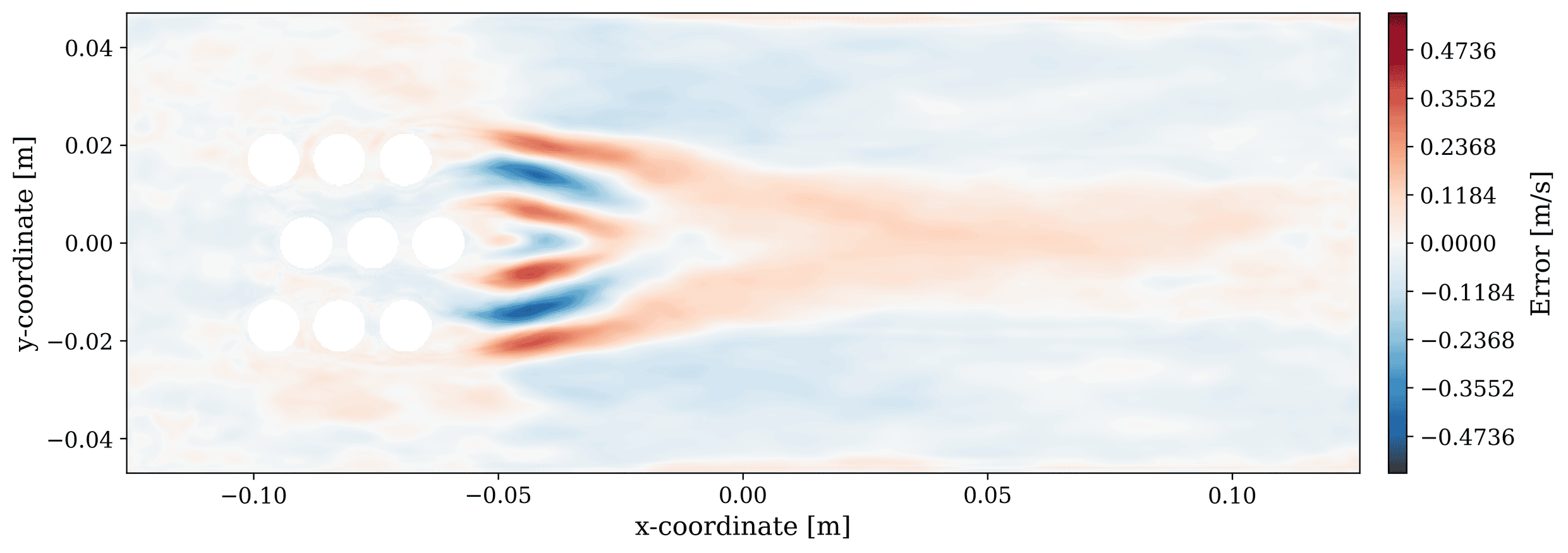} &
        \includegraphics[width=0.45\textwidth,valign=c]{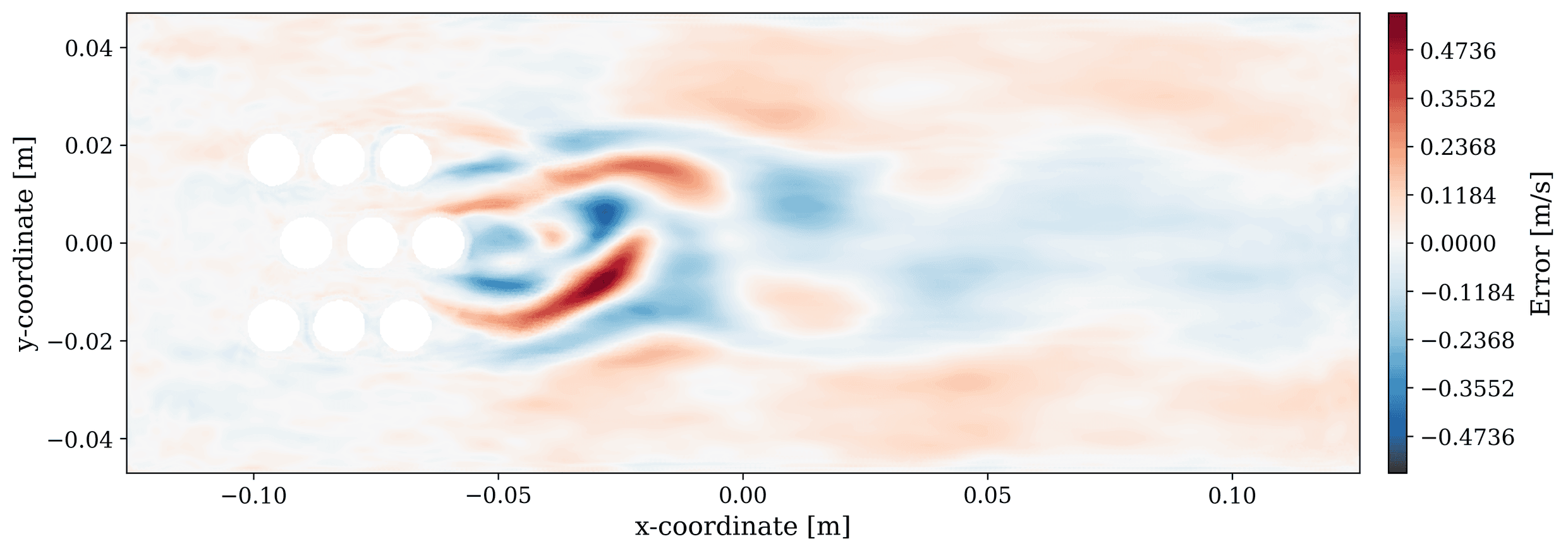} &
        \includegraphics[width=0.45\textwidth,valign=c]{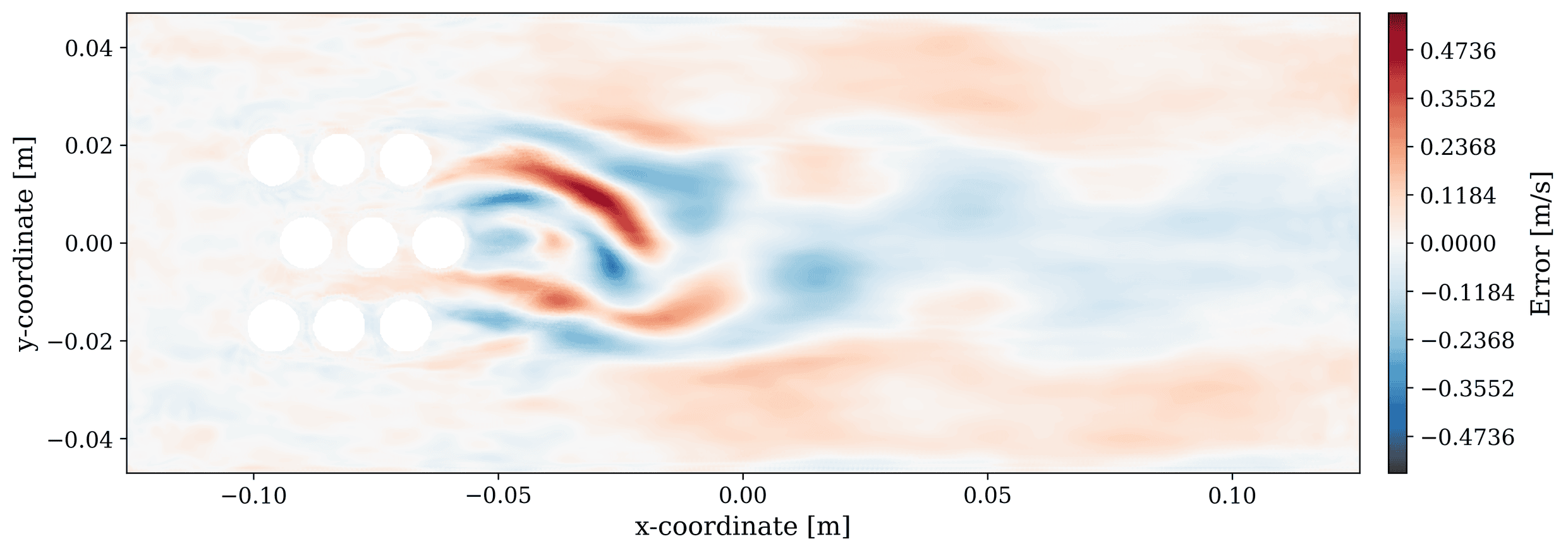} \\
    \end{tabular}
    }%
    \caption{Predicted velocity field by the MscaleFNO at inlet velocity 0.7 m/s. Top: reference CFD solution, middle: predicted field, bottom: absolute error at $t = 2$, $50$, and $100$.}
    \label{fig:mscalefno_velocity_inlet070}
\end{figure}

Figure~\ref{fig:mscalefno_velocity_inlet040} and Figure~\ref{fig:mscalefno_velocity_inlet070} show the prediction results of the MscaleFNO for inlet velocities of 0.4 m/s and 0.7 m/s, respectively. Despite incorporating the multi-scale methodology to address spectral bias, the MscaleFNO still fails to reproduce the periodic vortex structures. Similar to the standard FNO, the MscaleFNO predictions capture the overall velocity magnitude and the global flow pattern, but the predicted fields correspond to a mean flow field rather than the instantaneous oscillatory flow. This suggests that, for the K\'{a}rm\'{a}n vortex flow within the HCSG, the spectral bias in the FNO framework cannot be fully resolved by input scaling alone, and the periodic flow structures arising from inherent flow instability pose a fundamentally different challenge compared to the oscillatory functions studied in previous MscaleFNO applications.

Overall, the multi-scale L-DeepONet, with both MLP-based AE and CAE, successfully captures the periodic K\'{a}rm\'{a}n vortex dynamics within the HCSG. By contrast, the FNO-based models, including the MscaleFNO, predict only the time-averaged mean flow and fail to reproduce the periodic vortex streets, even though they accurately capture the global velocity scale.

\subsection{Quantitative Comparison and Pressure Drop Analysis}
\label{subsec6.3}

To complement the qualitative flow field comparison, the prediction accuracy of all four models was quantitatively evaluated using the relative $L^2$ error across all time steps and the velocity time histories at representative probe locations shown in Figure~\ref{fig:unstructured_probe_locations}.

\begin{figure}[H]
    \centering
    \begin{subfigure}[t]{0.48\textwidth}
        \centering
        \includegraphics[width=\textwidth]{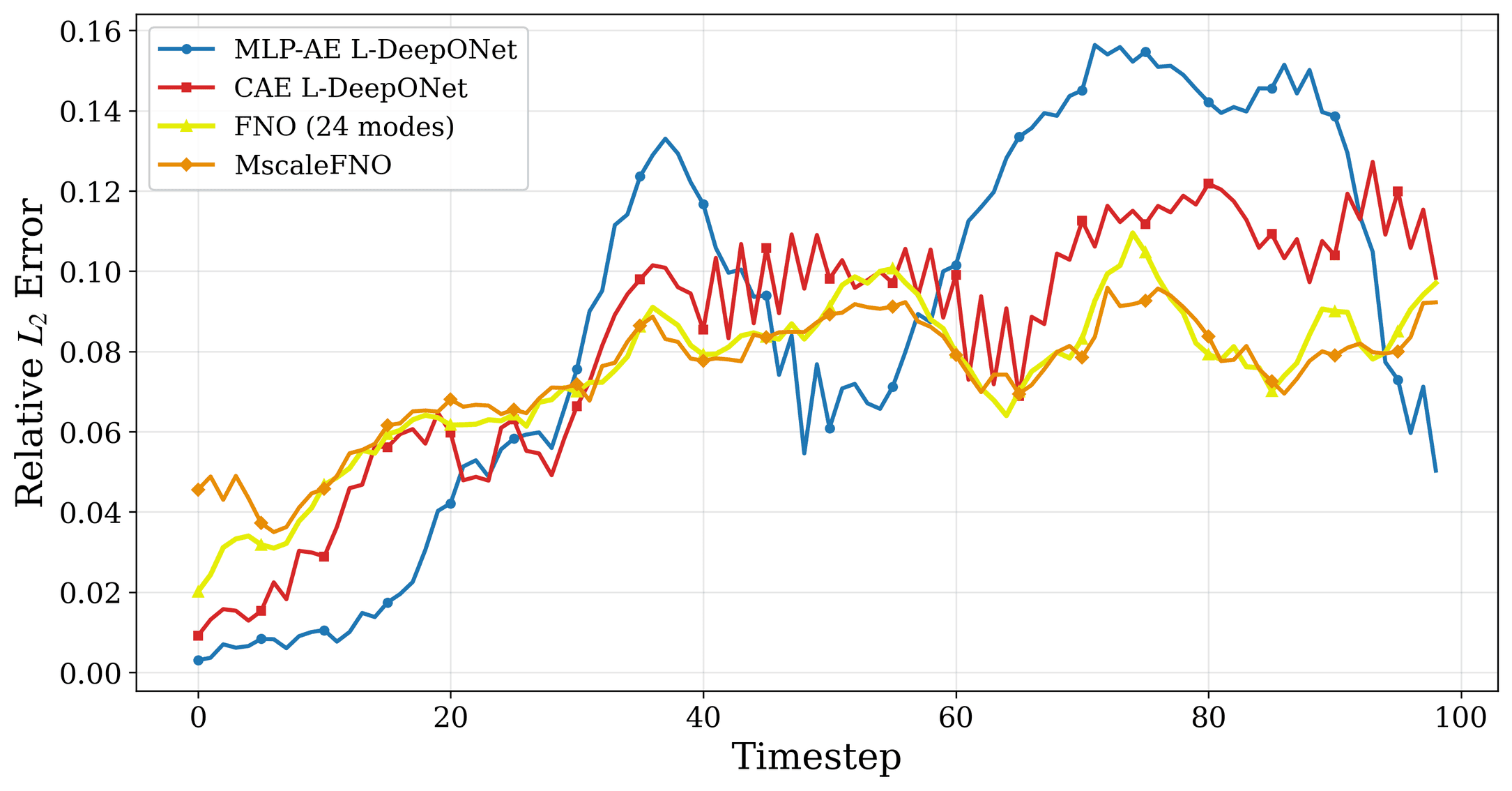}
        \caption{Velocity, inlet 0.4 m/s}
        \label{fig:rel_l2_err_velocity_040}
    \end{subfigure}
    \hfill
    \begin{subfigure}[t]{0.48\textwidth}
        \centering
        \includegraphics[width=\textwidth]{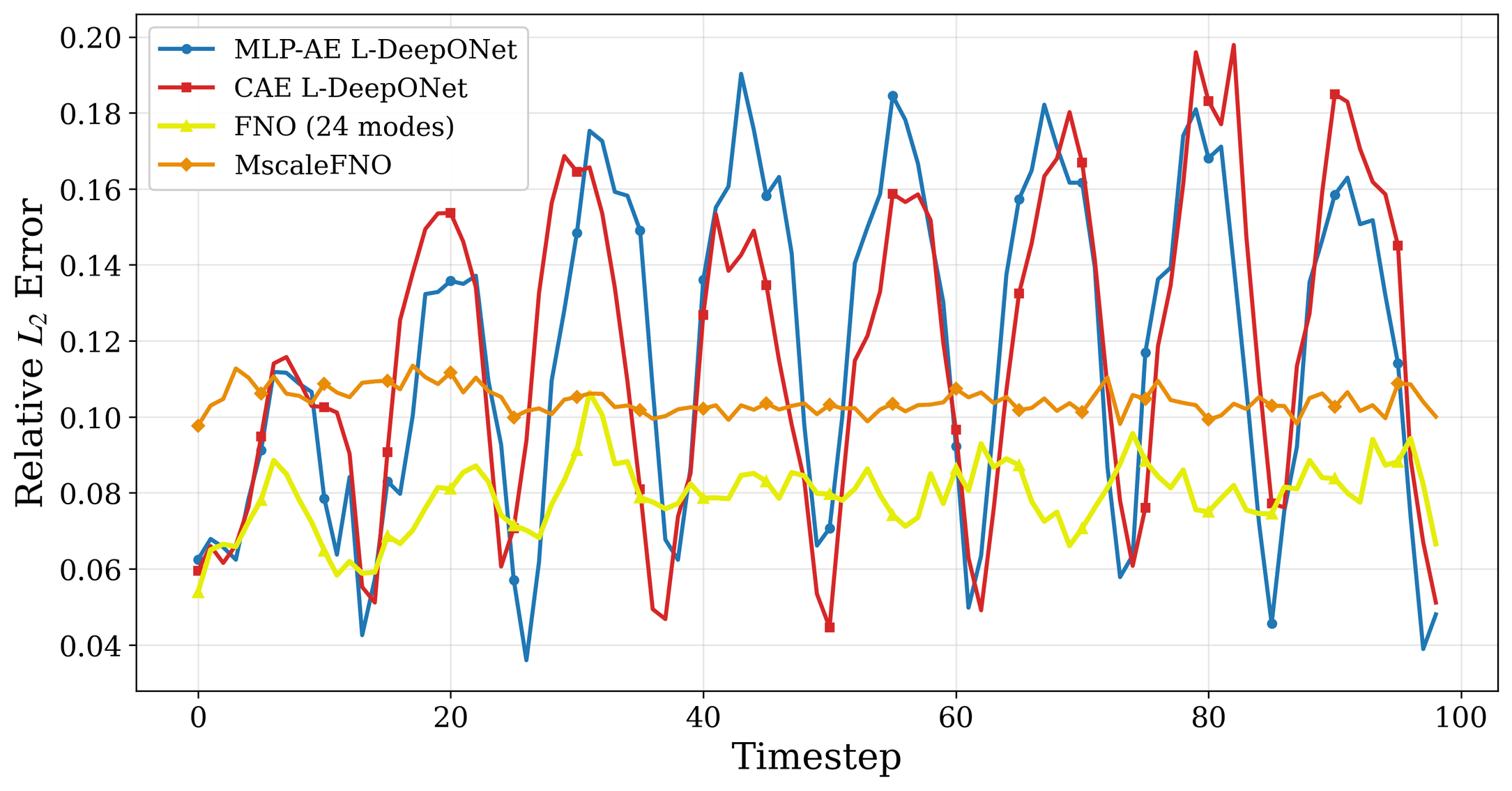}
        \caption{Velocity, inlet 0.7 m/s}
        \label{fig:rel_l2_err_velocity_070}
    \end{subfigure}

    \vspace{1em}

    \begin{subfigure}[t]{0.48\textwidth}
        \centering
        \includegraphics[width=\textwidth]{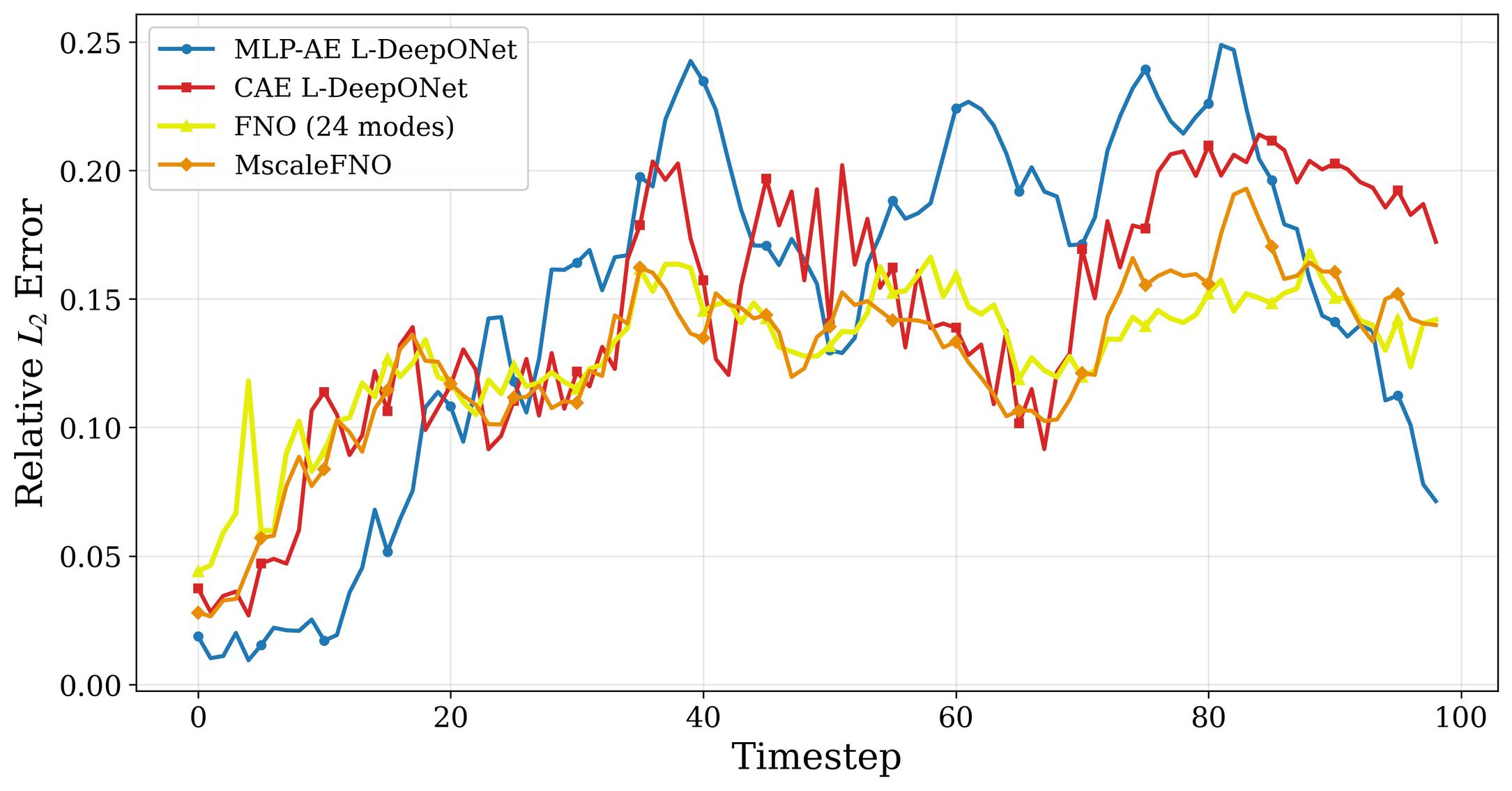}
        \caption{Pressure, inlet 0.4 m/s}
        \label{fig:rel_l2_err_pressure_040}
    \end{subfigure}
    \hfill
    \begin{subfigure}[t]{0.48\textwidth}
        \centering
        \includegraphics[width=\textwidth]{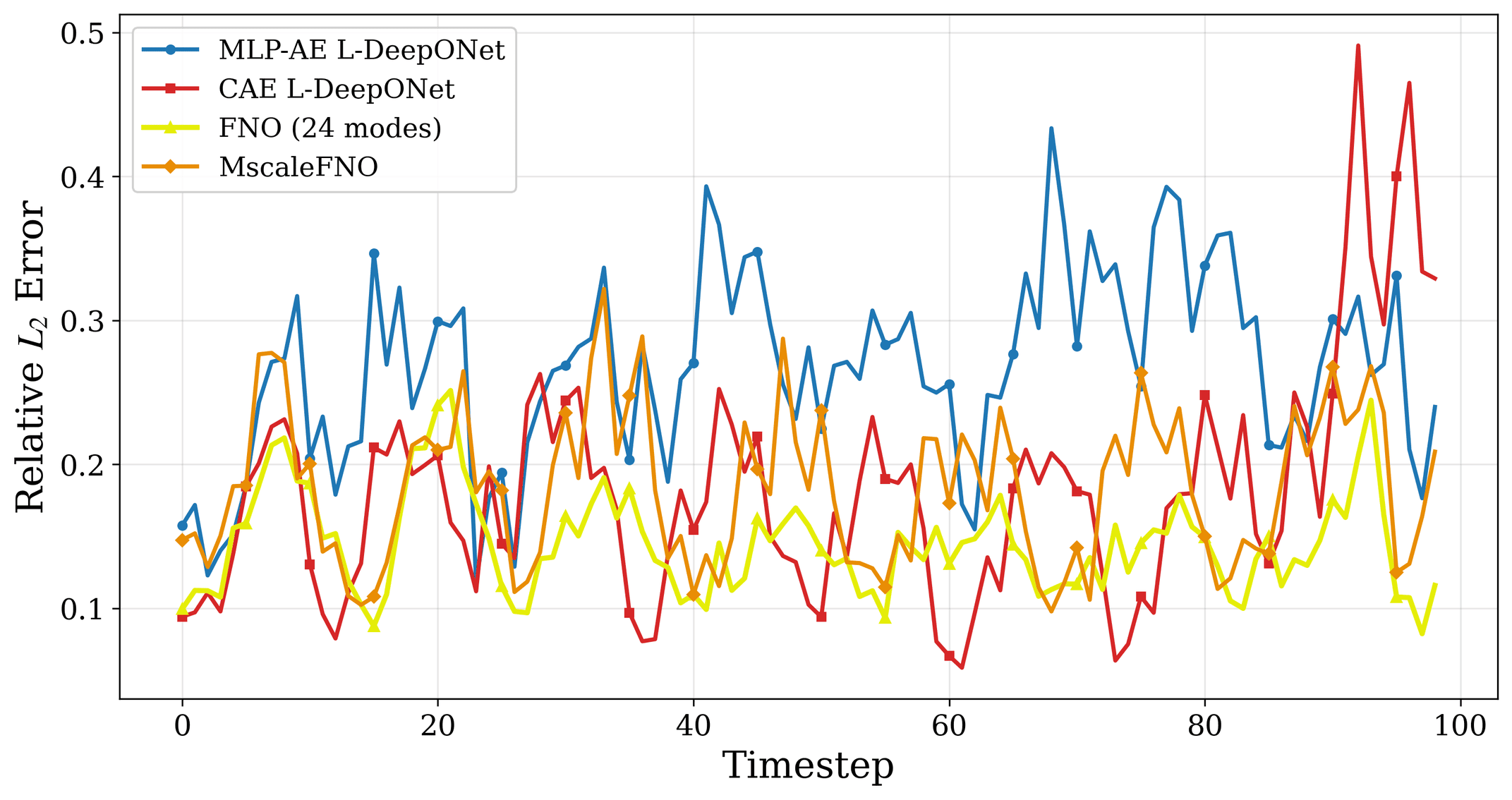}
        \caption{Pressure, inlet 0.7 m/s}
        \label{fig:rel_l2_err_pressure_070}
    \end{subfigure}
    \caption{Time evolution of relative $L^2$ errors for velocity magnitude (a, b) and pressure (c, d) at inlet velocities of 0.4 m/s and 0.7 m/s.}
    \label{fig:relative_l2_errors}
\end{figure}

Figure~\ref{fig:relative_l2_errors} presents the time evolution of the relative $L^2$ error for both velocity and pressure fields across all four models. For the velocity field (Figure~\ref{fig:rel_l2_err_velocity_040} and Figure~\ref{fig:rel_l2_err_velocity_070}), the L-DeepONet models exhibit oscillatory error patterns that reflect the periodic nature of the predicted flow. In contrast, the FNO-based models show relatively flat and lower error curves despite their inability to capture the K\'{a}rm\'{a}n vortex streets. This paradoxical result can be explained by the fact that the FNO-based models predict only the time-averaged mean flow, which accounts for the dominant component of the velocity field. Since the relative $L^2$ error is computed over the entire spatial domain, accurately predicting this mean component alone yields a low error value, while the absence of oscillatory fluctuations does not significantly increase the global metric. The L-DeepONet models, by attempting to reproduce the instantaneous periodic variations, inevitably introduce additional error in the phase and amplitude of the vortex structures, resulting in higher but oscillatory error values. For the pressure field (Figure~\ref{fig:rel_l2_err_pressure_040} and Figure~\ref{fig:rel_l2_err_pressure_070}), similar trends are observed. The FNO-based models again show flat error curves, while the L-DeepONet models exhibit larger fluctuations in the pressure error, particularly at the higher inlet velocity of 0.7 m/s.

Table~\ref{tab:rl2e_summary} summarizes the mean relative $L^2$ error across all time steps for both velocity and pressure fields. As discussed above, the FNO-based models achieve the lowest mean error values for both fields, while the L-DeepONet variants exhibit slightly higher values due to the additional error introduced by predicting the instantaneous periodic variations.

\begin{table}[H]
\centering
\caption{Mean relative $L^2$ error for velocity and pressure fields.}
\label{tab:rl2e_summary}
\renewcommand{\arraystretch}{1.2}
\begin{tabular}{llcc}
\toprule
\textbf{Field} & \textbf{Model} & \textbf{Inlet 0.4 m/s} & \textbf{Inlet 0.7 m/s} \\
\midrule
\multirow{4}{*}{Velocity}
& MLP-AE L-DeepONet & 0.0872 & 0.1172 \\
& CAE L-DeepONet     & 0.0831 & 0.1175 \\
& FNO (24 modes)     & 0.0745 & 0.0791 \\
& MscaleFNO          & 0.0743 & 0.1043 \\
\midrule
\multirow{4}{*}{Pressure}
& MLP-AE L-DeepONet & 0.1484 & 0.2669 \\
& CAE L-DeepONet     & 0.1450 & 0.1819 \\
& FNO (24 modes)     & 0.1301 & 0.1448 \\
& MscaleFNO          & 0.1271 & 0.1837 \\
\bottomrule
\end{tabular}
\end{table}

\begin{figure}[H]
    \centering
    \makebox[\textwidth][c]{%
    \begin{minipage}{1.3\textwidth}
        \centering
        \begin{subfigure}[t]{0.32\textwidth}
            \centering
            \includegraphics[width=\textwidth]{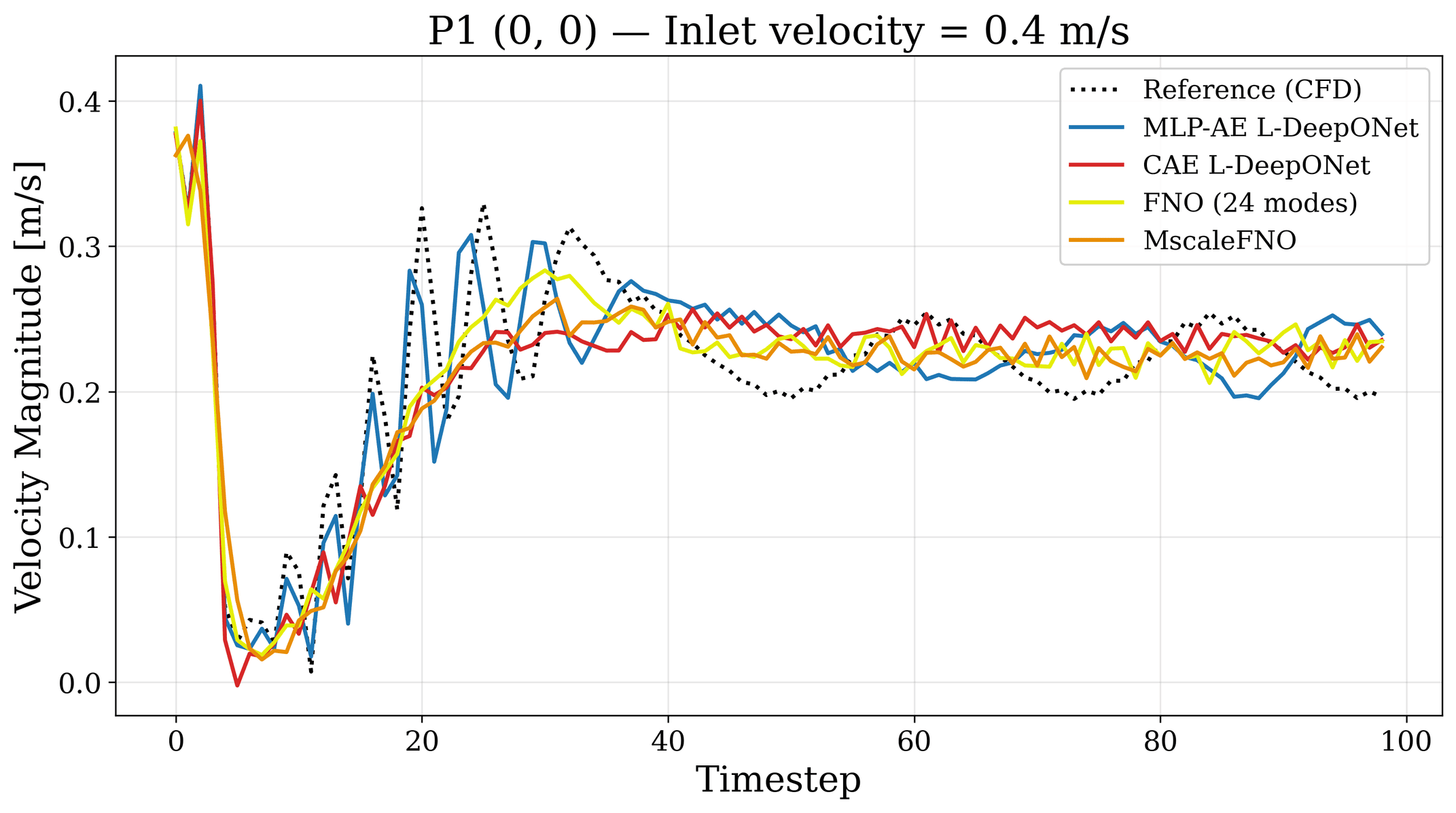}
            \caption{P1 (0, 0)}
        \end{subfigure}
        \hfill
        \begin{subfigure}[t]{0.32\textwidth}
            \centering
            \includegraphics[width=\textwidth]{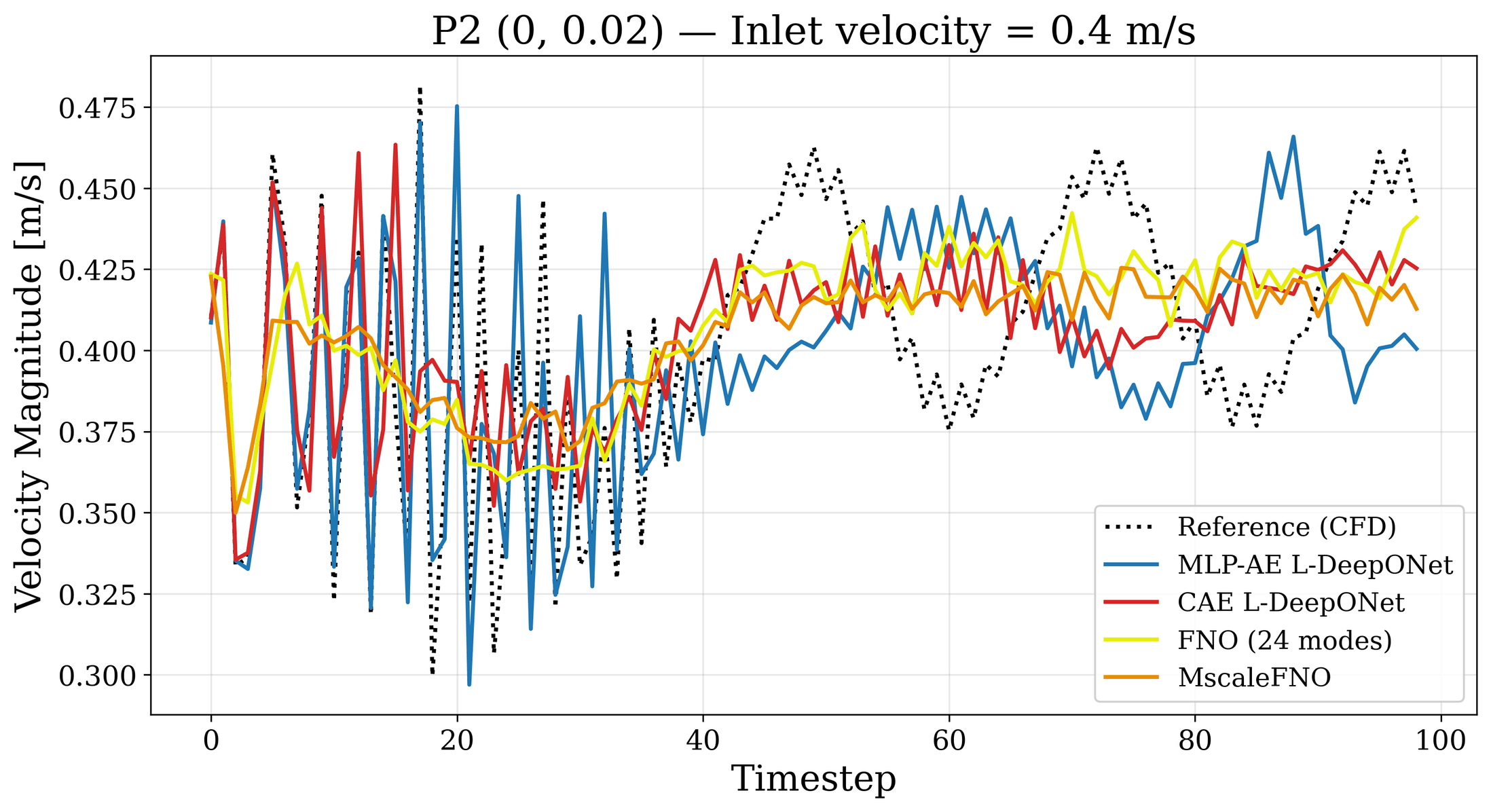}
            \caption{P2 (0, 0.02)}
        \end{subfigure}
        \hfill
        \begin{subfigure}[t]{0.32\textwidth}
            \centering
            \includegraphics[width=\textwidth]{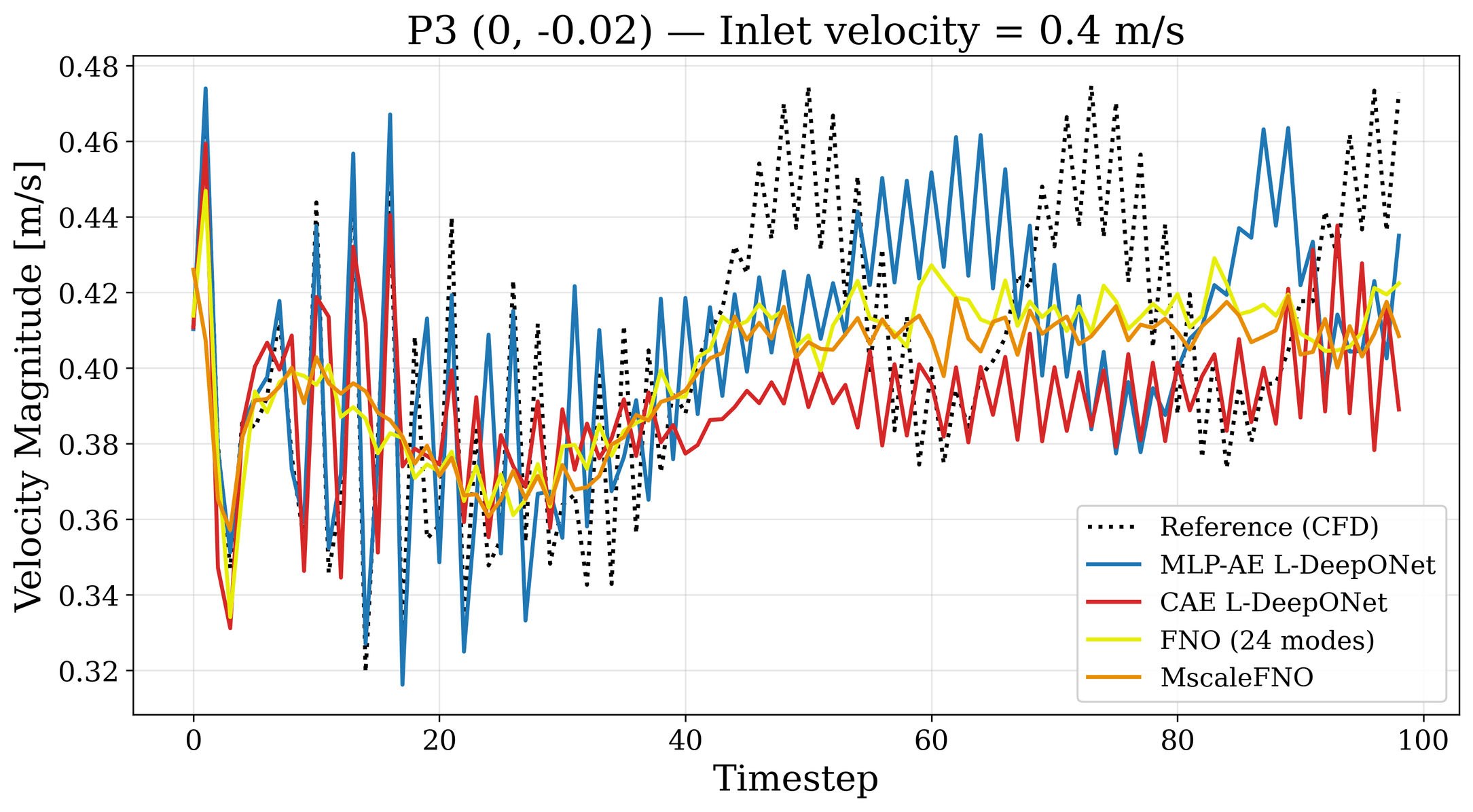}
            \caption{P3 (0, -0.02)}
        \end{subfigure}
        
        \vspace{4pt}
        
        \begin{subfigure}[t]{0.32\textwidth}
            \centering
            \includegraphics[width=\textwidth]{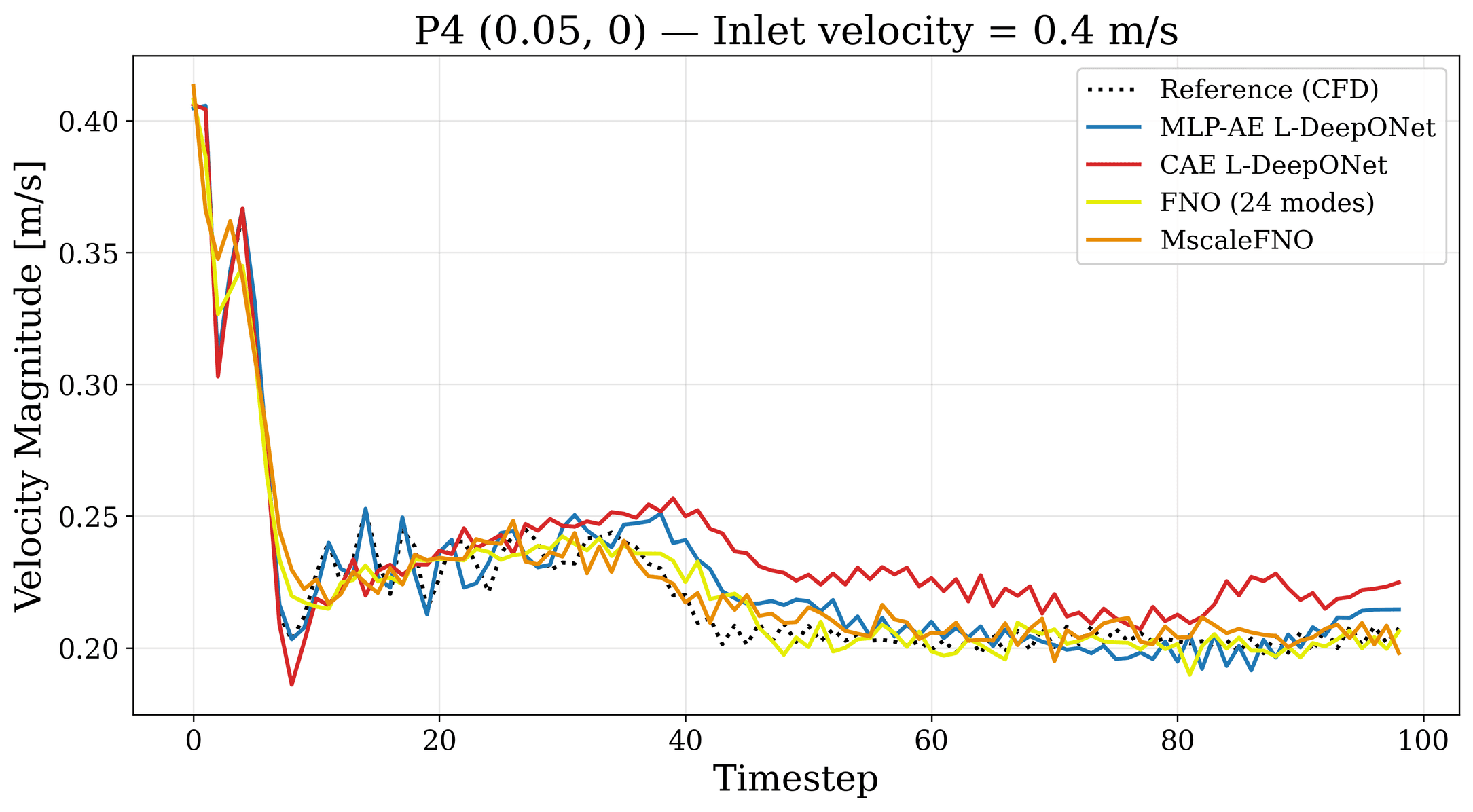}
            \caption{P4 (0.05, 0)}
        \end{subfigure}
        \hfill
        \begin{subfigure}[t]{0.32\textwidth}
            \centering
            \includegraphics[width=\textwidth]{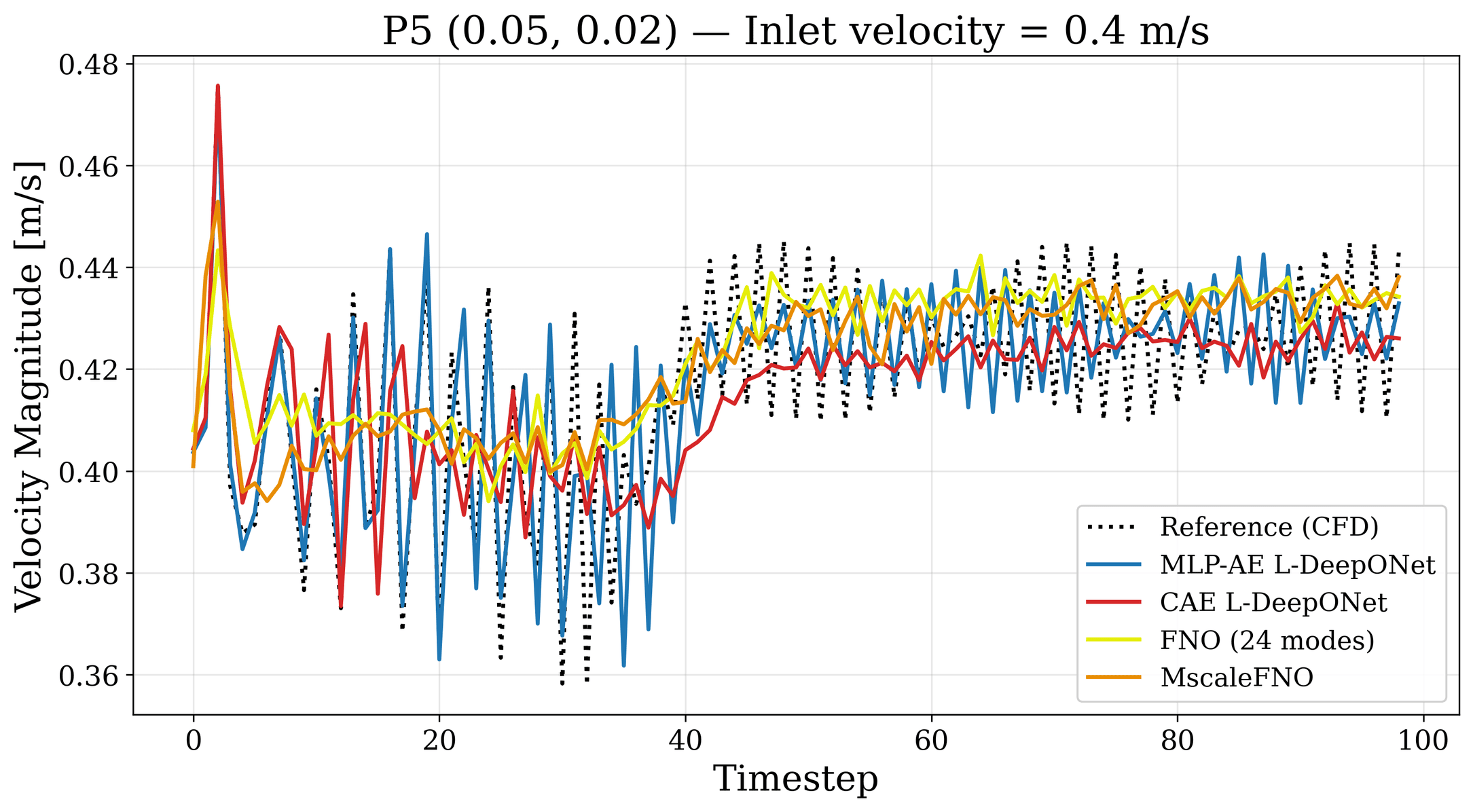}
            \caption{P5 (0.05, 0.02)}
        \end{subfigure}
        \hfill
        \begin{subfigure}[t]{0.32\textwidth}
            \centering
            \includegraphics[width=\textwidth]{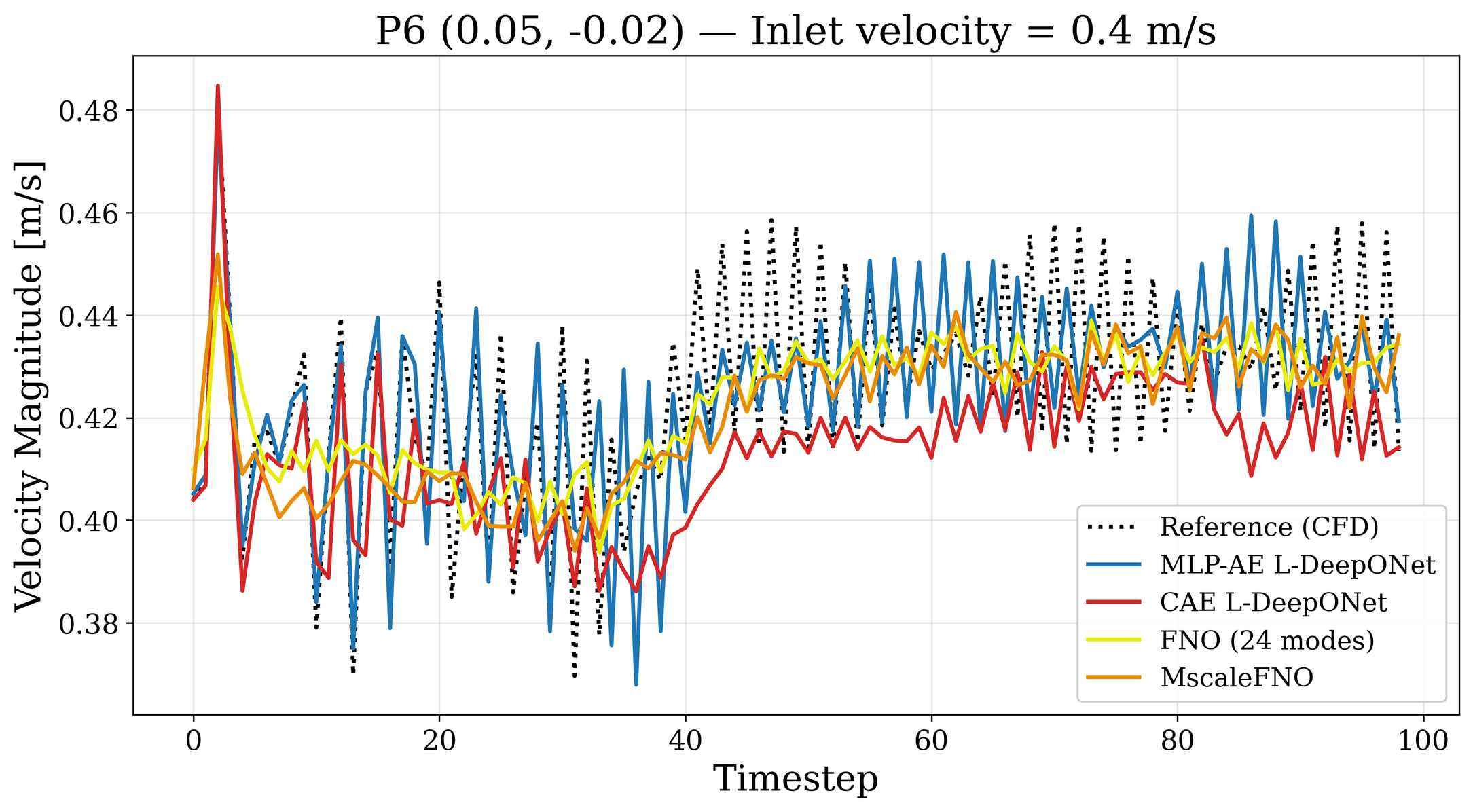}
            \caption{P6 (0.05, -0.02)}
        \end{subfigure}
    \end{minipage}
    }%
    \caption{Velocity time histories at six probe locations for inlet velocity 0.4 m/s.}
    \label{fig:probe_velocity_inlet040}
\end{figure}

\begin{figure}[H]
    \centering
    \makebox[\textwidth][c]{%
    \begin{minipage}{1.3\textwidth}
        \centering
        \begin{subfigure}[t]{0.32\textwidth}
            \centering
            \includegraphics[width=\textwidth]{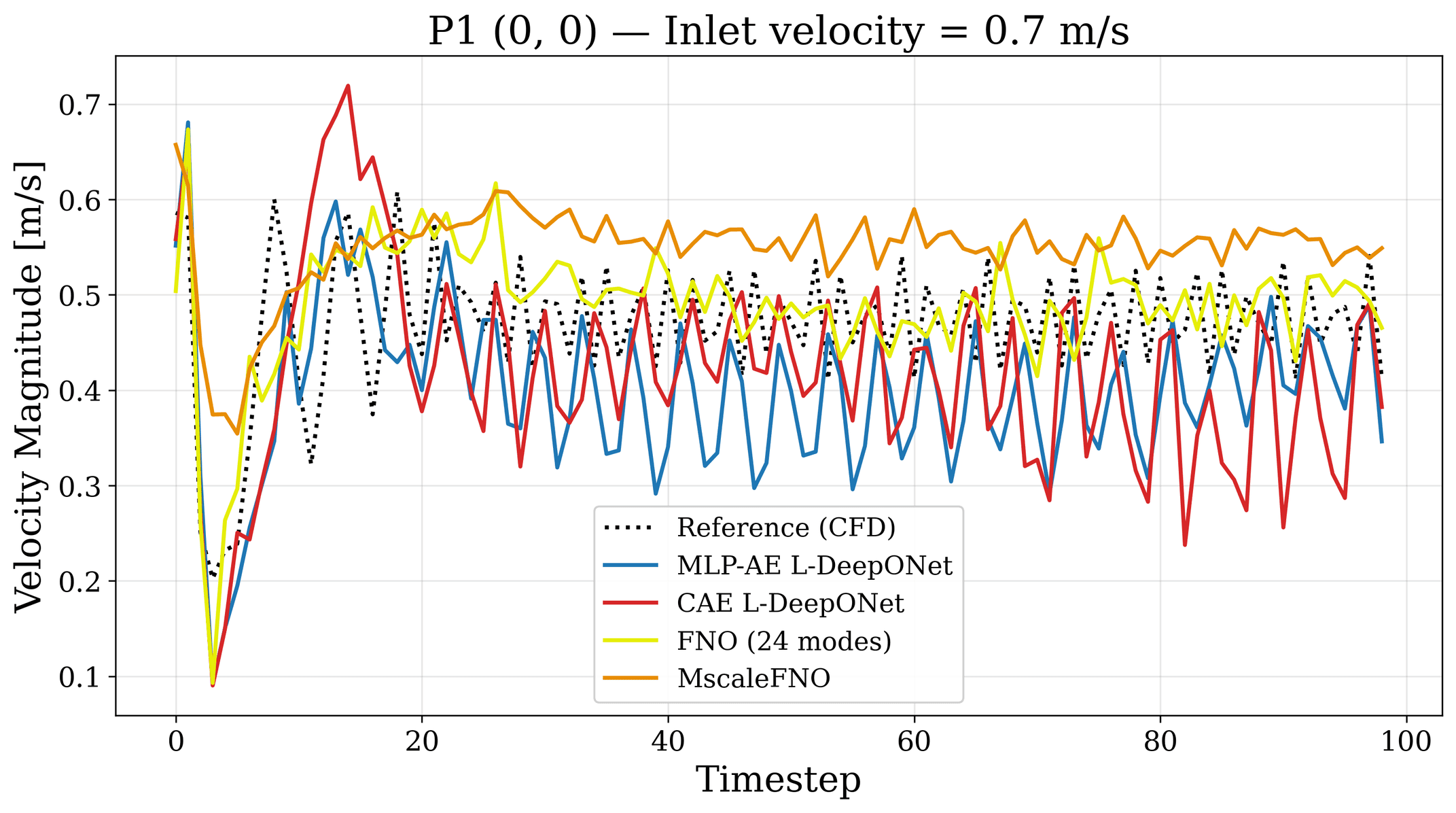}
            \caption{P1 (0, 0)}
        \end{subfigure}
        \hfill
        \begin{subfigure}[t]{0.32\textwidth}
            \centering
            \includegraphics[width=\textwidth]{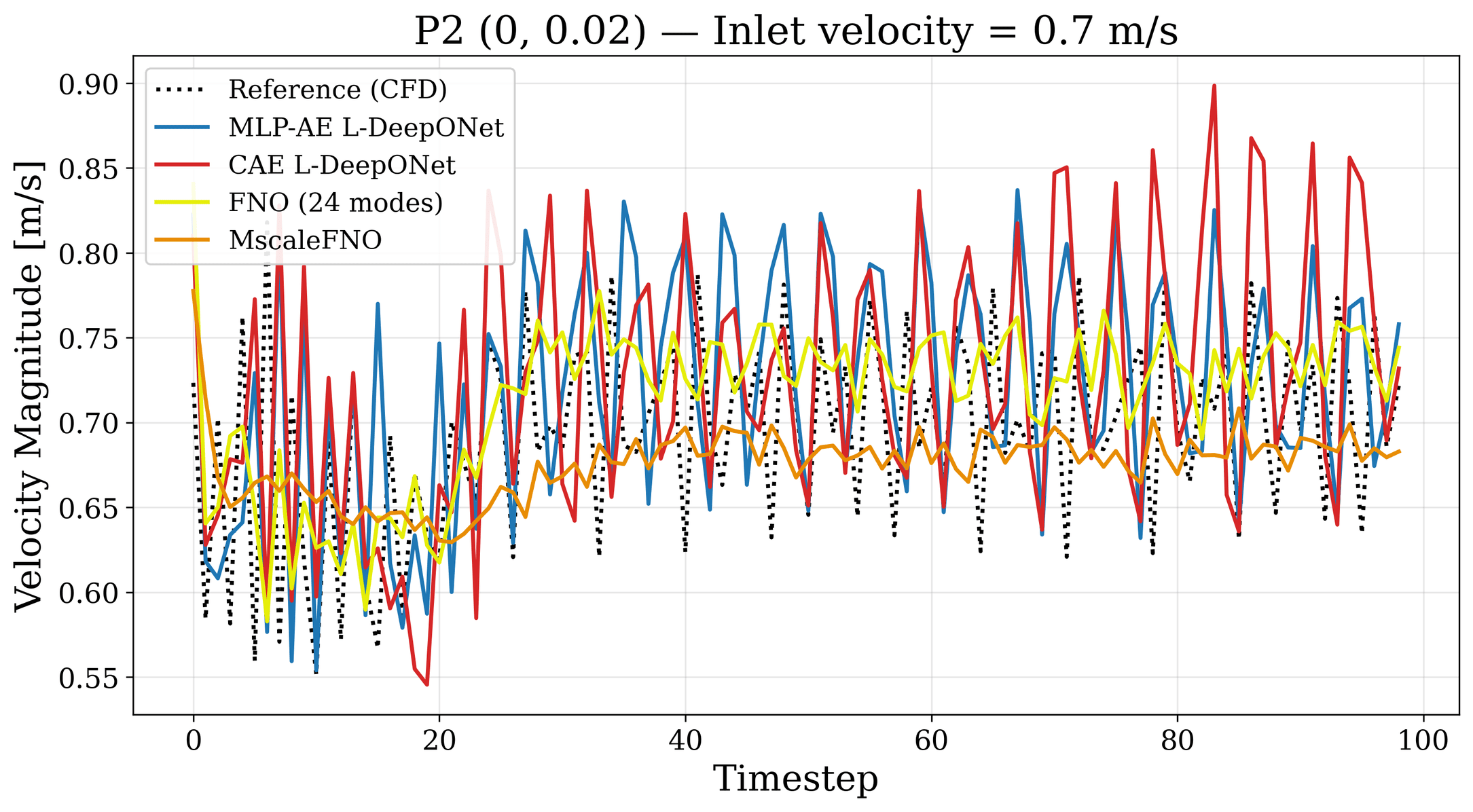}
            \caption{P2 (0, 0.02)}
        \end{subfigure}
        \hfill
        \begin{subfigure}[t]{0.32\textwidth}
            \centering
            \includegraphics[width=\textwidth]{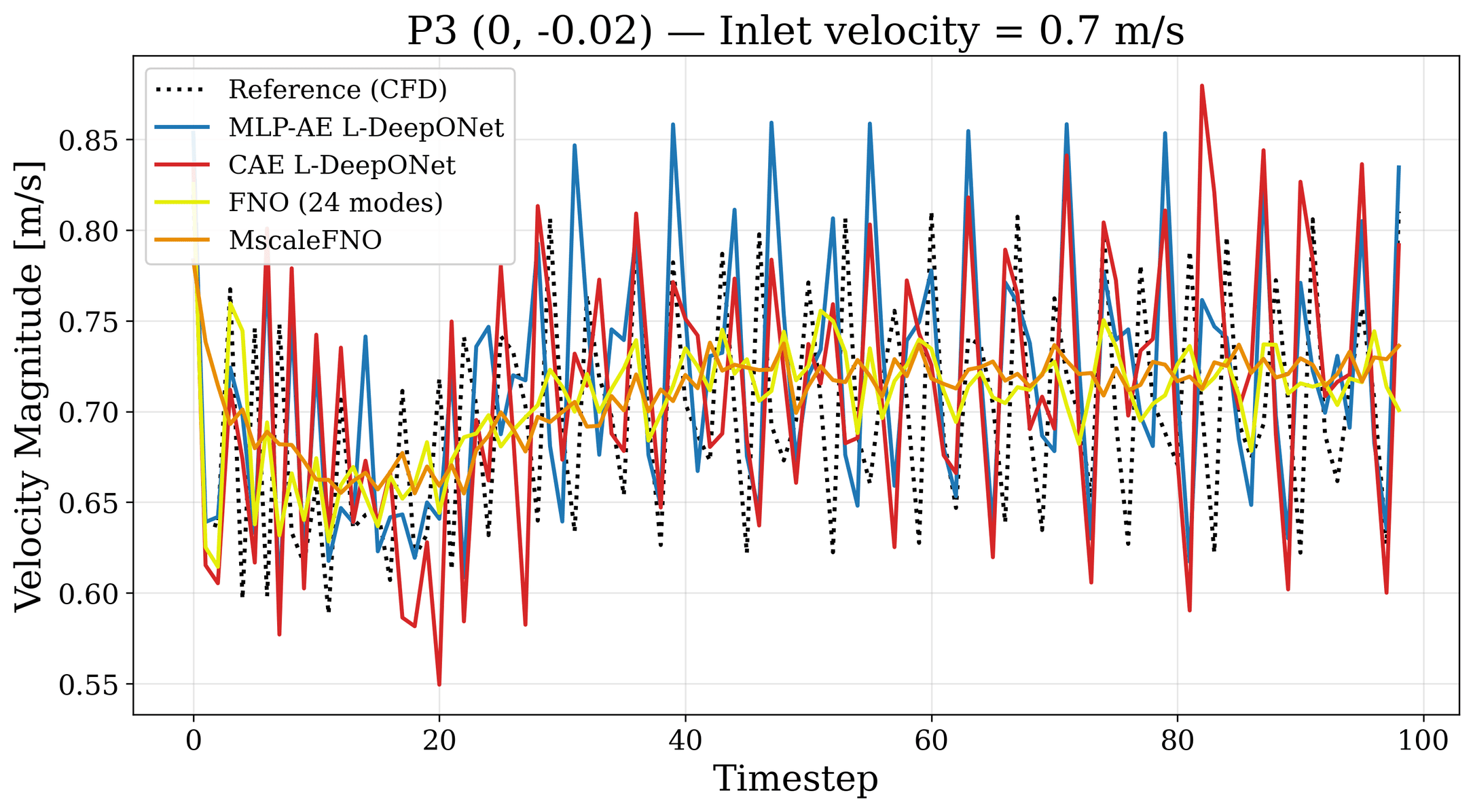}
            \caption{P3 (0, -0.02)}
        \end{subfigure}
        
        \vspace{4pt}
        
        \begin{subfigure}[t]{0.32\textwidth}
            \centering
            \includegraphics[width=\textwidth]{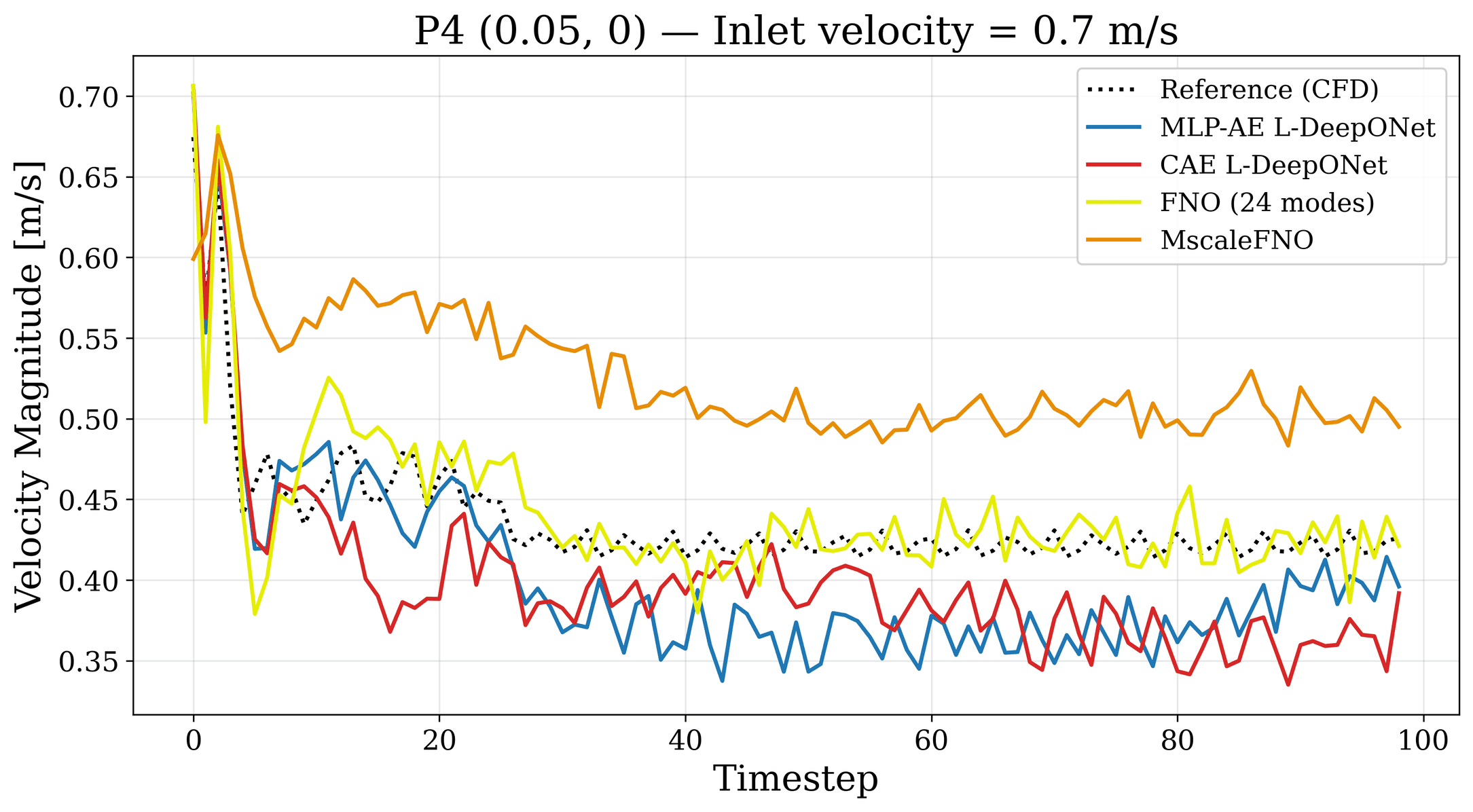}
            \caption{P4 (0.05, 0)}
        \end{subfigure}
        \hfill
        \begin{subfigure}[t]{0.32\textwidth}
            \centering
            \includegraphics[width=\textwidth]{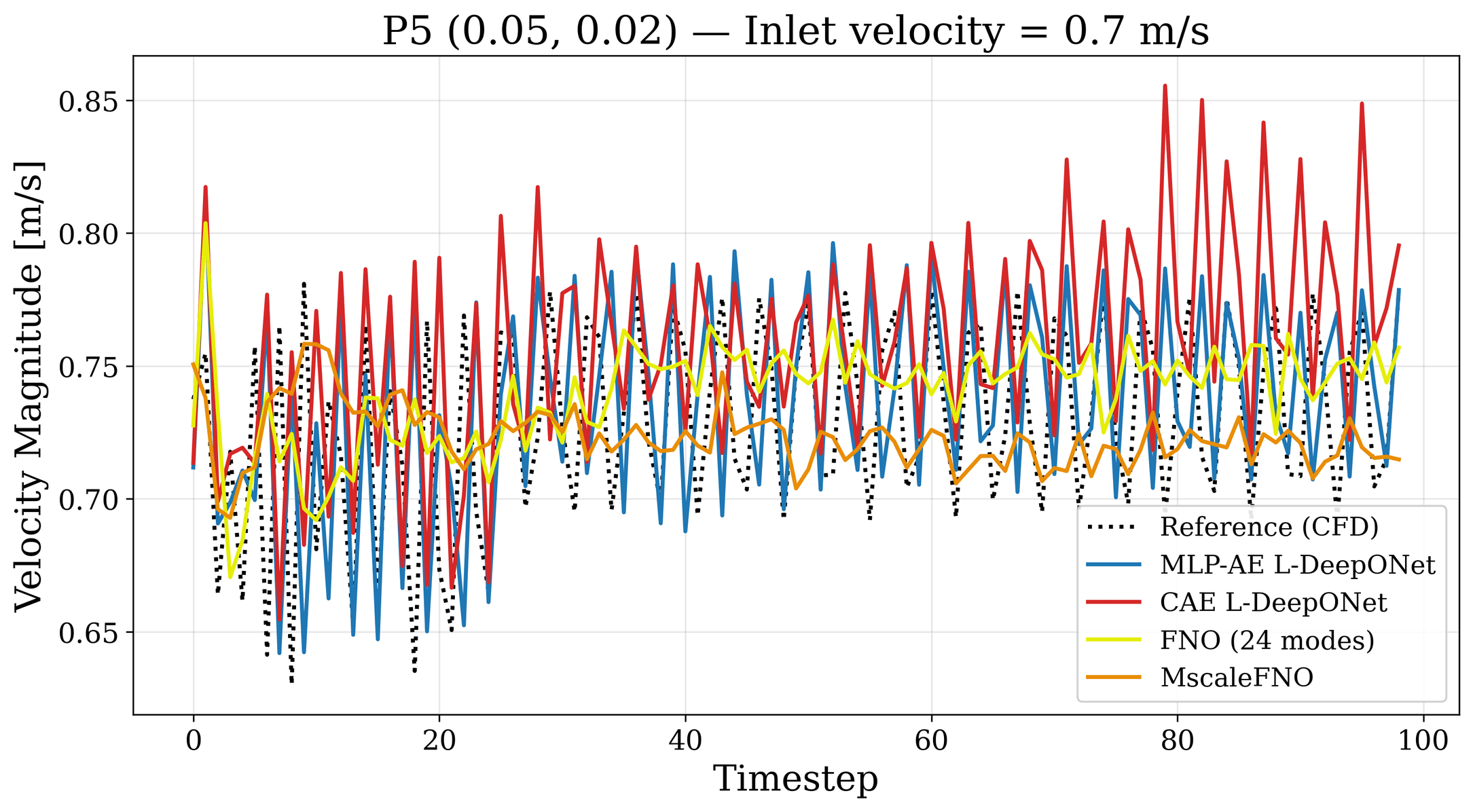}
            \caption{P5 (0.05, 0.02)}
        \end{subfigure}
        \hfill
        \begin{subfigure}[t]{0.32\textwidth}
            \centering
            \includegraphics[width=\textwidth]{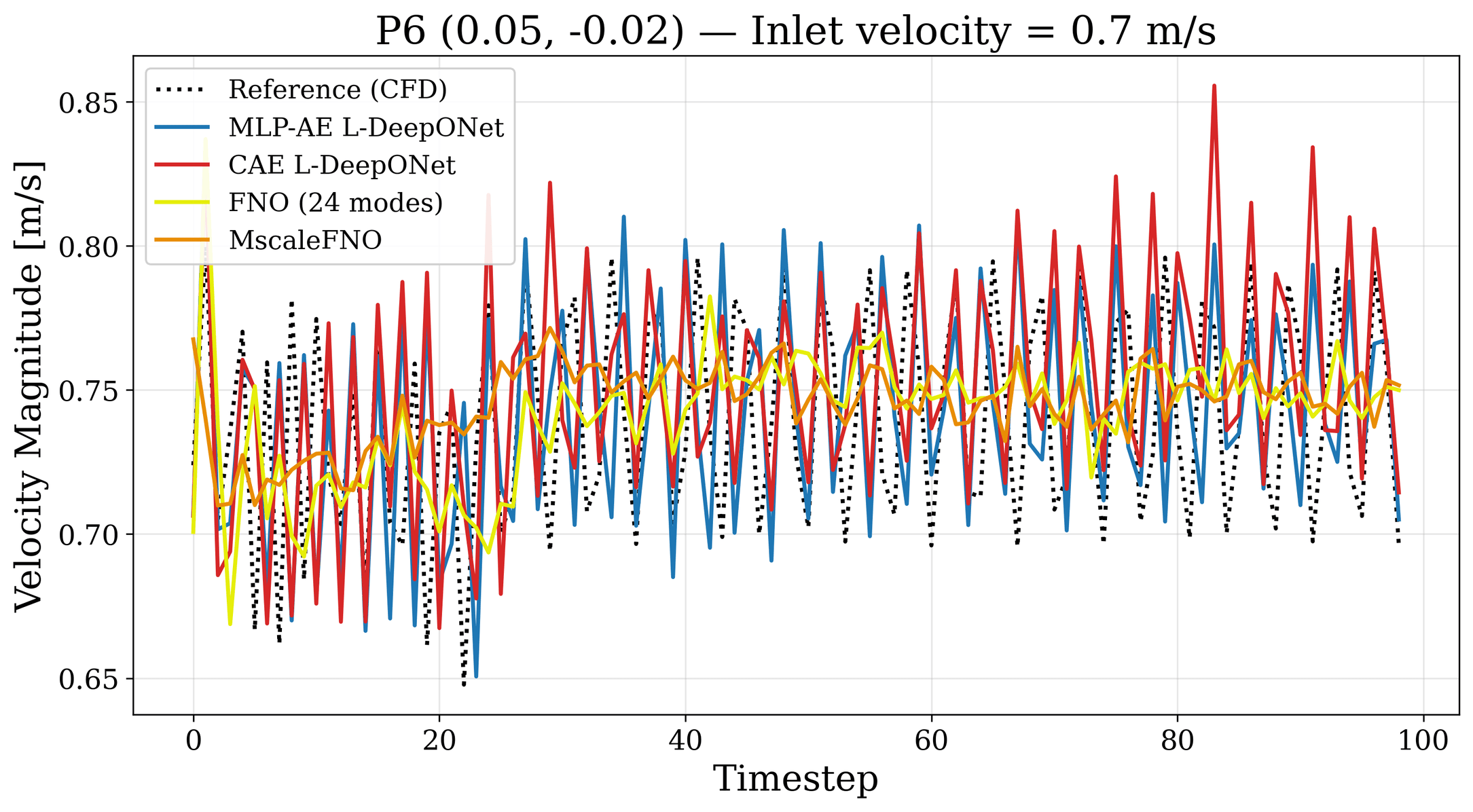}
            \caption{P6 (0.05, -0.02)}
        \end{subfigure}
    \end{minipage}
    }%
    \caption{Velocity time histories at six probe locations for inlet velocity 0.7 m/s.}
    \label{fig:probe_velocity_inlet070}
\end{figure}

To further evaluate the models' ability to capture the temporal dynamics of the flow, velocity time histories at six representative probe locations were compared against the reference CFD solutions. Figure~\ref{fig:probe_velocity_inlet040} and Figure~\ref{fig:probe_velocity_inlet070} present the probe-level velocity comparisons for inlet velocities of 0.4 m/s and 0.7 m/s, respectively. The multi-scale L-DeepONet models (using MLP-based AE and CAE) successfully reproduce the periodic oscillations observed in the reference solutions at all probe locations, closely following the phase and amplitude of the vortex-induced velocity fluctuations. In contrast, the FNO and MscaleFNO predictions appear as smoothed curves that follow the mean trend but lack the oscillatory variations, further confirming that the FNO-based models are unable to resolve the high-frequency flow dynamics despite achieving low global relative $L^2$ error values.

Although the velocity field contours in Section~\ref{subsec6.2} and the probe time histories in Figures~\ref{fig:probe_velocity_inlet040} and~\ref{fig:probe_velocity_inlet070} demonstrate that the L-DeepONet models reproduce the periodic K\'{a}rm\'{a}n vortex streets more faithfully than the FNO-based models, the relative $L^2$ errors reported in Table~\ref{tab:rl2e_summary} suggest that the FNO-based models are more accurate. This discrepancy arises because the relative $L^2$ error is evaluated pointwise at each timestep, so any temporal phase shift between the predicted and the reference signals is penalized even when the underlying waveform is correctly captured. To verify whether the larger $L^2$ errors of the L-DeepONet models reflect a genuine deficiency in capturing the flow dynamics or merely a phase shift in the predicted vortex streets, we compute z-score normalized dynamic time warping (DTW) distances at the same six probe locations on the fully-developed regime ($t = 40 \sim 99$). Each signal is z-score normalized prior to DTW so that the comparison reflects waveform shape rather than absolute magnitude.The DTW distance between two normalized signals $\tilde{x} = (\tilde{x}_1, \ldots, \tilde{x}_N)$ and $\tilde{y} = (\tilde{y}_1, \ldots, \tilde{y}_N)$ is defined as
\begin{equation}
    D_{\text{DTW}}(\tilde{x}, \tilde{y}) = \min_{\pi \in \mathcal{A}_w} \sum_{(i,j) \in \pi} |\tilde{x}_i - \tilde{y}_j|,
    \label{eq:dtw}
\end{equation}
where $\mathcal{A}_w$ denotes the set of valid warping paths constrained by $|i - j| \leq w$. In this study, $w = 17$ is used.

Tables~\ref{tab:dtw_inlet040} and~\ref{tab:dtw_inlet070} report the z-score DTW distances for each (probe, model) combination at inlet velocities of 0.4 m/s and 0.7 m/s, respectively. Averaged over the six probes, MLP-AE L-DeepONet records the smallest mean DTW for both inlet conditions (4.825 at 0.4 m/s and 5.205 at 0.7 m/s), corresponding to a 19.6\% improvement over the second-ranked model FNO (6.000) at 0.4 m/s and a 6.4\% improvement over the second-ranked model CAE L-DeepONet (5.562) at 0.7 m/s. The advantage is consistent across probe locations, with MLP-AE L-DeepONet ranking first at five of six probes at 0.4 m/s and at four of six probes at 0.7 m/s. In particular, at the wake shear layer probes P5 and P6, where the K\'{a}rm\'{a}n vortex street is most strongly developed, MLP-AE L-DeepONet attains the smallest DTW under both inlet conditions. These results resolve the apparent discrepancy between the qualitative observations and the $L^2$ ranking. Once the temporal phase shift is accounted for, MLP-AE L-DeepONet emerges as the most accurate model under both inlet conditions, demonstrating that the larger relative $L^2$ error of the L-DeepONet models originates predominantly from temporal phase shift rather than from a genuine deficiency in learning the flow dynamics. This conclusion is robust to the choice of the warping band radius. The model ranking remains unchanged for $w \in \{10, 13, 15, 17, 20\}$, and although the FNO-based models and CAE L-DeepONet exhibit only minor variation in DTW with respect to $w$ (less than 2\% reduction from $w = 10$ to $w = 20$ at 0.4 m/s), MLP-AE L-DeepONet undergoes a 19.0\% reduction over the same range, which itself indicates that its prediction error contains a substantially larger phase-shift component than the other models.
 
 
\begin{table}[H]
\centering
\caption{Z-score normalized DTW distances at inlet velocity 0.4 m/s with Sakoe--Chiba band radius $w = 17$. The smallest value in each row is shown in bold.}
\label{tab:dtw_inlet040}
\renewcommand{\arraystretch}{1.2}
\resizebox{\textwidth}{!}{%
\begin{tabular}{lcccc}
\toprule
\textbf{Probe} & \textbf{MLP-AE L-DeepONet} & \textbf{CAE L-DeepONet} & \textbf{FNO (24 modes)} & \textbf{MscaleFNO} \\
\midrule
P1                      & \textbf{4.765} & 5.856          & 6.542          & 5.892          \\
P2                      & \textbf{4.725} & 5.194          & 5.934          & 6.702          \\
P3                      & \textbf{4.013} & 6.725          & 5.257          & 6.077          \\
P4                      & 5.026          & 5.335          & \textbf{4.916} & 6.077          \\
P5                      & \textbf{5.271} & 7.843          & 6.872          & 7.428          \\
P6                      & \textbf{5.152} & 5.786          & 6.480          & 6.539          \\
\midrule
\textbf{Mean (P1--P6)}  & \textbf{4.825} & 6.123          & 6.000          & 6.453          \\
\bottomrule
\end{tabular}}
\end{table}

 
\begin{table}[H]
\centering
\caption{Z-score normalized DTW distances at inlet velocity 0.7 m/s with Sakoe--Chiba band radius $w = 17$. The smallest value in each row is shown in bold.}
\label{tab:dtw_inlet070}
\renewcommand{\arraystretch}{1.2}
\resizebox{\textwidth}{!}{%
\begin{tabular}{lcccc}
\toprule
\textbf{Probe} & \textbf{MLP-AE L-DeepONet} & \textbf{CAE L-DeepONet} & \textbf{FNO (24 modes)} & \textbf{MscaleFNO} \\
\midrule
P1                      & \textbf{5.710} & 5.992          & 5.817          & 5.873          \\
P2                      & 5.668          & 5.567          & \textbf{5.390} & 5.761          \\
P3                      & \textbf{3.723} & 4.695          & 6.434          & 5.584          \\
P4                      & 6.254          & 6.772          & \textbf{5.583} & 6.904          \\
P5                      & \textbf{4.988} & 5.395          & 5.505          & 6.083          \\
P6                      & \textbf{4.887} & 4.952          & 5.971          & 5.837          \\
\midrule
\textbf{Mean (P1--P6)}  & \textbf{5.205} & 5.562          & 5.784          & 6.007          \\
\bottomrule
\end{tabular}}
\end{table}

Beyond the flow field predictions, the pressure drop across the HCSG tube bundle was also evaluated, as it is a key parameter for reactor thermal-hydraulic design and operational monitoring. The pressure drop was calculated as the difference between the area-averaged pressures at the inlet and outlet boundaries at each time step. Table~\ref{tab:pressure_drop_error} summarizes the mean relative error of the pressure drop predictions for all four models.

\begin{table}[H]
\centering
\caption{Mean relative error (\%) of pressure drop predictions.}
\label{tab:pressure_drop_error}
\renewcommand{\arraystretch}{1.2}
\begin{tabular}{lcc}
\toprule
\textbf{Model} & \textbf{Inlet 0.4 m/s} & \textbf{Inlet 0.7 m/s} \\
\midrule
MLP-AE L-DeepONet & 1.24 & 2.27 \\
CAE L-DeepONet    & 1.93 & 1.92 \\
FNO (24 modes)    & 0.85 & 2.67 \\
MscaleFNO         & 1.75 & 4.24 \\
\bottomrule
\end{tabular}
\end{table}

As shown in Table~\ref{tab:pressure_drop_error}, all four models predict the pressure drop with mean relative errors below 5\%, demonstrating sufficient accuracy for engineering applications. At the 0.4 m/s inlet velocity, the FNO achieves the best performance, while at the 0.7 m/s inlet velocity, the L-DeepONet models maintain more consistent accuracy. In particular, the MscaleFNO exhibits the largest error at 0.7 m/s, suggesting that the multi-scale input scaling may introduce additional uncertainty in pressure predictions at higher flow velocities.

\begin{figure}[H]
    \centering
    \begin{subfigure}[t]{0.48\textwidth}
        \centering
        \includegraphics[width=\textwidth]{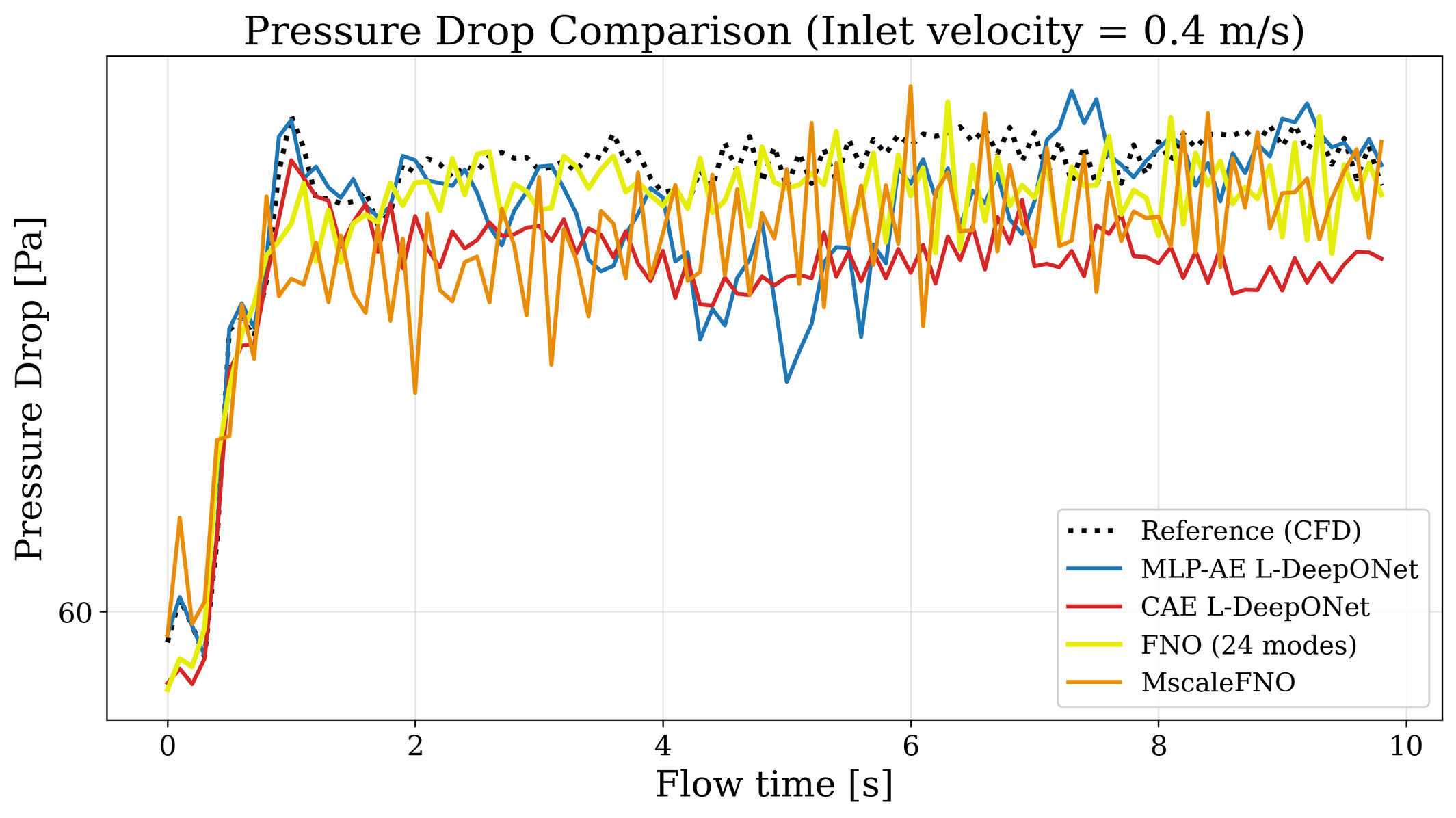}
        \caption{Pressure drop}
    \end{subfigure}
    \hfill
    \begin{subfigure}[t]{0.48\textwidth}
        \centering
        \includegraphics[width=\textwidth]{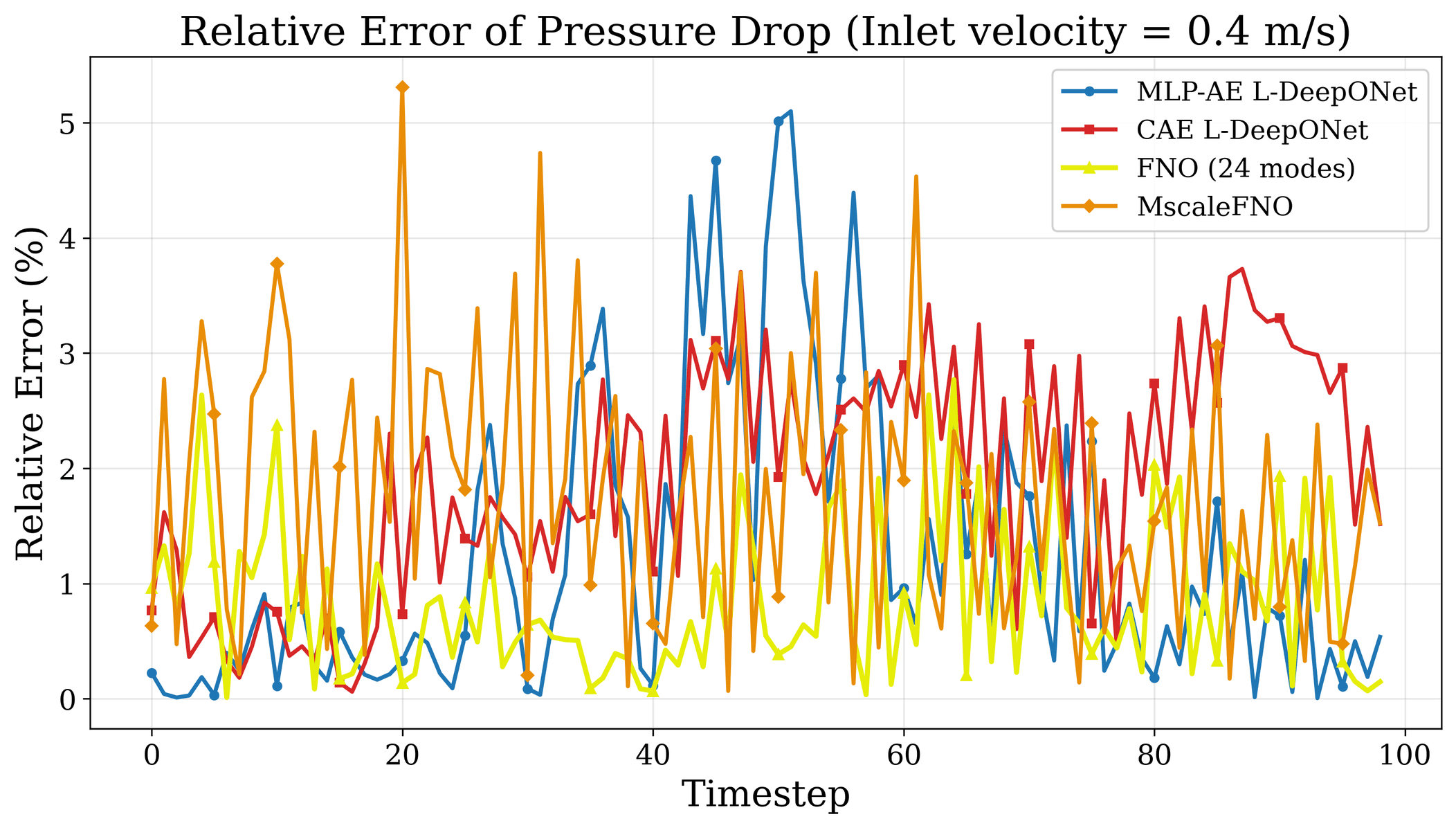}
        \caption{Relative error}
    \end{subfigure}
    \caption{Pressure drop comparison (a) and relative error (b) for inlet velocity 0.4 m/s.}
    \label{fig:pressure_drop_inlet040}
\end{figure}

\begin{figure}[H]
    \centering
    \begin{subfigure}[t]{0.48\textwidth}
        \centering
        \includegraphics[width=\textwidth]{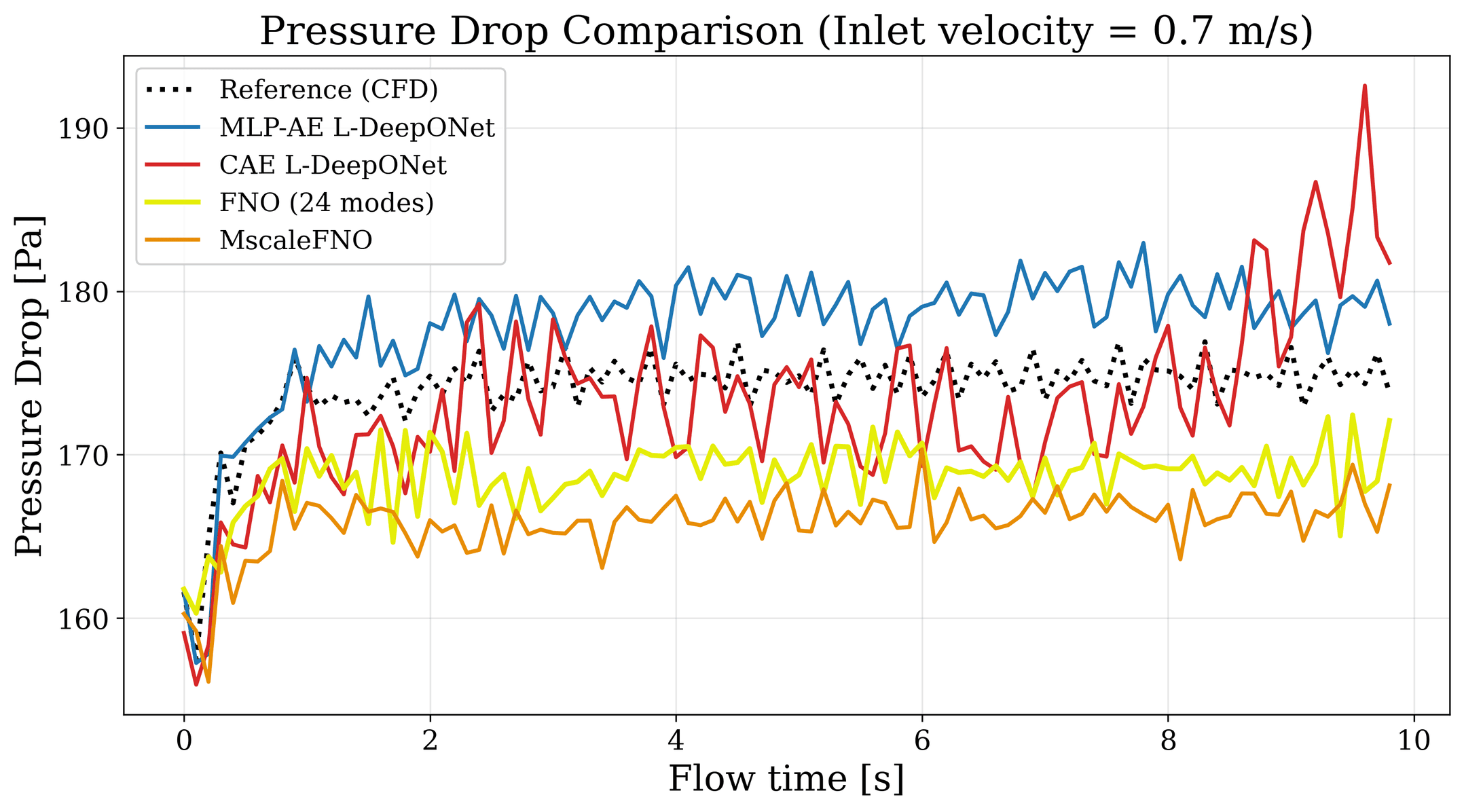}
        \caption{Pressure drop}
    \end{subfigure}
    \hfill
    \begin{subfigure}[t]{0.48\textwidth}
        \centering
        \includegraphics[width=\textwidth]{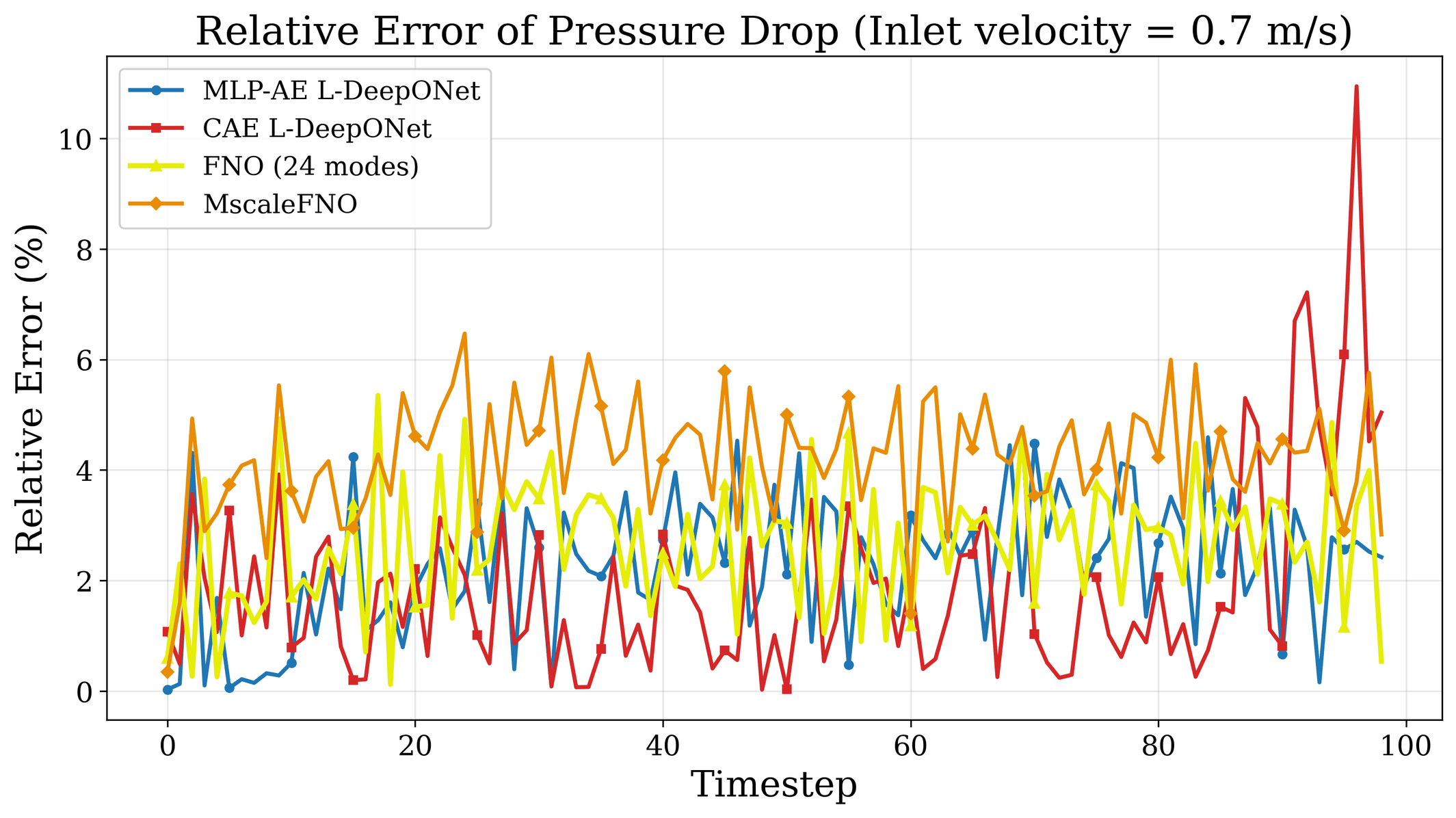}
        \caption{Relative error}
    \end{subfigure}
    \caption{Pressure drop comparison (a) and relative error (b) for inlet velocity 0.7 m/s.}
    \label{fig:pressure_drop_inlet070}
\end{figure}

Figure~\ref{fig:pressure_drop_inlet040} and Figure~\ref{fig:pressure_drop_inlet070} present the time evolution of the predicted pressure drop and its relative error for inlet velocities of 0.4 m/s and 0.7 m/s, respectively. At 0.4 m/s, all models closely follow the reference pressure drop curve with small deviations. At 0.7 m/s, the predictions exhibit larger fluctuations, particularly for the MscaleFNO. The L-DeepONet models demonstrate consistent pressure drop prediction across both inlet velocity conditions, indicating their robustness for engineering-level system monitoring.

\subsection{Discussion}
\label{sec6.4}

The training loss curves presented in ~\ref{app1} provide insight into the convergence behavior of each model architecture. The ROMs, including both the MLP-based AE and the CAE, exhibit stable and consistent convergence for both velocity and pressure fields. The training and validation losses decrease smoothly without significant oscillation or divergence, indicating that the latent representations learned by these AEs faithfully preserve the essential features of the original flow fields. This confirms that the dimensionality reduction stage does not introduce a bottleneck in the overall prediction pipeline, and any limitations observed in the final predictions can be attributed to the subsequent surrogate modeling stage rather than the encoding process.

A common trend observed across all surrogate models is that the velocity field requires more representational capacity than the pressure field. As shown in the loss curves, the velocity field exhibits slower convergence and greater fluctuation in the validation loss compared to the pressure field across all model architectures. This reflects the richer dynamics of the velocity field in the present configuration. The flow in the HCSG geometry generates Kármán vortex streets, producing strong periodic oscillations in the velocity field. The model is therefore required to capture both the phase and amplitude of the vortex-induced velocity oscillations across multiple time steps. The pressure field, while also affected by vortex streets, exhibits smoother temporal variations and less pronounced periodic content, making it a more straightforward learning target.

The multi-scale L-DeepONet successfully captured the periodic flow structures, as demonstrated in Section~\ref{subsec6.2}. This indicates that the multi-scale architecture possesses sufficient capacity to learn the high-frequency components of the flow. However, the multi-scale design expands the effective learning dimension through multiple trunk sub-networks and scale factors, and therefore requires a correspondingly larger training dataset. With the current dataset of 50 simulations, the training loss decreases steadily while the validation loss reaches a plateau, indicating that the dataset is not yet large enough relative to this expanded learning dimension. Therefore, acquiring additional high-fidelity CFD data is expected to further enhance the generalization performance of the multi-scale L-DeepONet. In this sense, the multi-scale L-DeepONet shows strong potential for DT applications that benefit from high-frequency resolution, such as flow visualization and instantaneous vortex tracking. Like the standard FNO, the MscaleFNO predicts the time-averaged mean flow rather than the periodic vortex structures. Although the multi-scale strategy showed limited effect in the FNO family, both models accurately predicted the mean flow and low-frequency components, making the FNO family well suited for applications focused on system-level monitoring.

The contrasting behavior between the FNO-based and L-DeepONet-based models can be further understood by examining the architectural differences and how each framework processes the flow field data. The FNO operates directly on the full structured grid and retains only the first $k$ Fourier modes in each Fourier layer, truncating higher-frequency components during training. This Fourier mode truncation inherently biases the model toward smooth, low-frequency solutions. While retaining more Fourier modes would in principle allow higher-frequency components to be captured, as shown in Section~\ref{subsec5.2}, simply increasing the number of Fourier modes is insufficient to overcome the spectral bias, and for high-fidelity CFD data the number of modes cannot be increased indefinitely due to memory and computational cost constraints. This behavior is consistent with the observations reported by You et al., where the standard FNO produced smoothed approximations of high-frequency oscillations ~\cite{you2025mscalefno}. On the other hand, the L-DeepONet employs multiple sub trunk networks, each receiving the temporal inputs scaled by a distinct factor, enabling each sub-network to learn basis functions at a different frequency band. This approach is consistent with the findings of Liu and Cai et al. and Wang et al., who demonstrated that applying multi-scale techniques to the trunk network significantly improves the ability to capture high-frequency oscillatory components~\cite{liu2021multiscale, wang2025multi}. The MscaleFNO, in contrast, applies its scaling factors to the input function rather than to the output basis representation. In the present operator-learning setting, the input function is an essentially smooth initial flow field, and input-level scaling alone does not directly induce frequency diversity in the output. The multi-scale L-DeepONet, by contrast, explicitly constructs basis functions across distinct frequency bands at the trunk-network level, so that output frequency diversity is guaranteed by the architecture itself. This architectural distinction provides a plausible explanation for why the multi-scale L-DeepONet captures the periodic vortex dynamics while the MscaleFNO converges to the time-averaged mean flow analogous to the standard FNO. As discussed in the preceding paragraph, the expressiveness of the multi-scale trunk structure also accounts for the data scaling behavior observed in the L-DeepONet loss curves. The distinction between these two model classes therefore reflects a fundamental difference in where and how multi-scale decomposition is applied within each framework.

\section{Conclusion}
\label{sec7}
This study contributes to neural operator-based surrogate modeling for nuclear thermal-hydraulics by developing a CFD-level transient surrogate framework for an SMR geometry. Four neural operator architectures for predicting transient thermal-hydraulic flow fields in a two-dimensional HCSG configuration were evaluated, including two L-DeepONet variants combined with ROMs, a standard FNO, and a MscaleFNO. All models were trained on CFD simulations of K\'{a}rm\'{a}n vortex streets and evaluated under interpolation and extrapolation test conditions.

The L-DeepONet models, operating in the latent space constructed by the AEs, successfully captured the instantaneous periodic vortex dynamics in both velocity and pressure fields. Probe-level comparisons confirmed that the L-DeepONet predictions closely follow the phase and amplitude of the vortex-induced oscillations observed in the reference CFD solutions. The ROMs demonstrated stable convergence during training, confirming that the latent representations faithfully preserve the essential features of the original flow fields without introducing a bottleneck in the prediction pipeline.

The FNO-based models, including the standard FNO and MscaleFNO, achieved the lowest mean relative $L^2$ errors among all models for both velocity and pressure fields. The qualitative analysis showed that both models predict the time-averaged mean flow rather than the periodic vortex structures. This behavior originates from the spectral bias inherent in Fourier-based neural operators, which favors smooth, low-frequency solutions, and as a result the FNO family provides stable predictions of the mean flow and the pressure drop. Since the relative $L^2$ error is dominated by the mean flow component, however, it does not fully reflect the absence of instantaneous flow features. To complement the $L^2$ metric, z-score normalized DTW distances, which are tolerant to temporal phase shifts, were computed at the same probe locations. The MLP-AE L-DeepONet recorded the lowest mean DTW distance under both inlet conditions, indicating that the larger relative $L^2$ error of the L-DeepONet models originates predominantly from temporal phase shift rather than from a genuine deficiency in learning the flow dynamics. In terms of faithfully reproducing the periodic structure of the K\'{a}rm\'{a}n vortex streets, the L-DeepONet variants, particularly the MLP-AE L-DeepONet, therefore outperformed the FNO-based models.

The multi-scale technique applied to the two frameworks exhibited distinct characteristics that translate into complementary strengths for different digital twin applications. The multi-scale L-DeepONet successfully captured the K\'{a}rm\'{a}n vortex structures and is therefore well suited for applications that require high-frequency resolution, such as flow visualization and instantaneous vortex tracking. The MscaleFNO, on the other hand, predicts the time-averaged mean flow analogous to the standard FNO, making the FNO family well suited for applications focused on system-level monitoring. The present study therefore provides a practical guideline for model selection based on the digital twin objective and the characteristics of the available CFD data.

Future work will focus on extending the proposed framework in three directions. First, expanding the training dataset through additional CFD simulations is expected to further enhance the generalization performance of the multi-scale L-DeepONet. Second, the framework will be extended to thermal-hydraulic simulations that include heat transfer, broadening its applicability to a wider range of physical phenomena in SMR components. Finally, the framework will be extended to three-dimensional geometries of small modular reactors to validate its applicability to practical nuclear thermal-hydraulic systems.

\section*{Acknowledgements}
This work was supported by the National Research Council of Science \& Technology (NST) grant funded by the Korea government (MSIT) (No. GTL24031-000).

\appendix
\section{Neural Network Configurations and Training Loss Curves}
\label{app1}

This appendix provides the detailed neural network configurations, training hyperparameters, and training loss curves for all surrogate models investigated in this study. Table~\ref{tab:hyperparameters_ae} summarizes the hyperparameters for the two ROMs: the MLP-based AE, which operates on the unstructured mesh with 28,877 nodes, and the CAE, which takes the interpolated structured grid of size $282 \times 756$ as input. Both AEs compress the flow field into a 256-dimensional latent space.

Table~\ref{tab:hyperparameters_don} presents the configurations for the L-DeepONet models. Three variants are compared: a vanilla L-DeepONet without multi-scale decomposition, and two multi-scale L-DeepONet models paired with the MLP-based AE and CAE, respectively. The multi-scale variants employ multiple trunk sub-networks, each associated with a distinct scale factor, to decompose the solution into components at different frequency scales.

Table~\ref{tab:hyperparameters_fno} lists the hyperparameters for the FNO and MscaleFNO models. Both models operate on the structured grid and use 24 Fourier modes in each spatial direction. The MscaleFNO consists of 6 sub-networks with learnable scaling factors, while the standard FNO uses a single network with a larger hidden channel width of 128.

Table~\ref{tab:model_summary_compact} provides a summary of the total number of trainable parameters, model sizes, GPU memory consumption, and training times for all architectures. All models were trained on a single NVIDIA A100 40GB GPU.

\begin{figure}[H]
    \centering
    \begin{subfigure}[t]{0.48\textwidth}
        \centering
        \includegraphics[width=\textwidth]{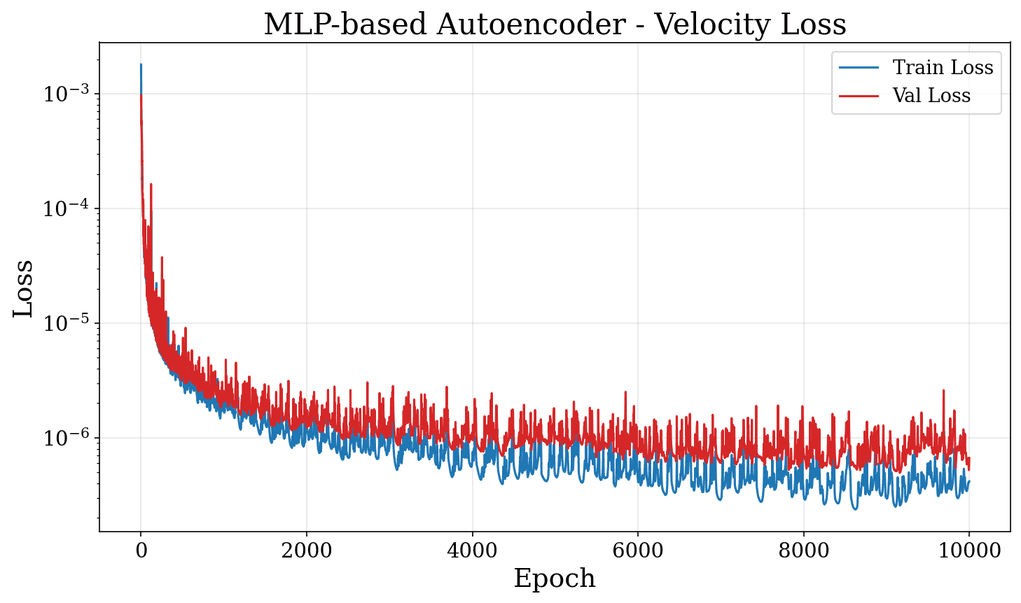}
        \caption{MLP-based AE (Velocity)}
        \label{fig:loss_mlp_ae_vel}
    \end{subfigure}
    \hfill
    \begin{subfigure}[t]{0.48\textwidth}
        \centering
        \includegraphics[width=\textwidth]{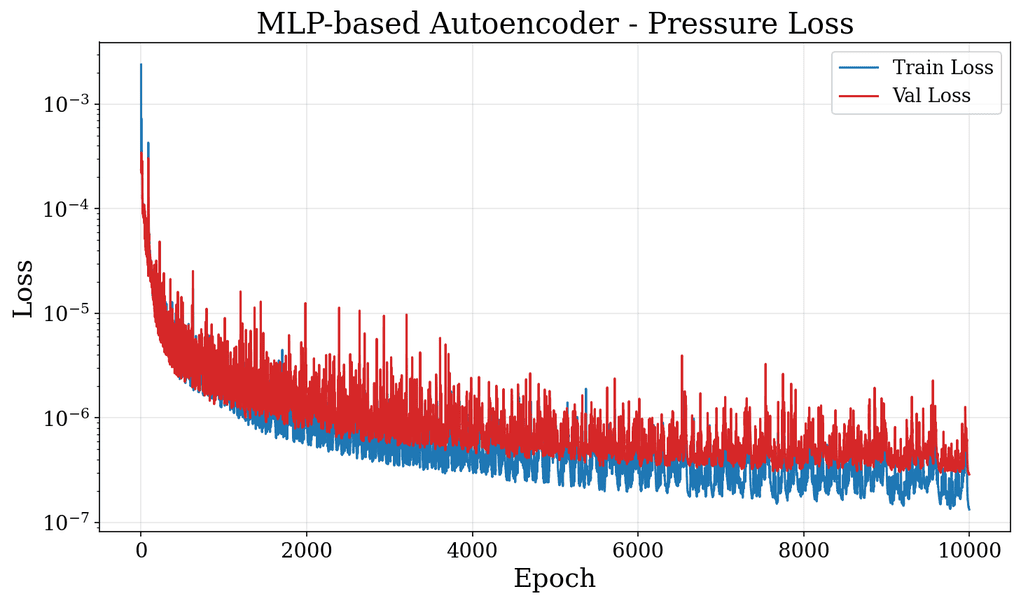}
        \caption{MLP-based AE (Pressure)}
        \label{fig:loss_mlp_ae_pres}
    \end{subfigure}
    \vspace{1em}
    \begin{subfigure}[t]{0.48\textwidth}
        \centering
        \includegraphics[width=\textwidth]{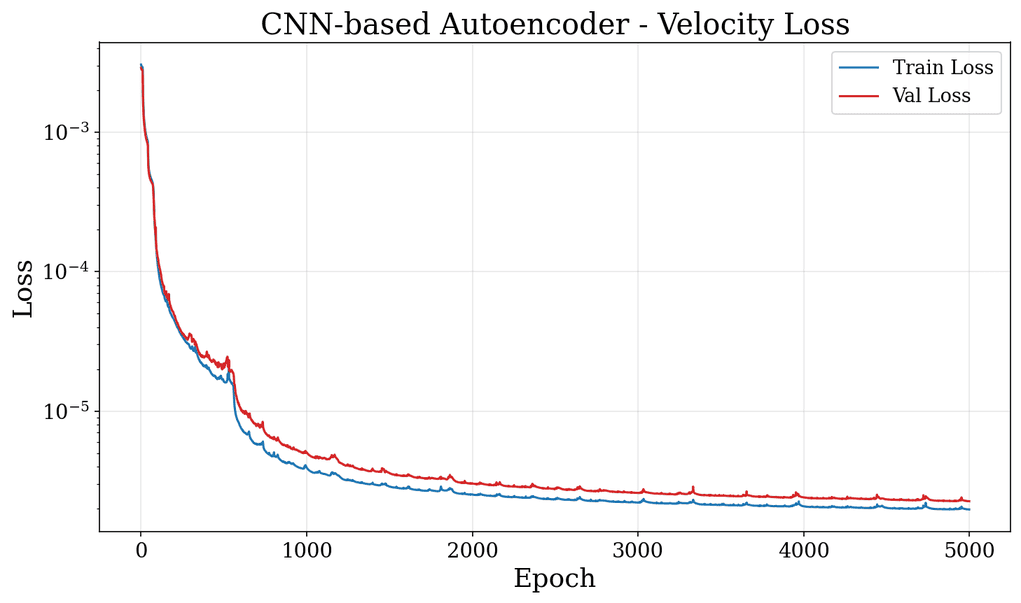}
        \caption{CAE (Velocity)}
        \label{fig:loss_cae_ae_vel}
    \end{subfigure}
    \hfill
    \begin{subfigure}[t]{0.48\textwidth}
        \centering
        \includegraphics[width=\textwidth]{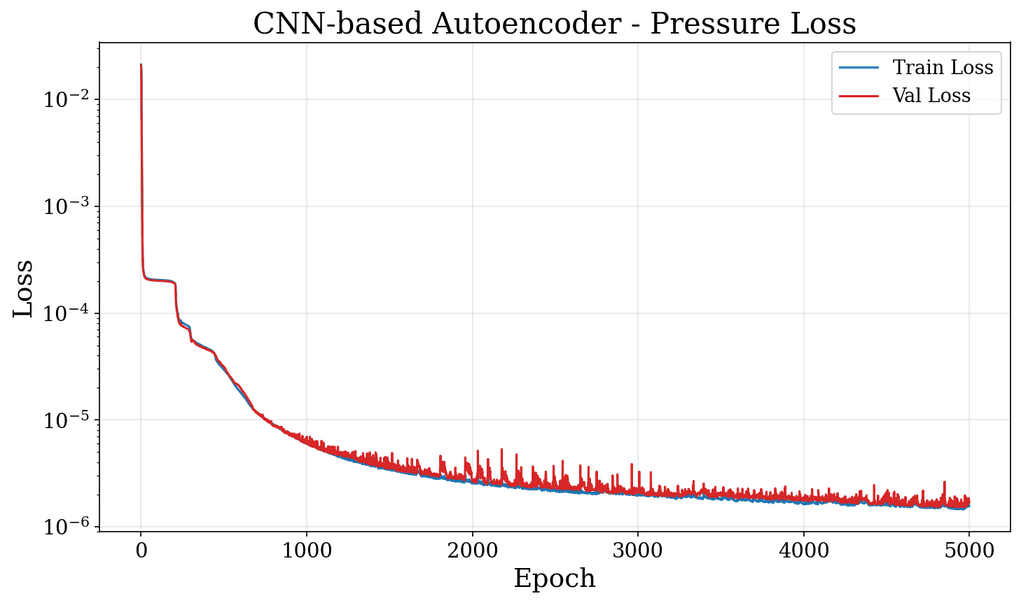}
        \caption{CAE (Pressure)}
        \label{fig:loss_cae_ae_pres}
    \end{subfigure}
    \vspace{1em}
    \begin{subfigure}[t]{0.48\textwidth}
        \centering
        \includegraphics[width=\textwidth]{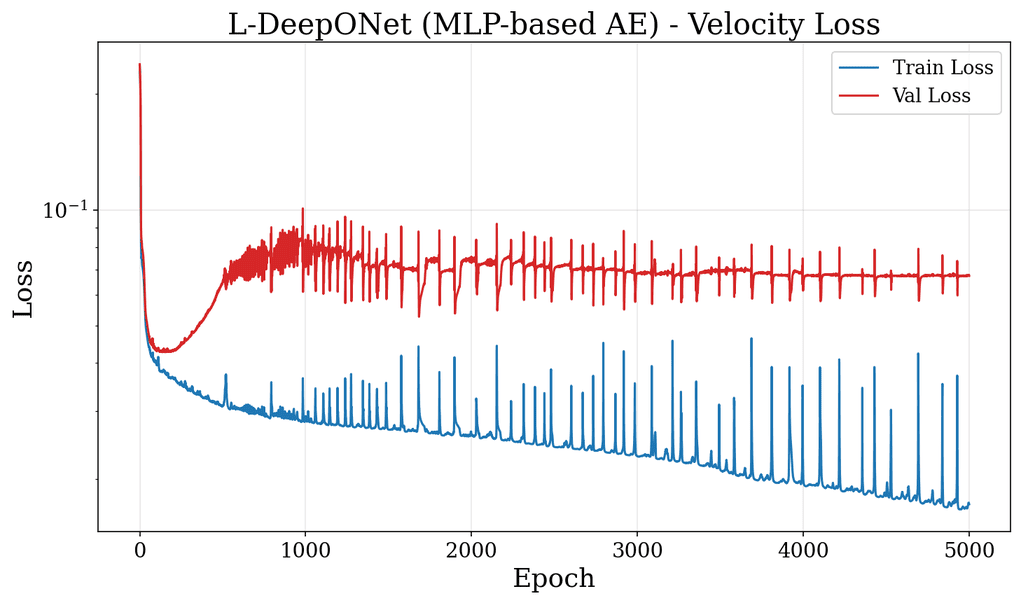}
        \caption{L-DeepONet with MLP-based AE (Velocity)}
        \label{fig:loss_mlp_ldon_vel}
    \end{subfigure}
    \hfill
    \begin{subfigure}[t]{0.48\textwidth}
        \centering
        \includegraphics[width=\textwidth]{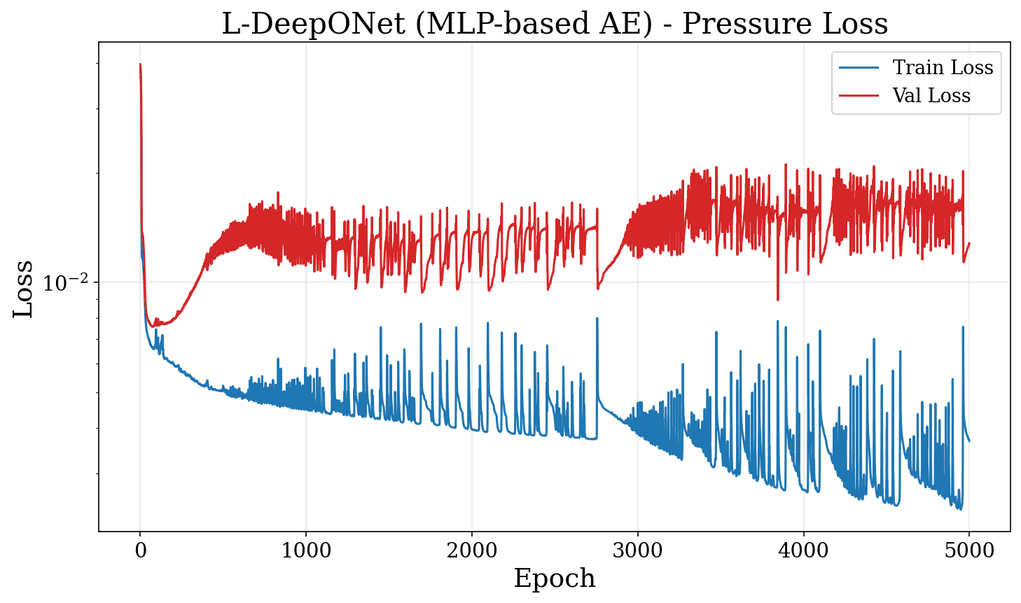}
        \caption{L-DeepONet with MLP-based AE (Pressure)}
        \label{fig:loss_mlp_ldon_pres}
    \end{subfigure}
    \vspace{1em}
    \begin{subfigure}[t]{0.48\textwidth}
        \centering
        \includegraphics[width=\textwidth]{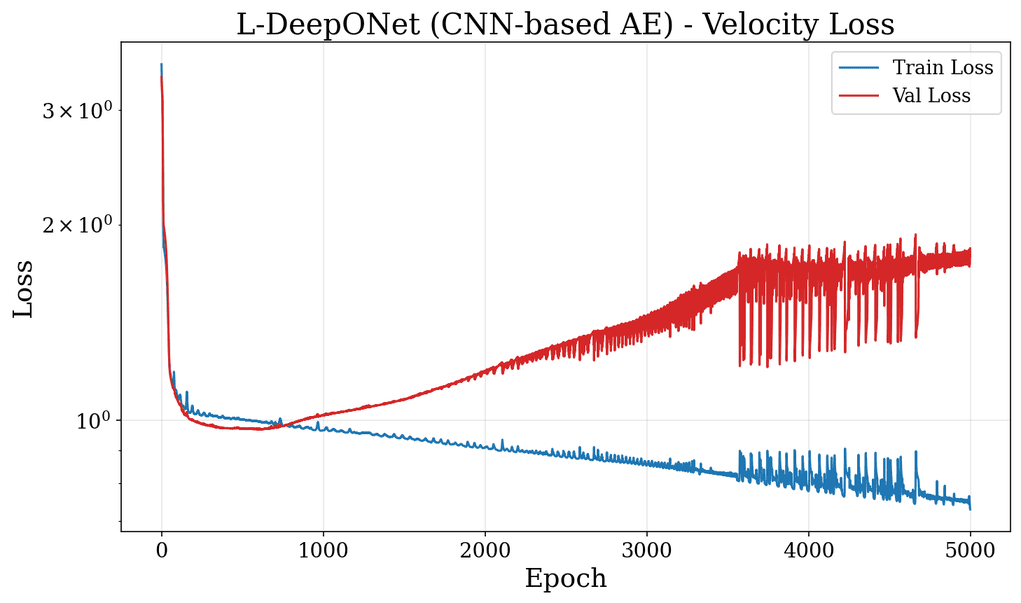}
        \caption{L-DeepONet with CAE (Velocity)}
        \label{fig:loss_cae_ldon_vel}
    \end{subfigure}
    \hfill
    \begin{subfigure}[t]{0.48\textwidth}
        \centering
        \includegraphics[width=\textwidth]{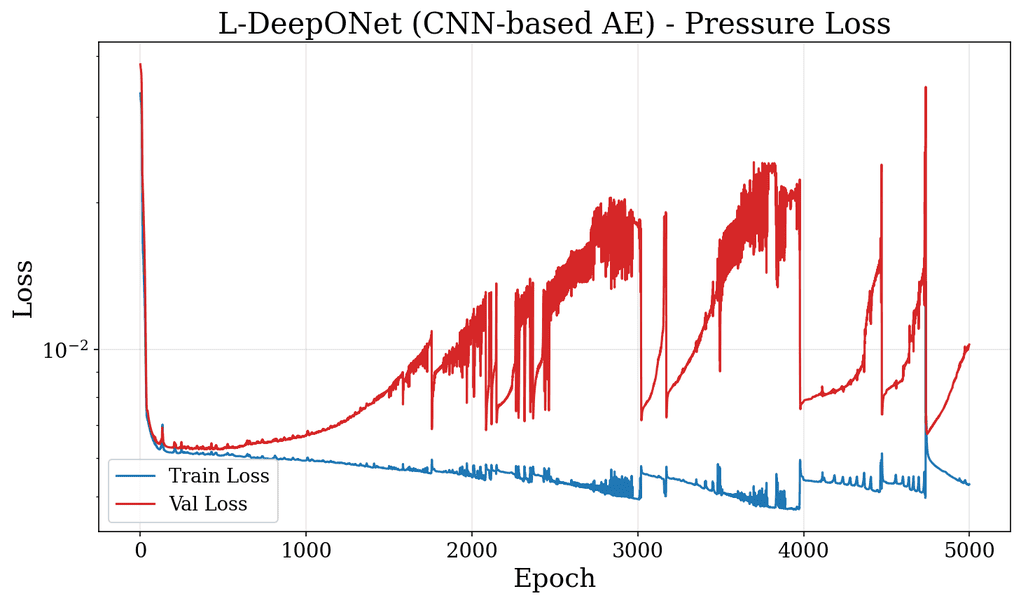}
        \caption{L-DeepONet with CAE (Pressure)}
        \label{fig:loss_cae_ldon_pres}
    \end{subfigure}
    \caption{Training and validation loss curves for AEs and L-DeepONet models on the velocity and pressure fields.}
    \label{fig:loss_curves_ae_ldon}
\end{figure}

\begin{figure}[H]
    \centering
    \begin{subfigure}[t]{0.48\textwidth}
        \centering
        \includegraphics[width=\textwidth]{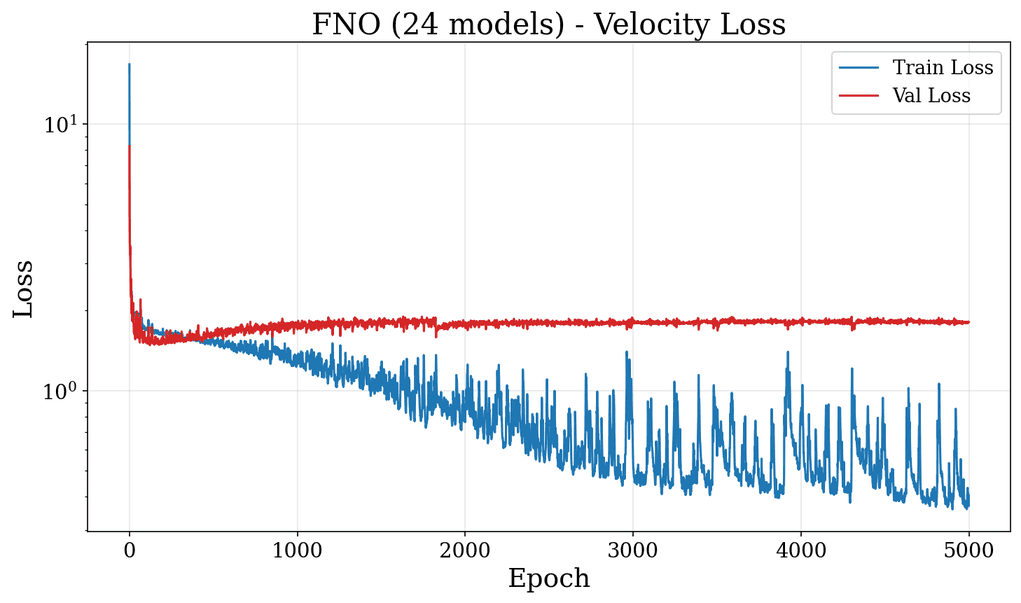}
        \caption{FNO (Velocity)}
        \label{fig:loss_fno_vel}
    \end{subfigure}
    \hfill
    \begin{subfigure}[t]{0.48\textwidth}
        \centering
        \includegraphics[width=\textwidth]{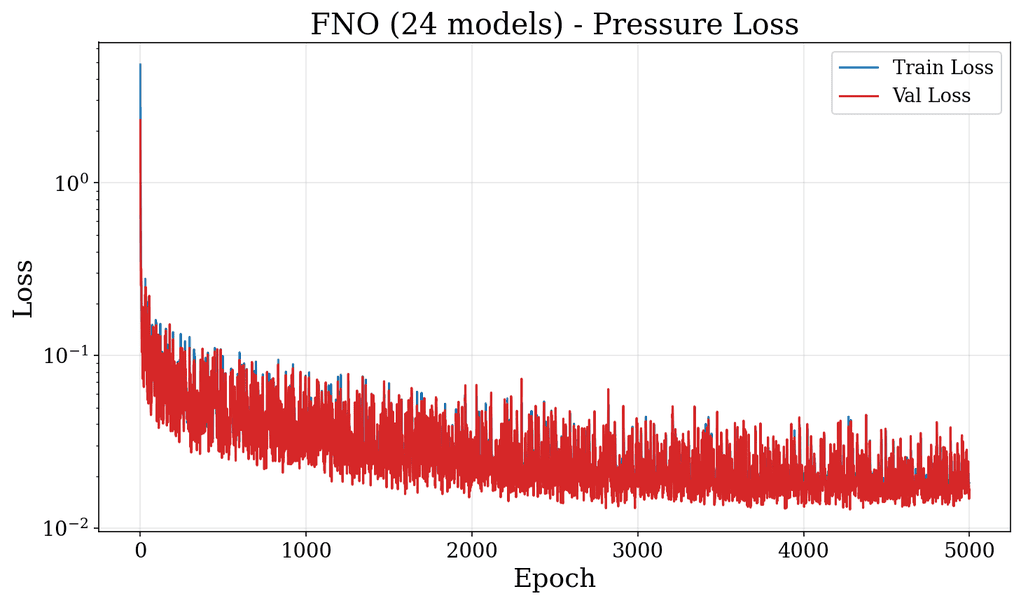}
        \caption{FNO (Pressure)}
        \label{fig:loss_fno_pres}
    \end{subfigure}
    \vspace{1em}
    \begin{subfigure}[t]{0.48\textwidth}
        \centering
        \includegraphics[width=\textwidth]{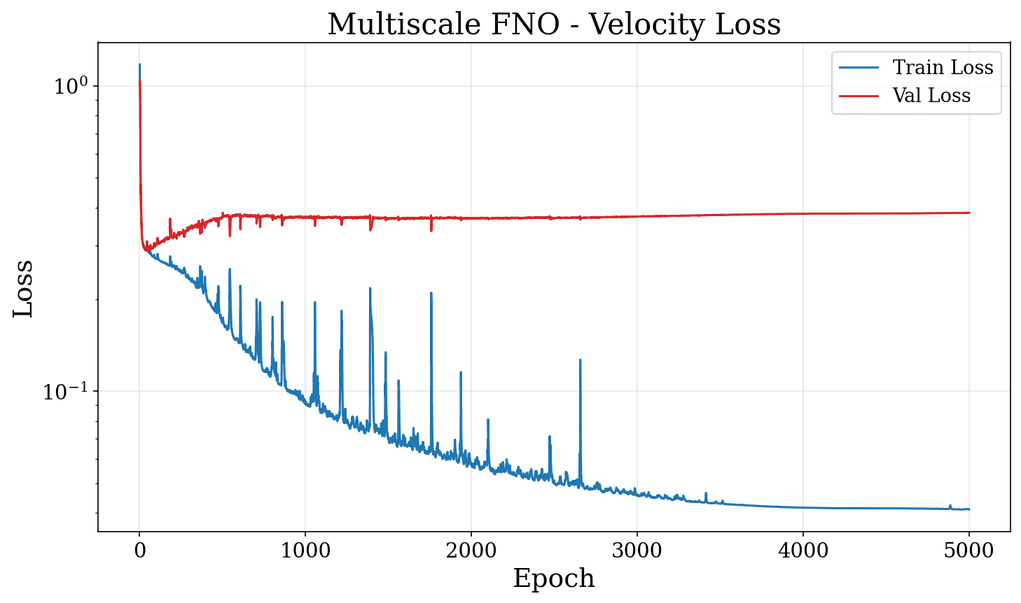}
        \caption{MscaleFNO (Velocity)}
        \label{fig:loss_msfno_vel}
    \end{subfigure}
    \hfill
    \begin{subfigure}[t]{0.48\textwidth}
        \centering
        \includegraphics[width=\textwidth]{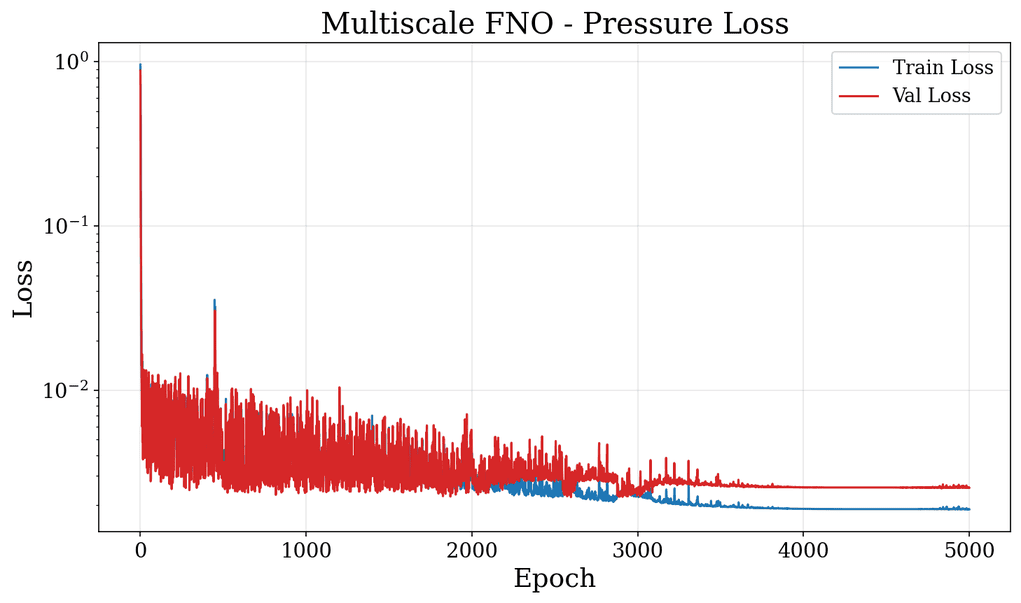}
        \caption{MscaleFNO (Pressure)}
        \label{fig:loss_msfno_pres}
    \end{subfigure}
    \caption{Training and validation loss curves for FNO and MscaleFNO models on the velocity and pressure fields.}
    \label{fig:loss_curves_fno_msfno}
\end{figure}

Figure~\ref{fig:loss_curves_ae_ldon} presents the training and validation loss curves for the AE and L-DeepONet models. The AEs exhibit stable convergence for both velocity and pressure fields, while the multi-scale L-DeepONet models show signs of overfitting, particularly in the velocity field, where the validation loss plateaus while the training loss continues to decrease. Figure~\ref{fig:loss_curves_fno_msfno} shows the corresponding loss curves for the FNO and MscaleFNO models, both of which demonstrate smooth convergence for the pressure field and relatively slower convergence for the velocity field.

\begin{table}[H]
\centering
\caption{Training hyperparameters for AE models.}
\label{tab:hyperparameters_ae}
\renewcommand{\arraystretch}{1.3}
\begin{tabular}{lcc}
\toprule
\textbf{Hyperparameter} & \textbf{MLP AE} & \textbf{CAE} \\
\midrule
Input dimension     & 28{,}877                  & $(282, 756, 1)$ \\
Latent dimension    & 256                       & 256 \\
Activation function & ReLU                      & ReLU \\
Optimizer           & AdamW                     & AdamW \\
Learning rate       & $1 \times 10^{-4}$        & $1 \times 10^{-4}$ \\
Batch size          & 64                        & 64 \\
Epochs              & 10{,}000                   & 5{,}000 \\
Loss function       & MSE                       & MSE \\
\bottomrule
\end{tabular}
\end{table}

\begin{table}[H]
\centering
\caption{Training hyperparameters for DeepONet models.}
\label{tab:hyperparameters_don}
\renewcommand{\arraystretch}{1.5}
\resizebox{\textwidth}{!}{%
\begin{tabular}{lccc}
\toprule
\textbf{Hyperparameter} & \textbf{Vanilla L-DeepONet} & \textbf{MLP AE + MS L-DeepONet} & \textbf{CAE + MS L-DeepONet} \\
\midrule
Input dimension         & 256           & 256                       & 256 \\
Basis functions $p$     & 16            & 16                        & 16 \\
Number of scales $S$    & --            & 10                        & 9 \\
Scale factors           & --            & 1, 10, 20, 30, 40,        & 1, 10, 20, 30, 40, \\
                        &               & 50, 100, 200, 300, 500    & 50, 100, 200, 300 \\
Branch net layers       & Dense $\times 6$ & Dense $\times 6$       & Dense $\times 6$ \\
Branch net activation   & ReLU          & ReLU                      & ReLU \\
Branch net output       & $256 \times 16$  & $256 \times 16 \times 10$ & $256 \times 16 \times 9$ \\
Trunk sub-nets          & 1             & 10                        & 9 \\
Trunk net layers        & Dense $\times 3$ & Dense $\times 3$       & Dense $\times 3$ \\
Trunk net activation    & Sin           & Sin                       & Sin \\
Optimizer               & Adam          & Adam                      & Adam \\
Learning rate           & $1 \times 10^{-4}$ (Exp. decay) & $1 \times 10^{-4}$ (Exp. decay) & $1 \times 10^{-4}$ (Exp. decay) \\
Batch size              & 8             & 8                         & 8 \\
Epochs                  & 30{,}000       & 5{,}000                   & 5{,}000 \\
Loss function           & MSE           & MSE                       & MSE \\
\bottomrule
\end{tabular}%
}
\end{table}

\begin{table}[H]
\centering
\caption{Training hyperparameters for FNO and MscaleFNO models.}
\label{tab:hyperparameters_fno}
\renewcommand{\arraystretch}{1.3}
\resizebox{\textwidth}{!}{%
\begin{tabular}{lcc}
\toprule
\textbf{Hyperparameter} & \textbf{FNO} & \textbf{MscaleFNO} \\
\midrule
Input dimension             & $(282, 756, 1)$               & $(282, 756, 1)$ \\
Fourier modes $(k_1, k_2)$  & $(24, 24)$                    & $(24, 24)$ \\
Hidden channel width        & 128                           & 16 \\
Number of Fourier layers    & 4                             & 4 \\
Number of sub-networks      & --                            & 6 \\
Scale factors               & --                            & 1, 40, 80, 100, 140, 200 \\
Activation function         & GELU                          & Sin \\
Optimizer                   & AdamW                         & AdamW \\
Weight decay                & $1 \times 10^{-4}$            & $1 \times 10^{-4}$ \\
Learning rate               & $1 \times 10^{-3}$            & $1 \times 10^{-3}$ \\
Batch size                  & 6                             & 6 \\
Epochs                      & 5{,}000                       & 5{,}000 \\
Loss function               & Relative $L^2$                & Relative $L^2$ \\
\bottomrule
\end{tabular}%
}
\end{table}

\begin{table}[htbp]
    \centering
    \caption{Summary of model parameters, sizes, and training costs for each architecture.}
    \label{tab:model_summary_compact}
    \renewcommand{\arraystretch}{1.3}
    \resizebox{\textwidth}{!}{%
    \begin{tabular}{llrrrr}
        \toprule
        \textbf{Data Type} & \textbf{Model} & \textbf{Parameters} & \textbf{Size (MB)} & \textbf{GPU Memory} & \textbf{Training Time} \\
        \midrule
        \multirow{2}{*}{Unstructured} 
        & MLP-based AE                    & 549,985,741 & 2,098.03 & $\approx$17 GB   & $\approx$5.5 h \\
        & Multi-scale L-DeepONet          & 7,040,608   & 26.86    & $\approx$1 GB    & $\approx$5 min \\
        \midrule
        \multirow{4}{*}{Structured} 
        & CAE                             & 28,965,409  & 110.49   & $\approx$17 GB   & $\approx$5.5 h \\
        & Multi-scale L-DeepONet          & 6,340,272   & 24.19    & $\approx$1 GB    & $\approx$5 min \\
        \cmidrule(l){2-6}
        & FNO                             & 20,672,483  & 78.86    & $\approx$29.9 GB & $\approx$5.9 h \\
        & MscaleFNO                       & 7,142,814   & 27.25    & $\approx$26.2 GB & $\approx$5.5 h \\
        \bottomrule
    \end{tabular}%
    }
\end{table}

\section{Pressure Field Prediction}
\label{app:pressure_field}

This appendix presents the predicted pressure field contours for all four surrogate models at selected time steps. The visualizations complement the velocity field comparisons discussed in Section~\ref{subsec6.2}, providing additional insight into each model's ability to capture the spatial distribution and temporal evolution of the pressure field under both interpolation (inlet velocity of 0.4 m/s) and extrapolation (inlet velocity of 0.7 m/s) conditions.

\begin{figure}[H]
    \centering
    \setlength{\tabcolsep}{1pt}
    \makebox[\textwidth][c]{%
    \begin{tabular}{c@{\hspace{4pt}}ccc}
        & \textbf{$t = 2$} & \textbf{$t = 50$} & \textbf{$t = 100$} \\
        \adjustbox{valign=c}{\rotatebox[origin=c]{90}{\small\textbf{Reference}}} &
        \includegraphics[width=0.45\textwidth,valign=c]{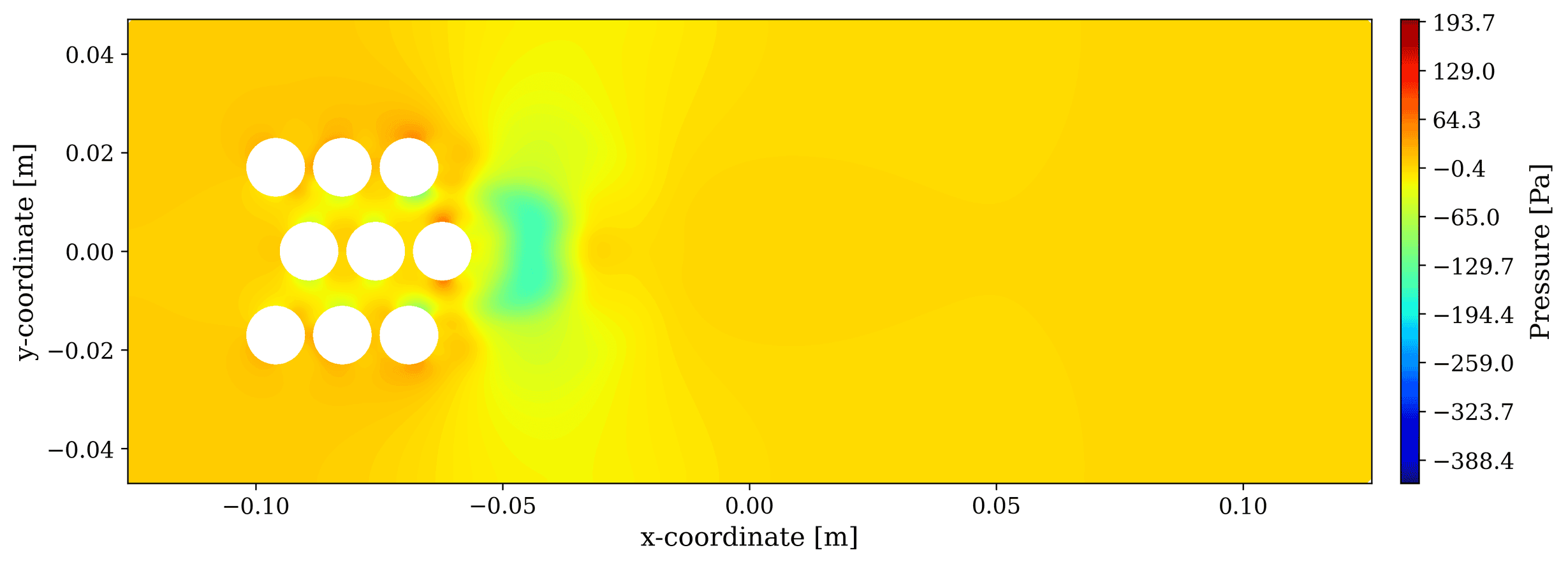} &
        \includegraphics[width=0.45\textwidth,valign=c]{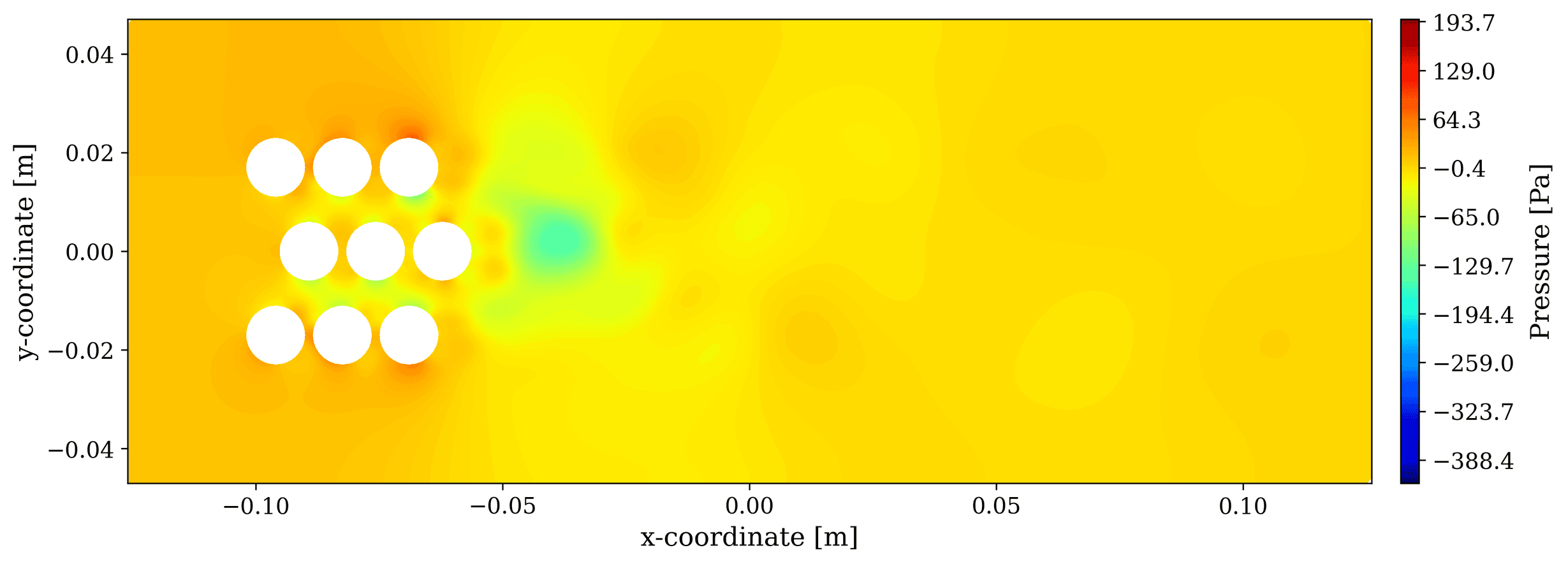} &
        \includegraphics[width=0.45\textwidth,valign=c]{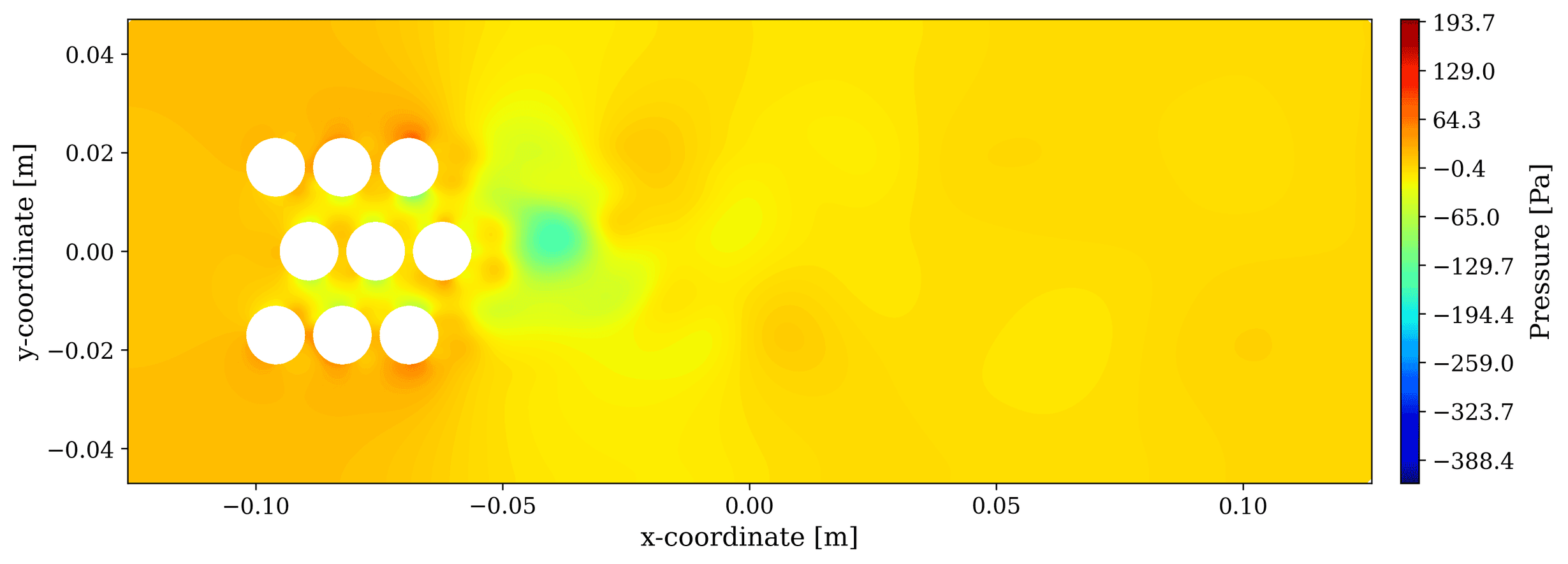} \\[2pt]
        \adjustbox{valign=c}{\rotatebox[origin=c]{90}{\small\textbf{Restored}}} &
        \includegraphics[width=0.45\textwidth,valign=c]{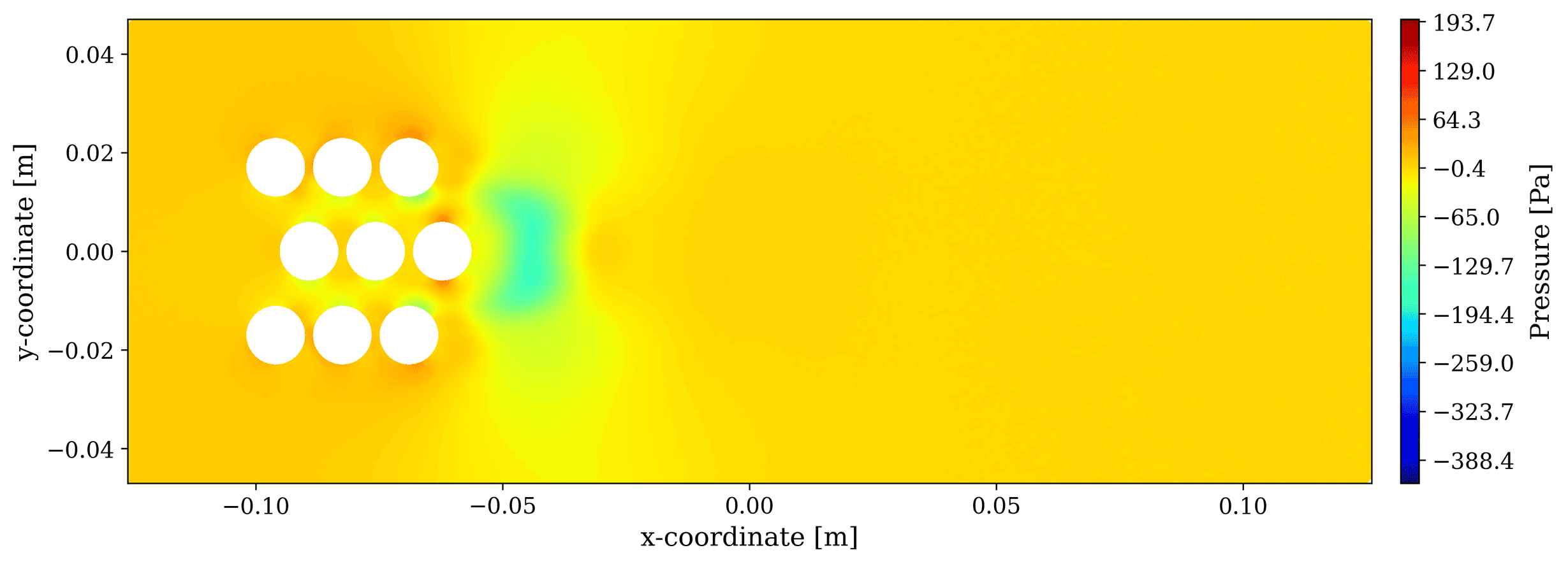} &
        \includegraphics[width=0.45\textwidth,valign=c]{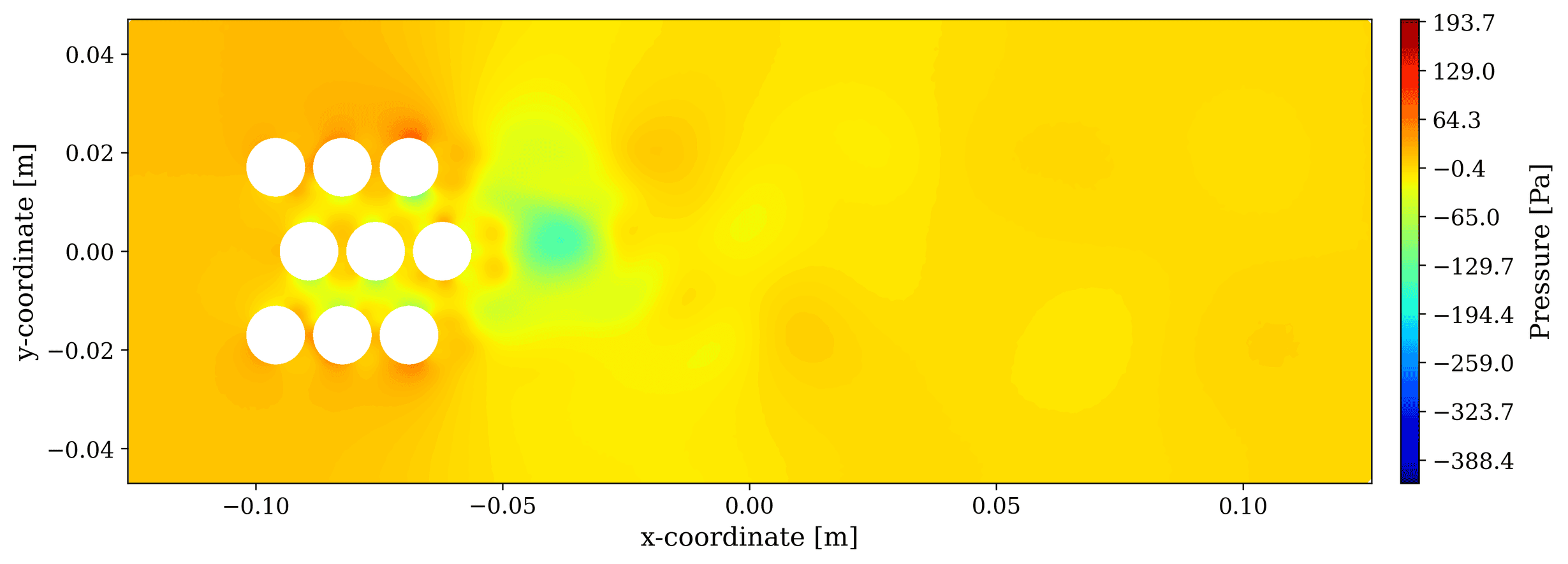} &
        \includegraphics[width=0.45\textwidth,valign=c]{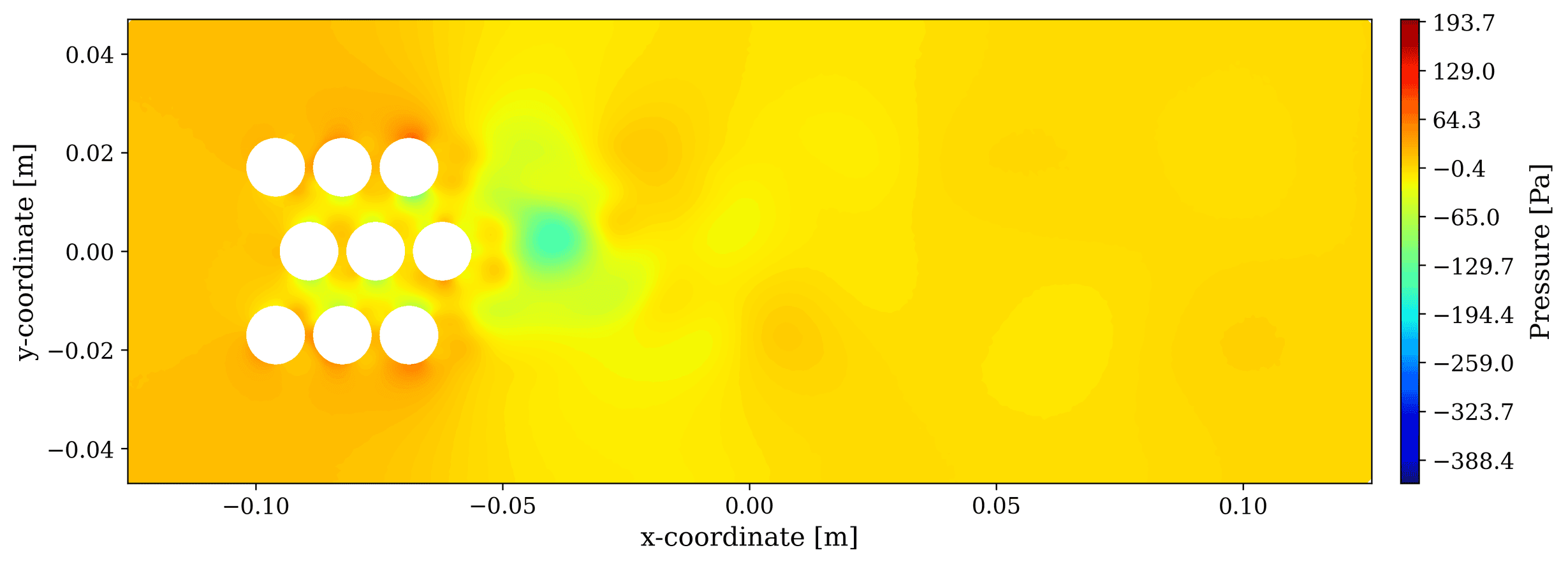} \\[2pt]
        \adjustbox{valign=c}{\rotatebox[origin=c]{90}{\small\textbf{Error}}} &
        \includegraphics[width=0.45\textwidth,valign=c]{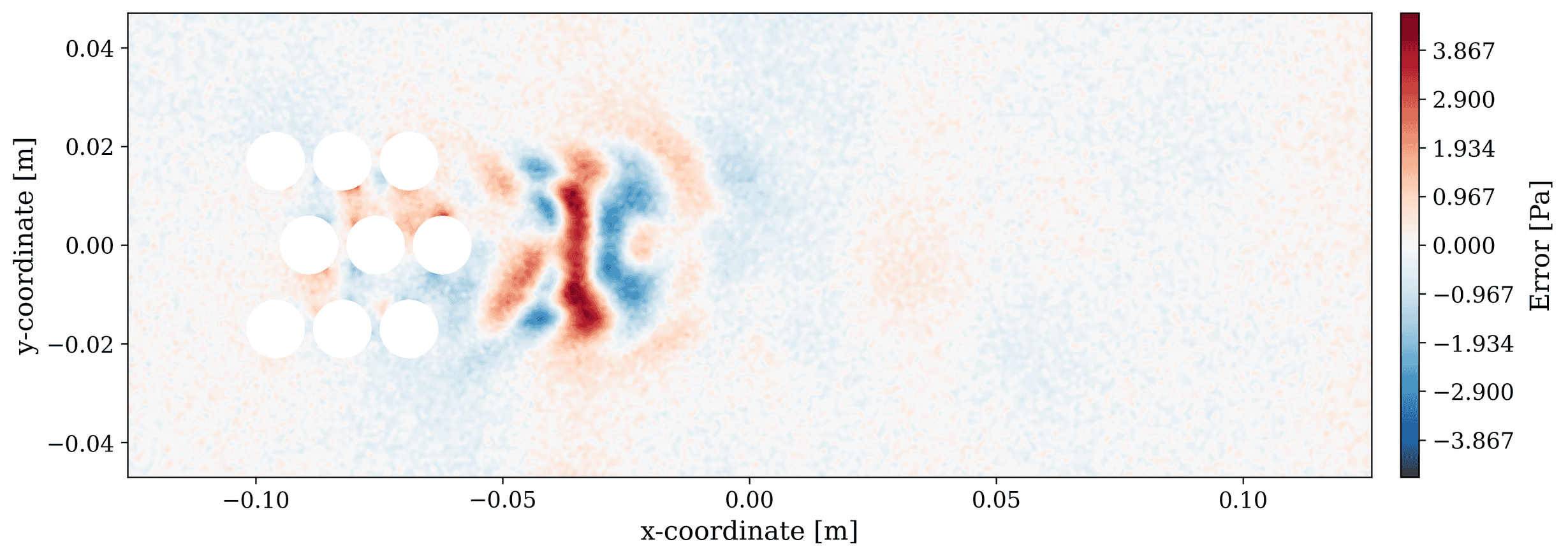} &
        \includegraphics[width=0.45\textwidth,valign=c]{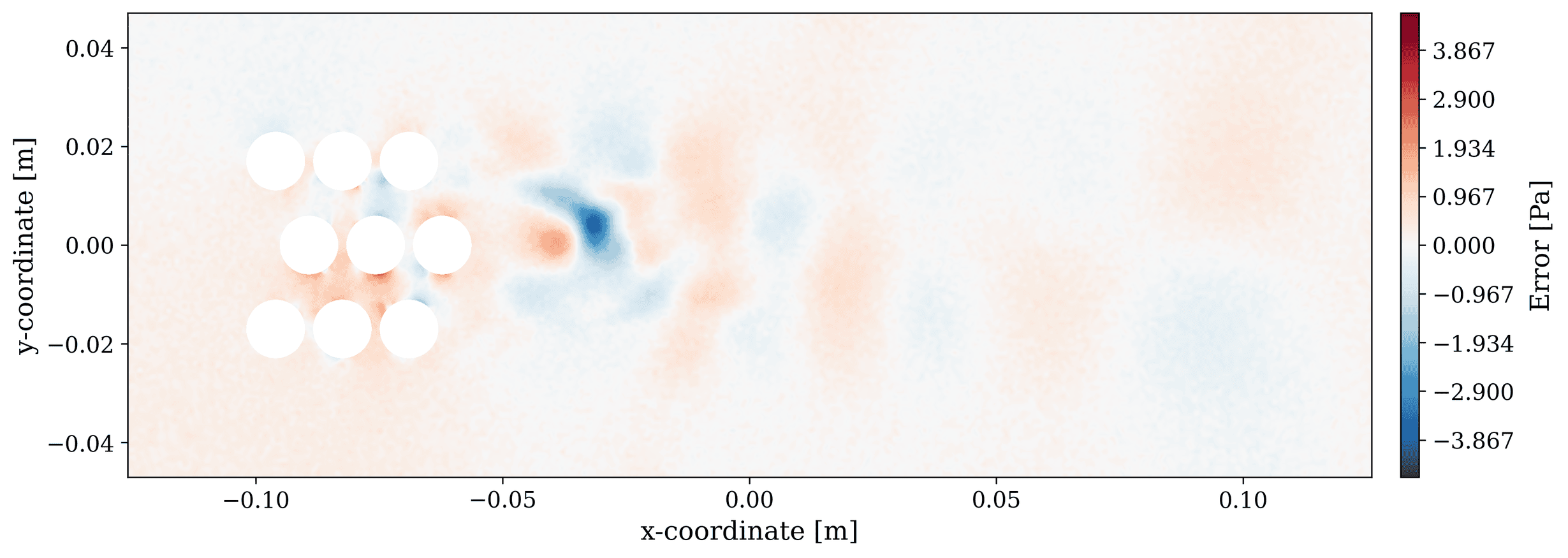} &
        \includegraphics[width=0.45\textwidth,valign=c]{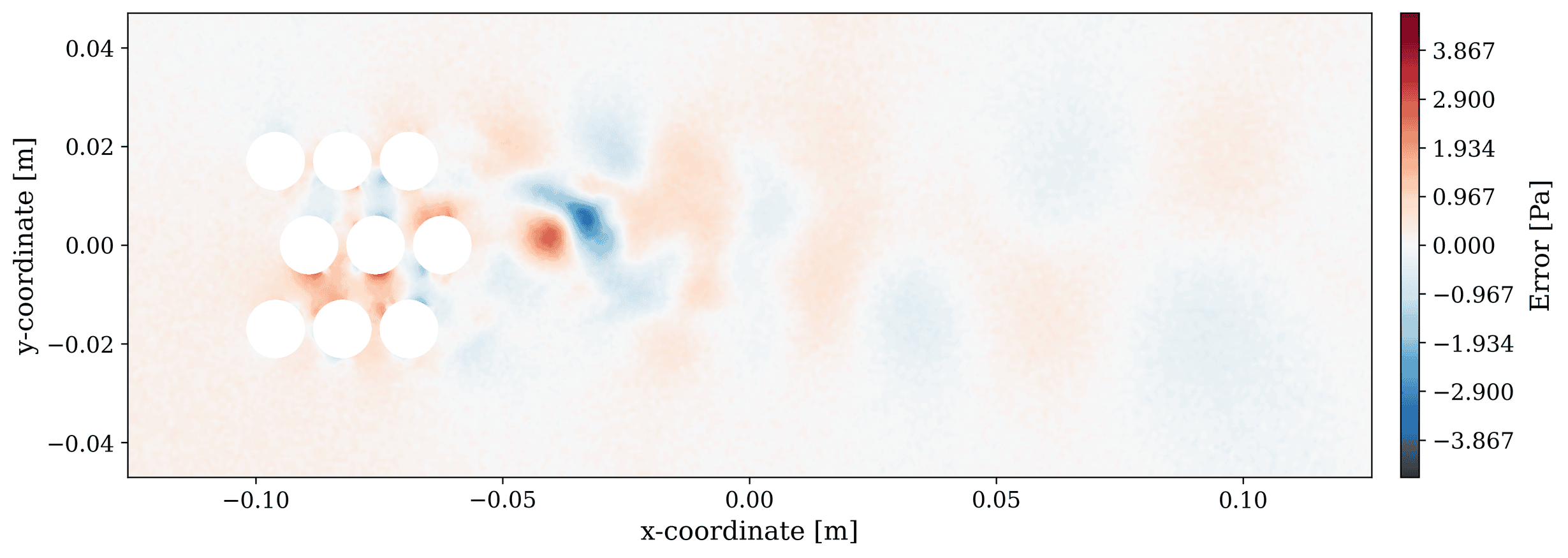} \\
    \end{tabular}
    }%
    \caption{MLP-based AE results for pressure field with inlet velocity 0.4 $m/s$. Reference, restored, and error at different timesteps.}
    \label{fig:ae_pressure_inlet040}
\end{figure}


\begin{figure}[H]
    \centering
    \setlength{\tabcolsep}{1pt}
    \makebox[\textwidth][c]{%
    \begin{tabular}{c@{\hspace{4pt}}ccc}
        & \textbf{$t = 2$} & \textbf{$t = 50$} & \textbf{$t = 100$} \\
        \adjustbox{valign=c}{\rotatebox[origin=c]{90}{\small\textbf{Reference}}} &
        \includegraphics[width=0.45\textwidth,valign=c]{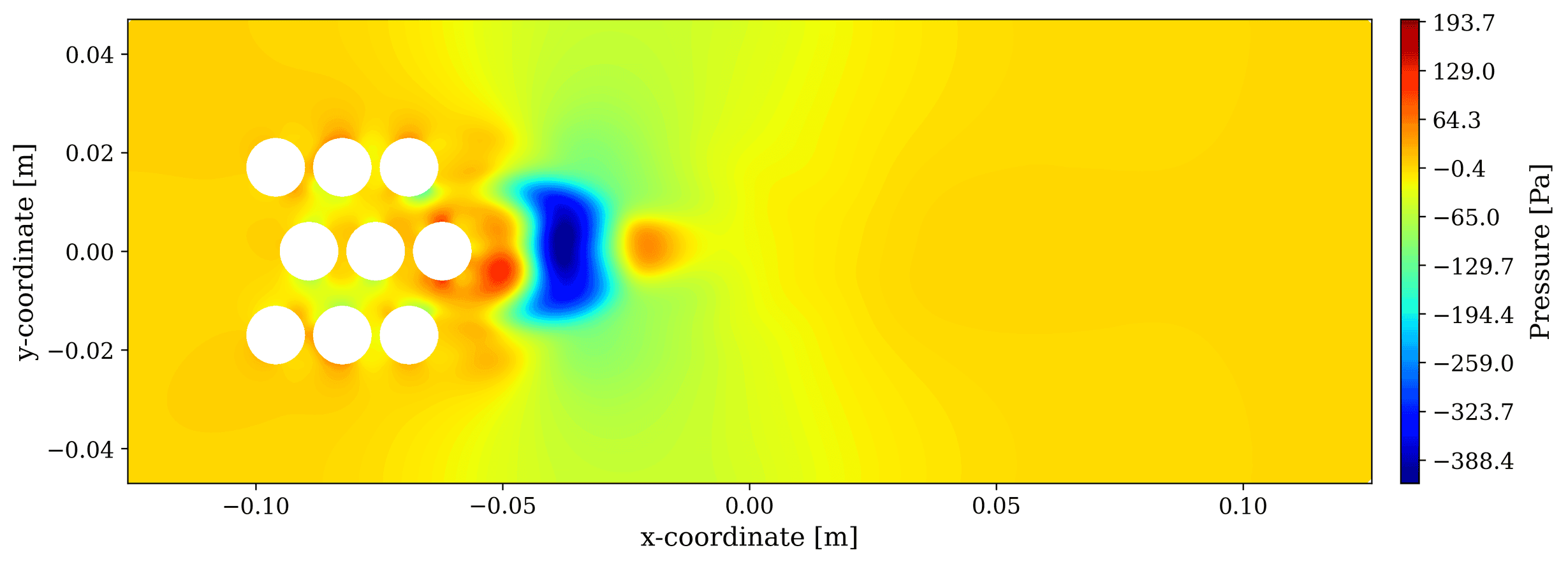} &
        \includegraphics[width=0.45\textwidth,valign=c]{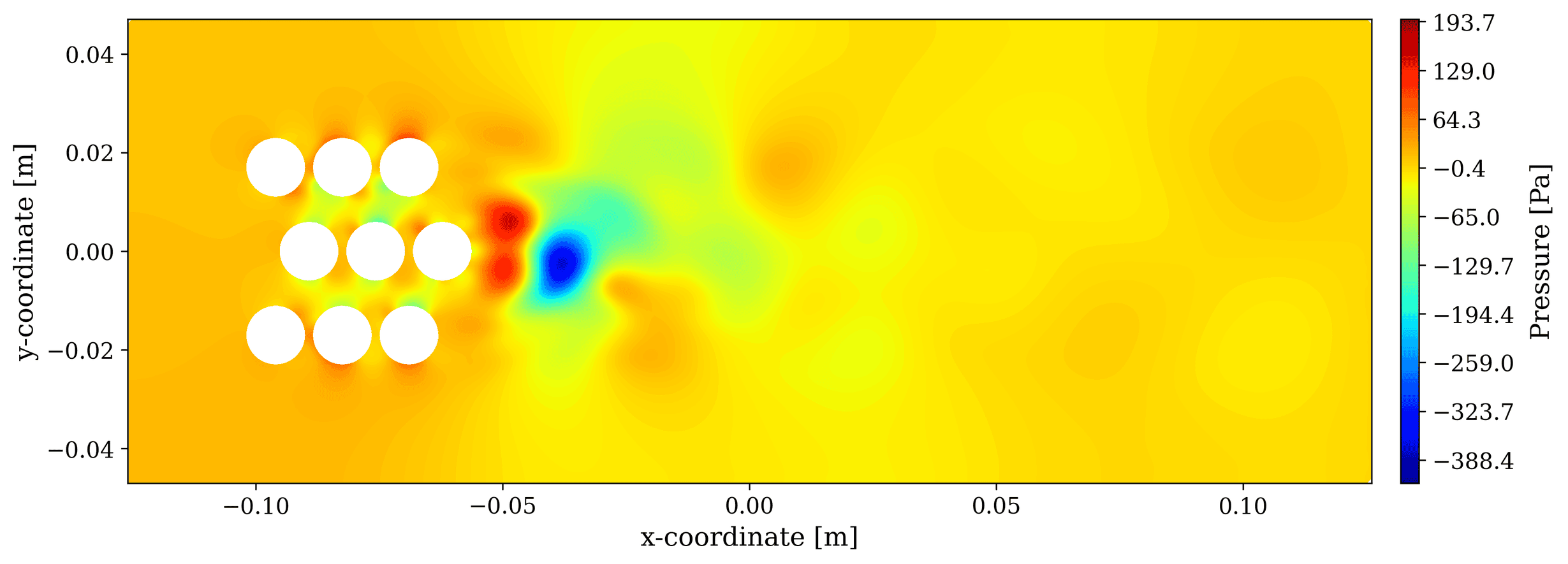} &
        \includegraphics[width=0.45\textwidth,valign=c]{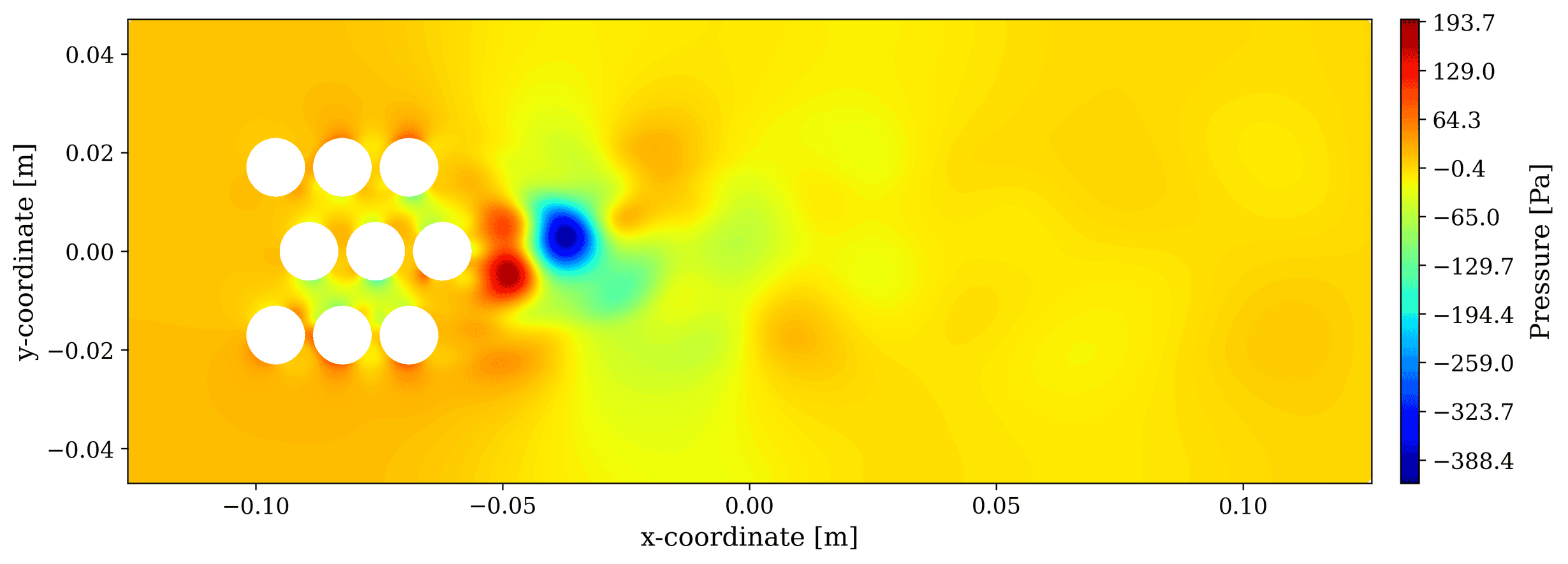} \\[2pt]
        \adjustbox{valign=c}{\rotatebox[origin=c]{90}{\small\textbf{Restored}}} &
        \includegraphics[width=0.45\textwidth,valign=c]{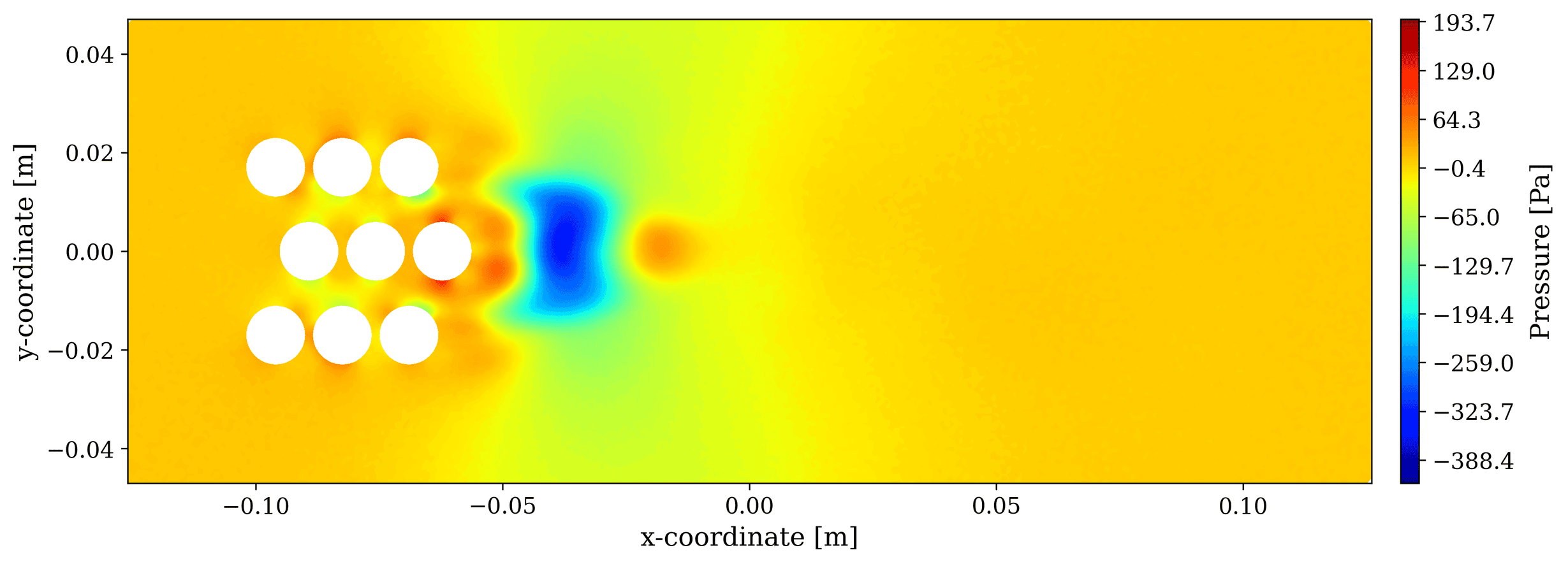} &
        \includegraphics[width=0.45\textwidth,valign=c]{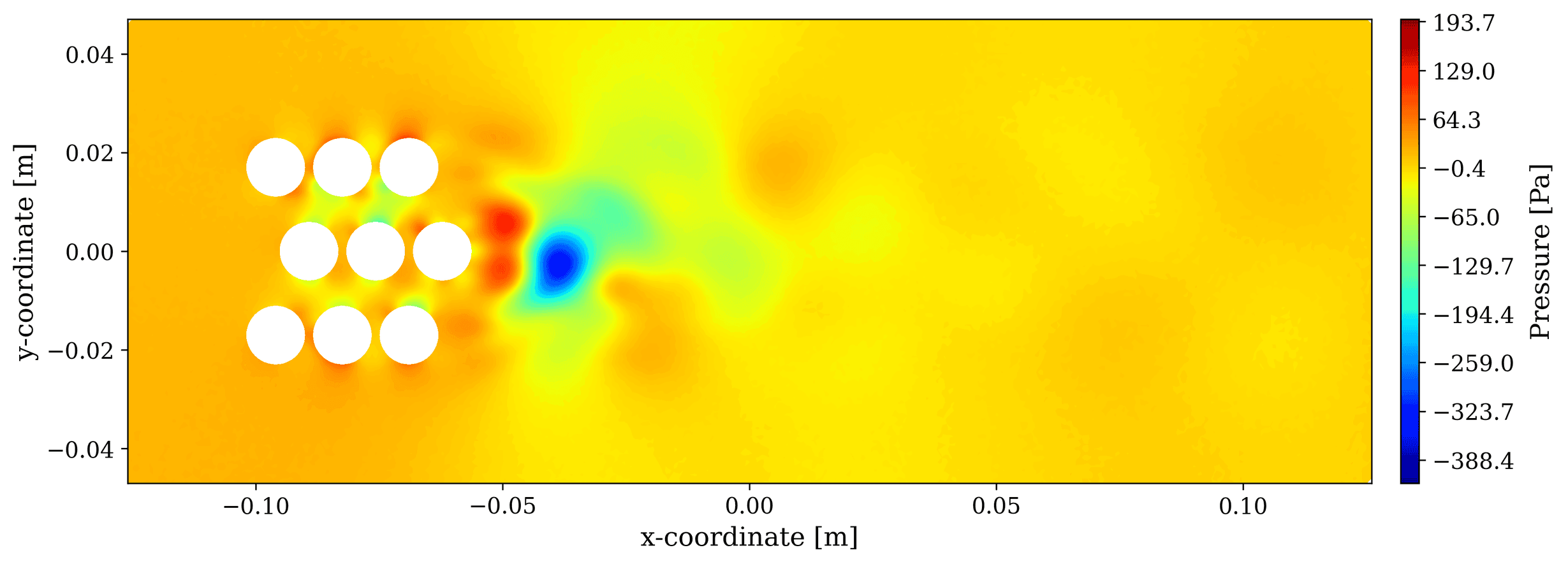} &
        \includegraphics[width=0.45\textwidth,valign=c]{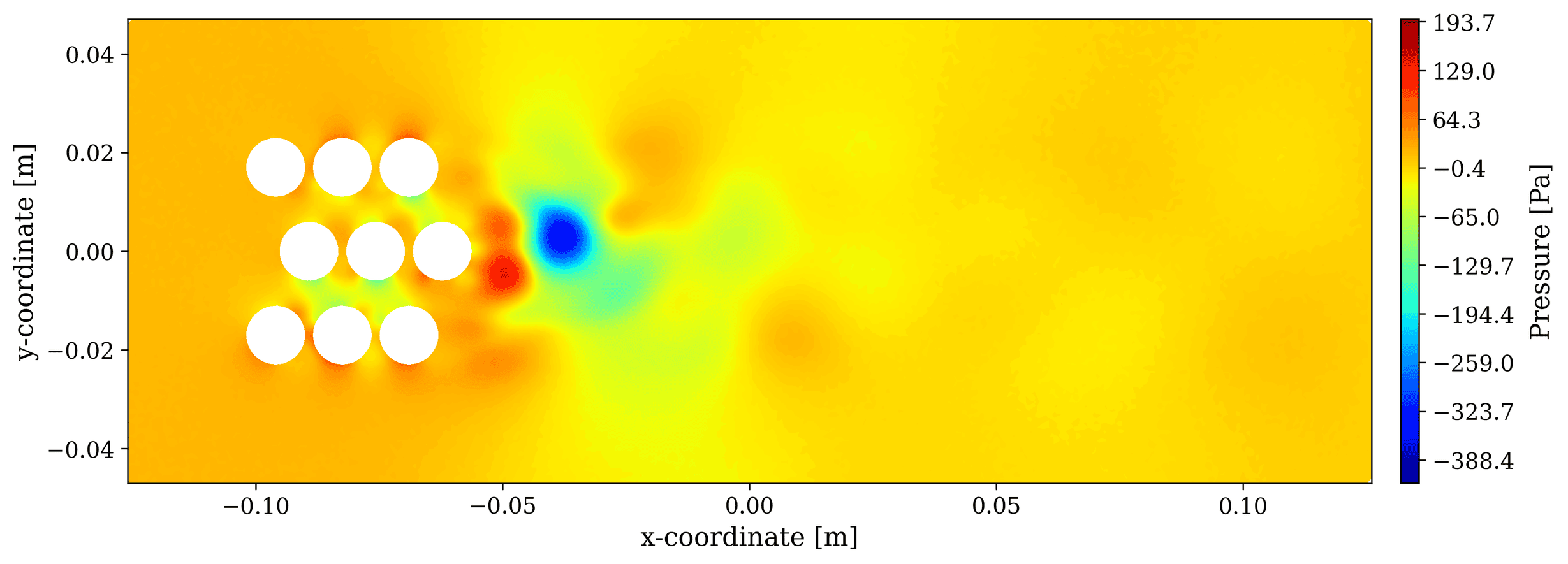} \\[2pt]
        \adjustbox{valign=c}{\rotatebox[origin=c]{90}{\small\textbf{Error}}} &
        \includegraphics[width=0.45\textwidth,valign=c]{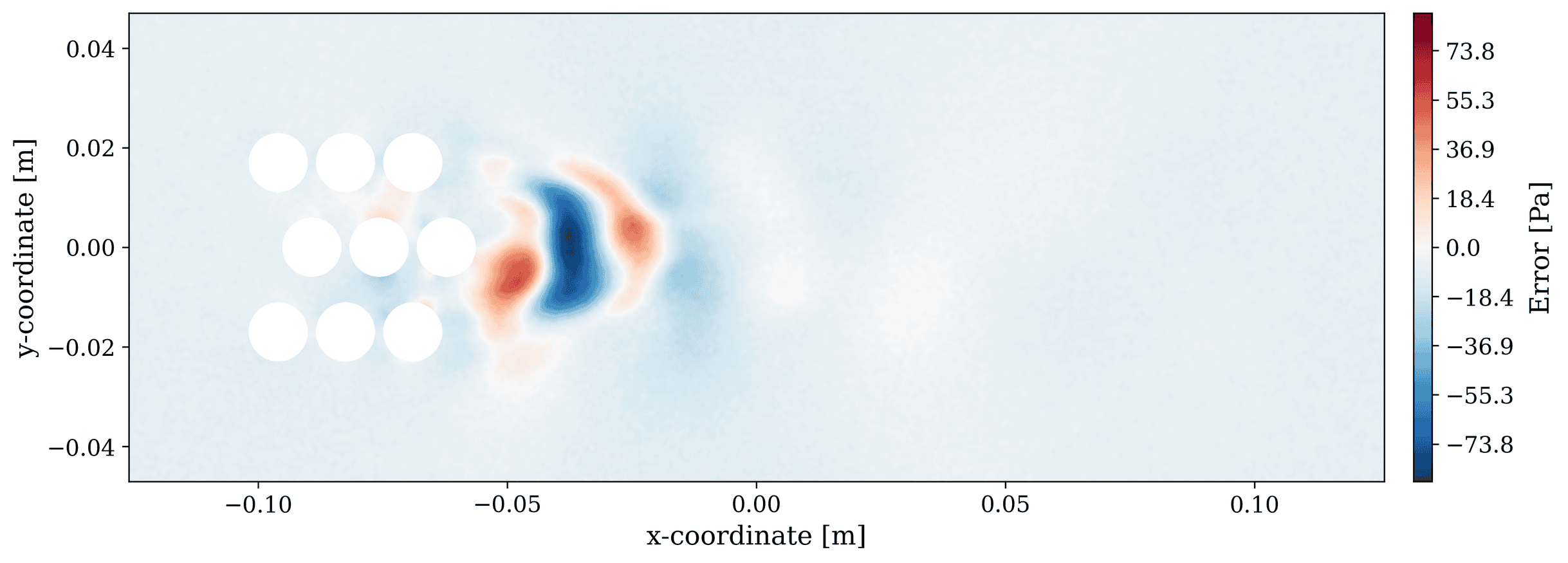} &
        \includegraphics[width=0.45\textwidth,valign=c]{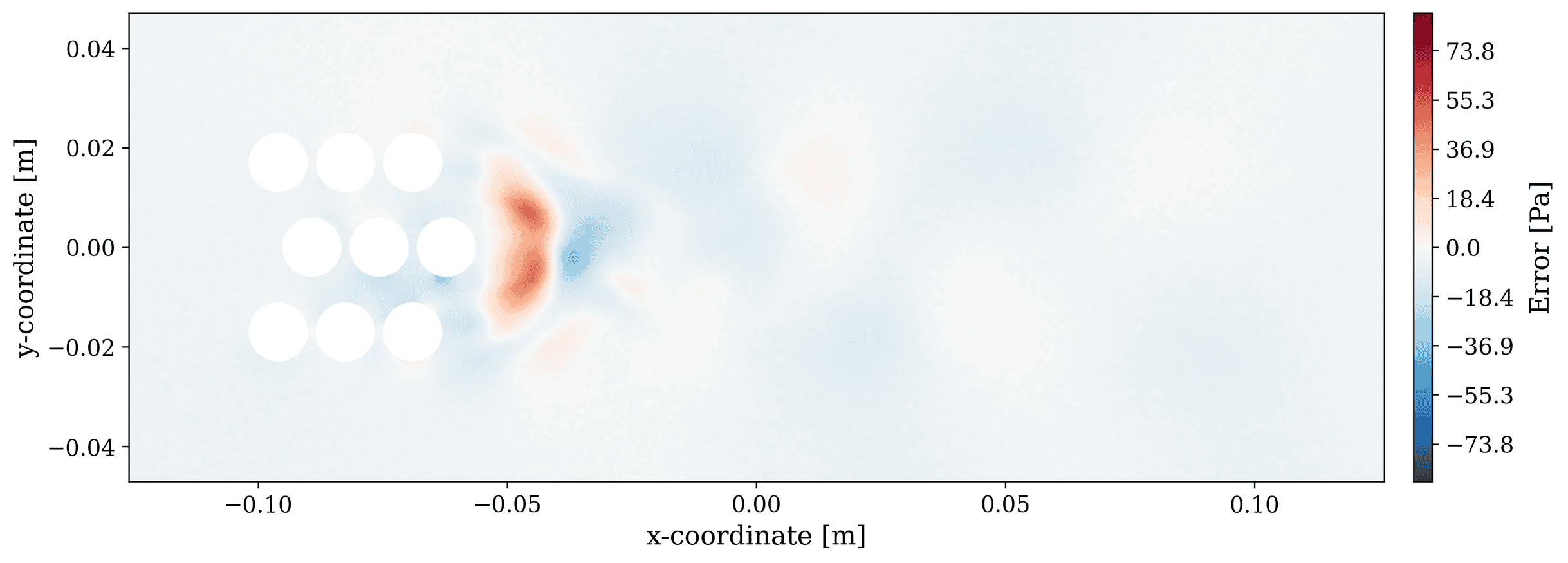} &
        \includegraphics[width=0.45\textwidth,valign=c]{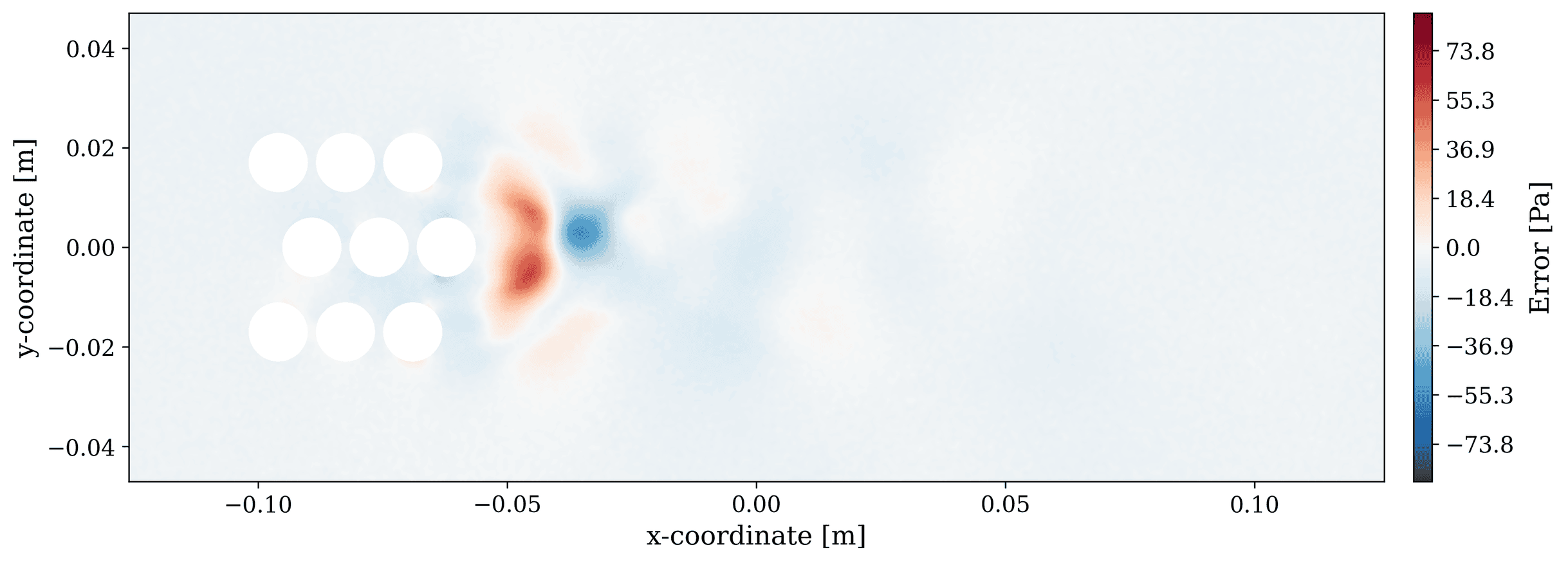} \\
    \end{tabular}
    }%
    \caption{MLP-based AE results for pressure field with inlet velocity 0.7 $m/s$. Reference, restored, and error at different timesteps.}
    \label{fig:ae_pressure_inlet070}
\end{figure}

\begin{figure}[H]
    \centering
    \setlength{\tabcolsep}{1pt}
    \makebox[\textwidth][c]{%
    \begin{tabular}{c@{\hspace{4pt}}ccc}
        & \textbf{$t = 2$} & \textbf{$t = 50$} & \textbf{$t = 100$} \\
        \adjustbox{valign=c}{\rotatebox[origin=c]{90}{\small\textbf{Reference}}} &
        \includegraphics[width=0.45\textwidth,valign=c]{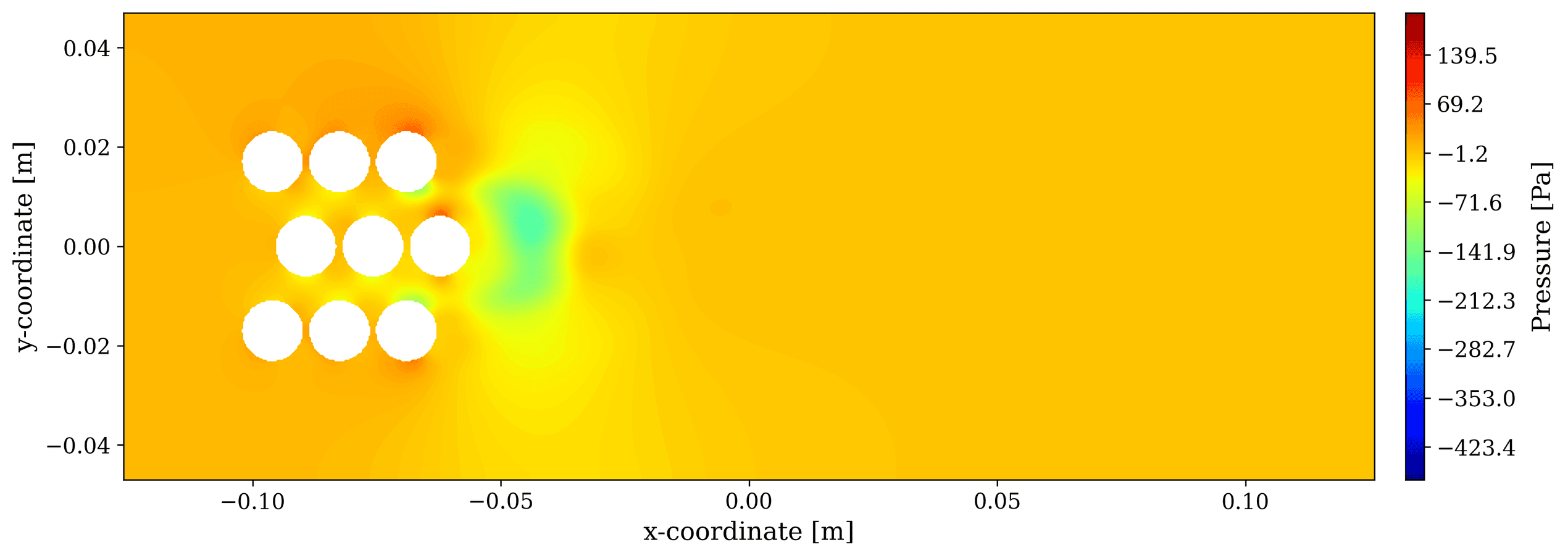} &
        \includegraphics[width=0.45\textwidth,valign=c]{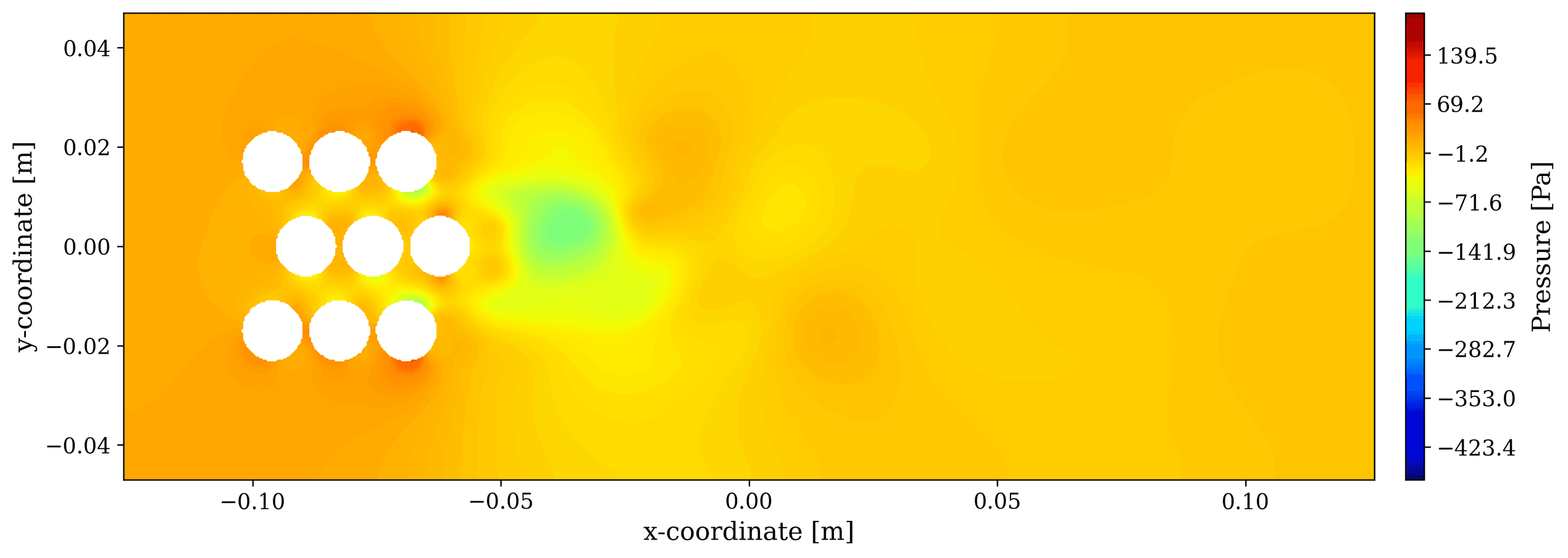} &
        \includegraphics[width=0.45\textwidth,valign=c]{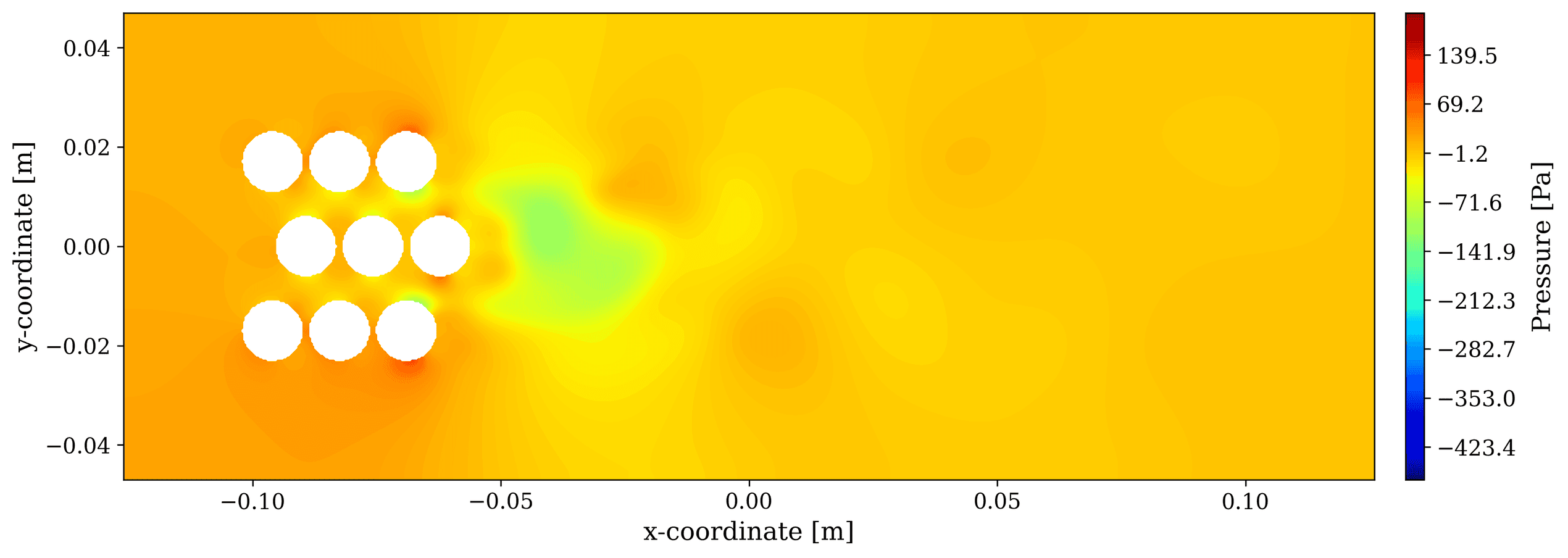} \\[2pt]
        \adjustbox{valign=c}{\rotatebox[origin=c]{90}{\small\textbf{Restored}}} &
        \includegraphics[width=0.45\textwidth,valign=c]{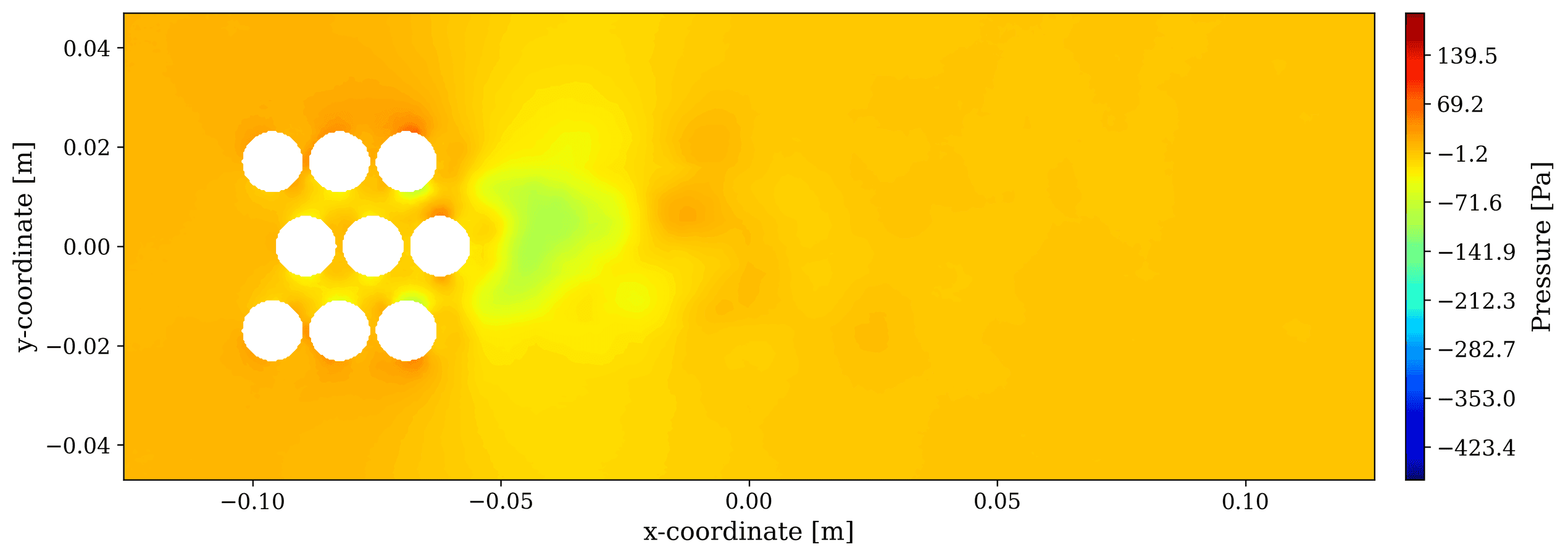} &
        \includegraphics[width=0.45\textwidth,valign=c]{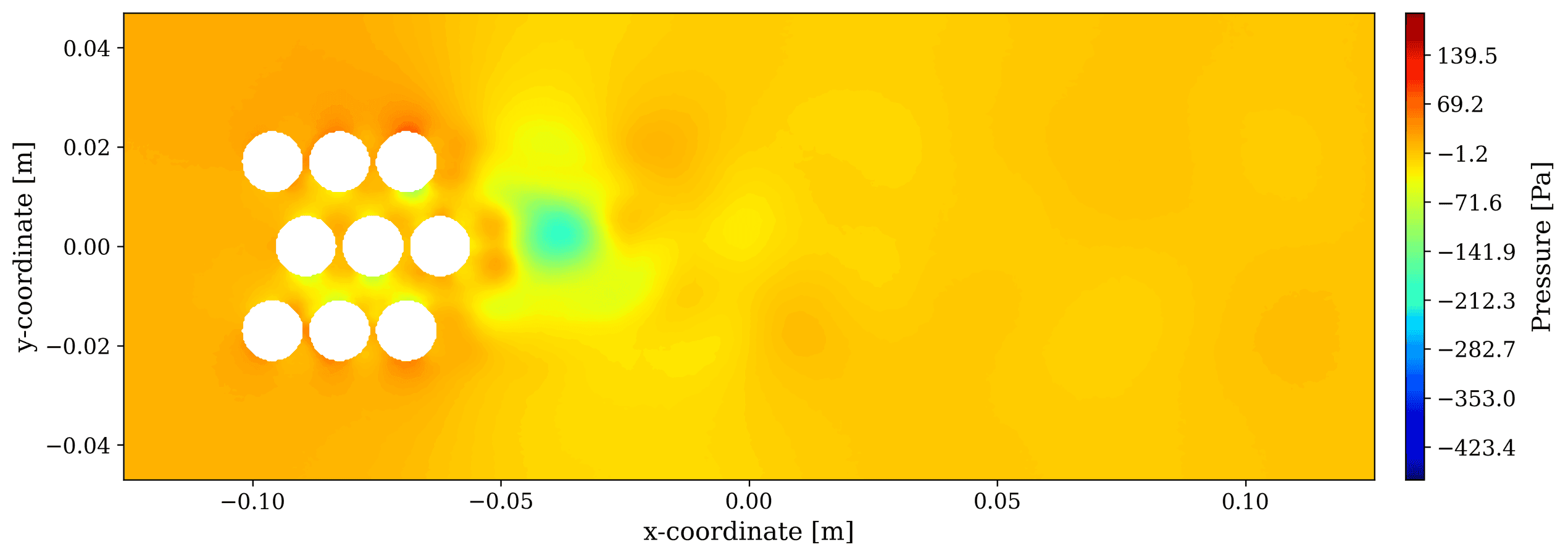} &
        \includegraphics[width=0.45\textwidth,valign=c]{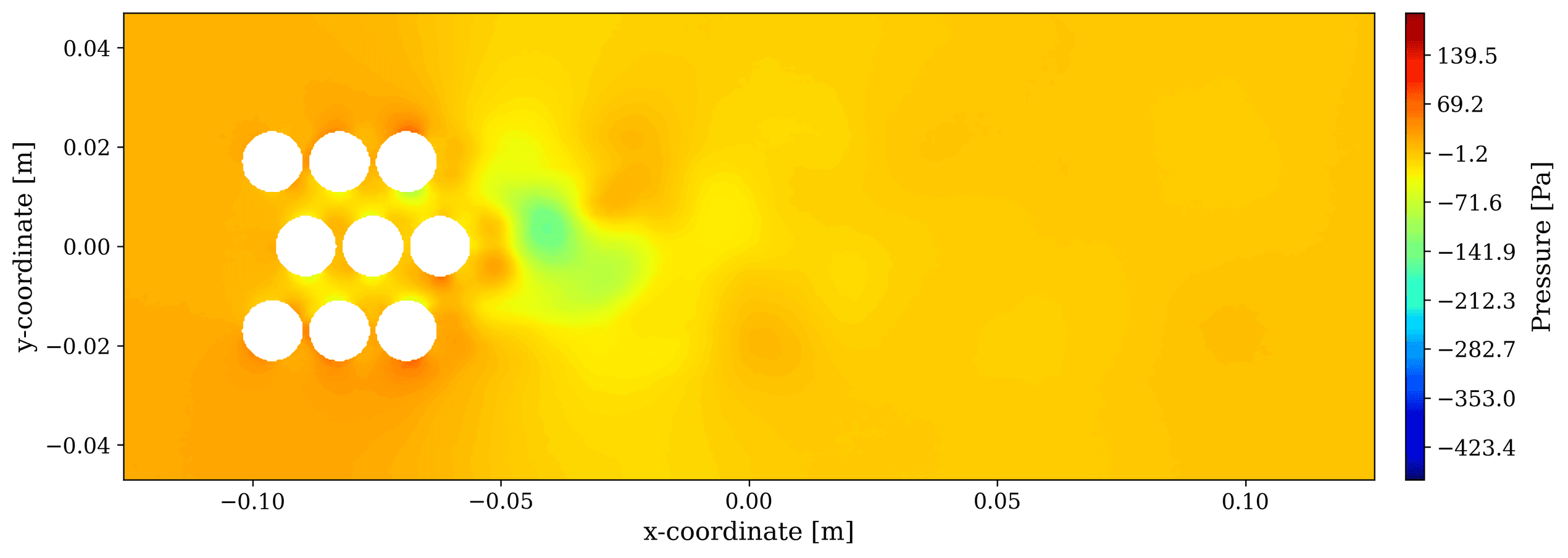} \\[2pt]
        \adjustbox{valign=c}{\rotatebox[origin=c]{90}{\small\textbf{Error}}} &
        \includegraphics[width=0.45\textwidth,valign=c]{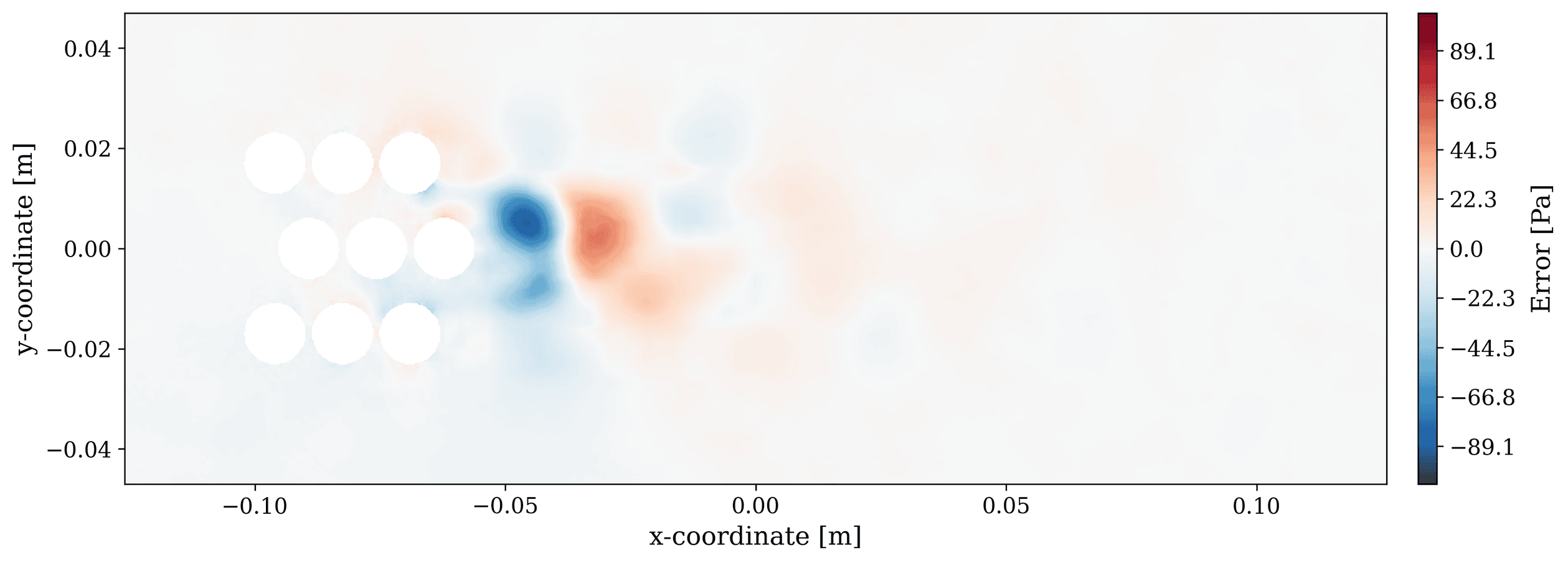} &
        \includegraphics[width=0.45\textwidth,valign=c]{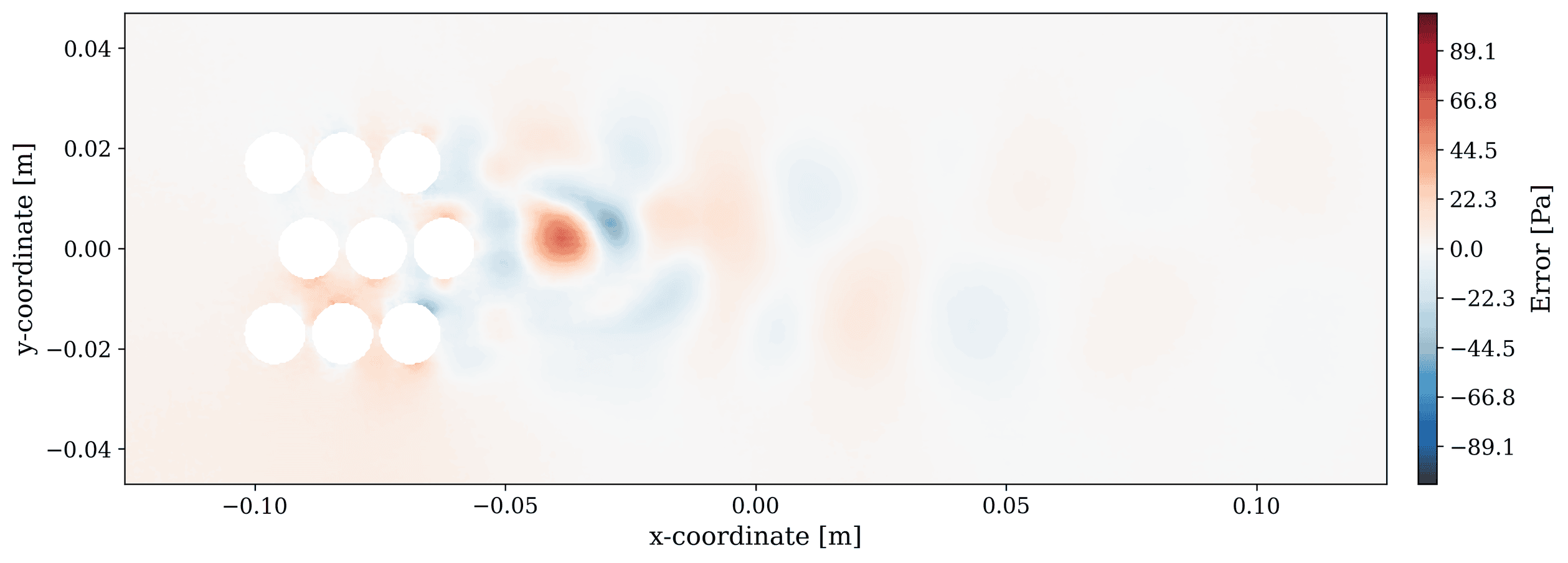} &
        \includegraphics[width=0.45\textwidth,valign=c]{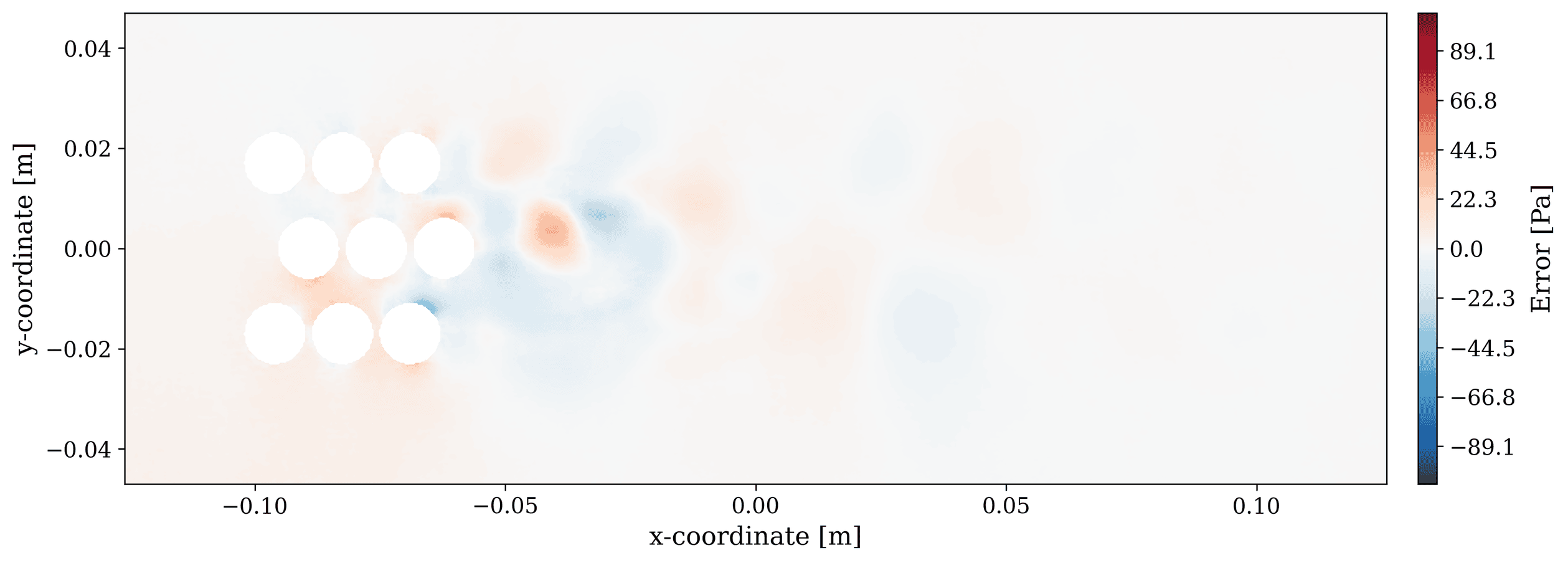} \\
    \end{tabular}
    }%
    \caption{CAE results for pressure field with inlet velocity 0.4 $m/s$. Reference, restored, and error at different timesteps.}
    \label{fig:cae_pressure_inlet040}
\end{figure}

\begin{figure}[H]
    \centering
    \setlength{\tabcolsep}{1pt}
    \makebox[\textwidth][c]{%
    \begin{tabular}{c@{\hspace{4pt}}ccc}
        & \textbf{$t = 2$} & \textbf{$t = 50$} & \textbf{$t = 100$} \\
        \adjustbox{valign=c}{\rotatebox[origin=c]{90}{\small\textbf{Reference}}} &
        \includegraphics[width=0.45\textwidth,valign=c]{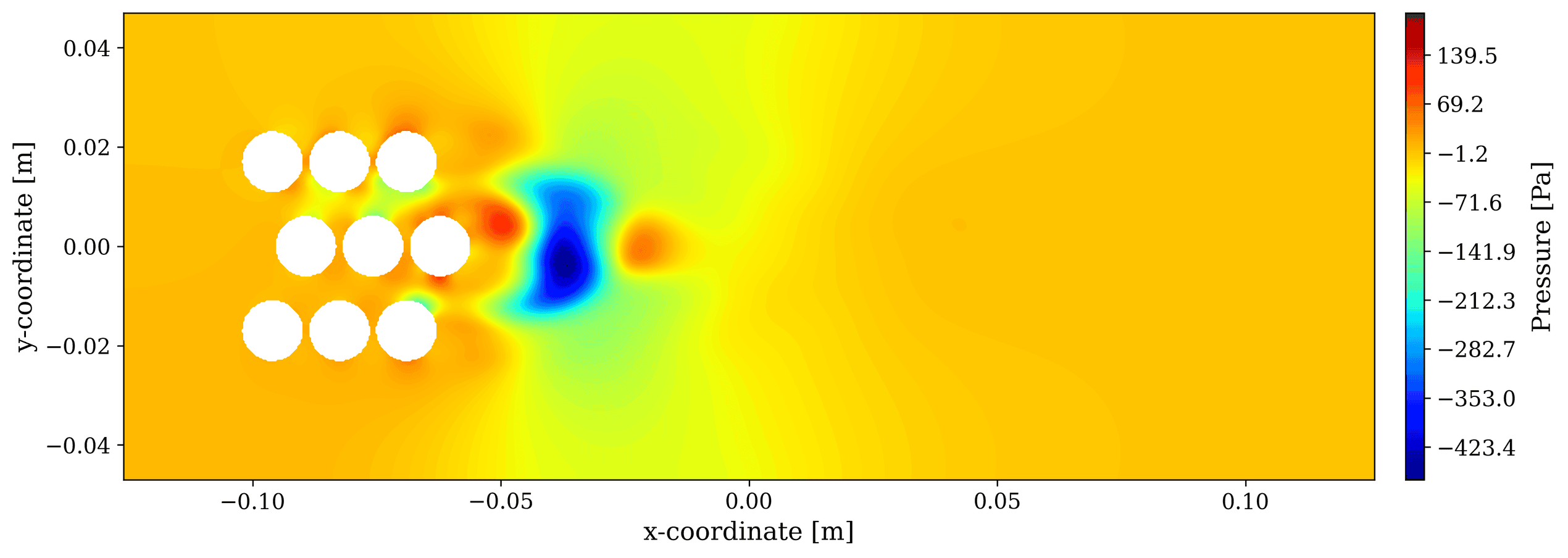} &
        \includegraphics[width=0.45\textwidth,valign=c]{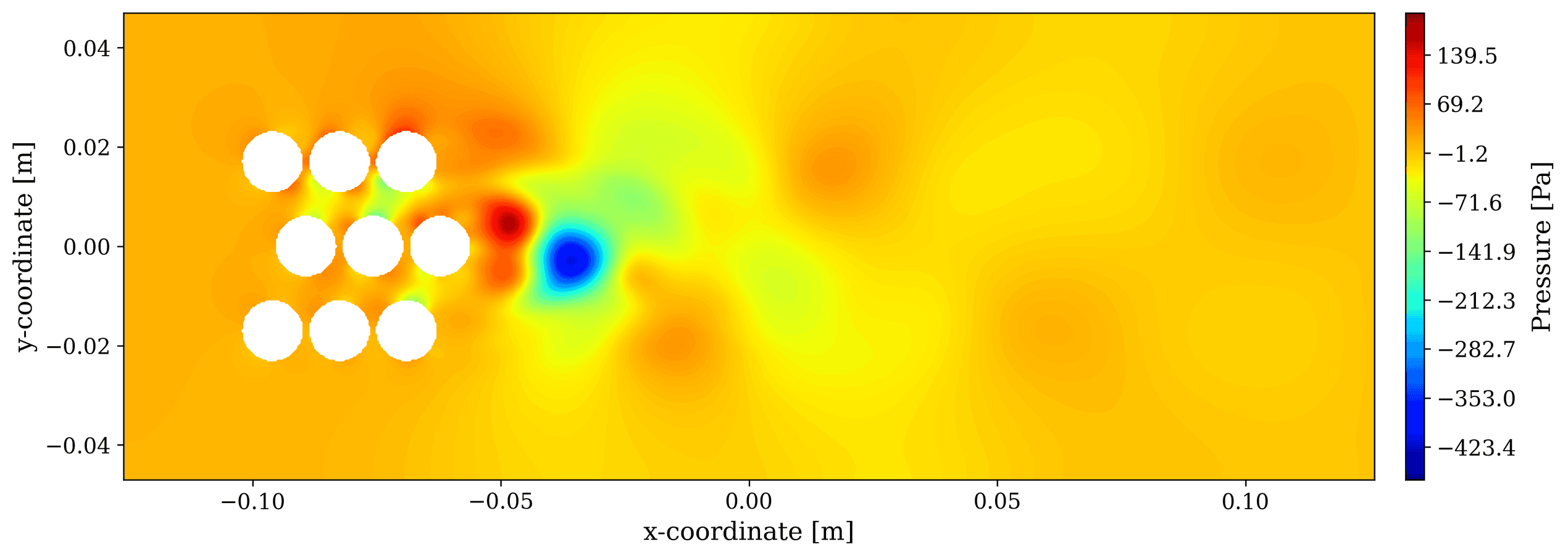} &
        \includegraphics[width=0.45\textwidth,valign=c]{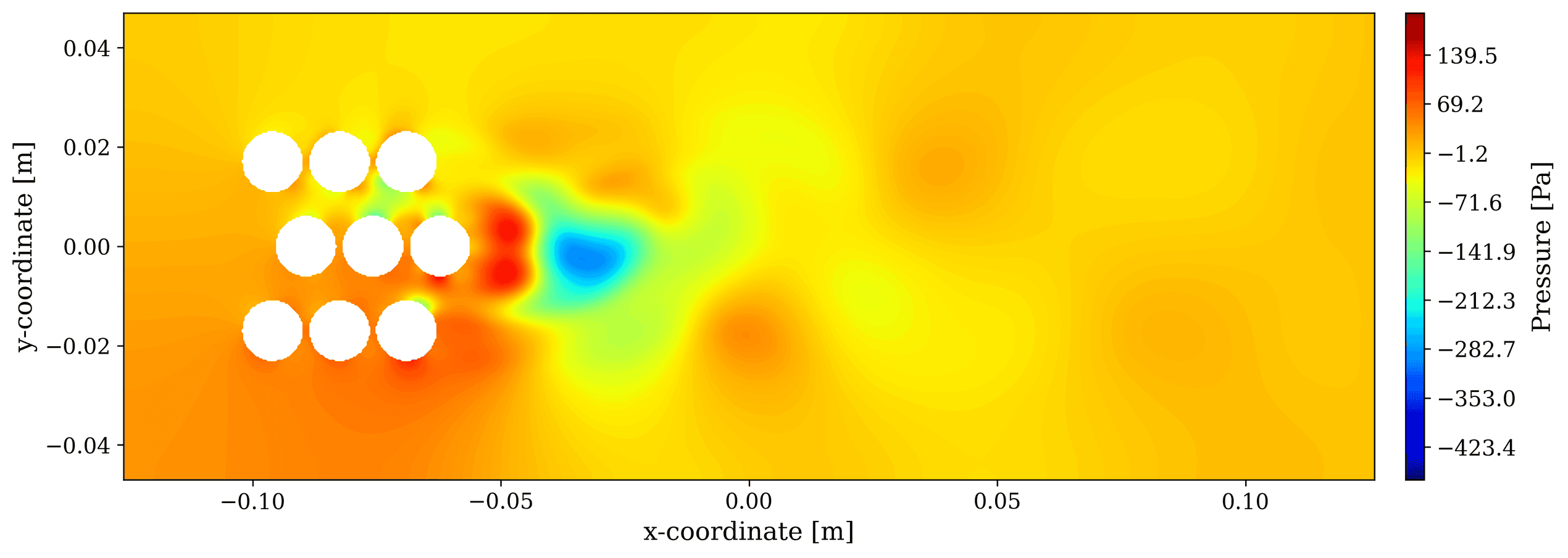} \\[2pt]
        \adjustbox{valign=c}{\rotatebox[origin=c]{90}{\small\textbf{Restored}}} &
        \includegraphics[width=0.45\textwidth,valign=c]{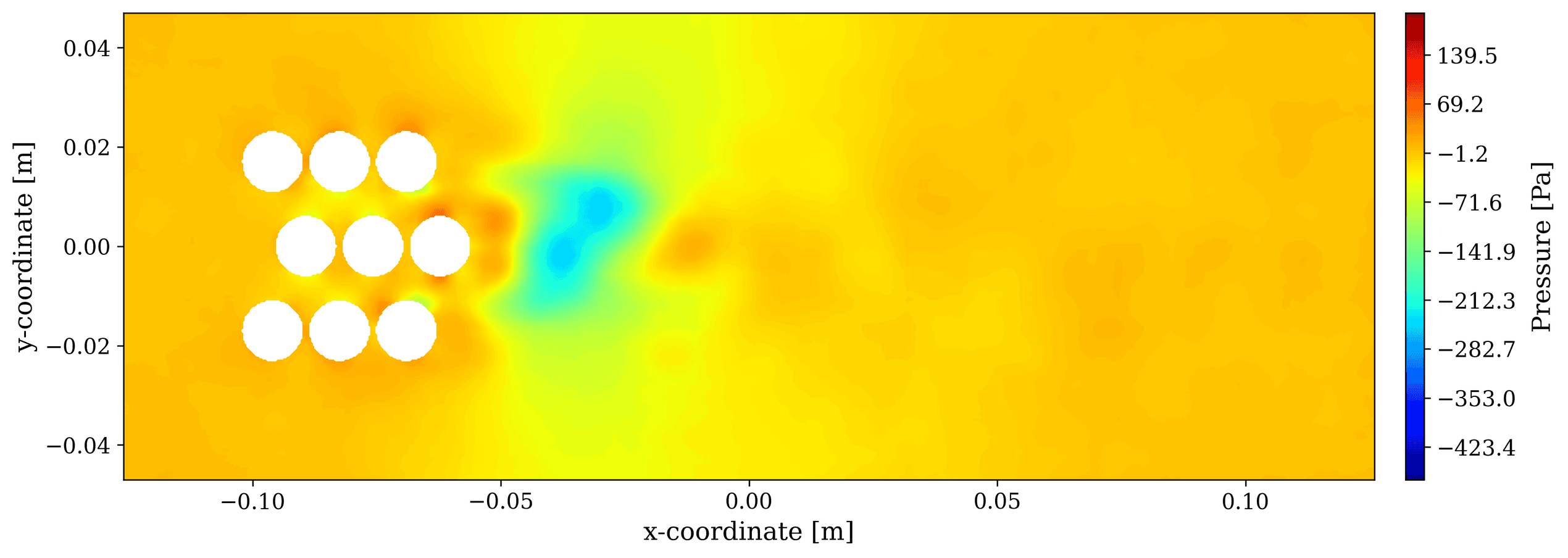} &
        \includegraphics[width=0.45\textwidth,valign=c]{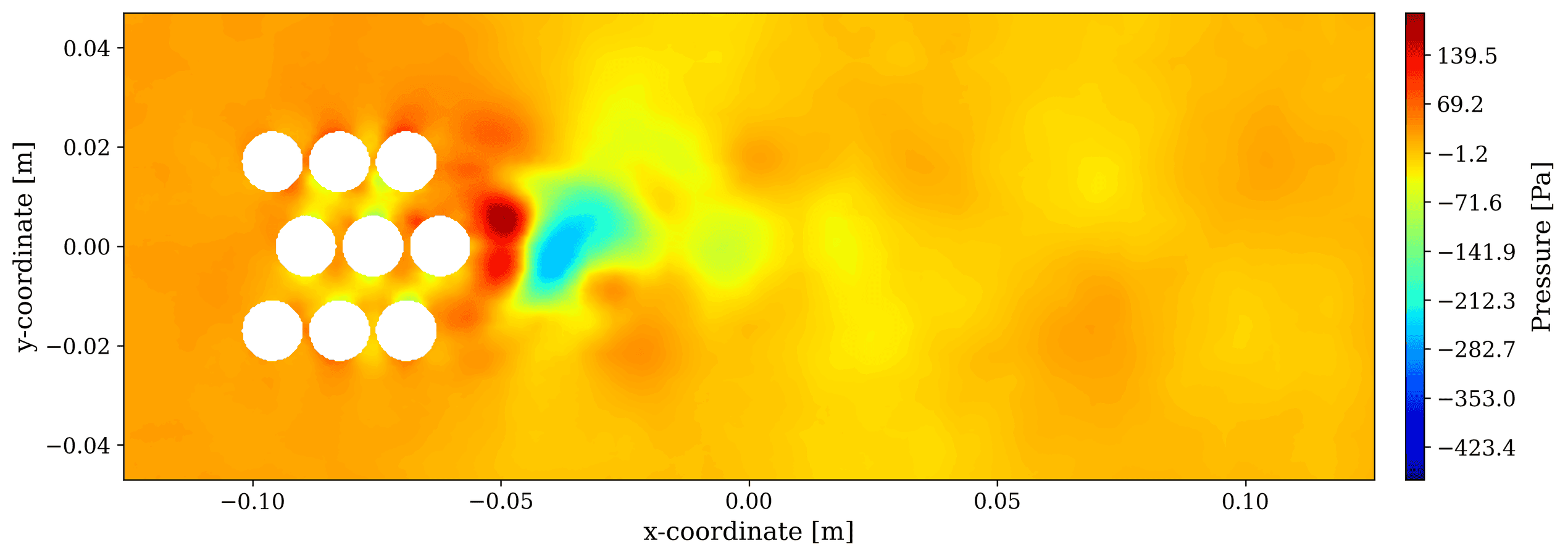} &
        \includegraphics[width=0.45\textwidth,valign=c]{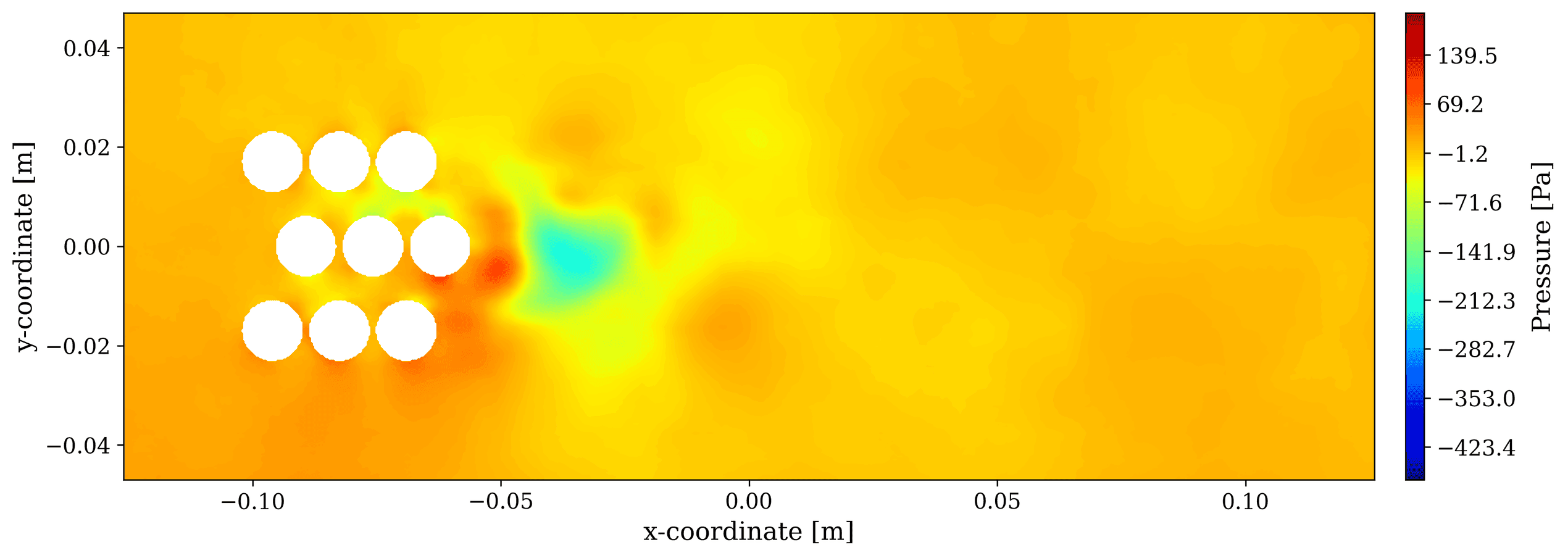} \\[2pt]
        \adjustbox{valign=c}{\rotatebox[origin=c]{90}{\small\textbf{Error}}} &
        \includegraphics[width=0.45\textwidth,valign=c]{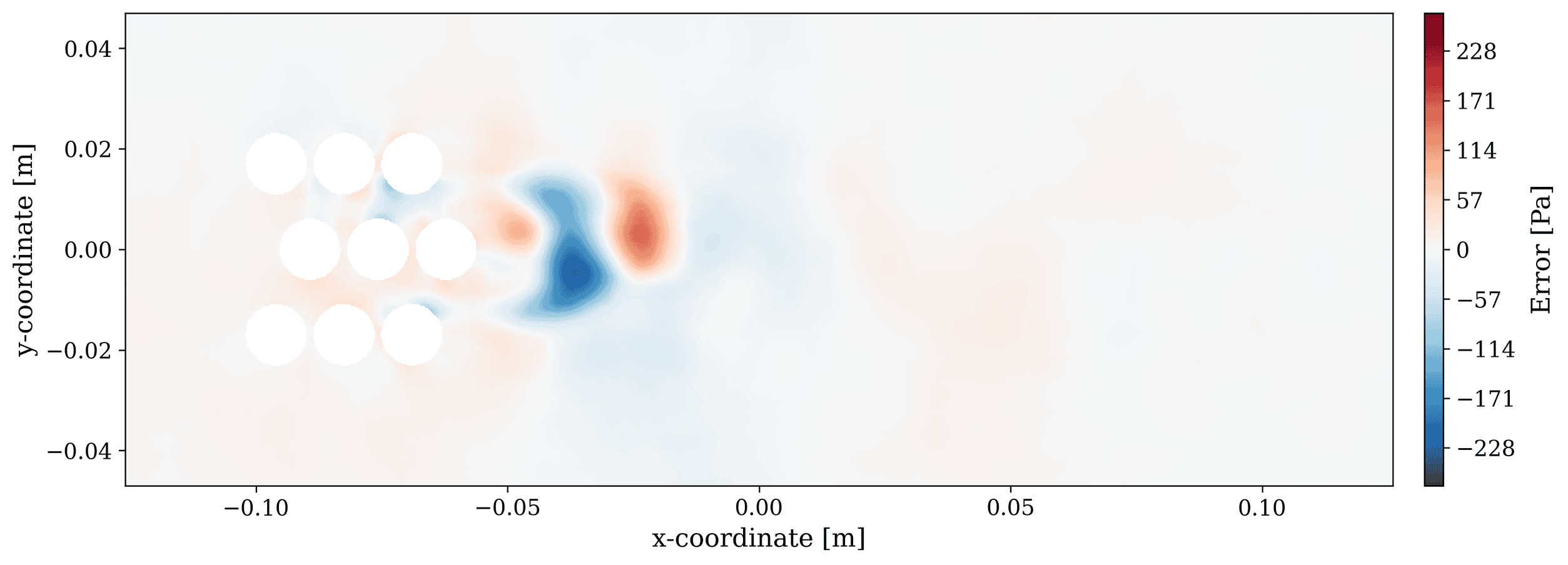} &
        \includegraphics[width=0.45\textwidth,valign=c]{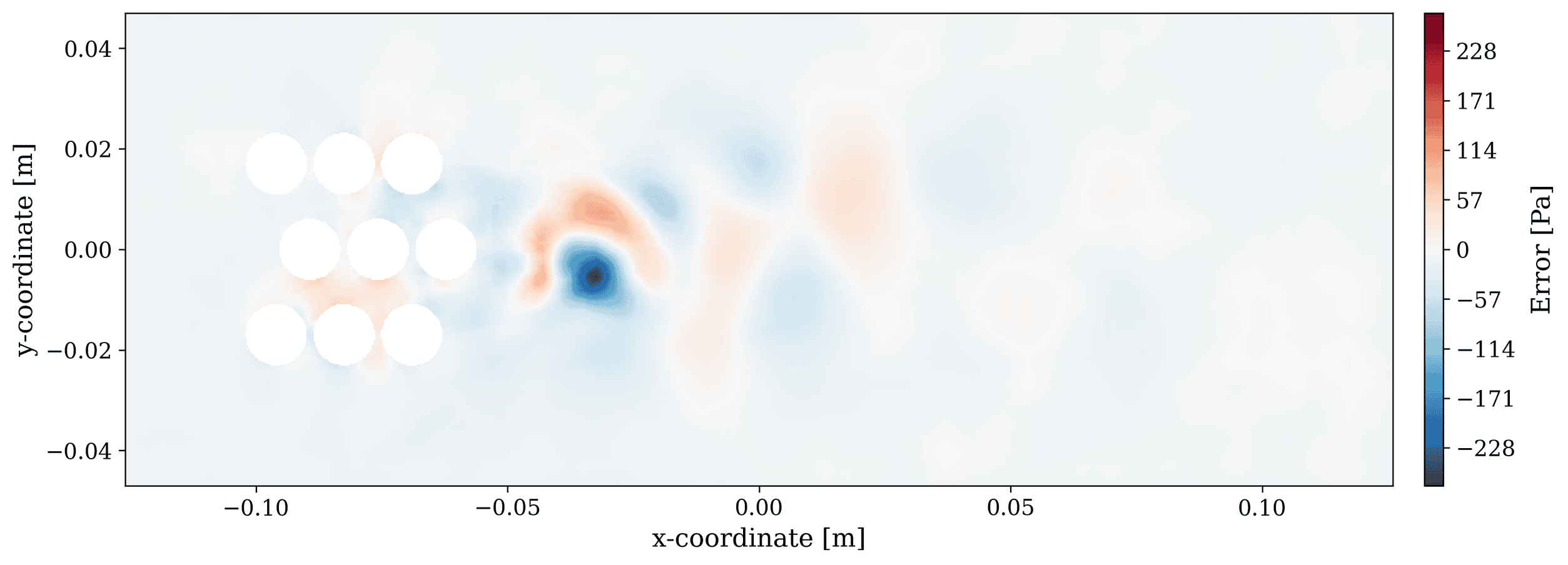} &
        \includegraphics[width=0.45\textwidth,valign=c]{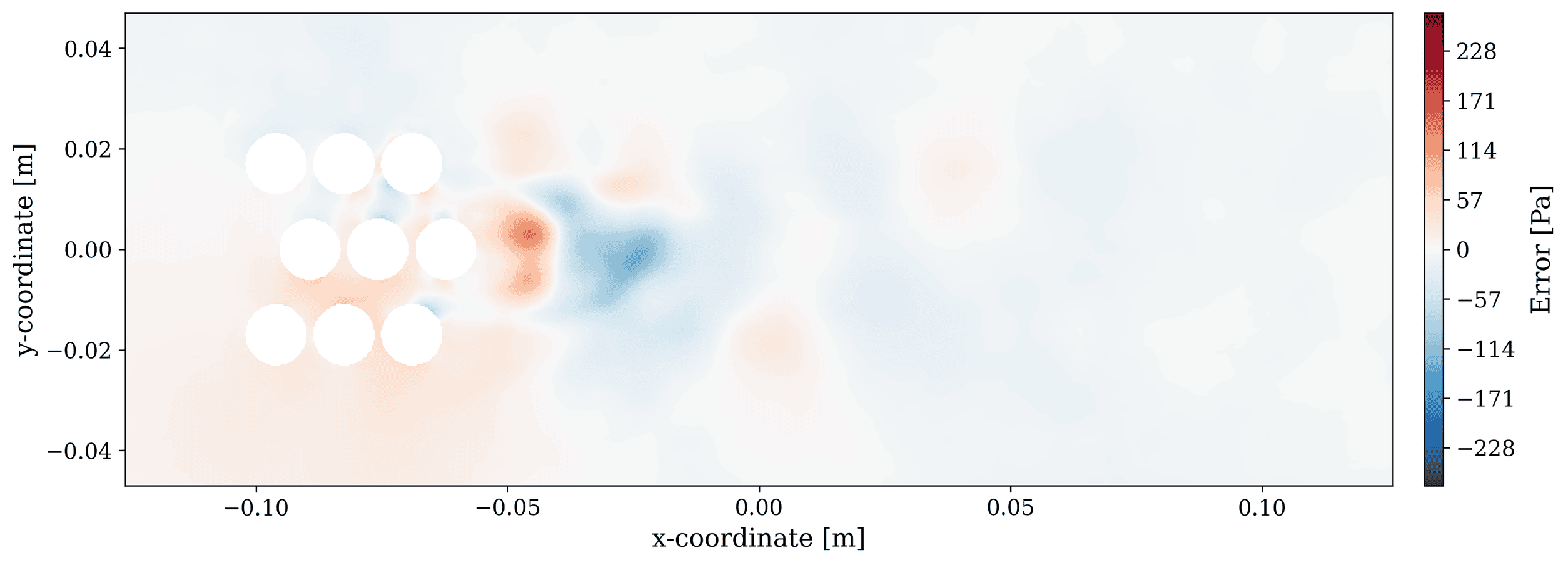} \\
    \end{tabular}
    }%
    \caption{CAE results for pressure field with inlet velocity 0.7 $m/s$. Reference, restored, and error at different timesteps.}
    \label{fig:cae_pressure_inlet070}
\end{figure}

\begin{figure}[H]
    \centering
    \setlength{\tabcolsep}{1pt}
    \makebox[\textwidth][c]{%
    \begin{tabular}{c@{\hspace{4pt}}ccc}
        & \textbf{$t = 2$} & \textbf{$t = 50$} & \textbf{$t = 100$} \\
        \adjustbox{valign=c}{\rotatebox[origin=c]{90}{\small\textbf{Reference}}} &
        \includegraphics[width=0.45\textwidth,valign=c]{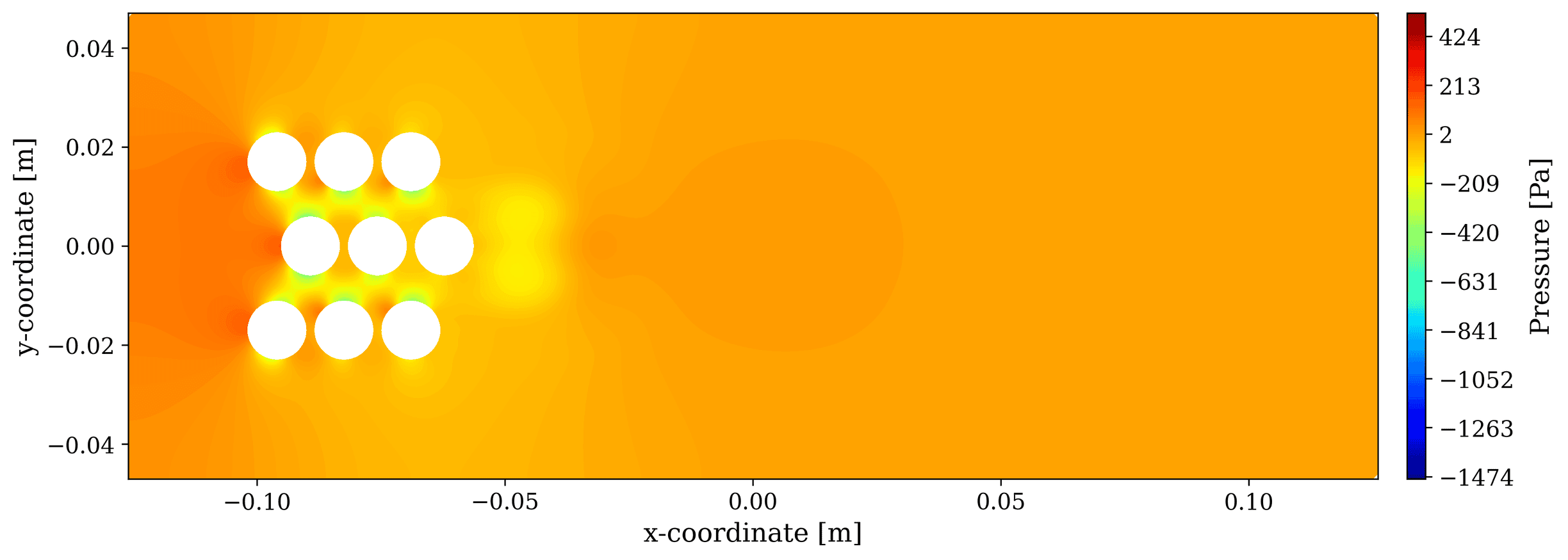} &
        \includegraphics[width=0.45\textwidth,valign=c]{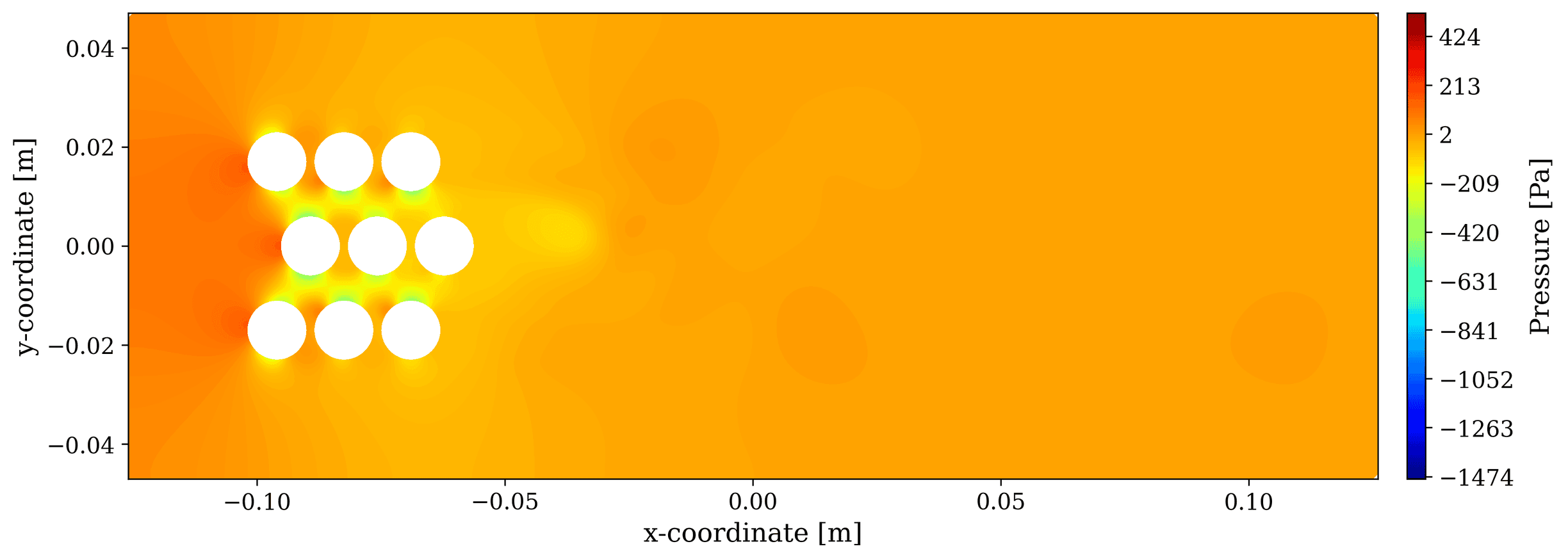} &
        \includegraphics[width=0.45\textwidth,valign=c]{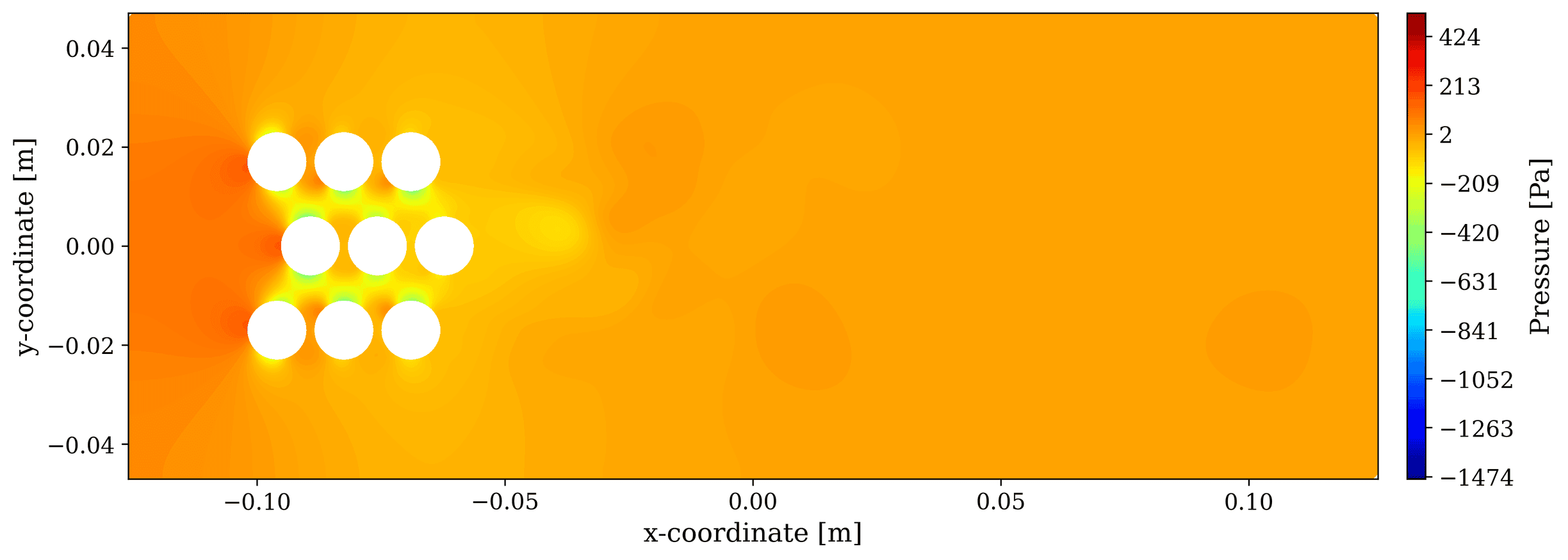} \\[2pt]
        \adjustbox{valign=c}{\rotatebox[origin=c]{90}{\small\textbf{Predicted}}} &
        \includegraphics[width=0.45\textwidth,valign=c]{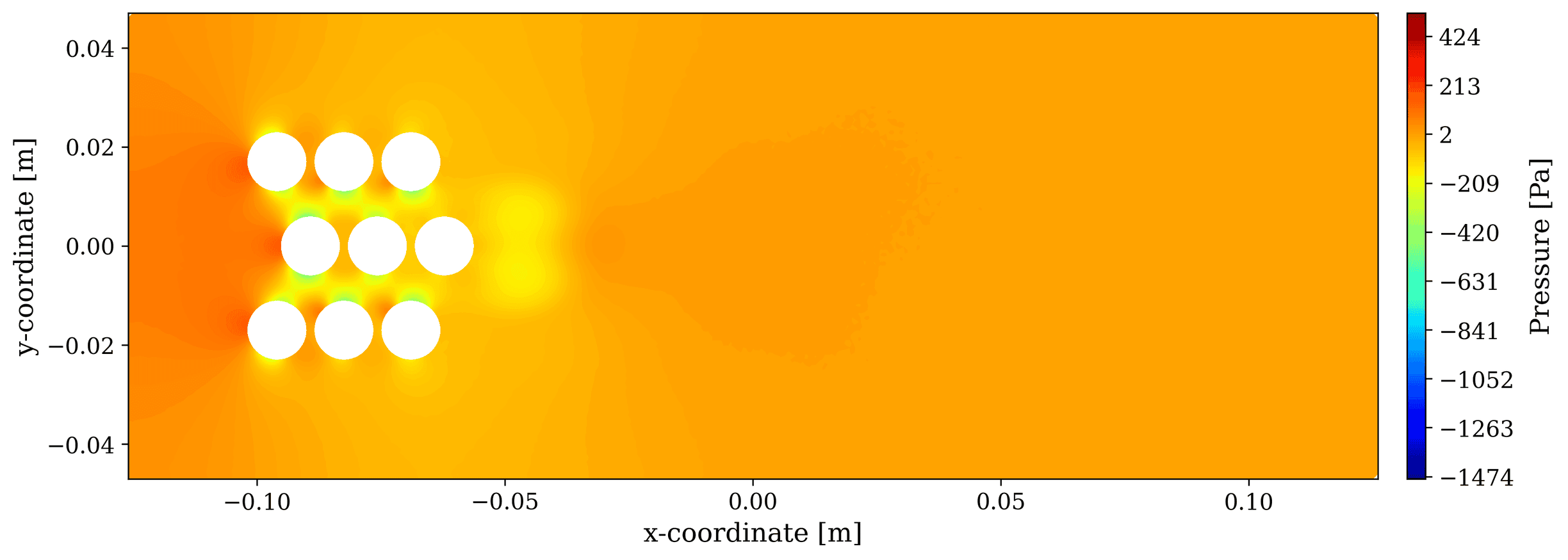} &
        \includegraphics[width=0.45\textwidth,valign=c]{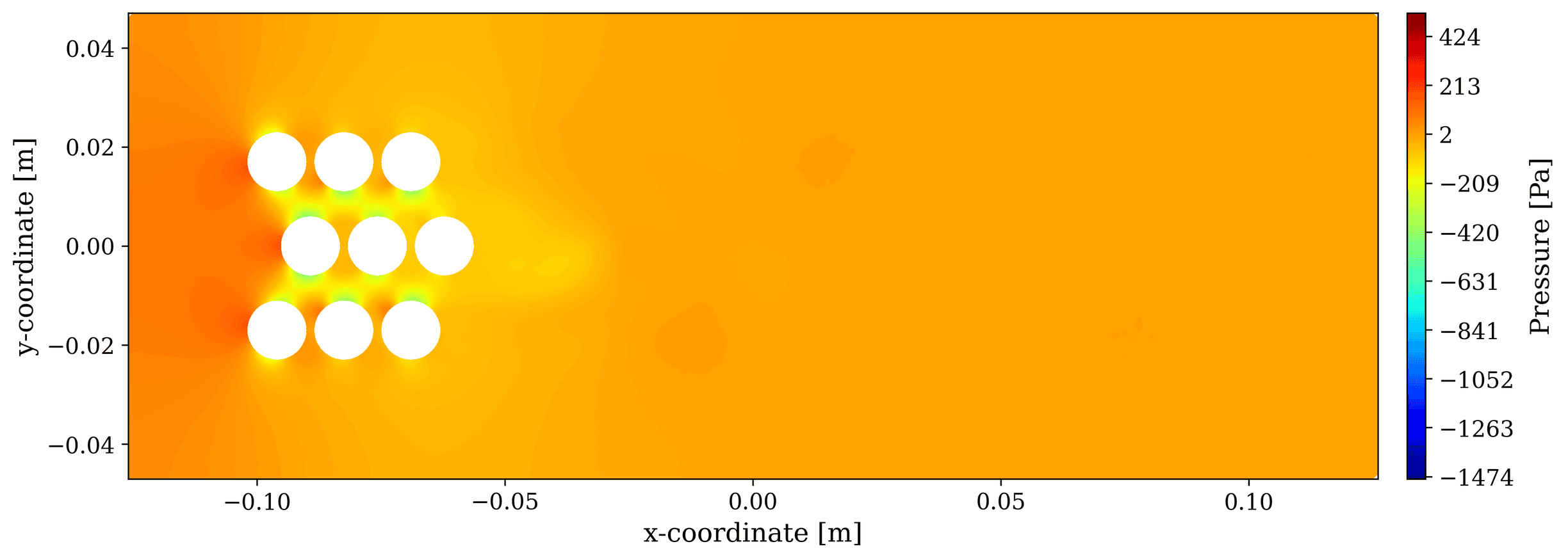} &
        \includegraphics[width=0.45\textwidth,valign=c]{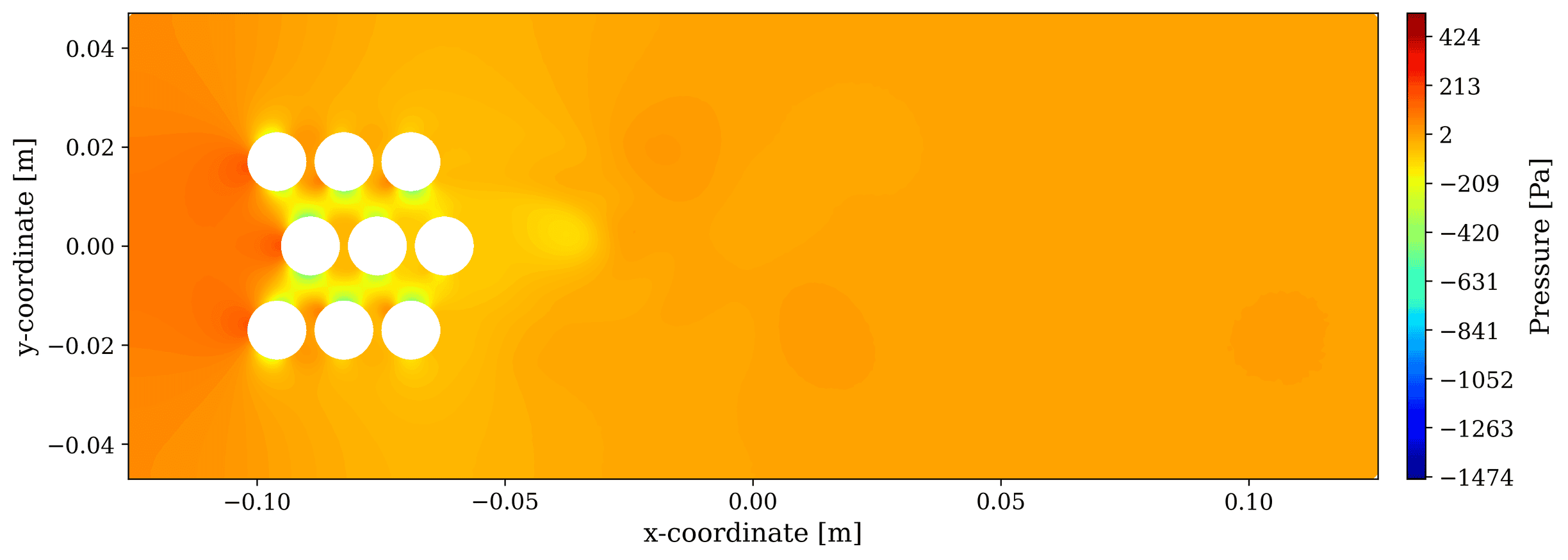} \\[2pt]
        \adjustbox{valign=c}{\rotatebox[origin=c]{90}{\small\textbf{Error}}} &
        \includegraphics[width=0.45\textwidth,valign=c]{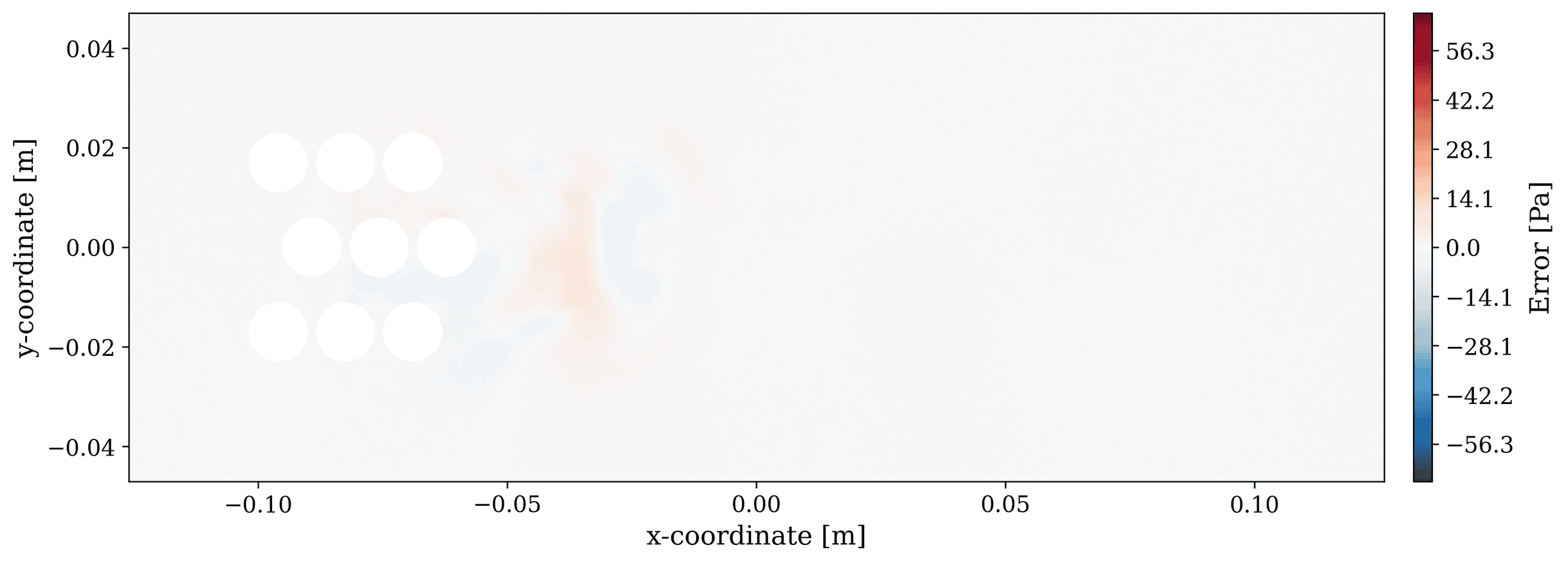} &
        \includegraphics[width=0.45\textwidth,valign=c]{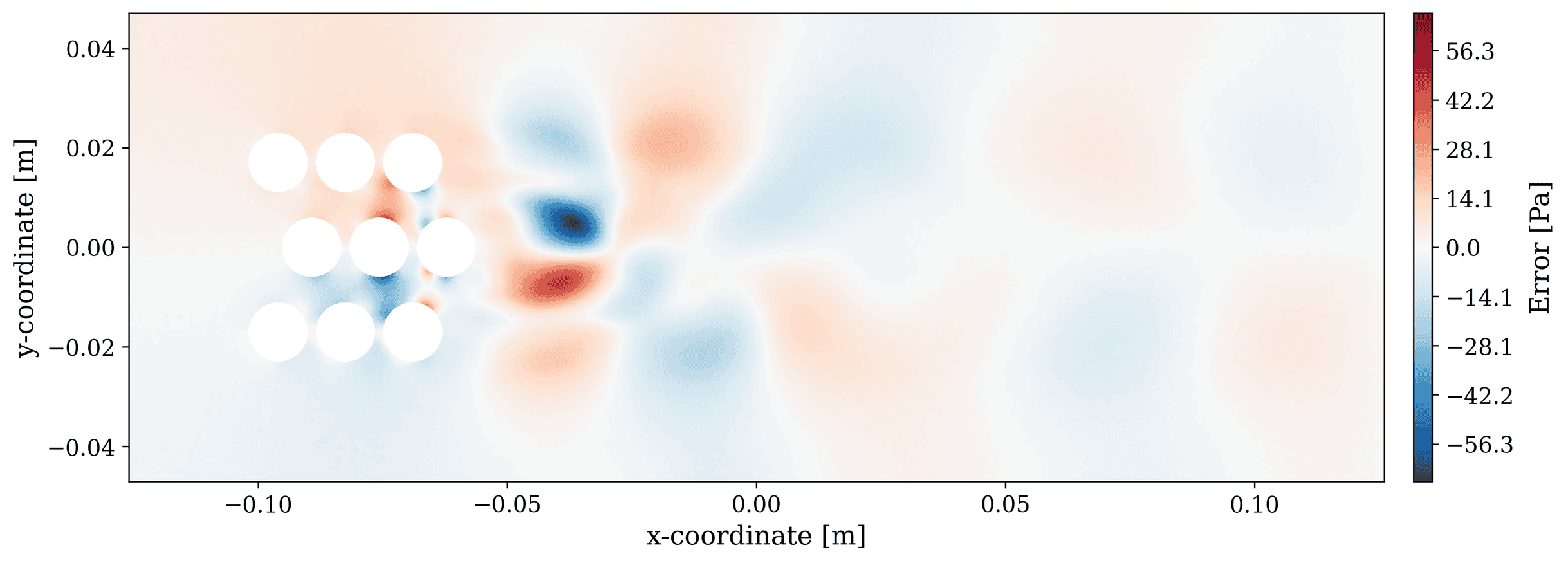} &
        \includegraphics[width=0.45\textwidth,valign=c]{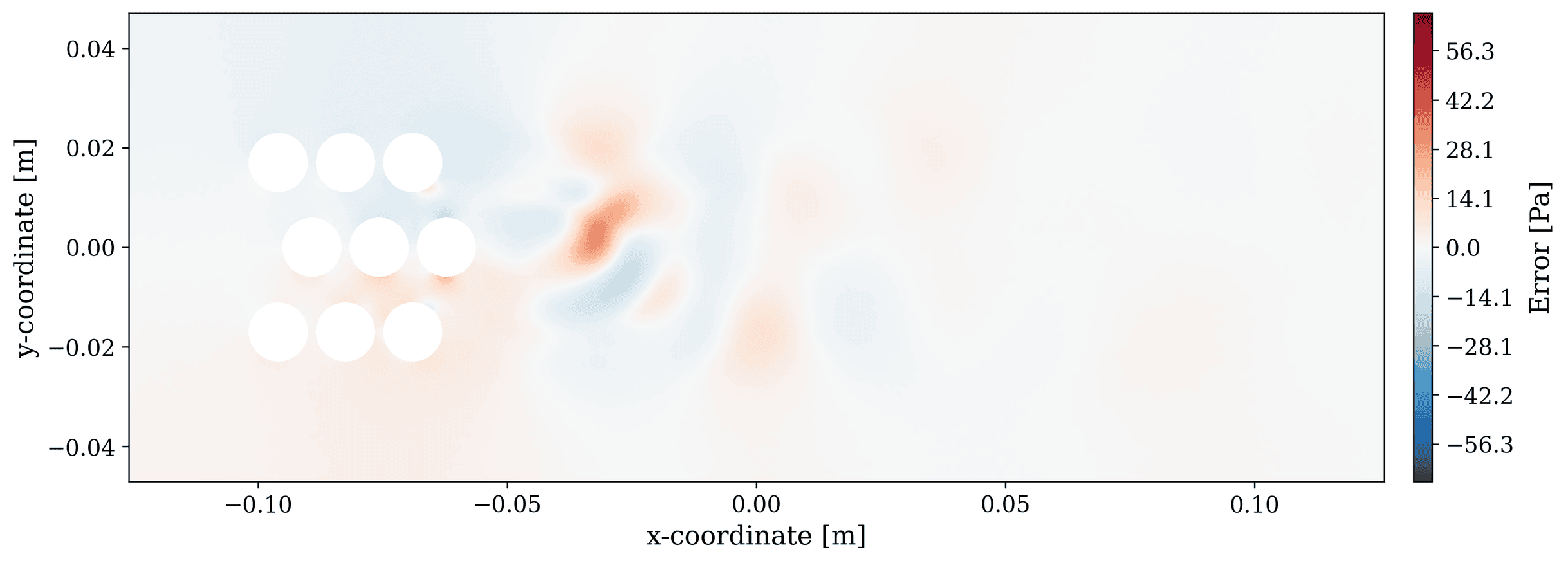} \\
    \end{tabular}
    }%
    \caption{MLP-based AE L-DeepONet with multi-scale results for pressure field with inlet velocity 0.4 $m/s$. Reference, predicted, and error at different timesteps.}
    \label{fig:ldon_unstructured_pressure_inlet040}
\end{figure}

\begin{figure}[H]
    \centering
    \setlength{\tabcolsep}{1pt}
    \makebox[\textwidth][c]{%
    \begin{tabular}{c@{\hspace{4pt}}ccc}
        & \textbf{$t = 2$} & \textbf{$t = 50$} & \textbf{$t = 100$} \\
        \adjustbox{valign=c}{\rotatebox[origin=c]{90}{\small\textbf{Reference}}} &
        \includegraphics[width=0.45\textwidth,valign=c]{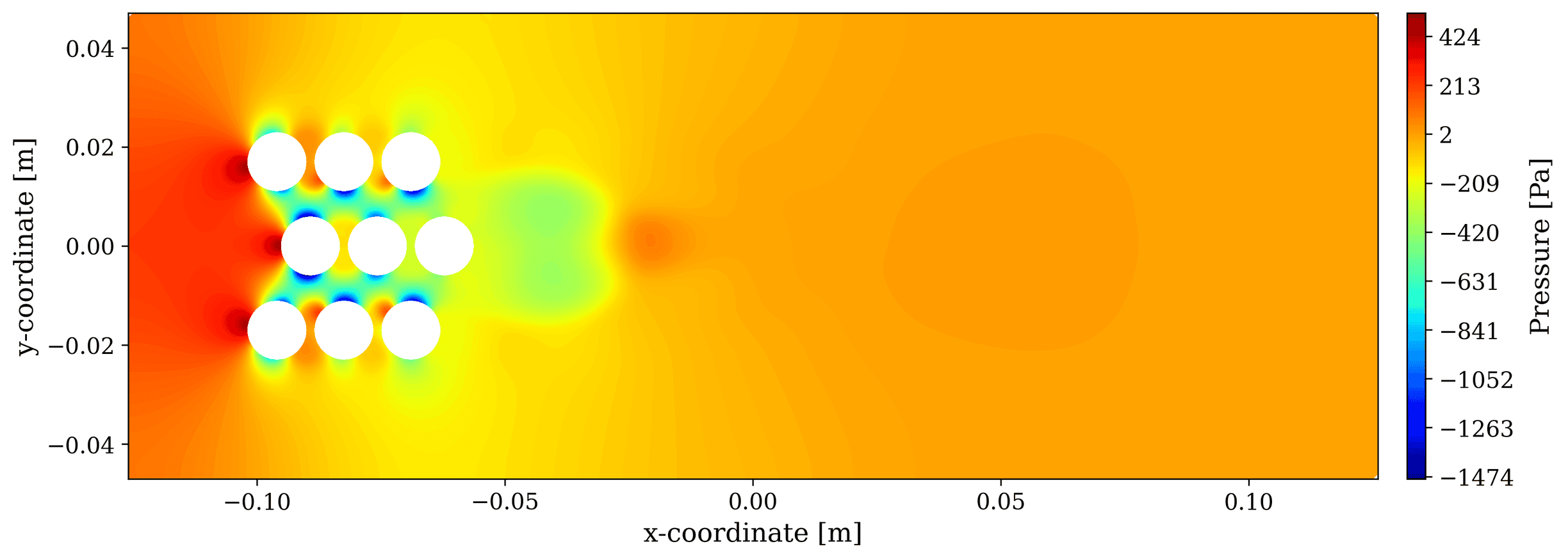} &
        \includegraphics[width=0.45\textwidth,valign=c]{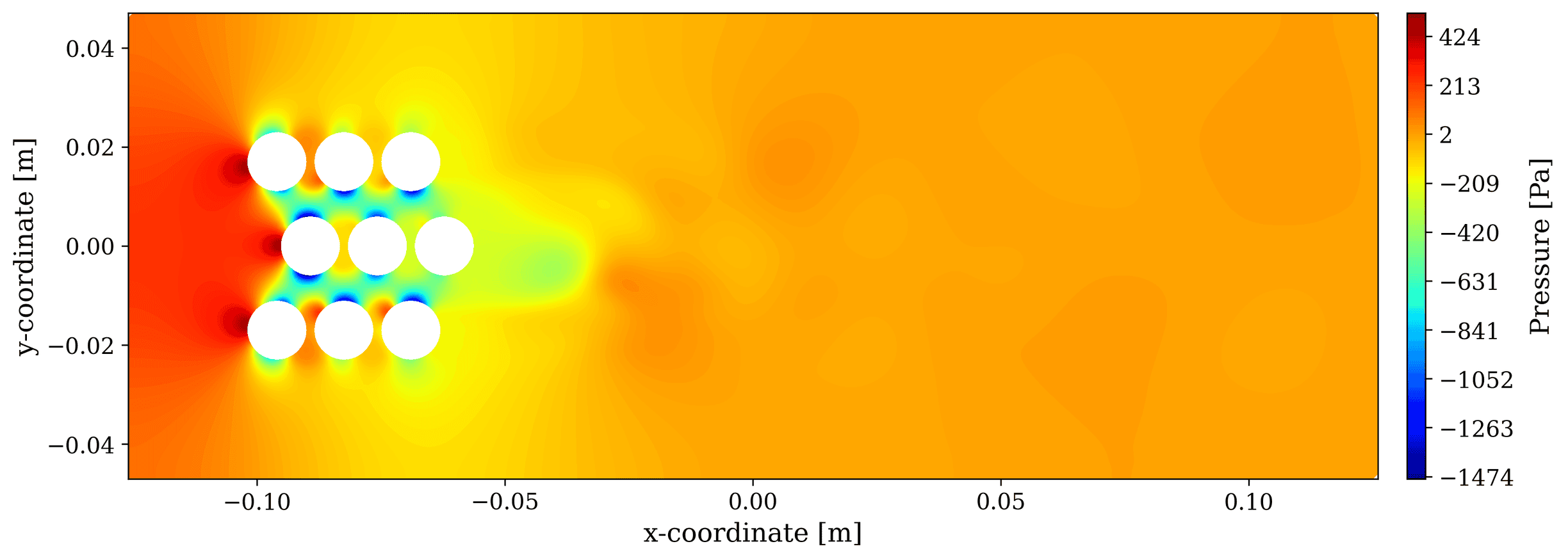} &
        \includegraphics[width=0.45\textwidth,valign=c]{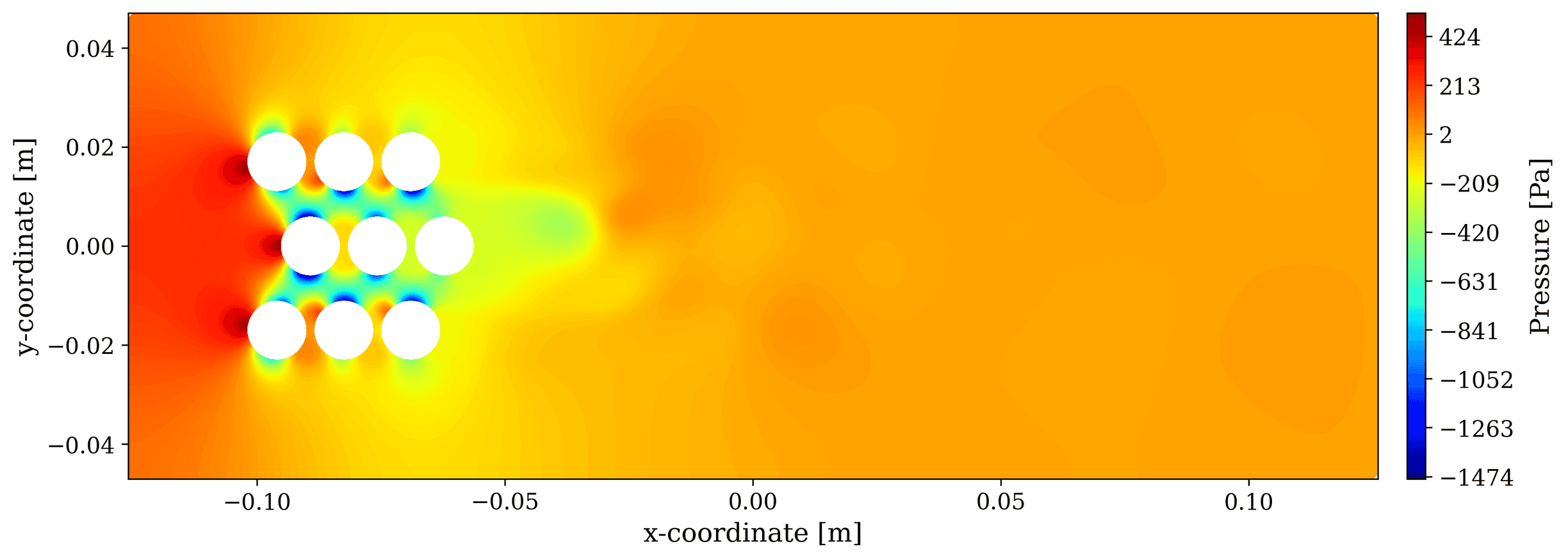} \\[2pt]
        \adjustbox{valign=c}{\rotatebox[origin=c]{90}{\small\textbf{Predicted}}} &
        \includegraphics[width=0.45\textwidth,valign=c]{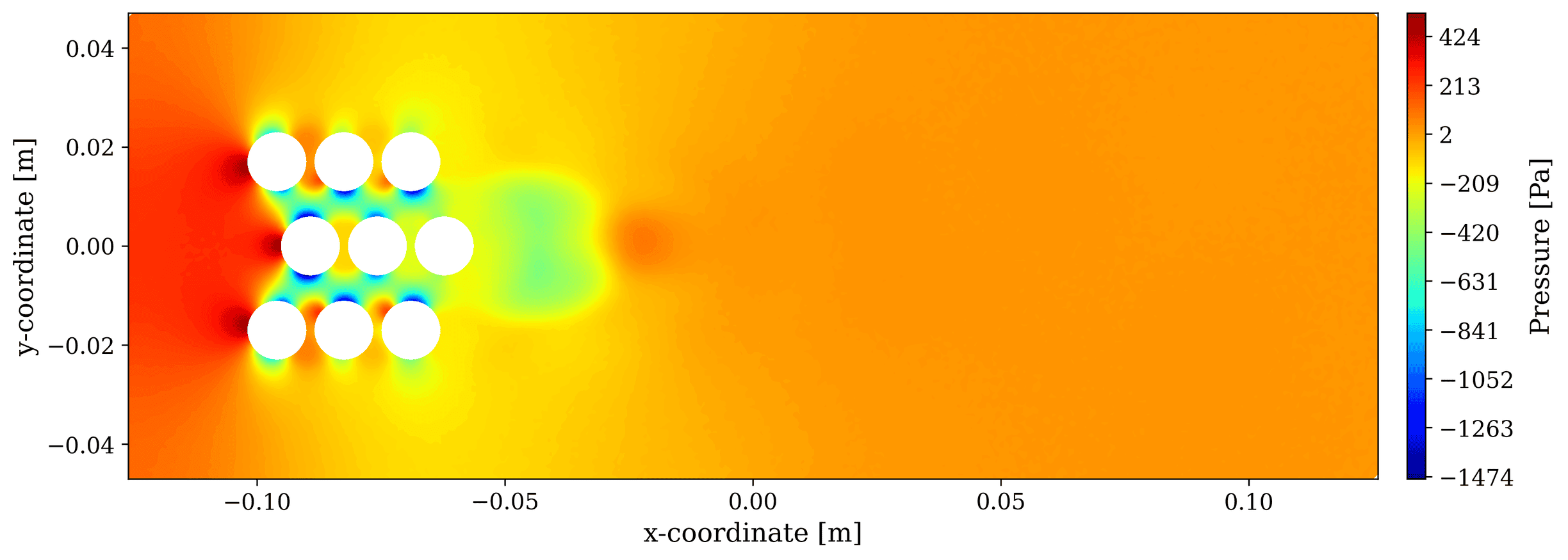} &
        \includegraphics[width=0.45\textwidth,valign=c]{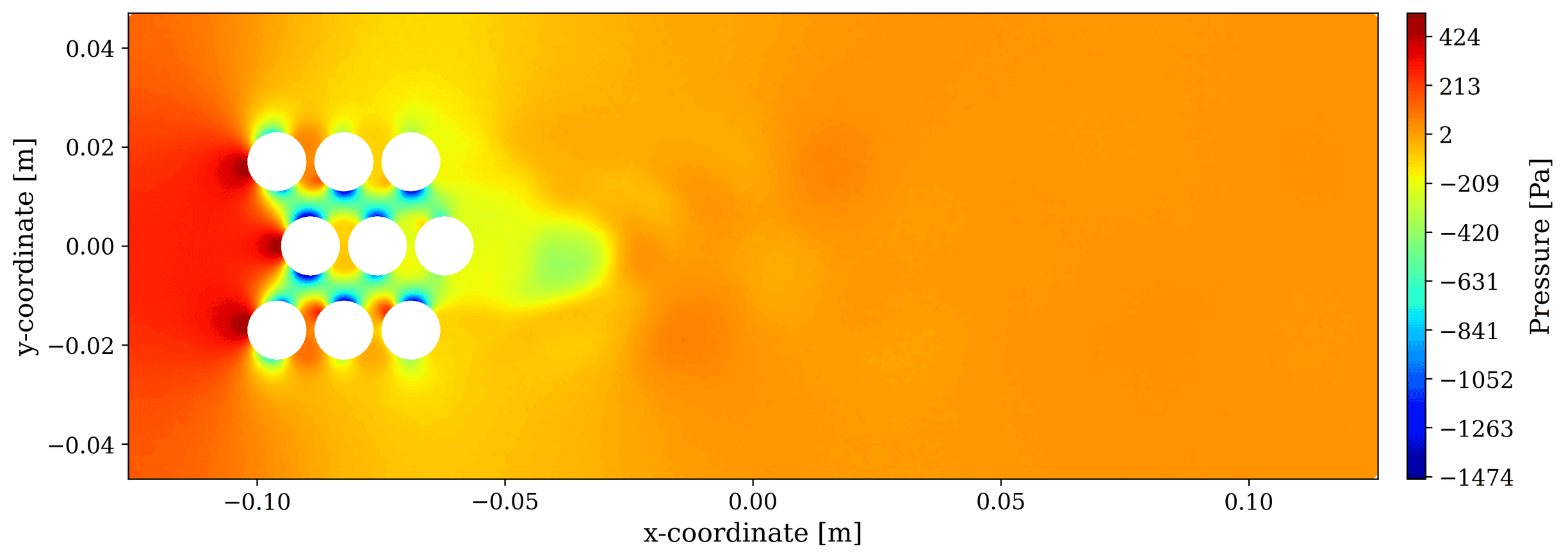} &
        \includegraphics[width=0.45\textwidth,valign=c]{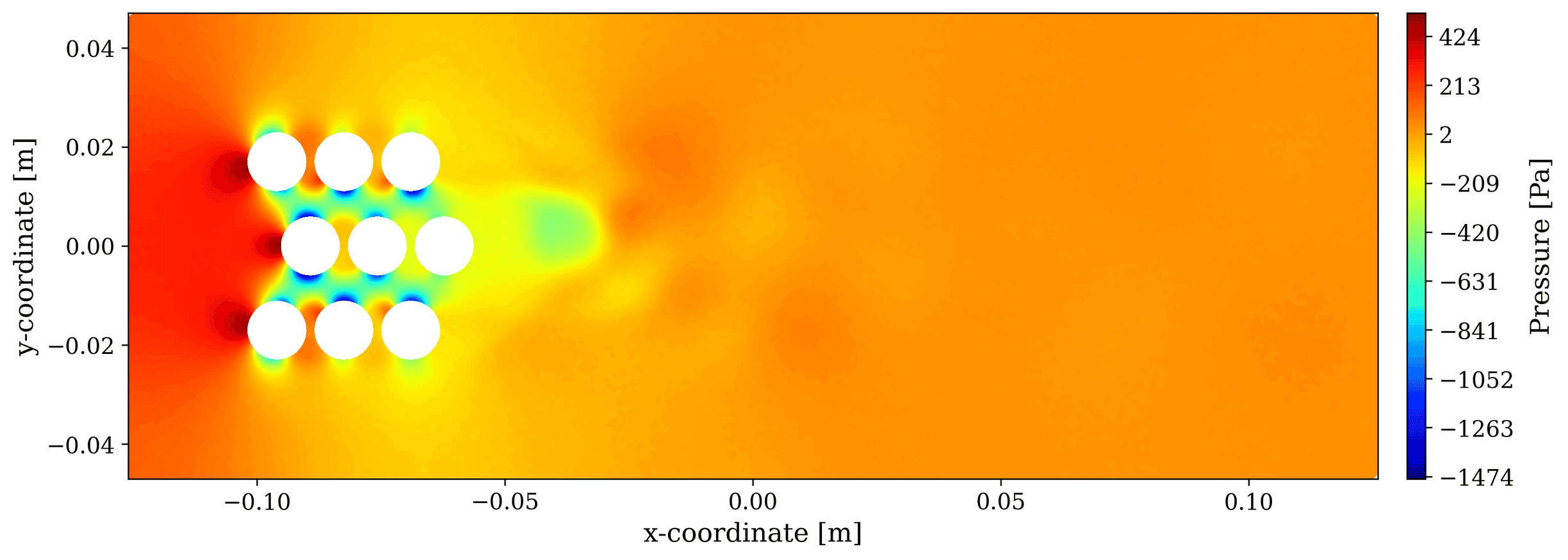} \\[2pt]
        \adjustbox{valign=c}{\rotatebox[origin=c]{90}{\small\textbf{Error}}} &
        \includegraphics[width=0.45\textwidth,valign=c]{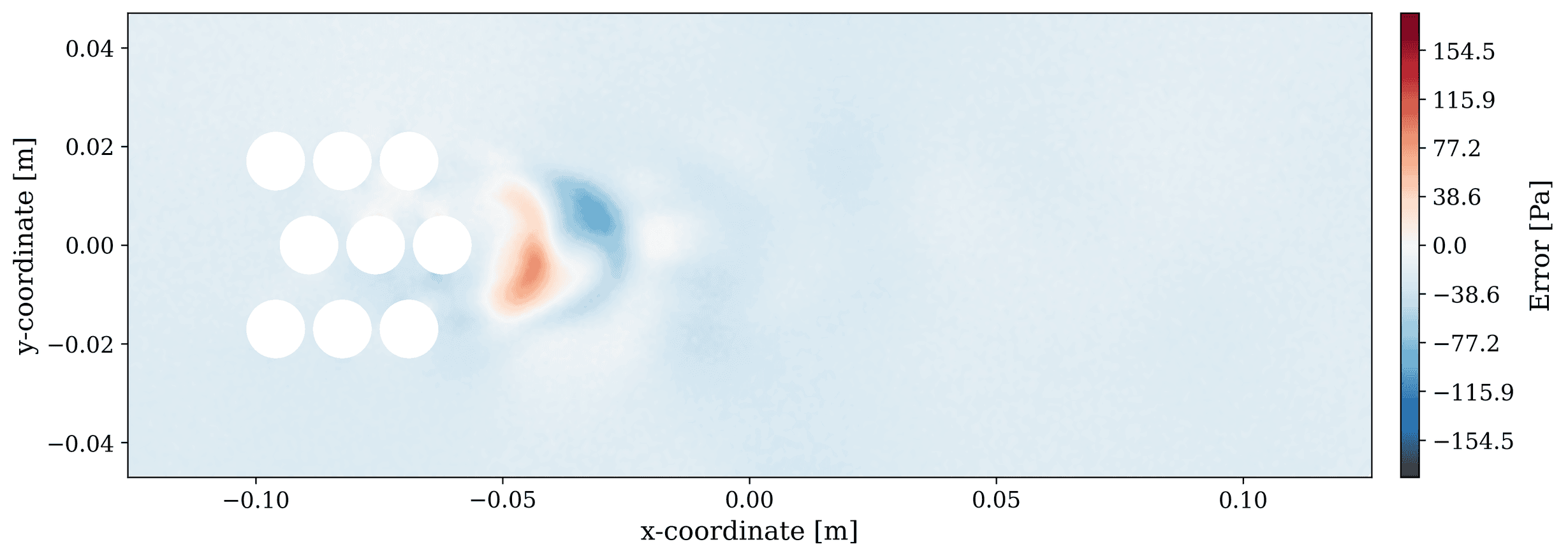} &
        \includegraphics[width=0.45\textwidth,valign=c]{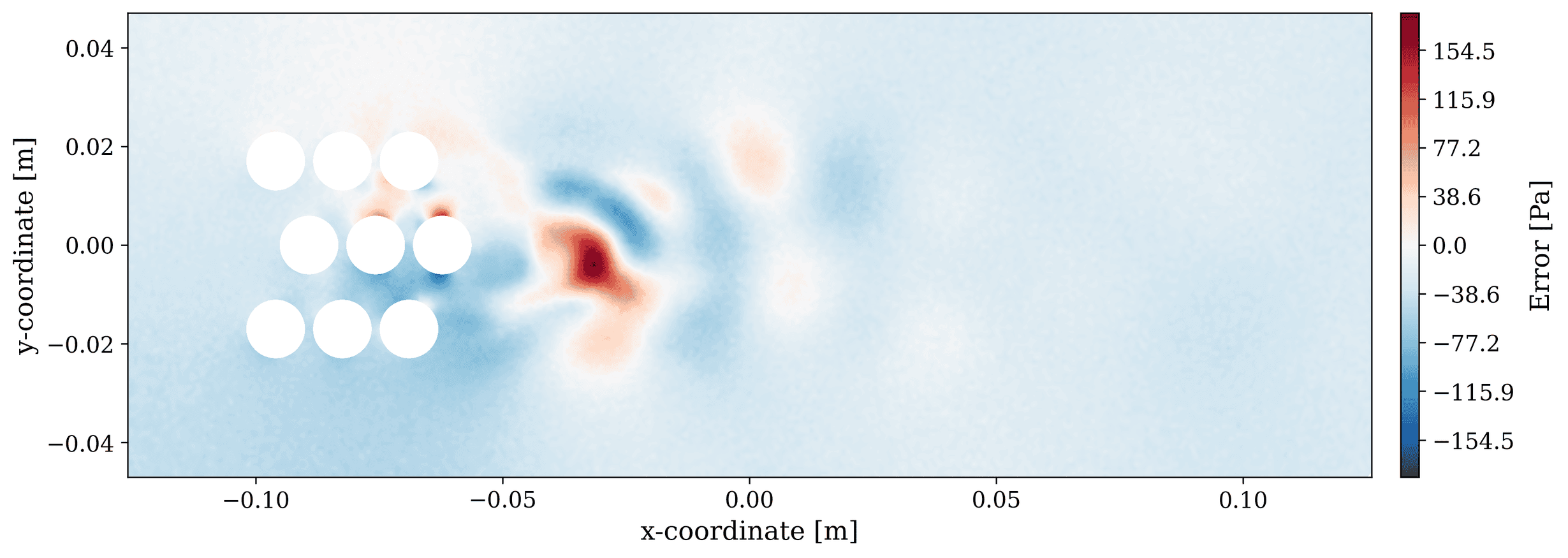} &
        \includegraphics[width=0.45\textwidth,valign=c]{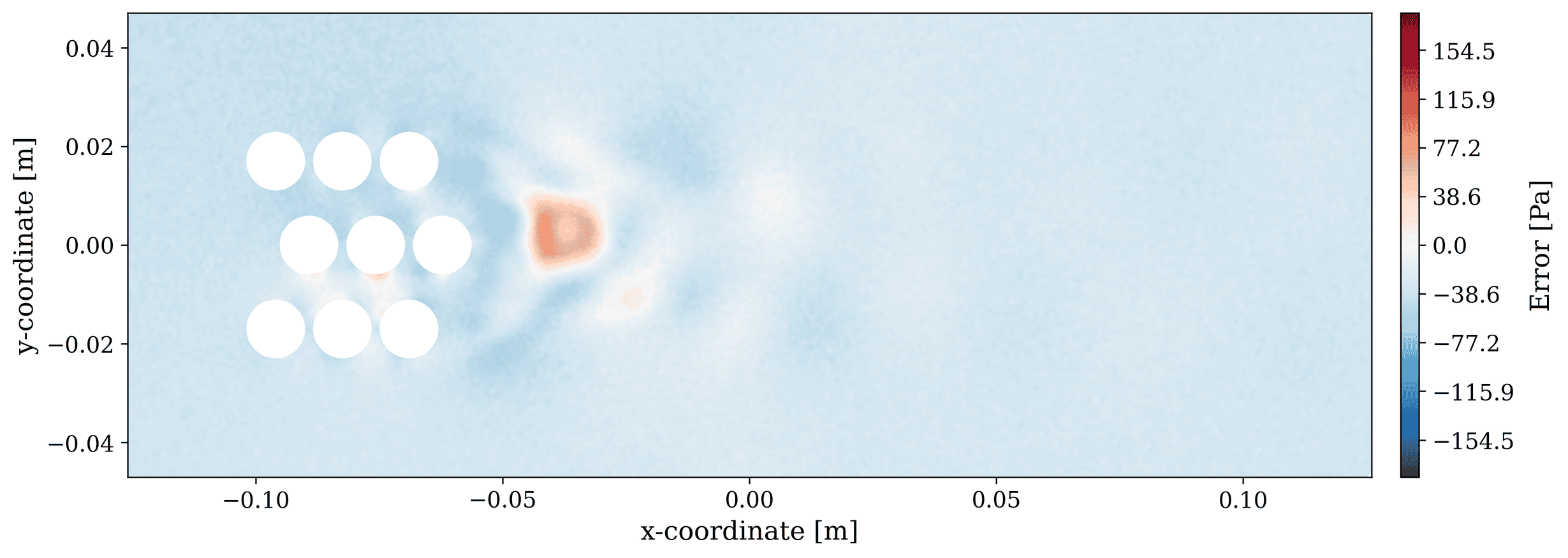} \\
    \end{tabular}
    }%
    \caption{MLP-based AE L-DeepONet results with multi-scale for pressure field with inlet velocity 0.7 $m/s$. Reference, predicted, and error at different timesteps.}
    \label{fig:ldon_unstructured_pressure_inlet070}
\end{figure}

\begin{figure}[H]
    \centering
    \setlength{\tabcolsep}{1pt}
    \makebox[\textwidth][c]{%
    \begin{tabular}{c@{\hspace{4pt}}ccc}
        & \textbf{$t = 2$} & \textbf{$t = 50$} & \textbf{$t = 100$} \\
        \adjustbox{valign=c}{\rotatebox[origin=c]{90}{\small\textbf{Reference}}} &
        \includegraphics[width=0.45\textwidth,valign=c]{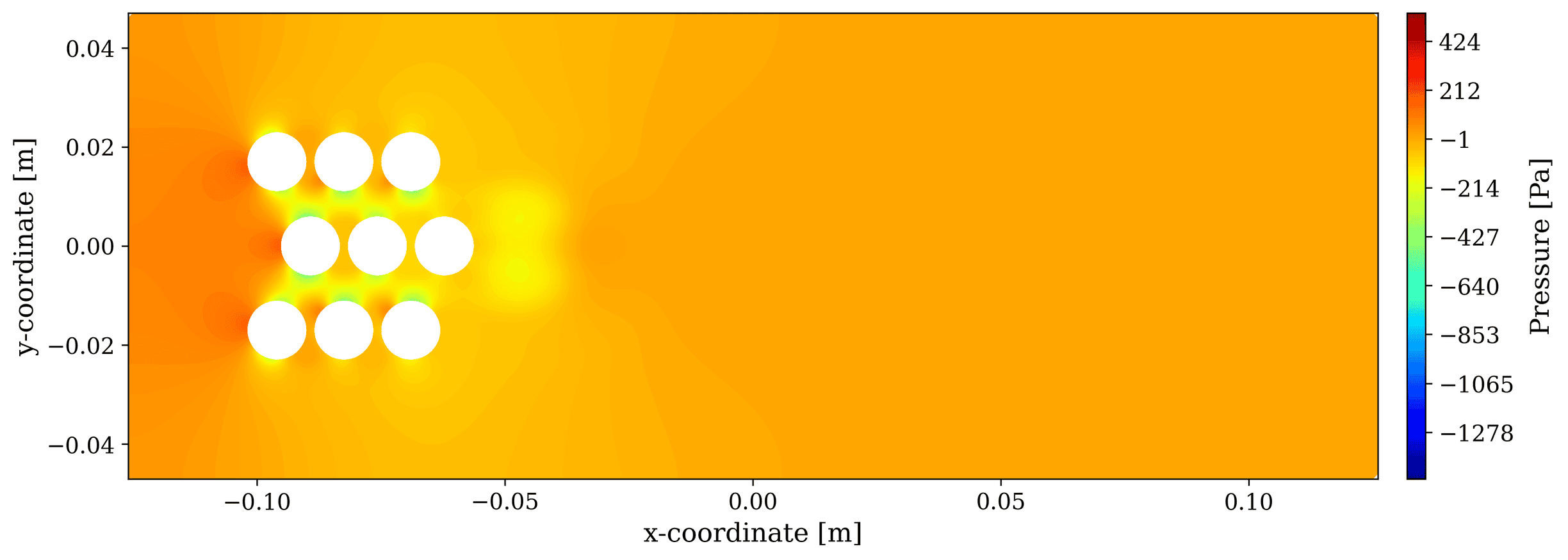} &
        \includegraphics[width=0.45\textwidth,valign=c]{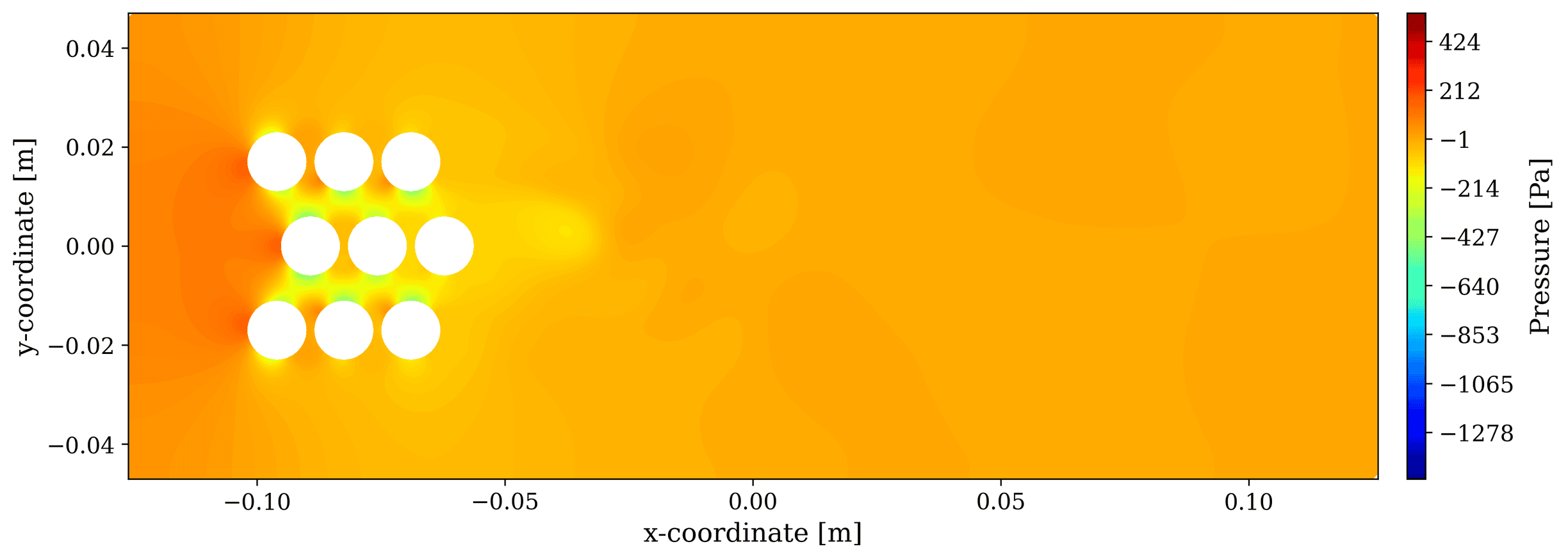} &
        \includegraphics[width=0.45\textwidth,valign=c]{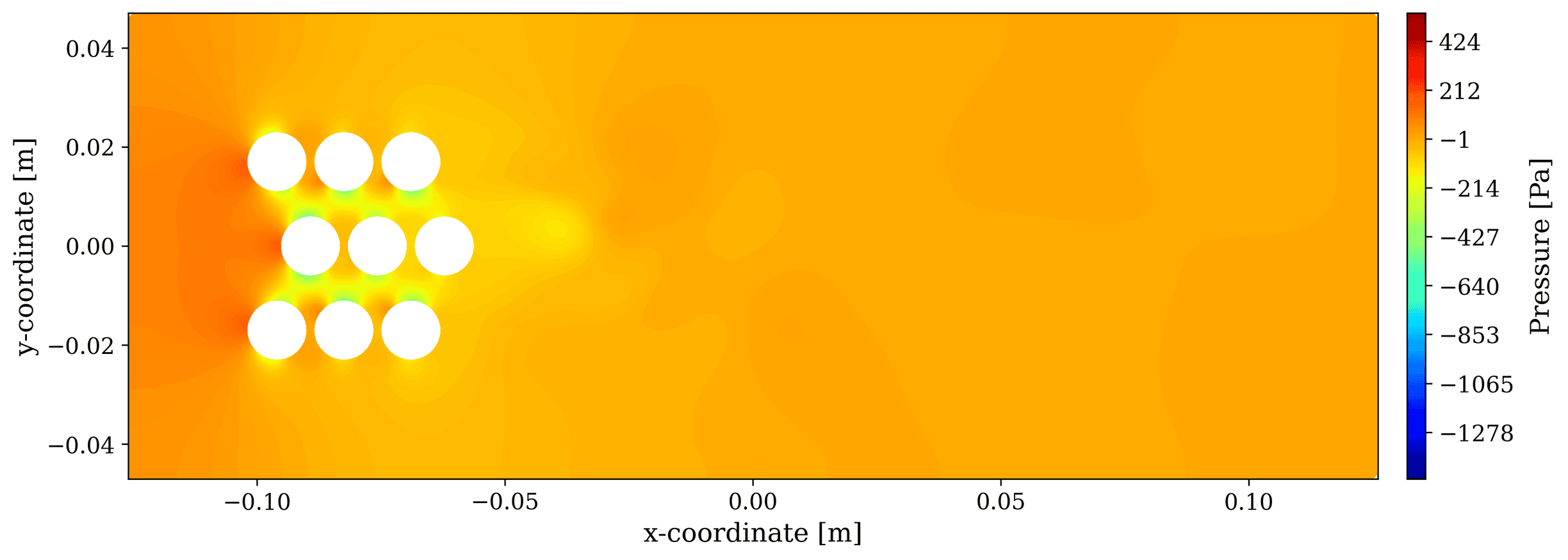} \\[2pt]
        \adjustbox{valign=c}{\rotatebox[origin=c]{90}{\small\textbf{Predicted}}} &
        \includegraphics[width=0.45\textwidth,valign=c]{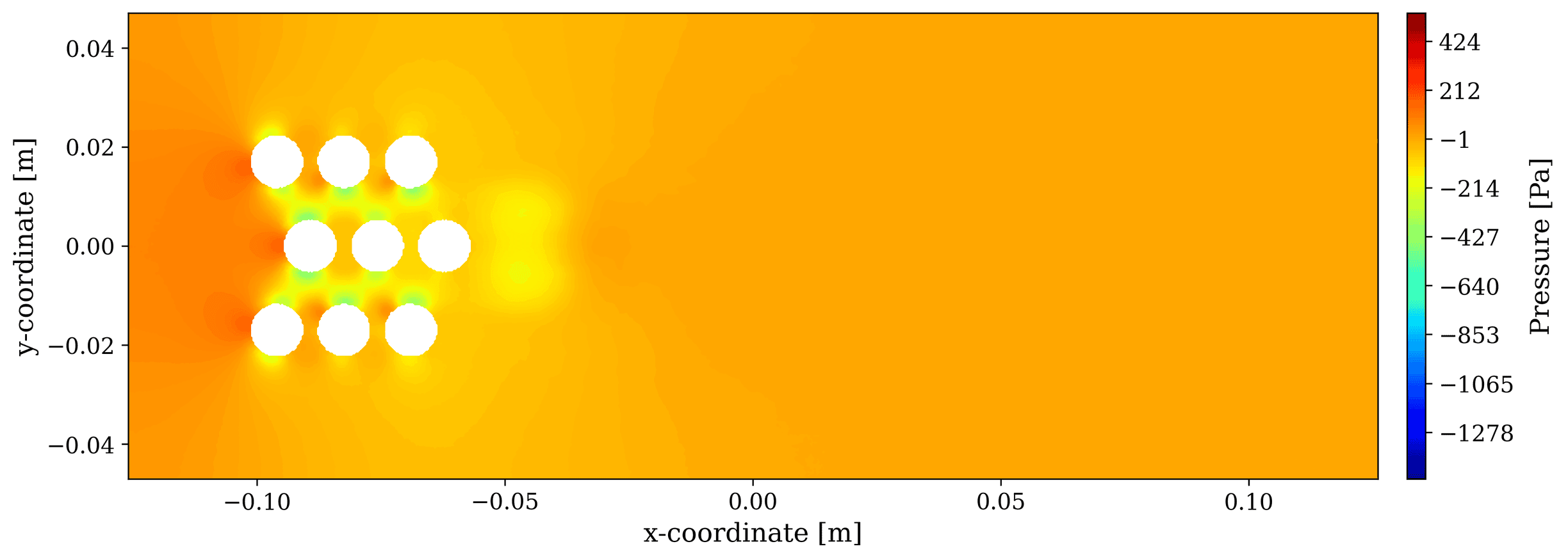} &
        \includegraphics[width=0.45\textwidth,valign=c]{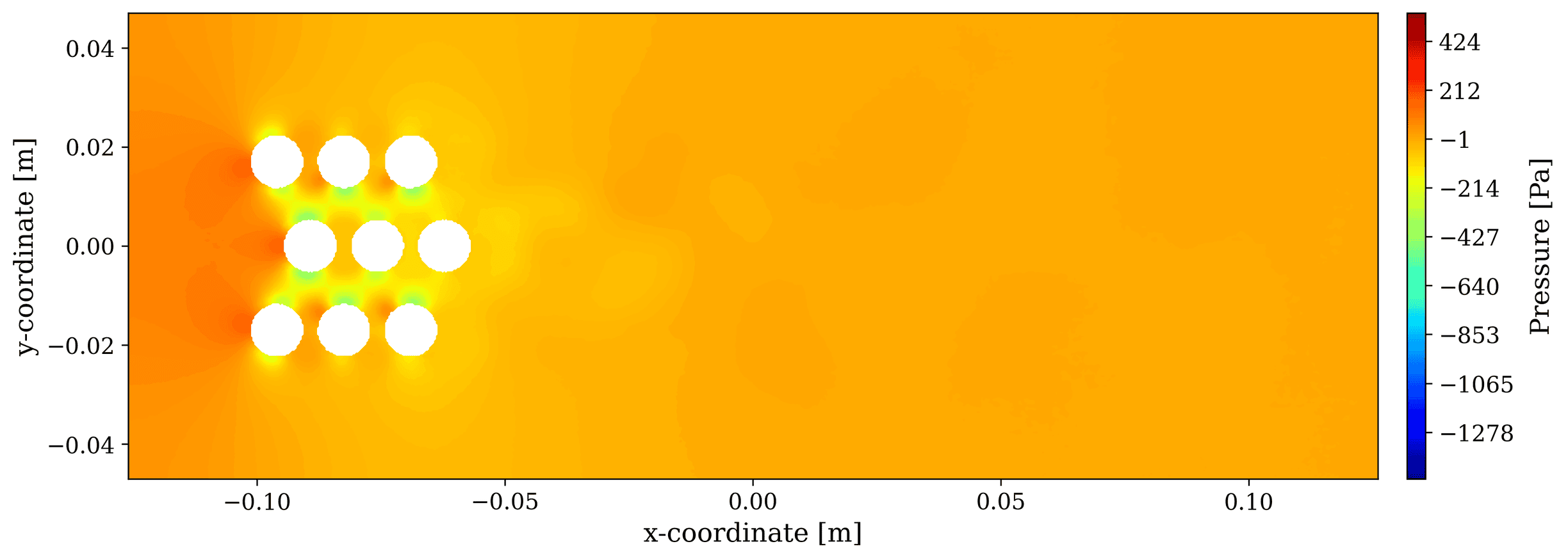} &
        \includegraphics[width=0.45\textwidth,valign=c]{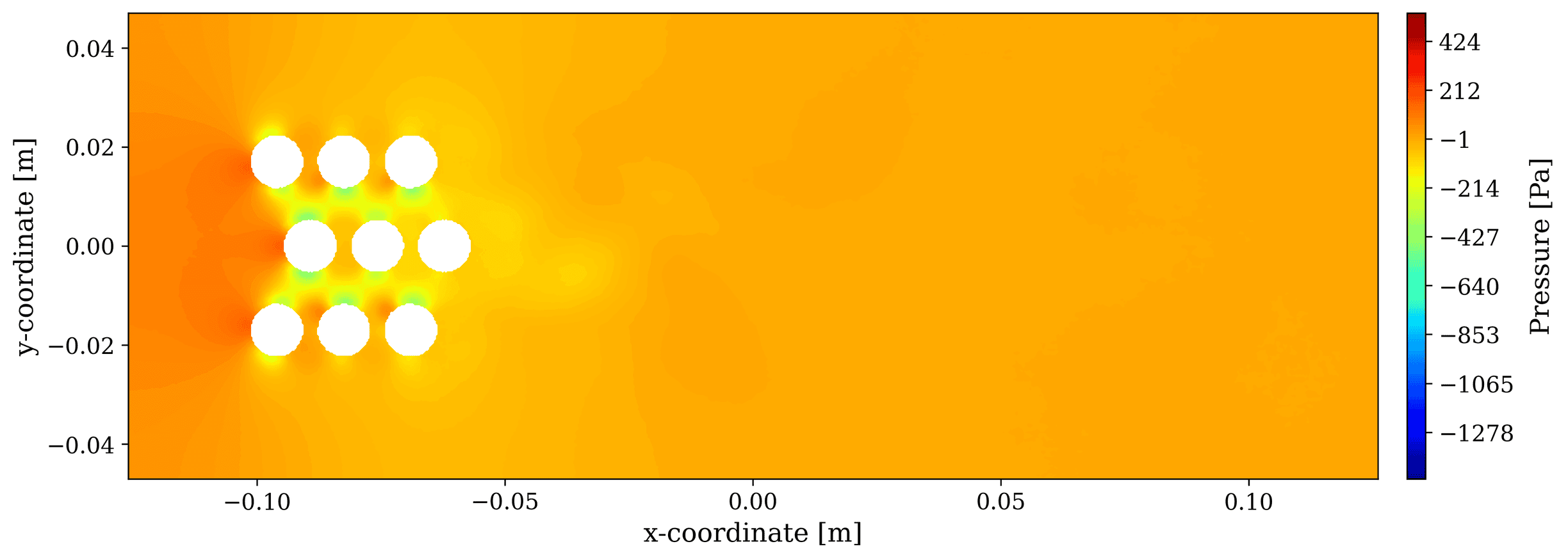} \\[2pt]
        \adjustbox{valign=c}{\rotatebox[origin=c]{90}{\small\textbf{Error}}} &
        \includegraphics[width=0.45\textwidth,valign=c]{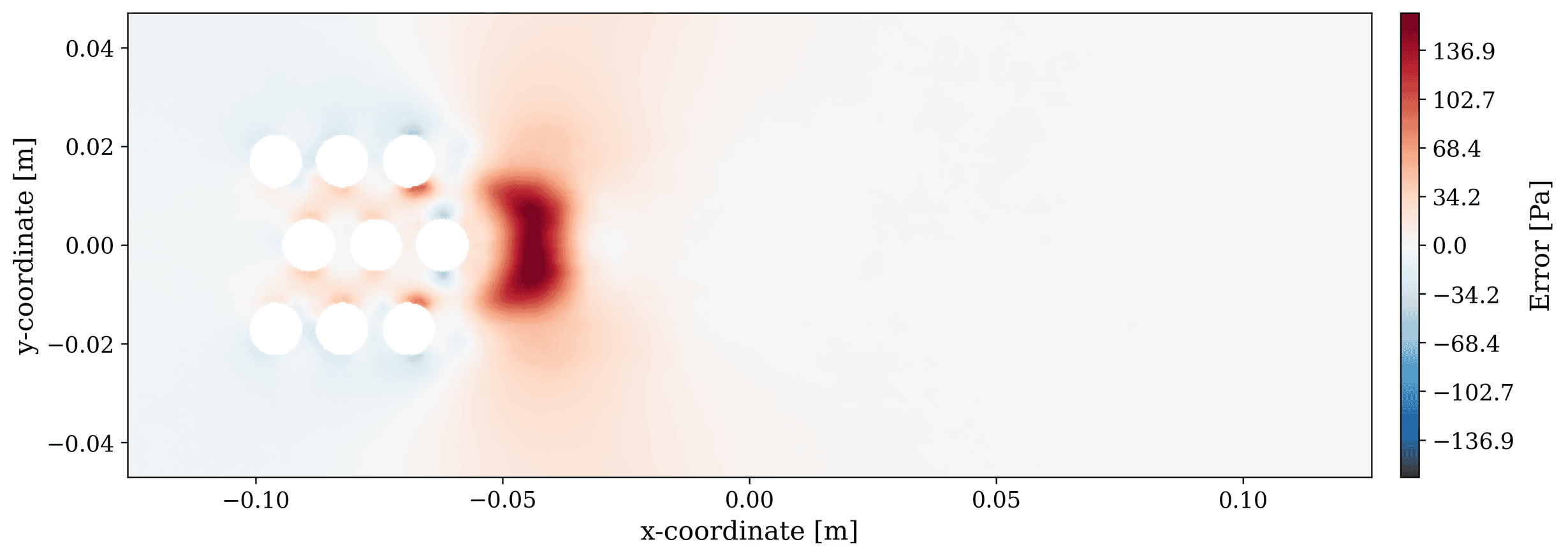} &
        \includegraphics[width=0.45\textwidth,valign=c]{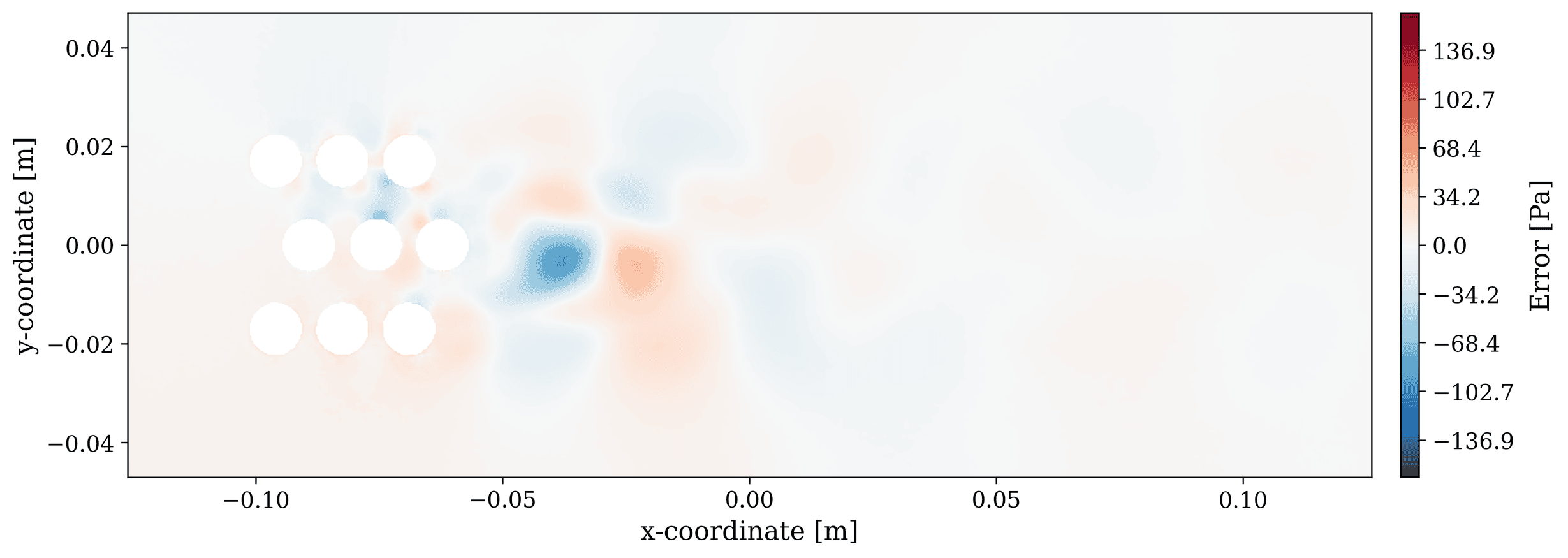} &
        \includegraphics[width=0.45\textwidth,valign=c]{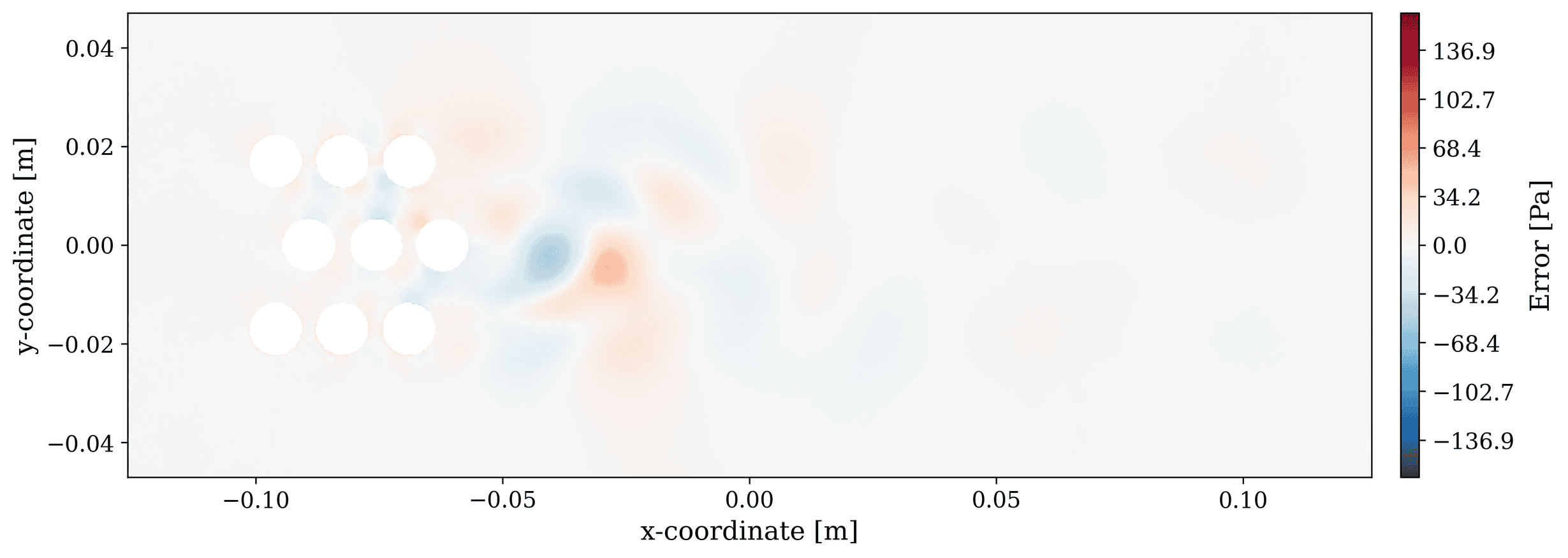} \\
    \end{tabular}
    }%
    \caption{CAE L-DeepONet results with multi-scale for pressure field with inlet velocity 0.4 $m/s$. Reference, predicted, and error at different timesteps.}
    \label{fig:cae_ldon_pressure_inlet040}
\end{figure}

\begin{figure}[H]
    \centering
    \setlength{\tabcolsep}{1pt}
    \makebox[\textwidth][c]{%
    \begin{tabular}{c@{\hspace{4pt}}ccc}
        & \textbf{$t = 2$} & \textbf{$t = 50$} & \textbf{$t = 100$} \\
        \adjustbox{valign=c}{\rotatebox[origin=c]{90}{\small\textbf{Reference}}} &
        \includegraphics[width=0.45\textwidth,valign=c]{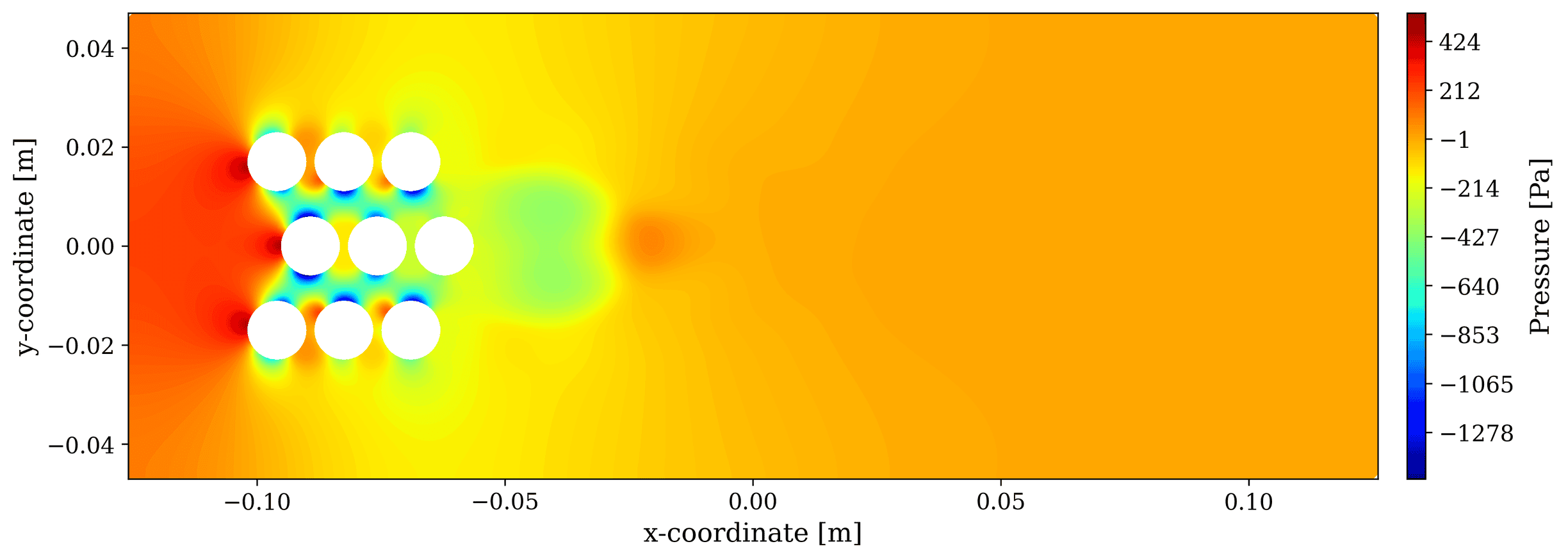} &
        \includegraphics[width=0.45\textwidth,valign=c]{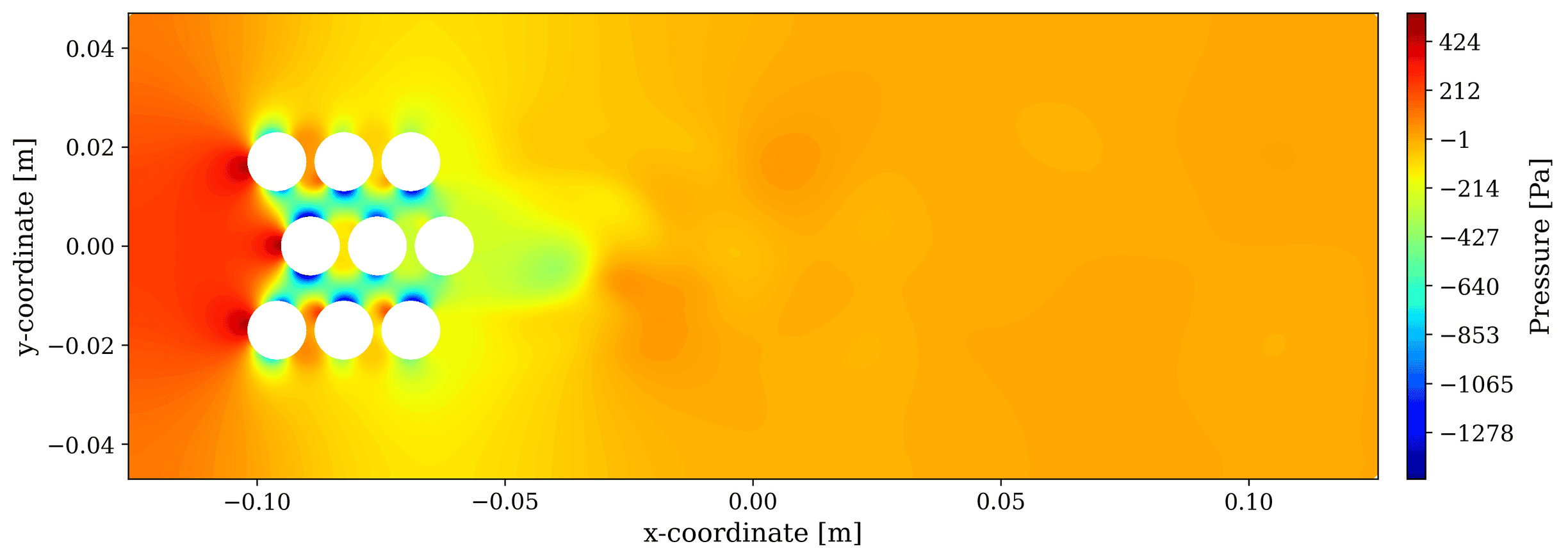} &
        \includegraphics[width=0.45\textwidth,valign=c]{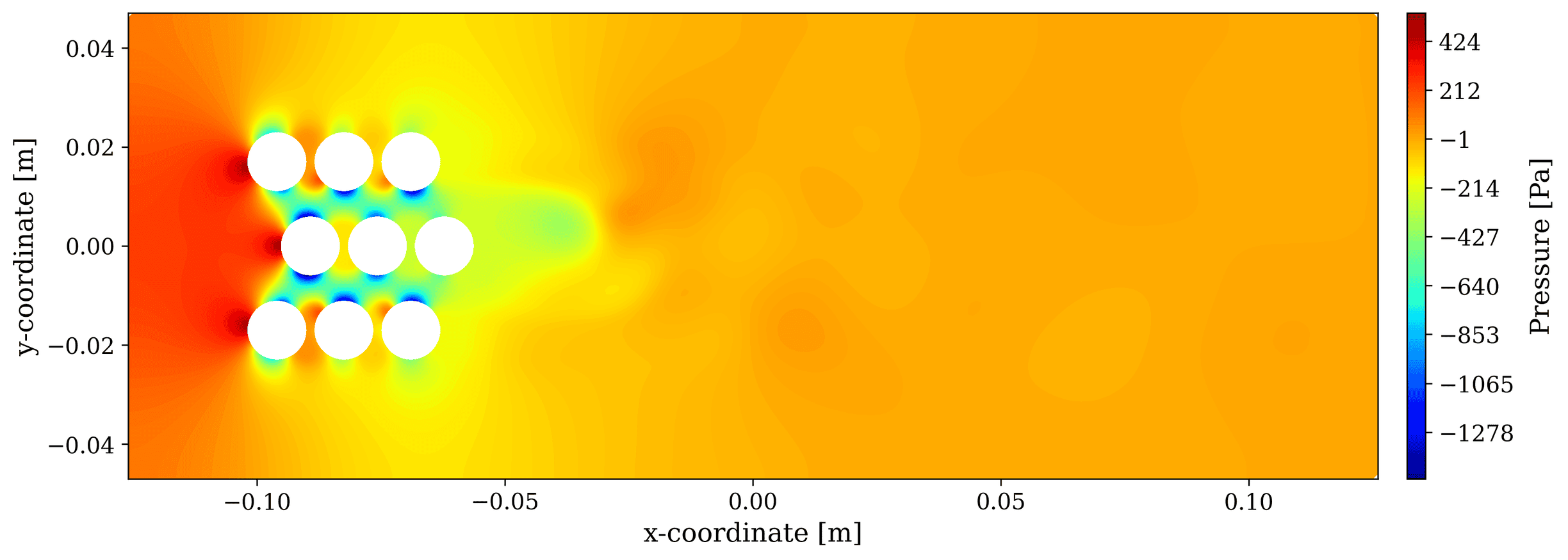} \\[2pt]
        \adjustbox{valign=c}{\rotatebox[origin=c]{90}{\small\textbf{Predicted}}} &
        \includegraphics[width=0.45\textwidth,valign=c]{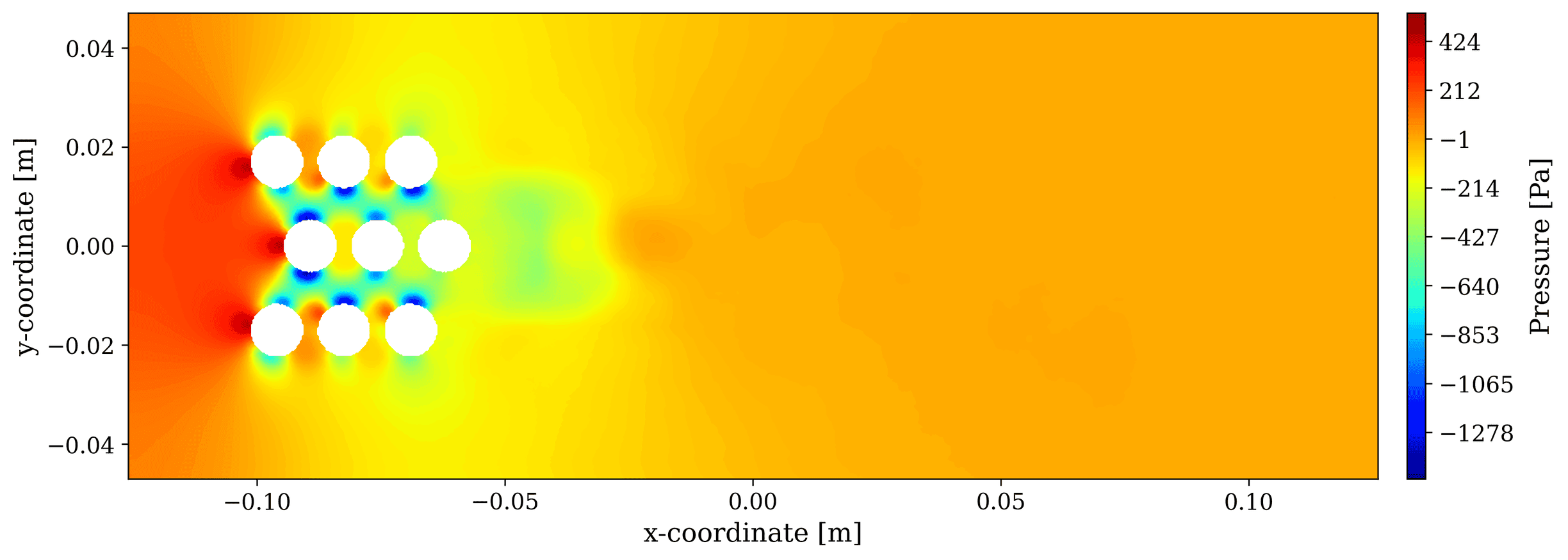} &
        \includegraphics[width=0.45\textwidth,valign=c]{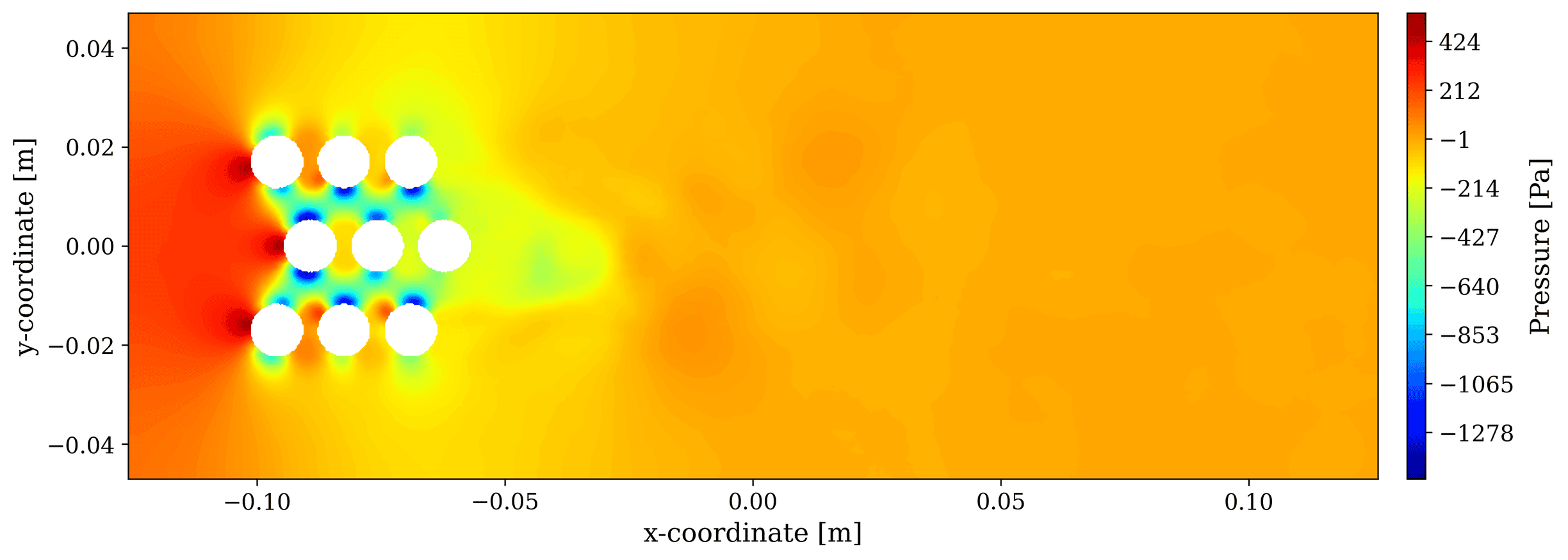} &
        \includegraphics[width=0.45\textwidth,valign=c]{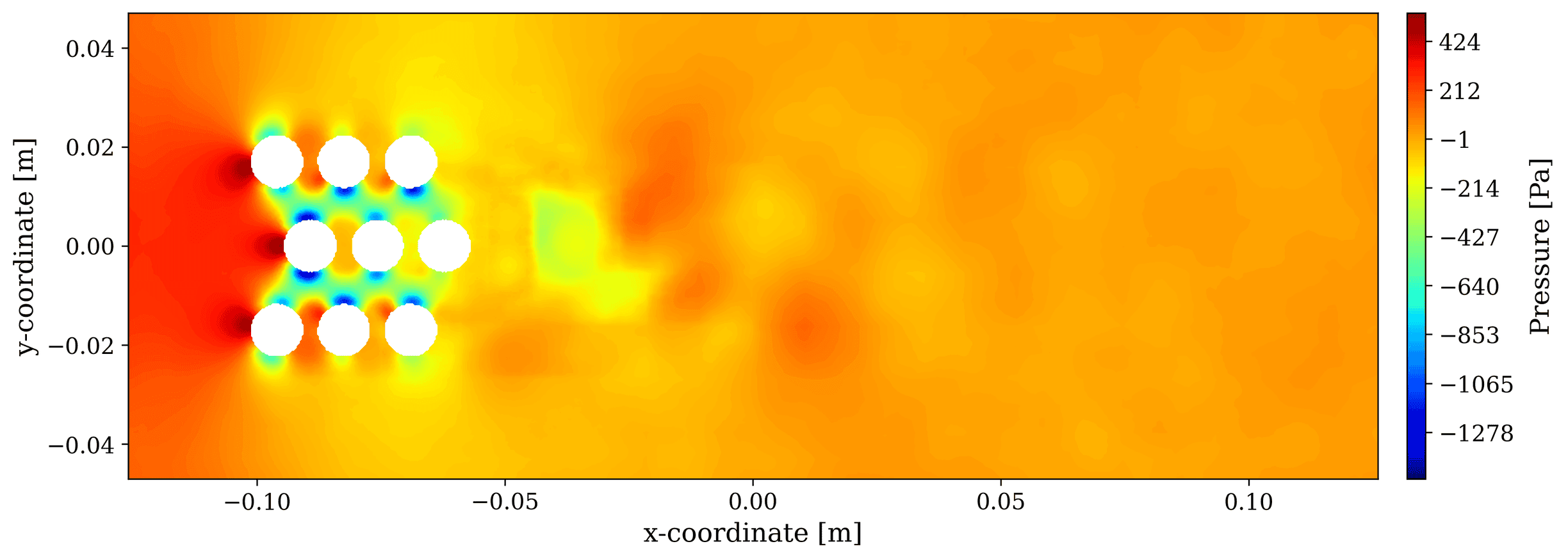} \\[2pt]
        \adjustbox{valign=c}{\rotatebox[origin=c]{90}{\small\textbf{Error}}} &
        \includegraphics[width=0.45\textwidth,valign=c]{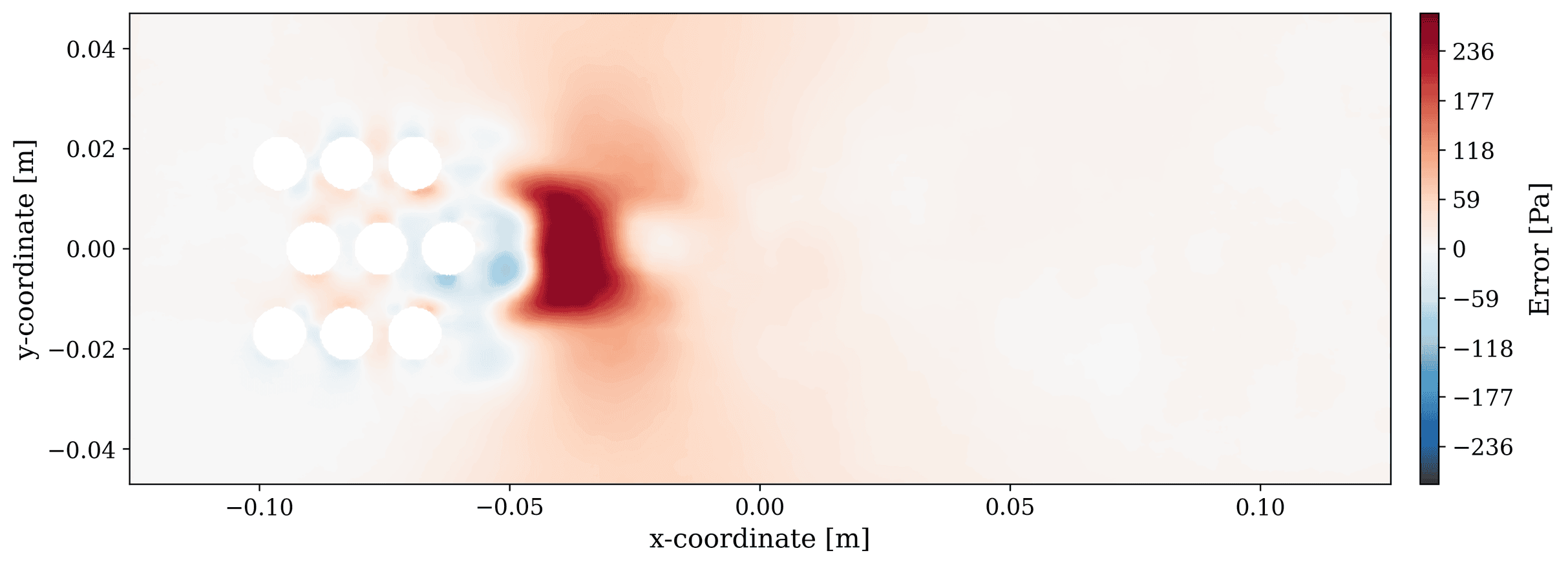} &
        \includegraphics[width=0.45\textwidth,valign=c]{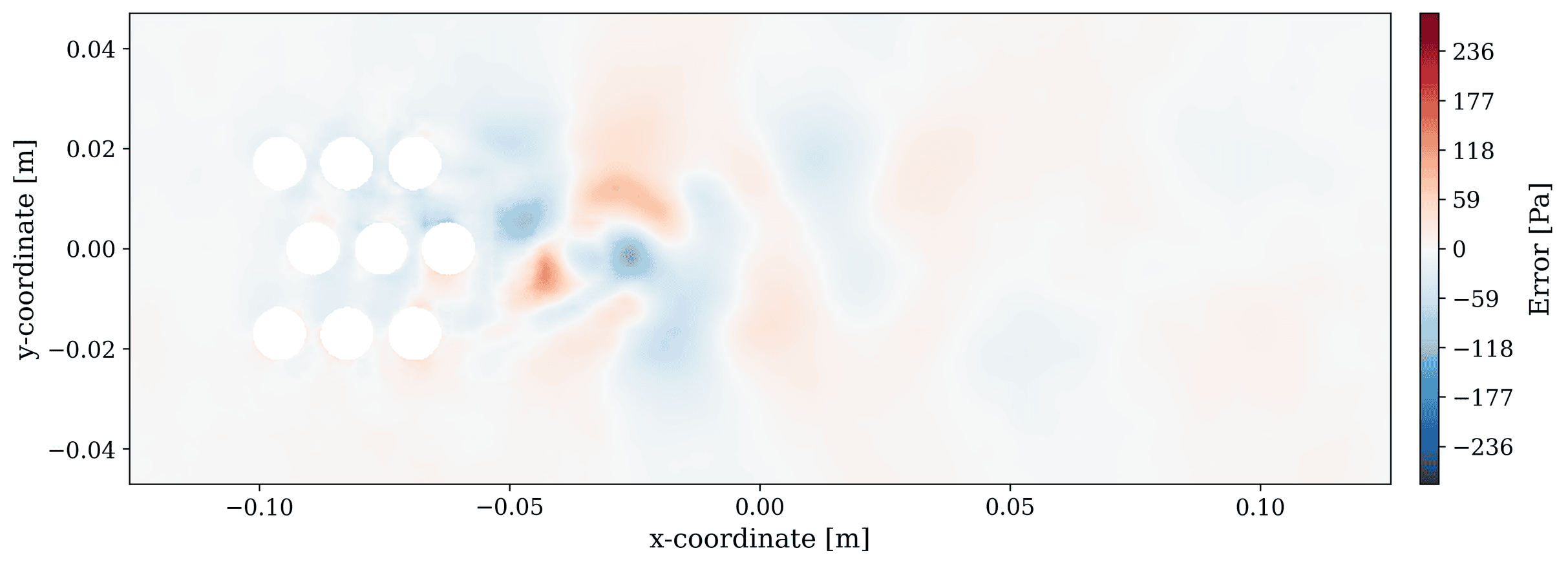} &
        \includegraphics[width=0.45\textwidth,valign=c]{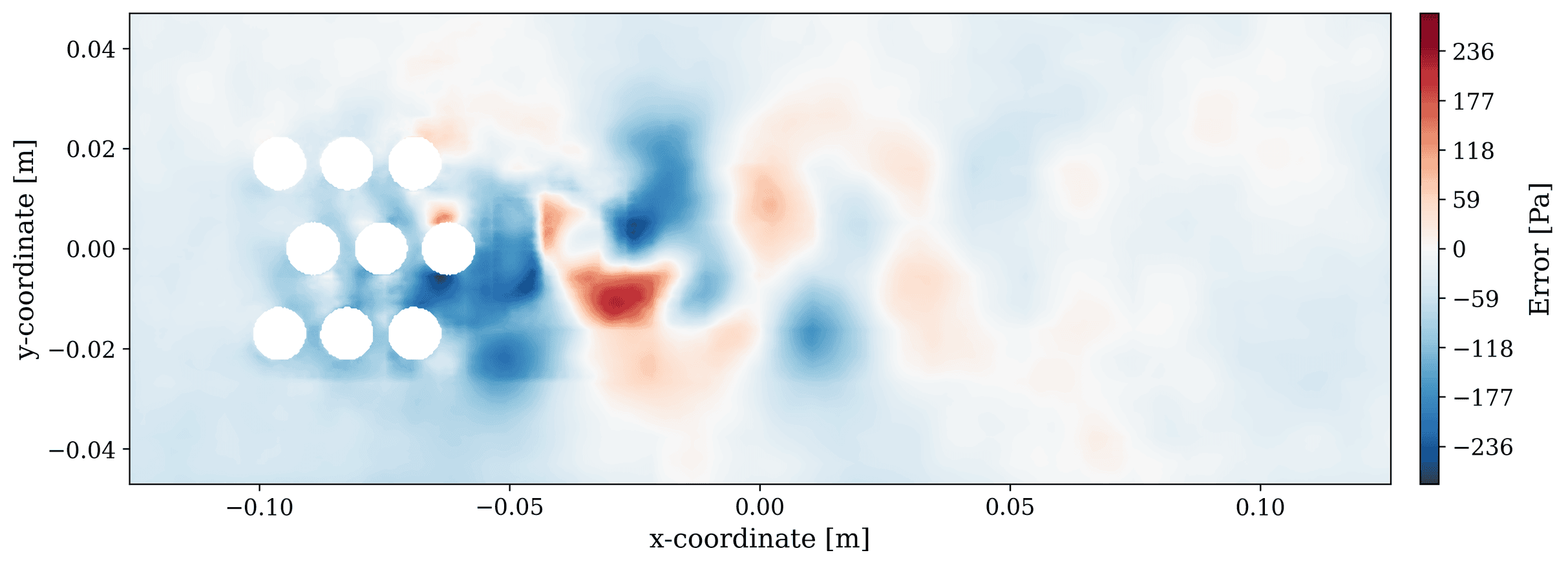} \\
    \end{tabular}
    }%
    \caption{CAE L-DeepONet results with multi-scale for pressure field with inlet velocity 0.7 $m/s$. Reference, predicted, and error at different timesteps.}
    \label{fig:cae_ldon_pressure_inlet070}
\end{figure}

\begin{figure}[H]
    \centering
    \setlength{\tabcolsep}{1pt}
    \makebox[\textwidth][c]{%
    \begin{tabular}{c@{\hspace{4pt}}ccc}
        & \textbf{$t = 2$} & \textbf{$t = 50$} & \textbf{$t = 100$} \\
        \adjustbox{valign=c}{\rotatebox[origin=c]{90}{\small\textbf{Reference}}} &
        \includegraphics[width=0.45\textwidth,valign=c]{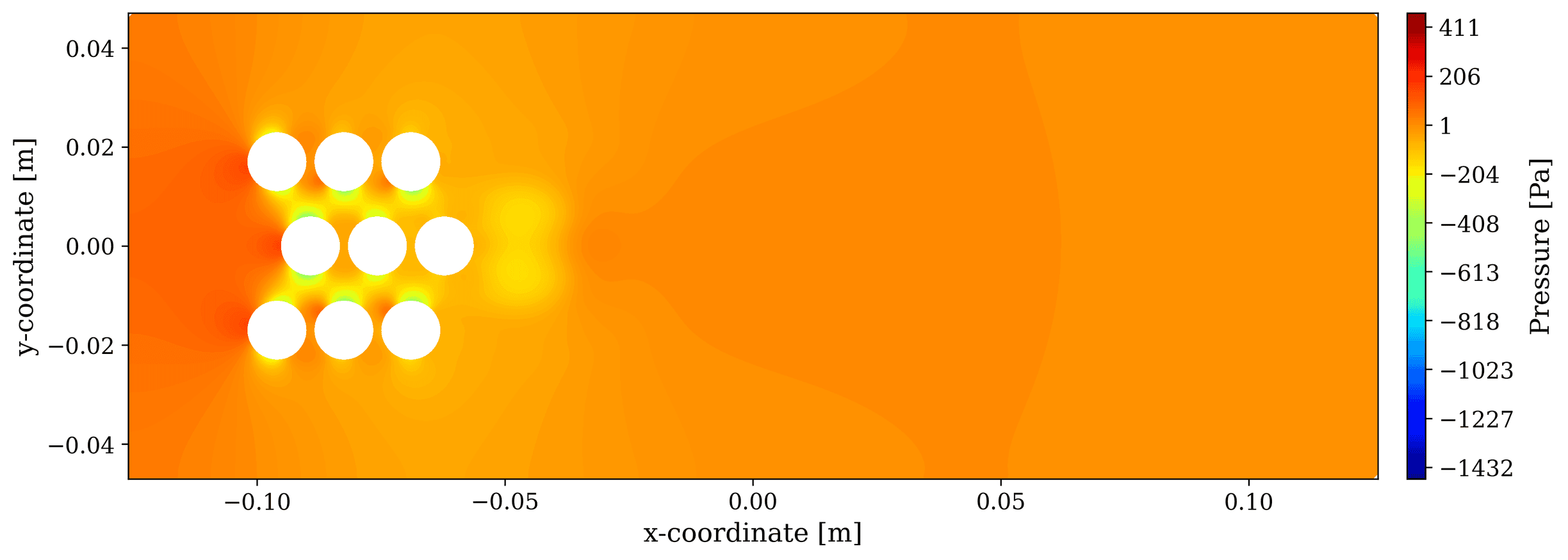} &
        \includegraphics[width=0.45\textwidth,valign=c]{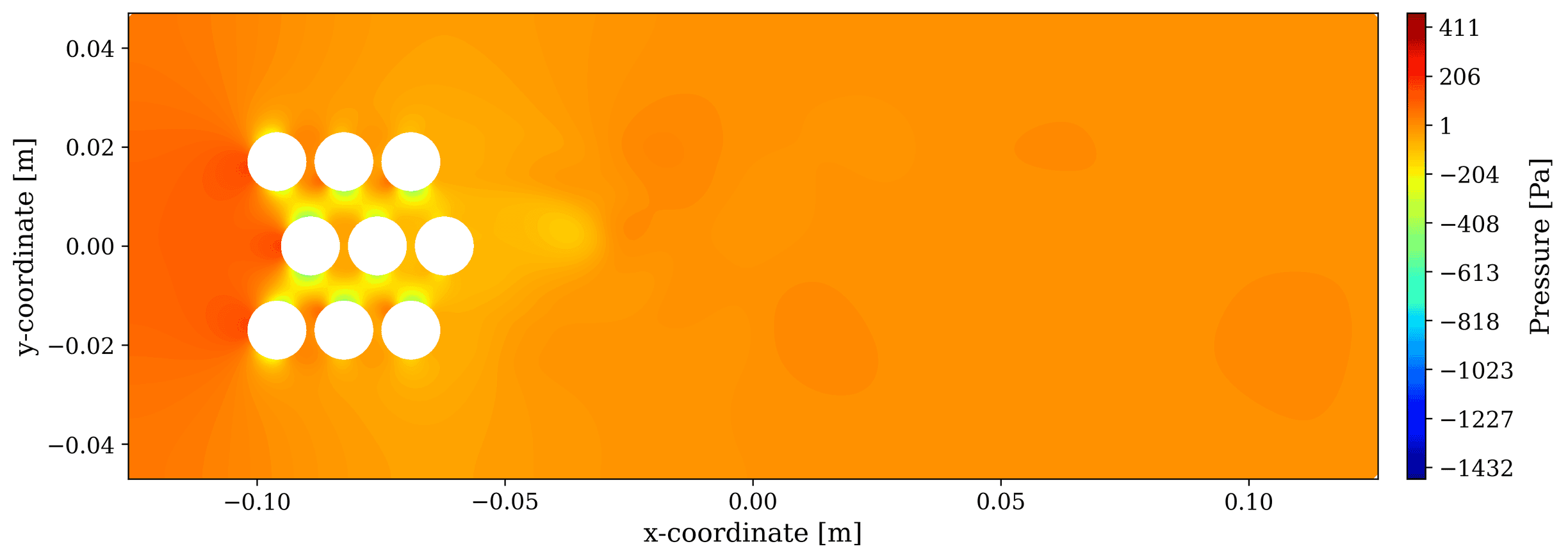} &
        \includegraphics[width=0.45\textwidth,valign=c]{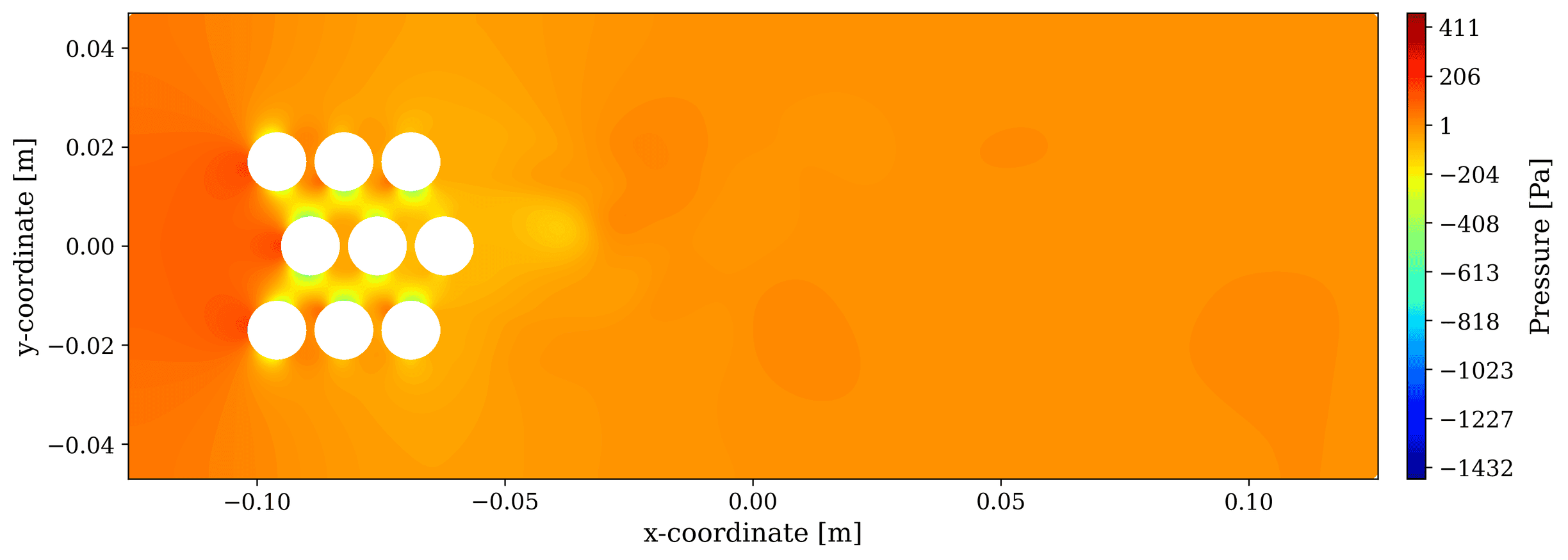} \\[2pt]
        \adjustbox{valign=c}{\rotatebox[origin=c]{90}{\small\textbf{Predicted}}} &
        \includegraphics[width=0.45\textwidth,valign=c]{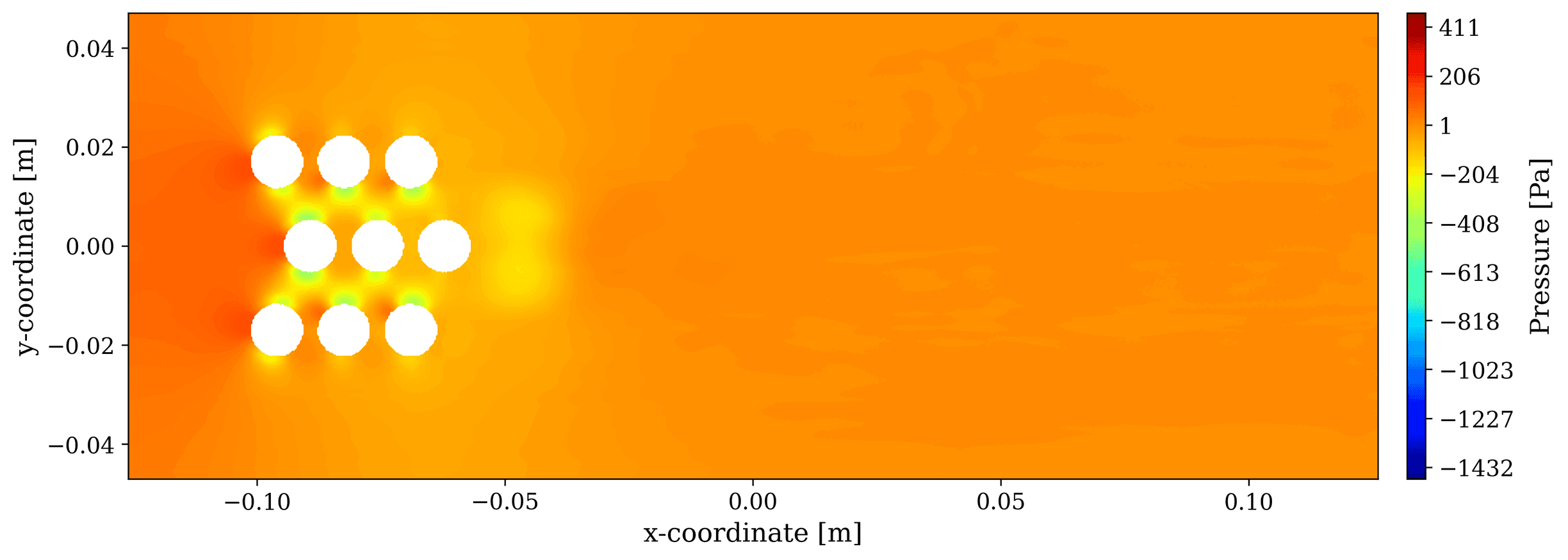} &
        \includegraphics[width=0.45\textwidth,valign=c]{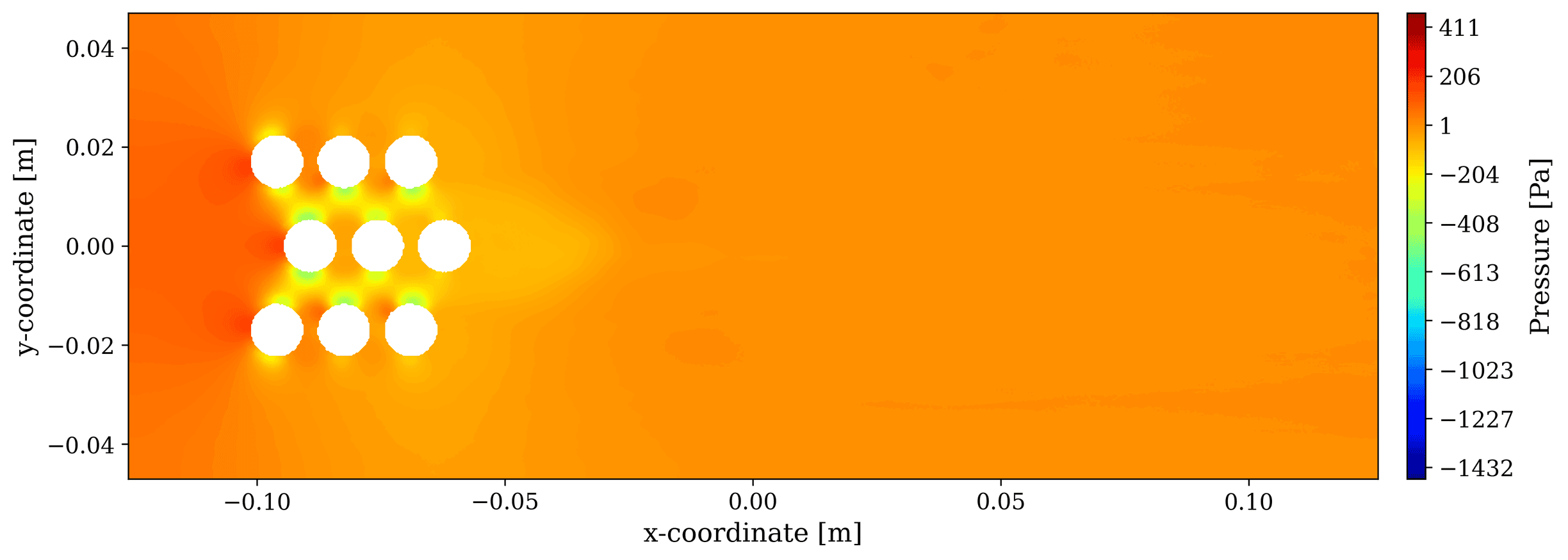} &
        \includegraphics[width=0.45\textwidth,valign=c]{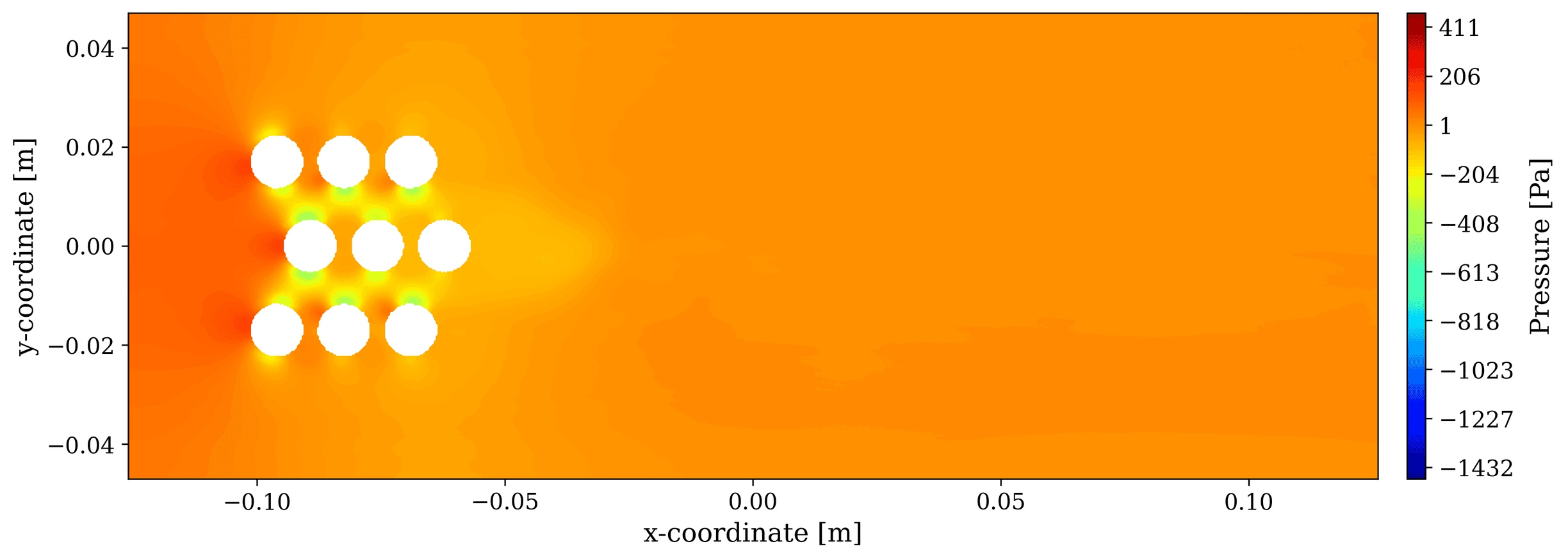} \\[2pt]
        \adjustbox{valign=c}{\rotatebox[origin=c]{90}{\small\textbf{Error}}} &
        \includegraphics[width=0.45\textwidth,valign=c]{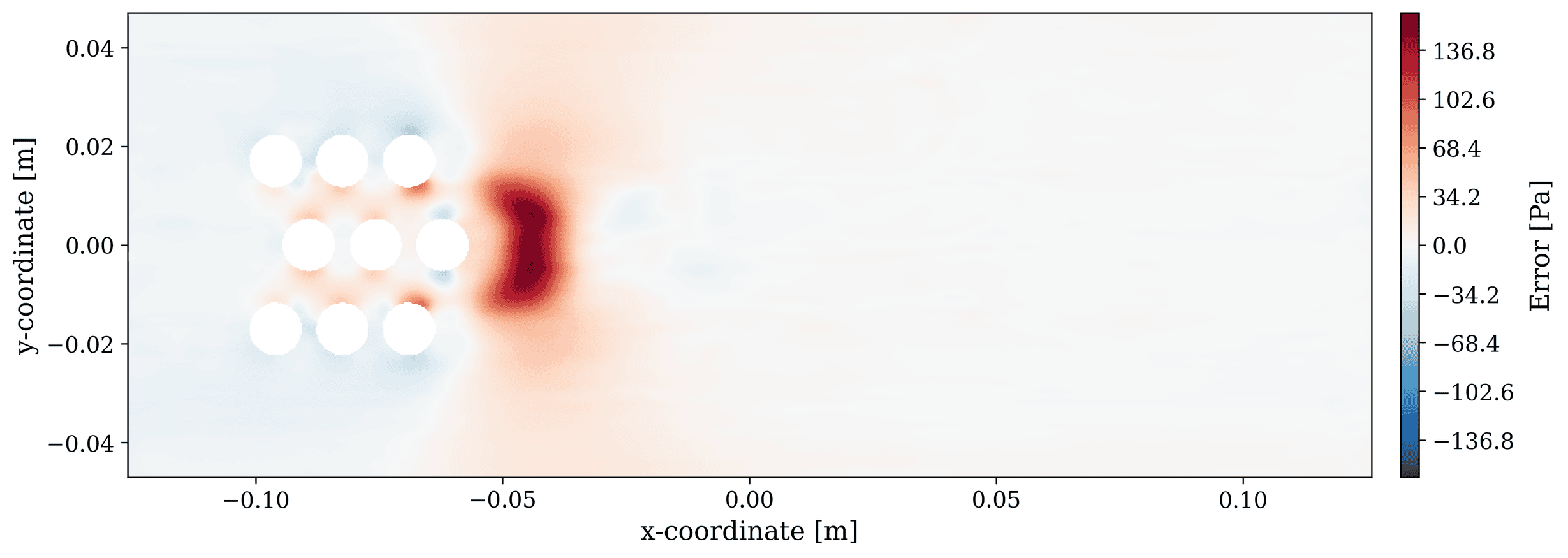} &
        \includegraphics[width=0.45\textwidth,valign=c]{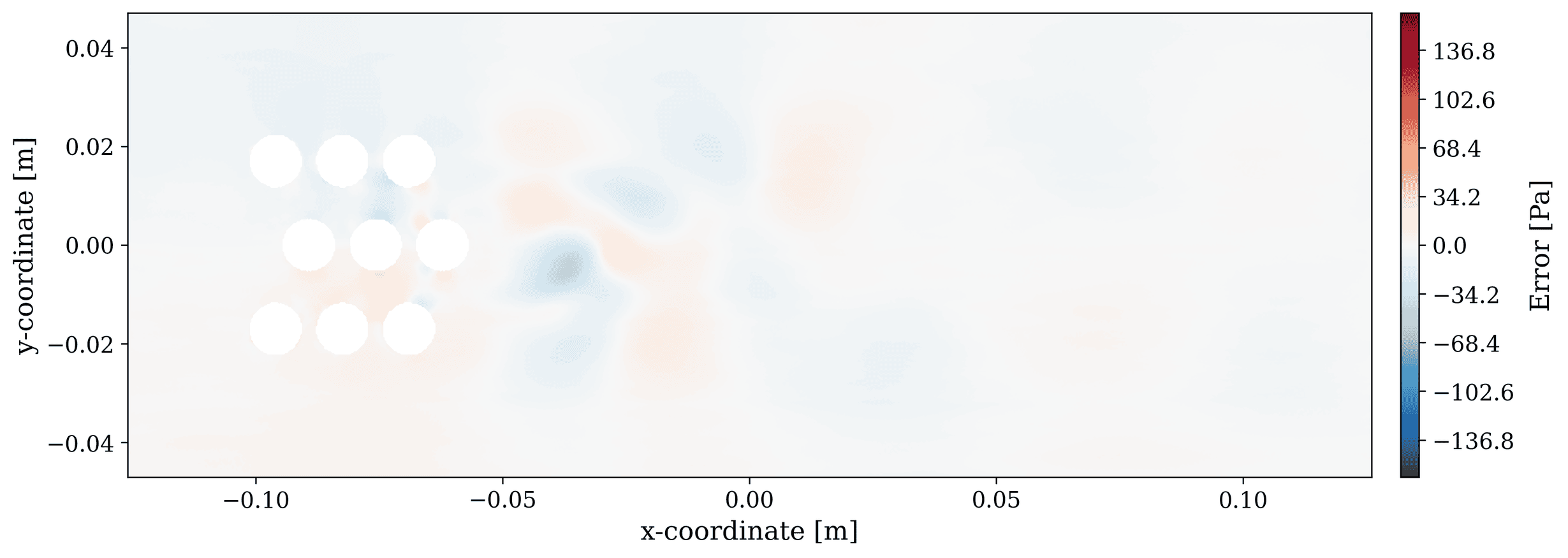} &
        \includegraphics[width=0.45\textwidth,valign=c]{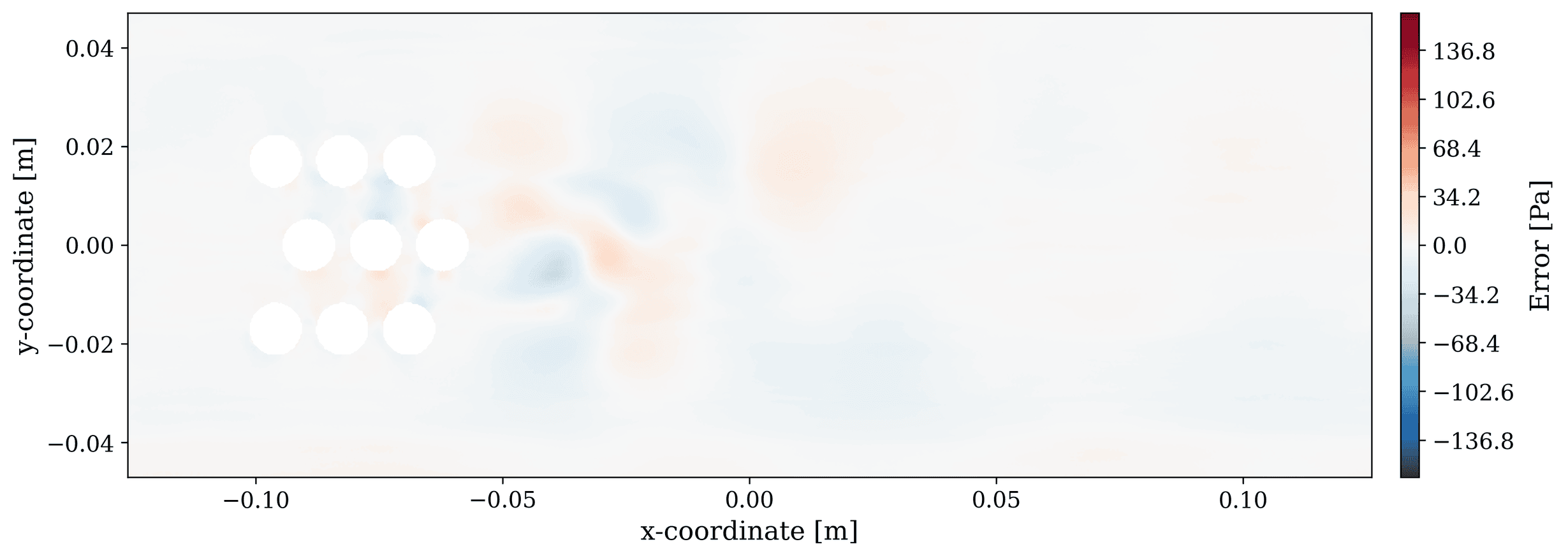} \\
    \end{tabular}
    }%
    \caption{FNO results for pressure field with inlet velocity 0.4 $m/s$. Reference, predicted, and error at different timesteps. The Fourier mode is 12.}
    \label{fig:fno_pressure_inlet040}
\end{figure}

\begin{figure}[H]
    \centering
    \setlength{\tabcolsep}{1pt}
    \makebox[\textwidth][c]{%
    \begin{tabular}{c@{\hspace{4pt}}ccc}
        & \textbf{$t = 2$} & \textbf{$t = 50$} & \textbf{$t = 100$} \\
        \adjustbox{valign=c}{\rotatebox[origin=c]{90}{\small\textbf{Reference}}} &
        \includegraphics[width=0.45\textwidth,valign=c]{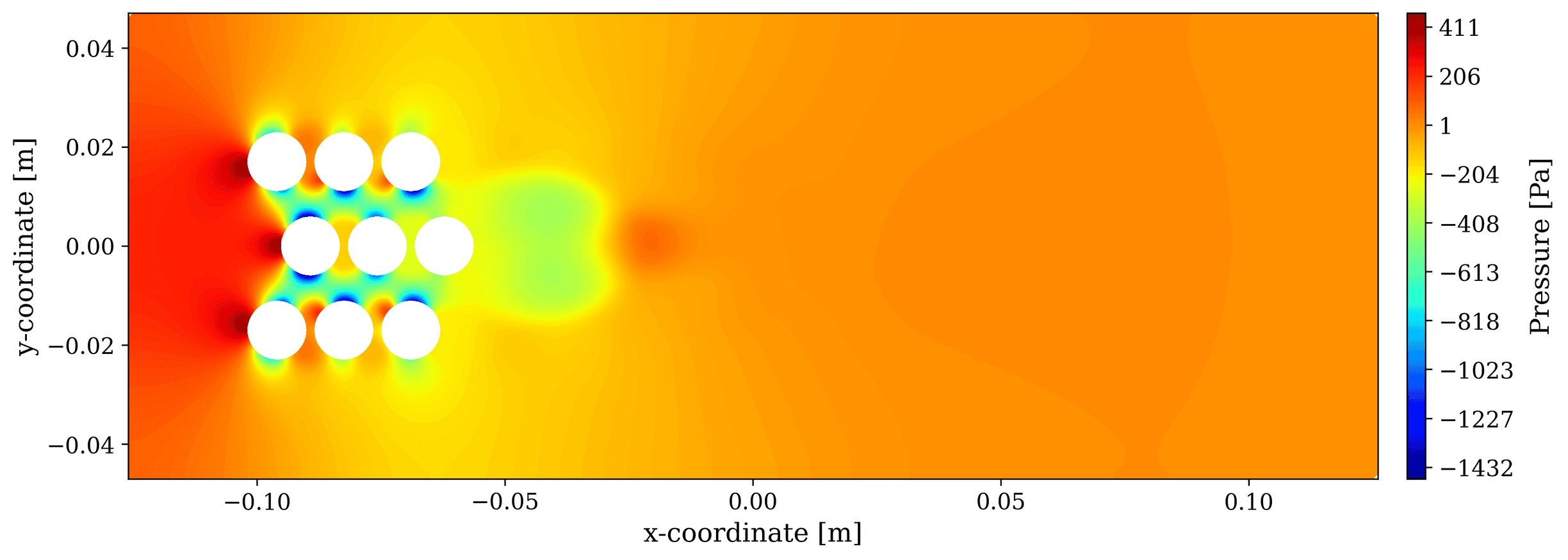} &
        \includegraphics[width=0.45\textwidth,valign=c]{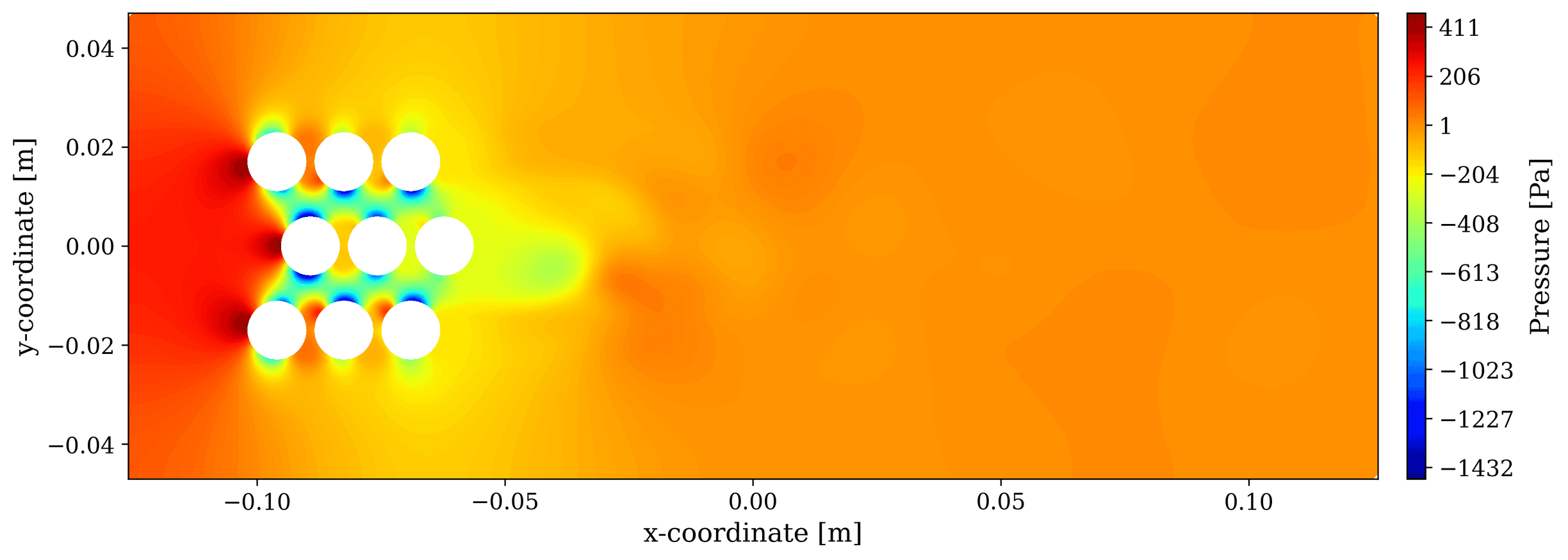} &
        \includegraphics[width=0.45\textwidth,valign=c]{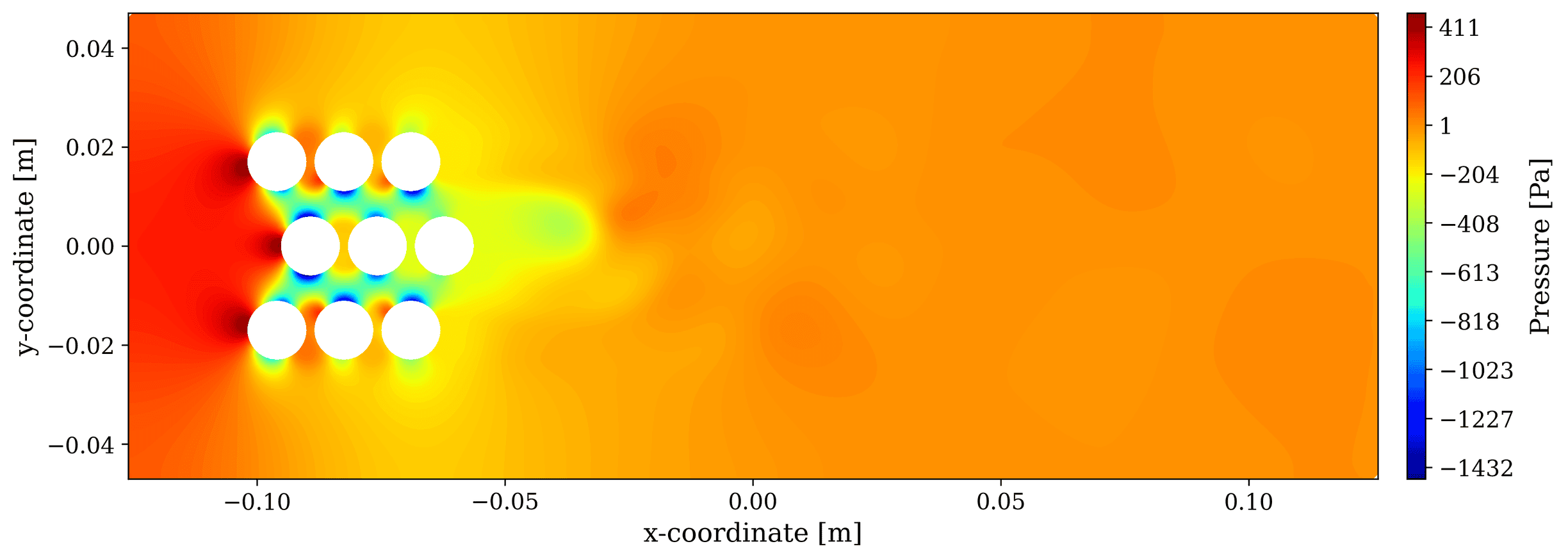} \\[2pt]
        \adjustbox{valign=c}{\rotatebox[origin=c]{90}{\small\textbf{Predicted}}} &
        \includegraphics[width=0.45\textwidth,valign=c]{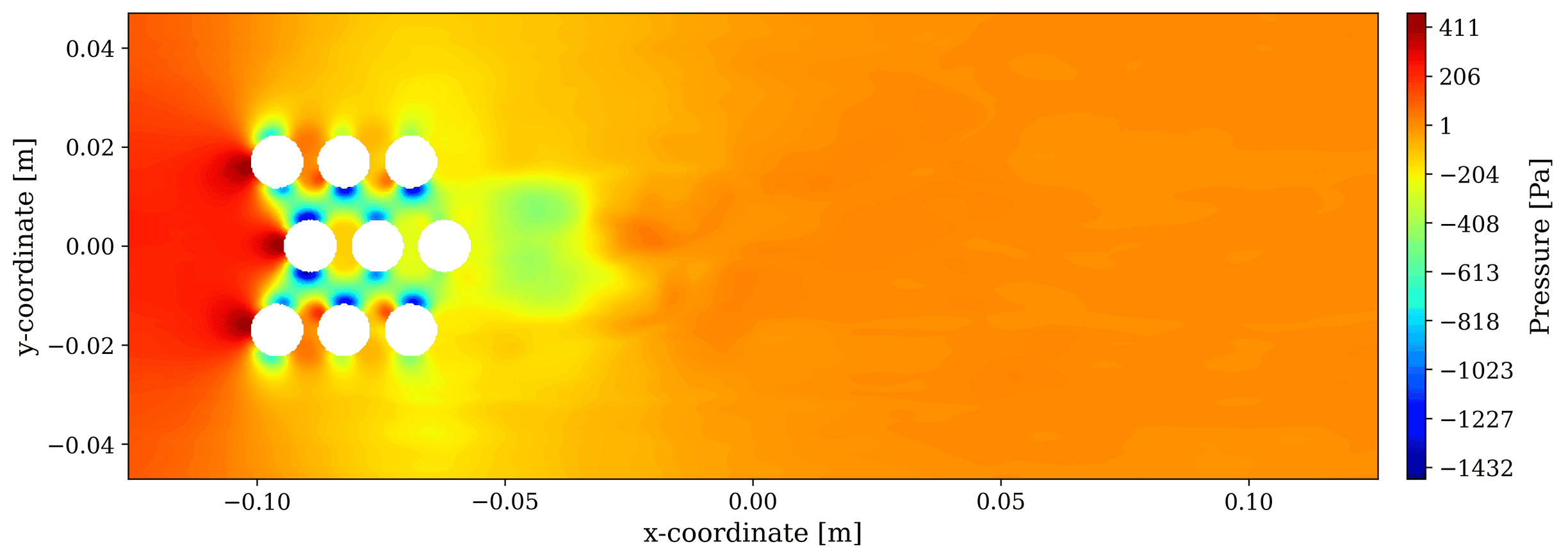} &
        \includegraphics[width=0.45\textwidth,valign=c]{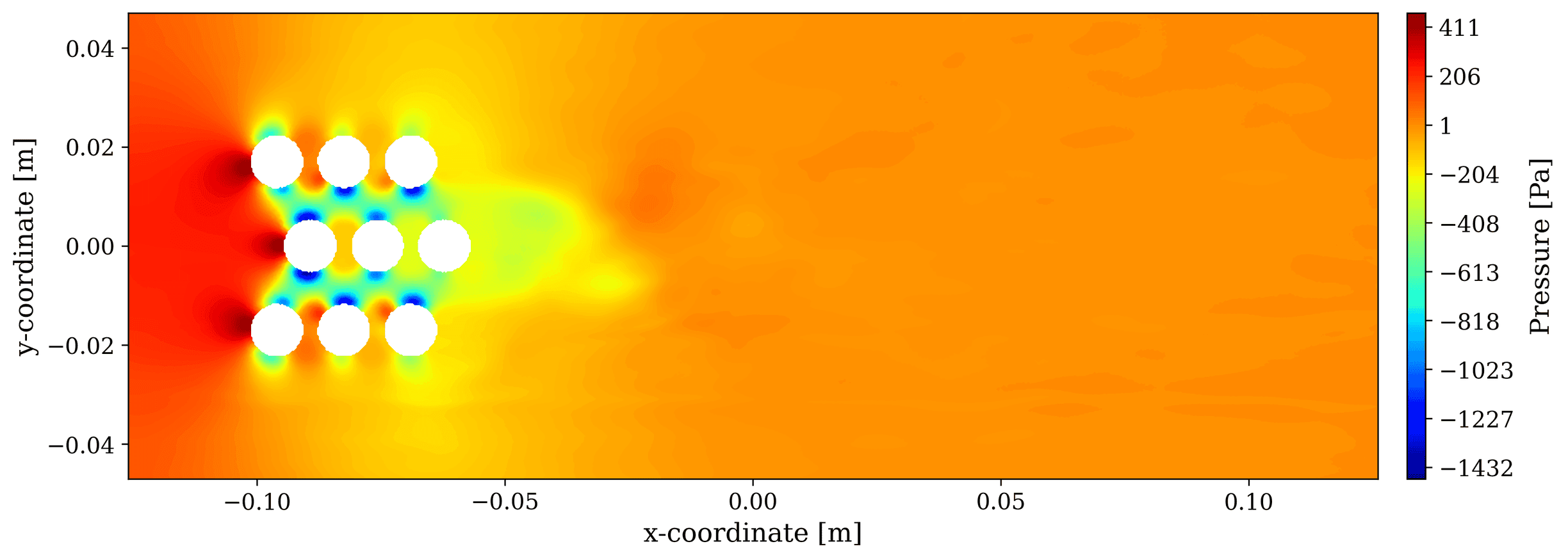} &
        \includegraphics[width=0.45\textwidth,valign=c]{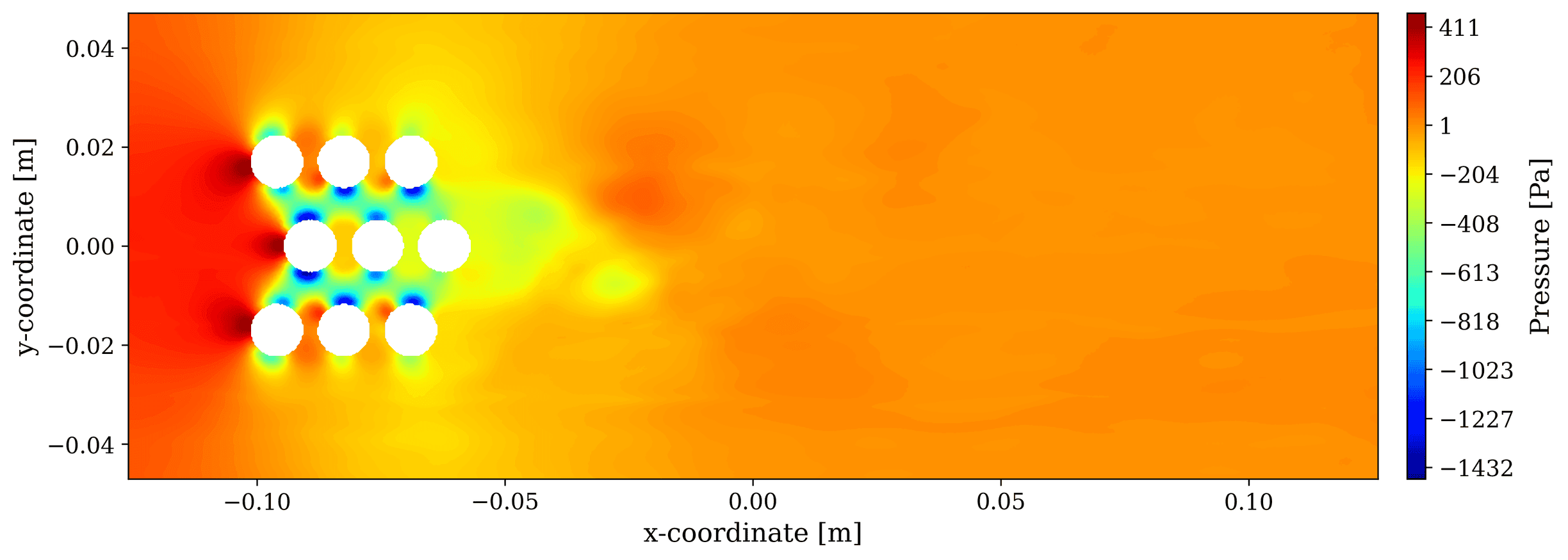} \\[2pt]
        \adjustbox{valign=c}{\rotatebox[origin=c]{90}{\small\textbf{Error}}} &
        \includegraphics[width=0.45\textwidth,valign=c]{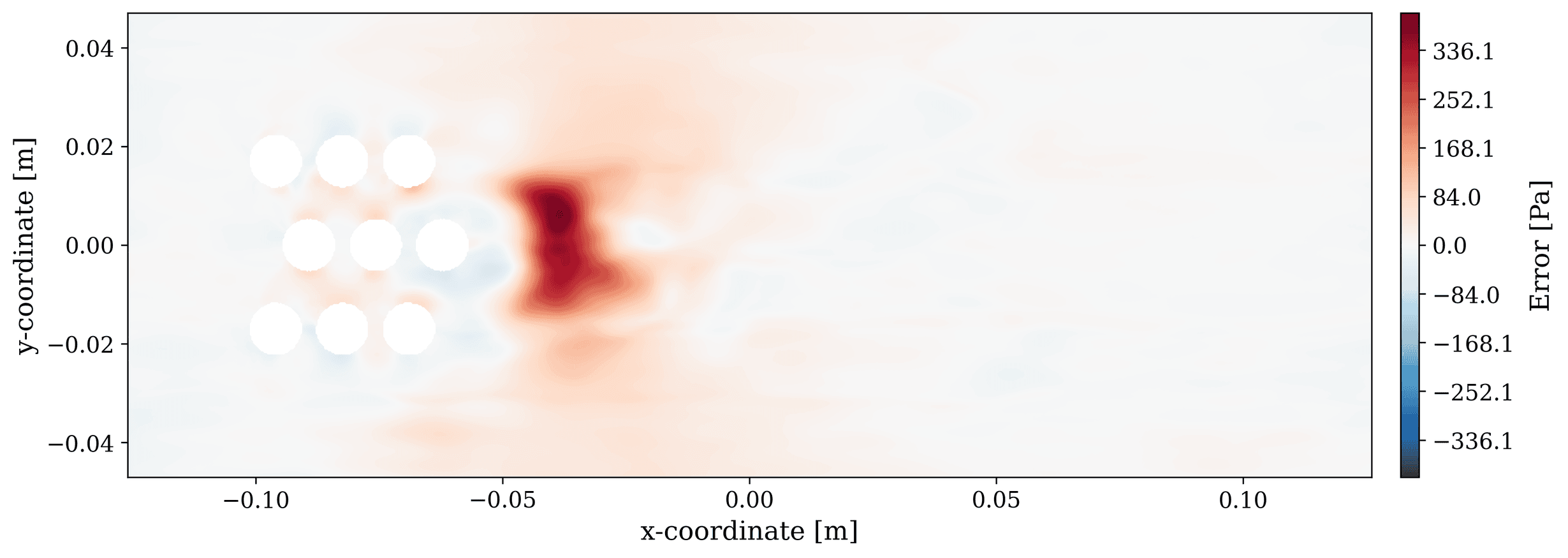} &
        \includegraphics[width=0.45\textwidth,valign=c]{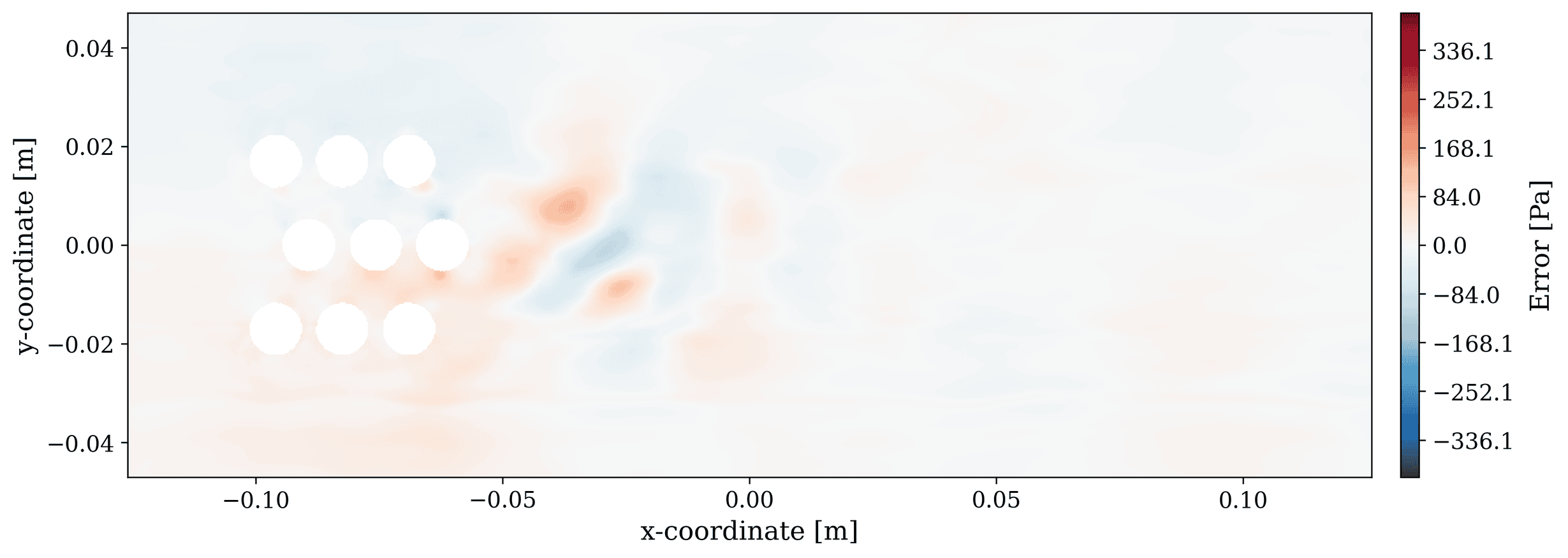} &
        \includegraphics[width=0.45\textwidth,valign=c]{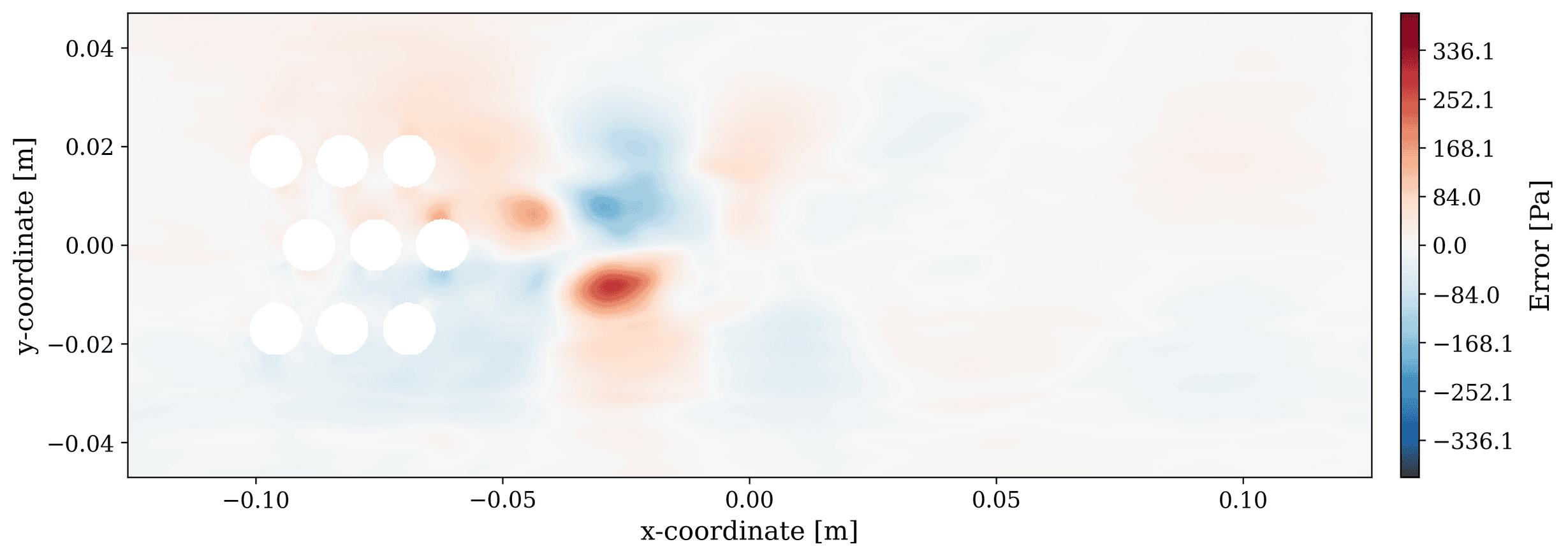} \\
    \end{tabular}
    }%
    \caption{FNO results for pressure field with inlet velocity 0.7 $m/s$. Reference, predicted, and error at different timesteps. The Fourier mode is 12.}
    \label{fig:fno_pressure_inlet070}
\end{figure}

\begin{figure}[H]
    \centering
    \setlength{\tabcolsep}{1pt}
    \makebox[\textwidth][c]{%
    \begin{tabular}{c@{\hspace{4pt}}ccc}
        & \textbf{$t = 2$} & \textbf{$t = 50$} & \textbf{$t = 100$} \\
        \adjustbox{valign=c}{\rotatebox[origin=c]{90}{\small\textbf{Reference}}} &
        \includegraphics[width=0.45\textwidth,valign=c]{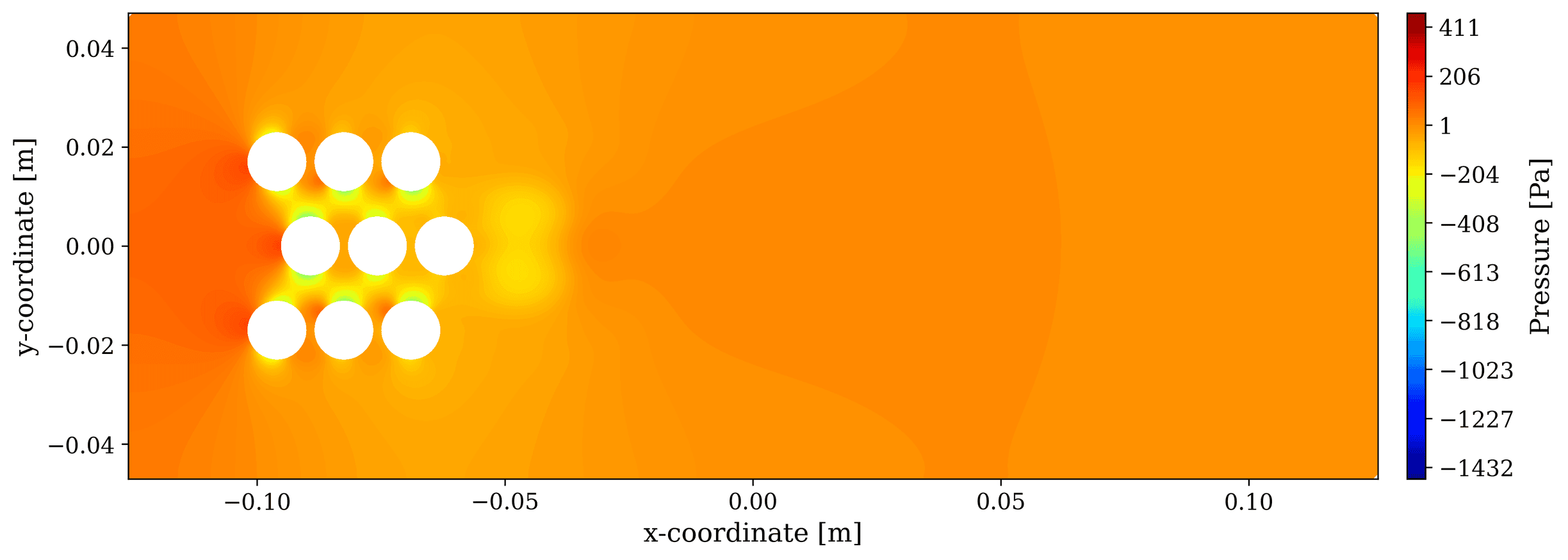} &
        \includegraphics[width=0.45\textwidth,valign=c]{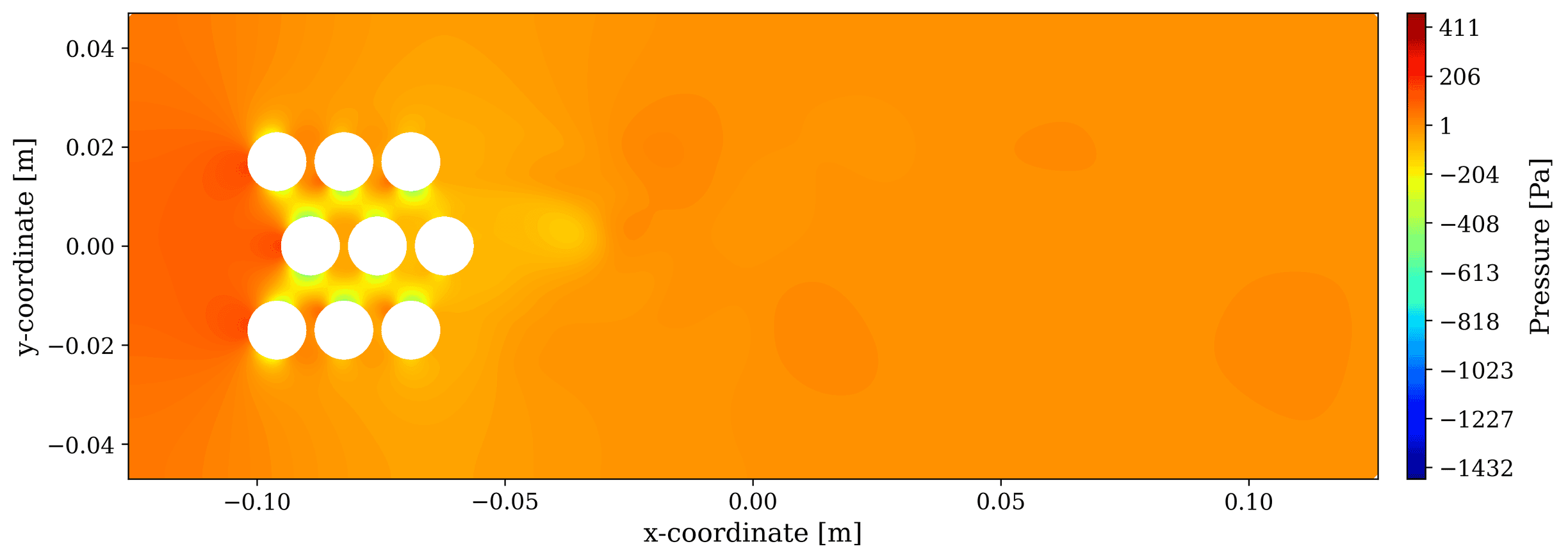} &
        \includegraphics[width=0.45\textwidth,valign=c]{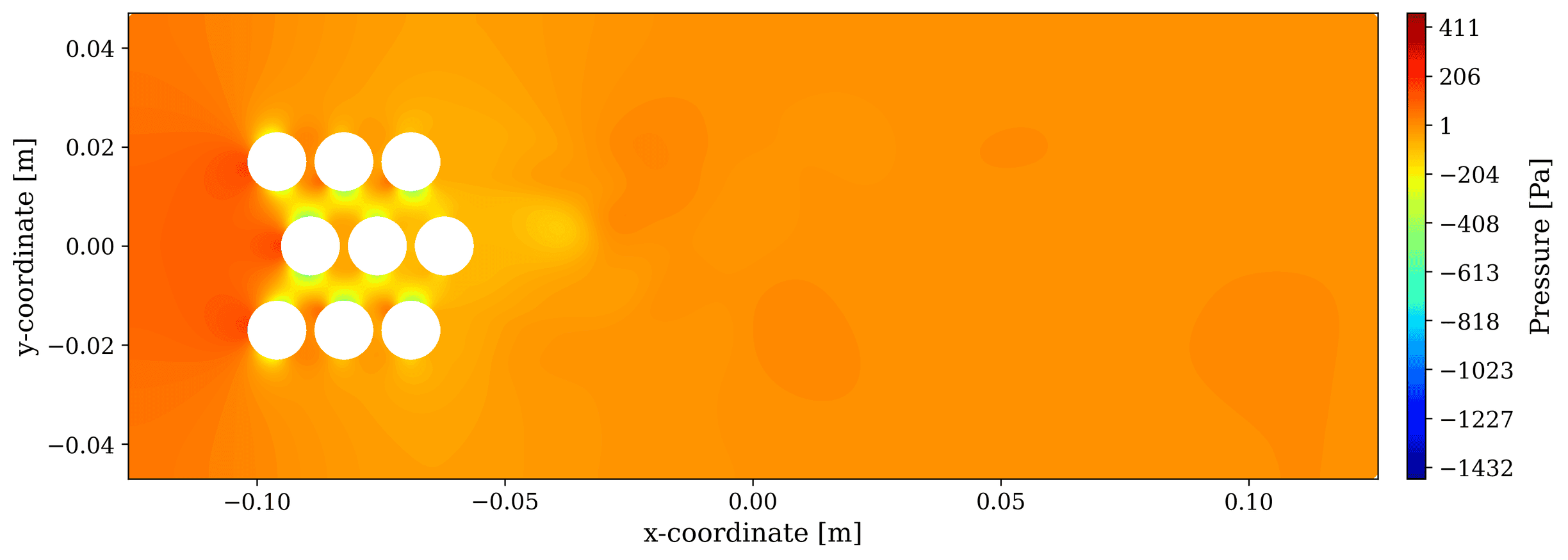} \\[2pt]
        \adjustbox{valign=c}{\rotatebox[origin=c]{90}{\small\textbf{Predicted}}} &
        \includegraphics[width=0.45\textwidth,valign=c]{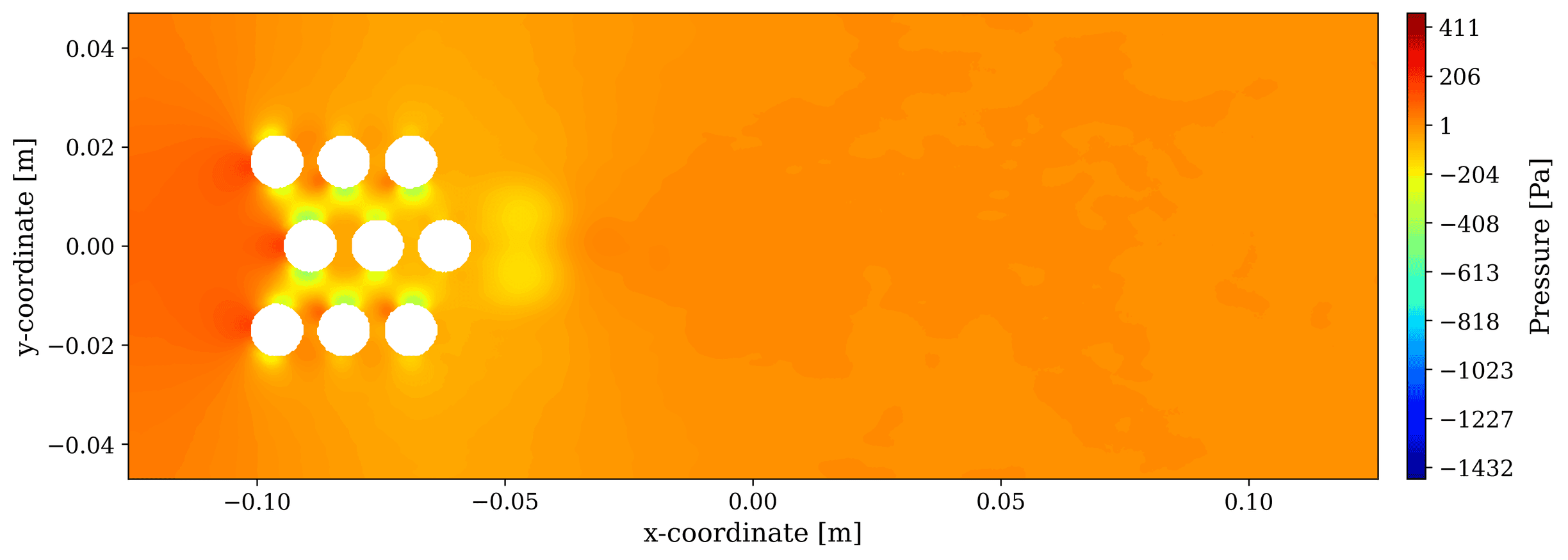} &
        \includegraphics[width=0.45\textwidth,valign=c]{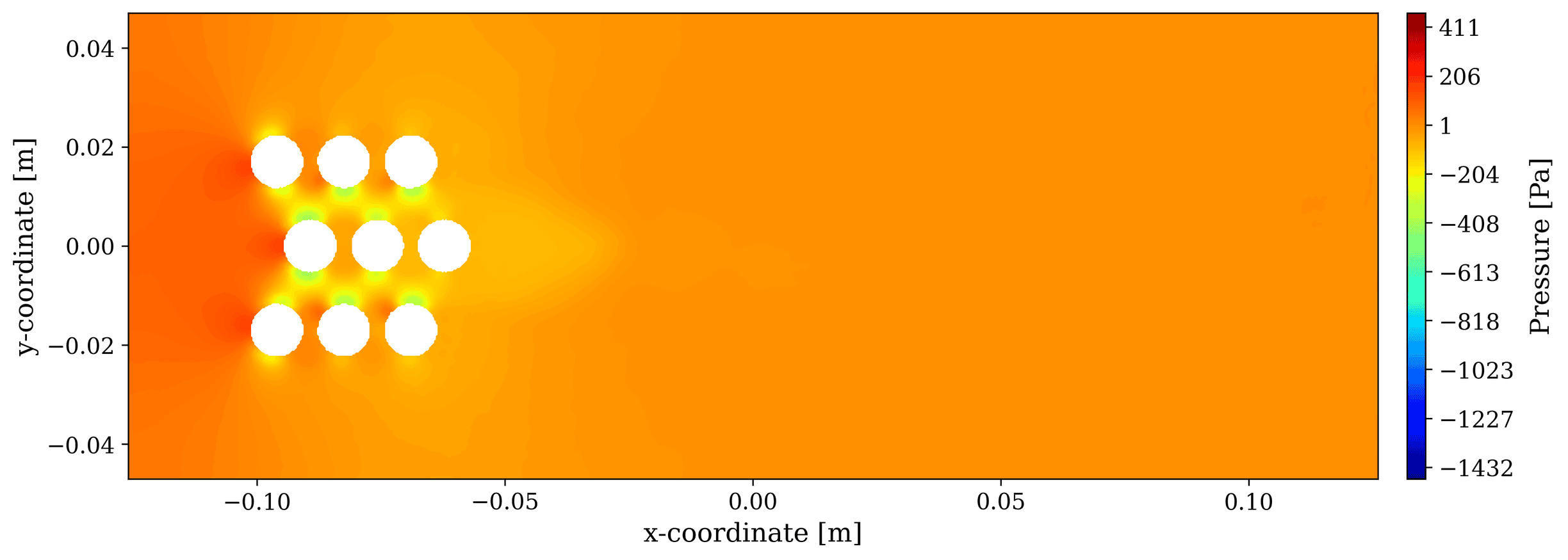} &
        \includegraphics[width=0.45\textwidth,valign=c]{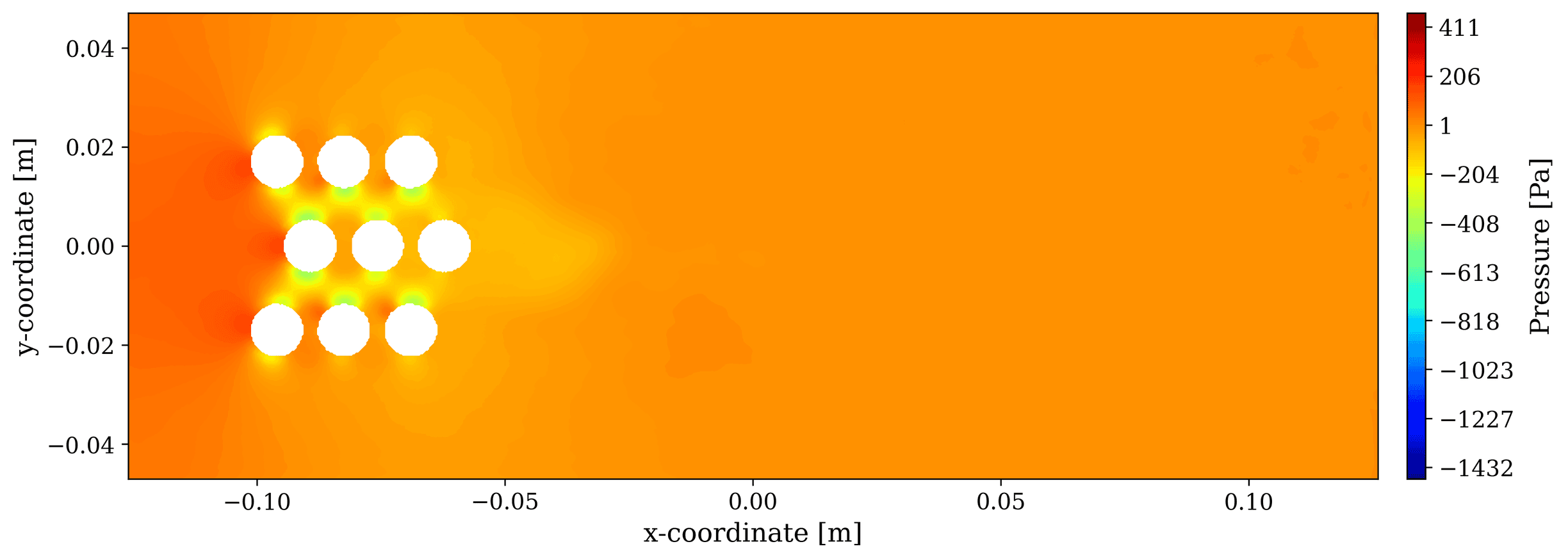} \\[2pt]
        \adjustbox{valign=c}{\rotatebox[origin=c]{90}{\small\textbf{Error}}} &
        \includegraphics[width=0.45\textwidth,valign=c]{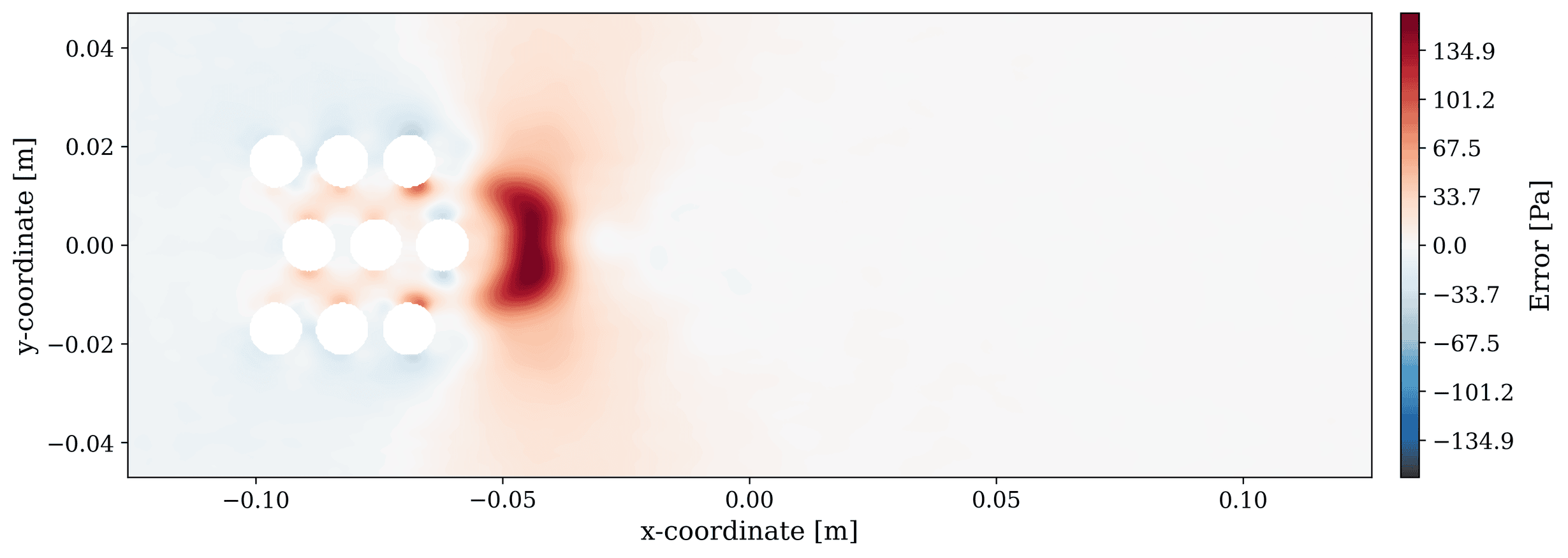} &
        \includegraphics[width=0.45\textwidth,valign=c]{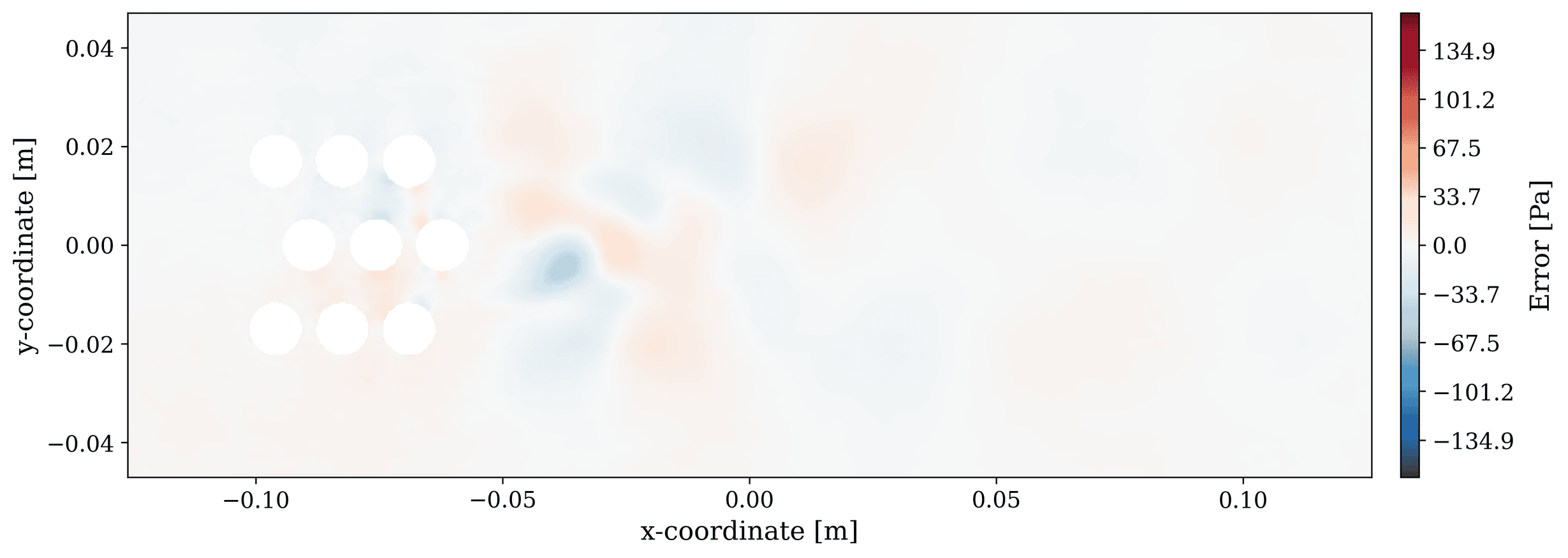} &
        \includegraphics[width=0.45\textwidth,valign=c]{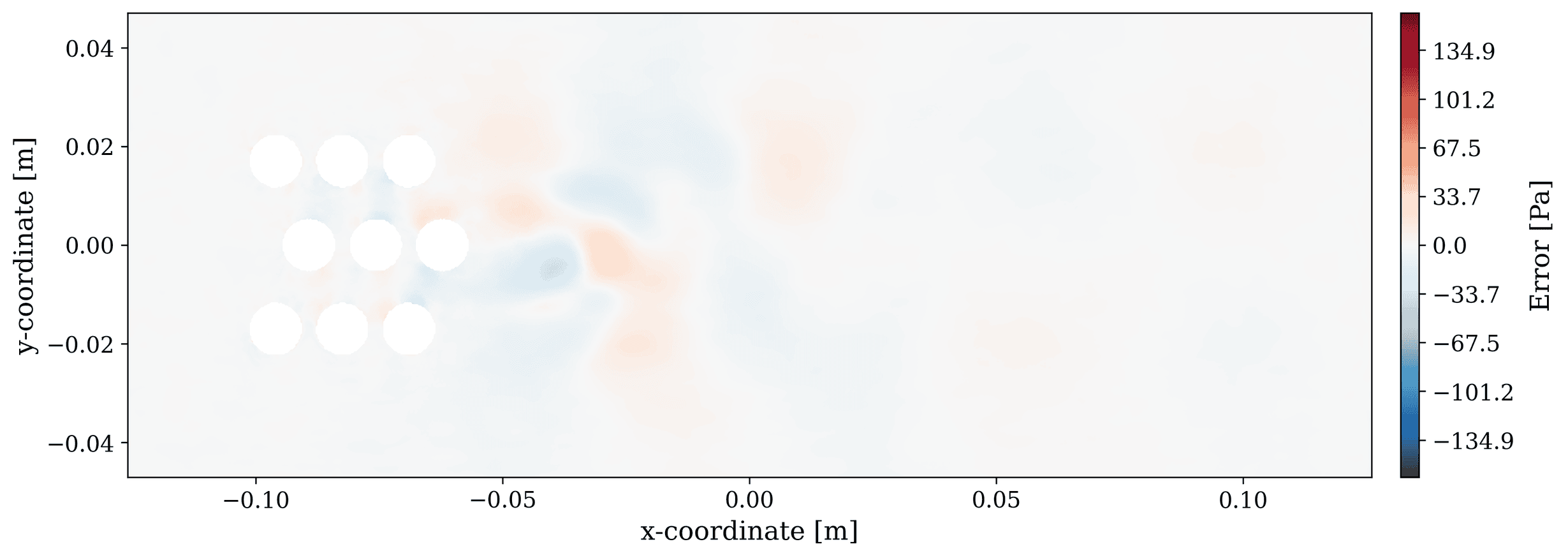} \\
    \end{tabular}
    }%
    \caption{MscaleFNO: pressure field prediction for inlet velocity 0.4 m/s. Top: reference CFD; middle: prediction; bottom: error.}
    \label{fig:mscalefno_pressure_inlet040}
\end{figure}

\begin{figure}[H]
    \centering
    \setlength{\tabcolsep}{1pt}
    \makebox[\textwidth][c]{%
    \begin{tabular}{c@{\hspace{4pt}}ccc}
        & \textbf{$t = 2$} & \textbf{$t = 50$} & \textbf{$t = 100$} \\
        \adjustbox{valign=c}{\rotatebox[origin=c]{90}{\small\textbf{Reference}}} &
        \includegraphics[width=0.45\textwidth,valign=c]{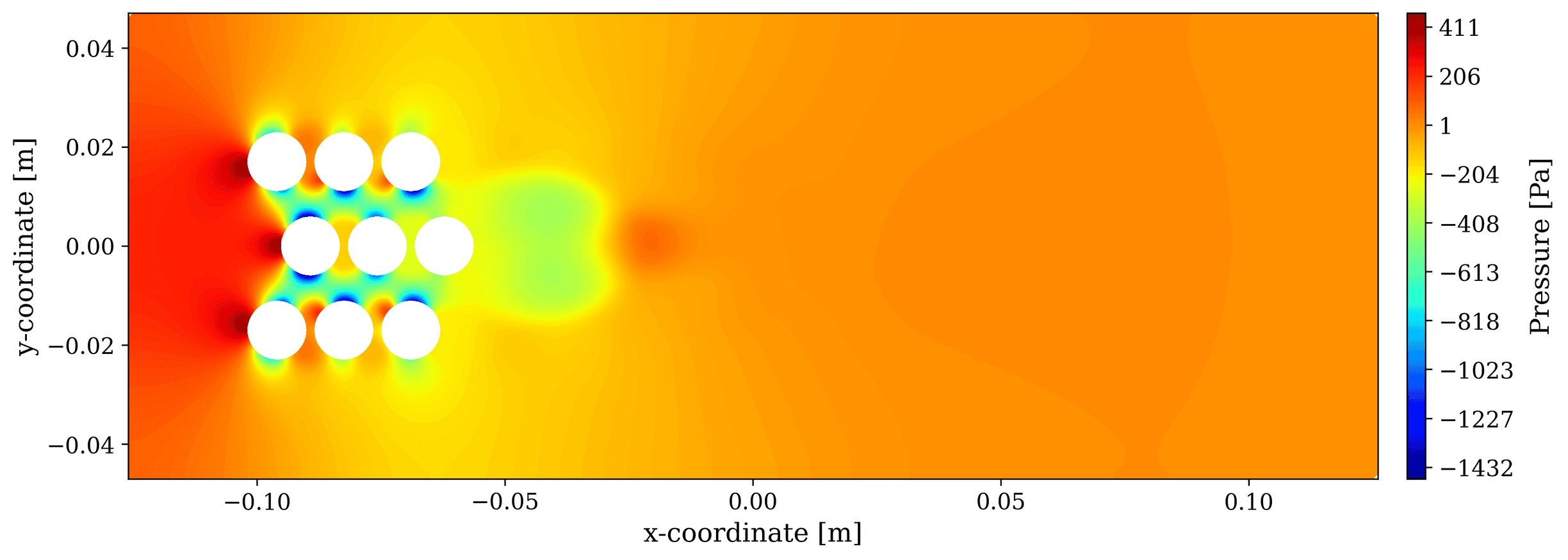} &
        \includegraphics[width=0.45\textwidth,valign=c]{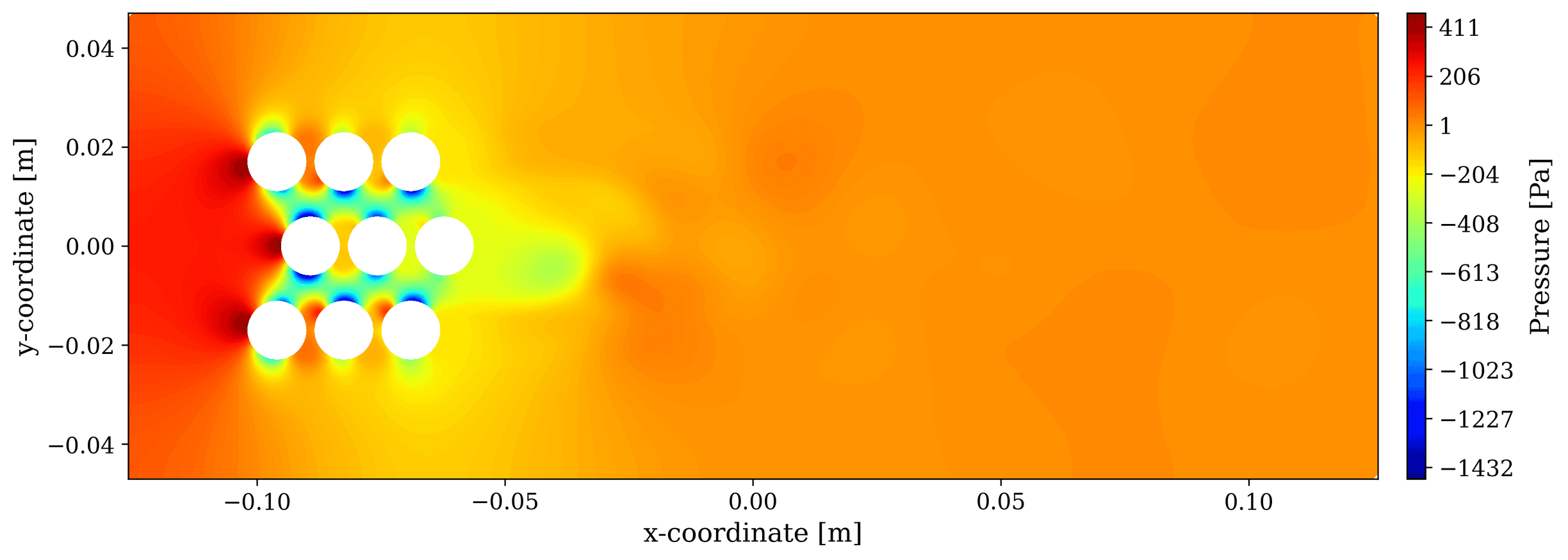} &
        \includegraphics[width=0.45\textwidth,valign=c]{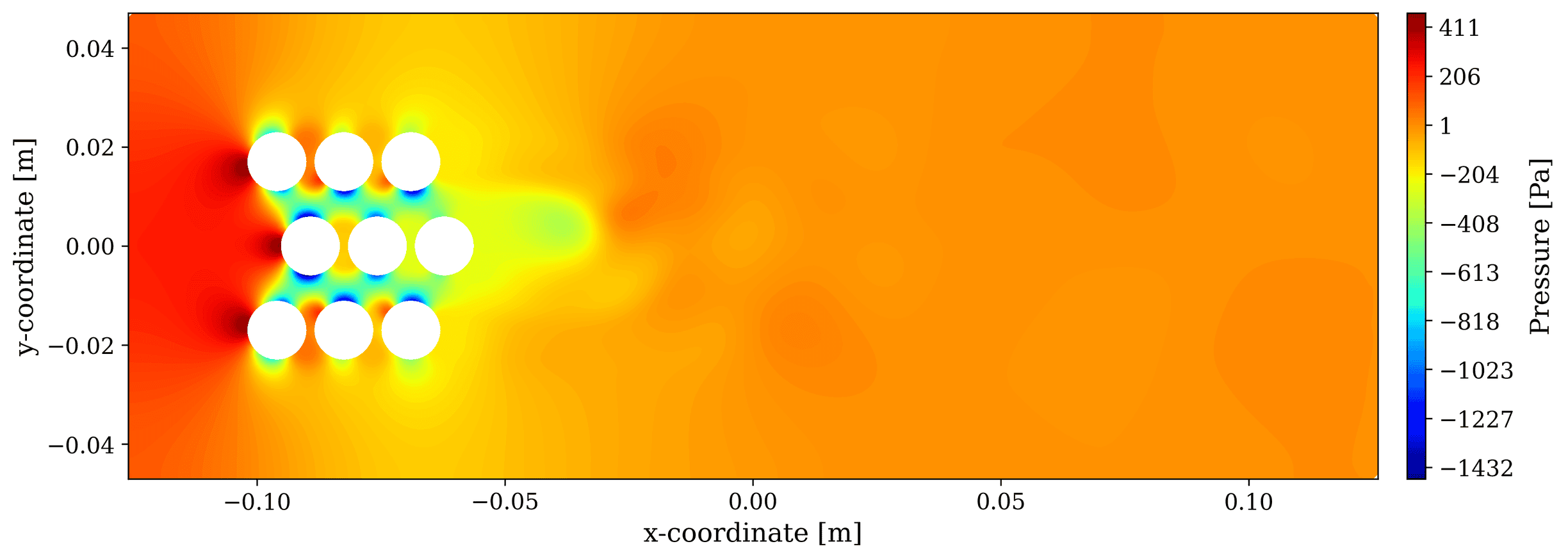} \\[2pt]
        \adjustbox{valign=c}{\rotatebox[origin=c]{90}{\small\textbf{Predicted}}} &
        \includegraphics[width=0.45\textwidth,valign=c]{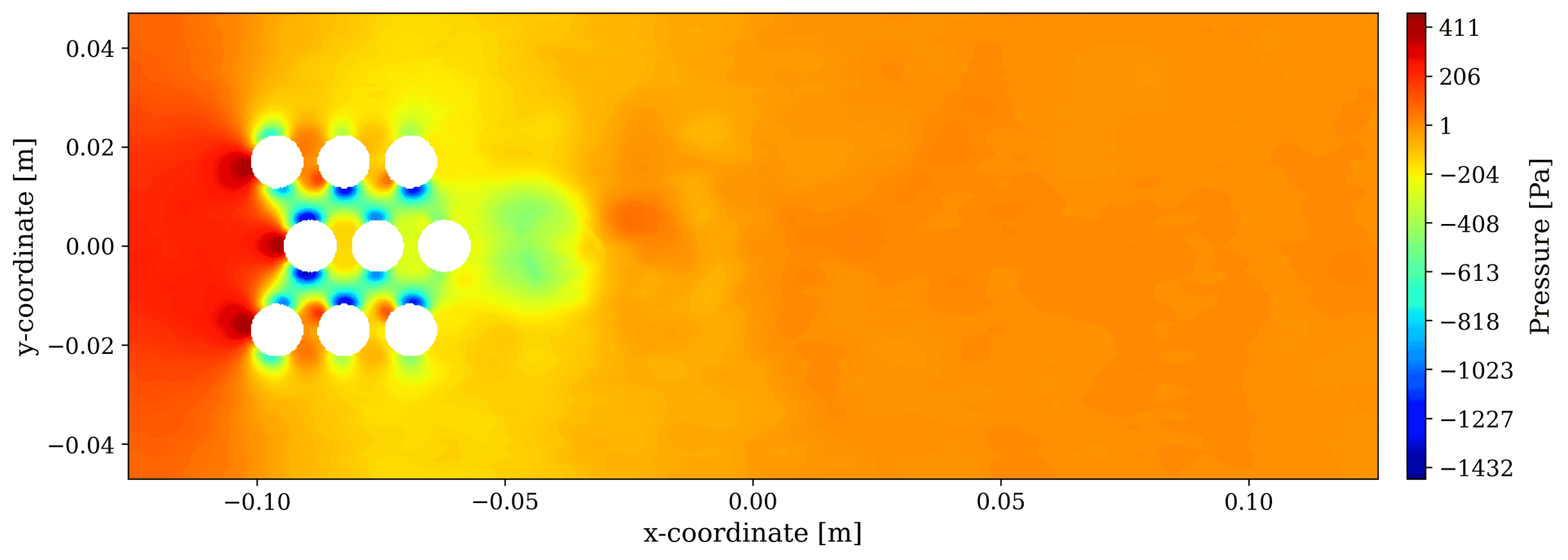} &
        \includegraphics[width=0.45\textwidth,valign=c]{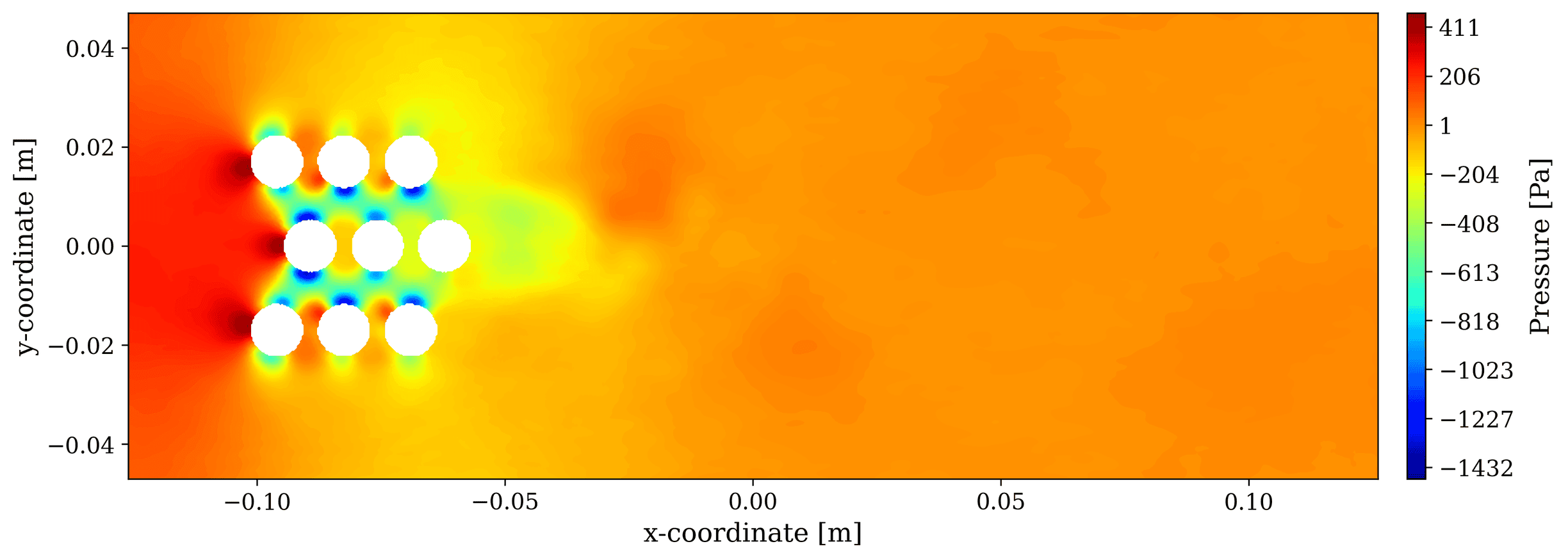} &
        \includegraphics[width=0.45\textwidth,valign=c]{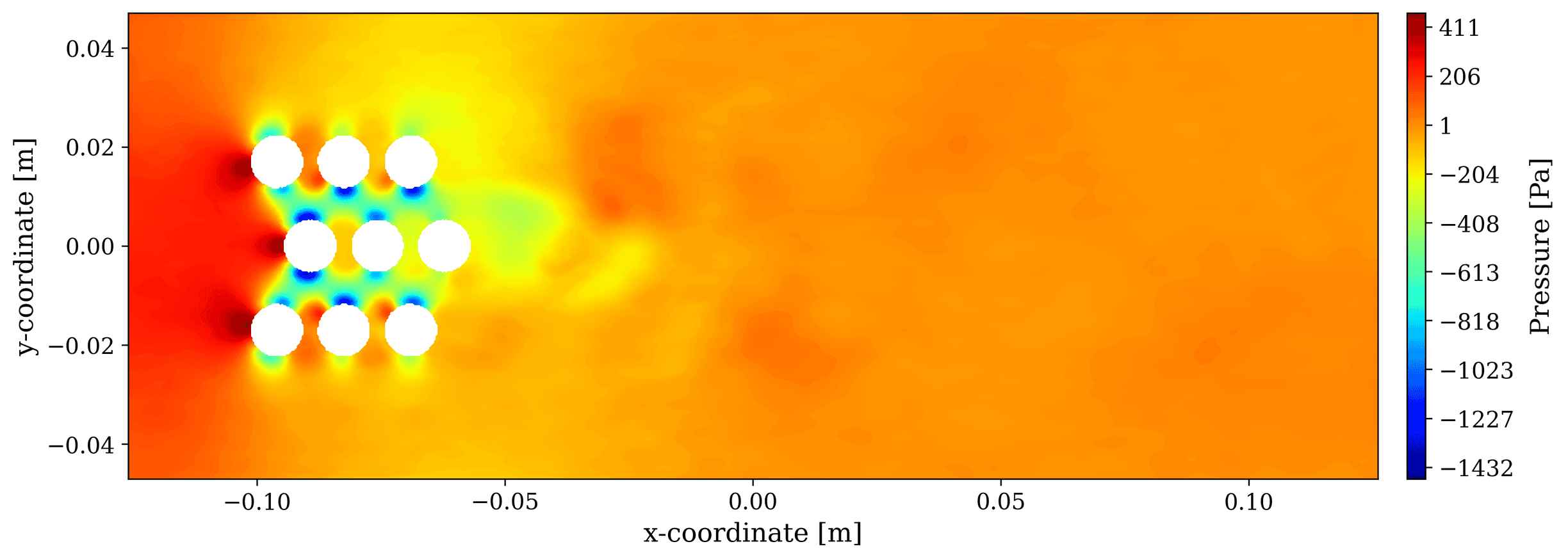} \\[2pt]
        \adjustbox{valign=c}{\rotatebox[origin=c]{90}{\small\textbf{Error}}} &
        \includegraphics[width=0.45\textwidth,valign=c]{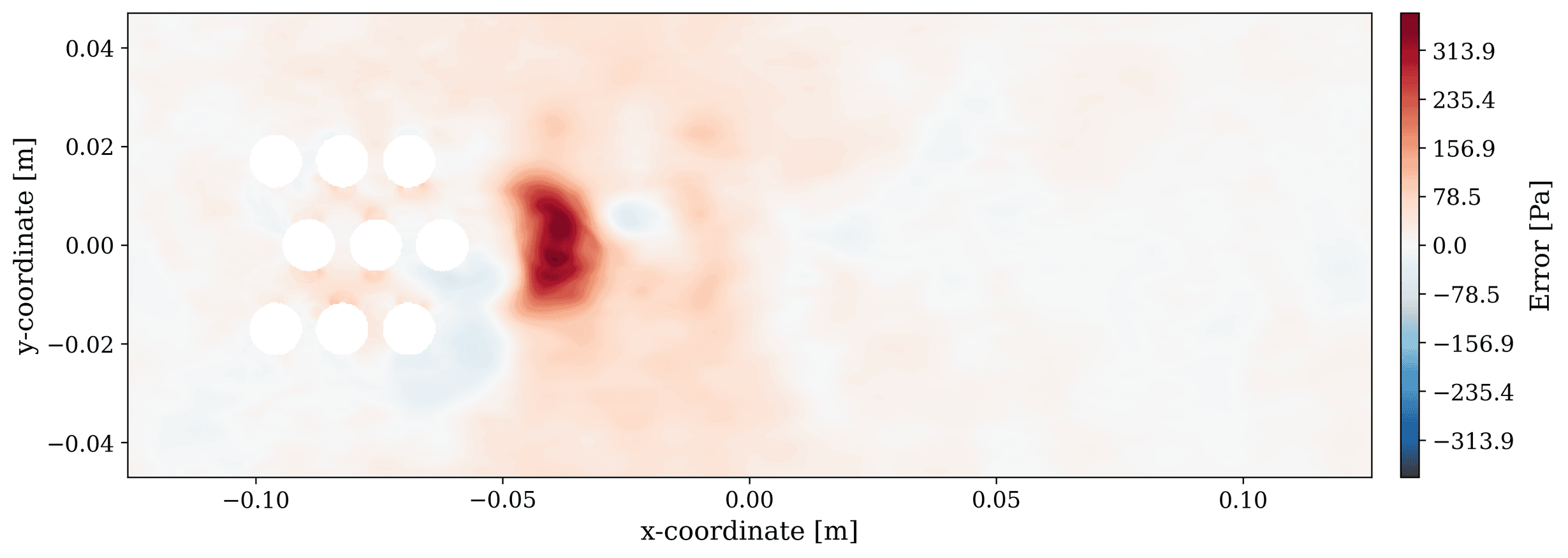} &
        \includegraphics[width=0.45\textwidth,valign=c]{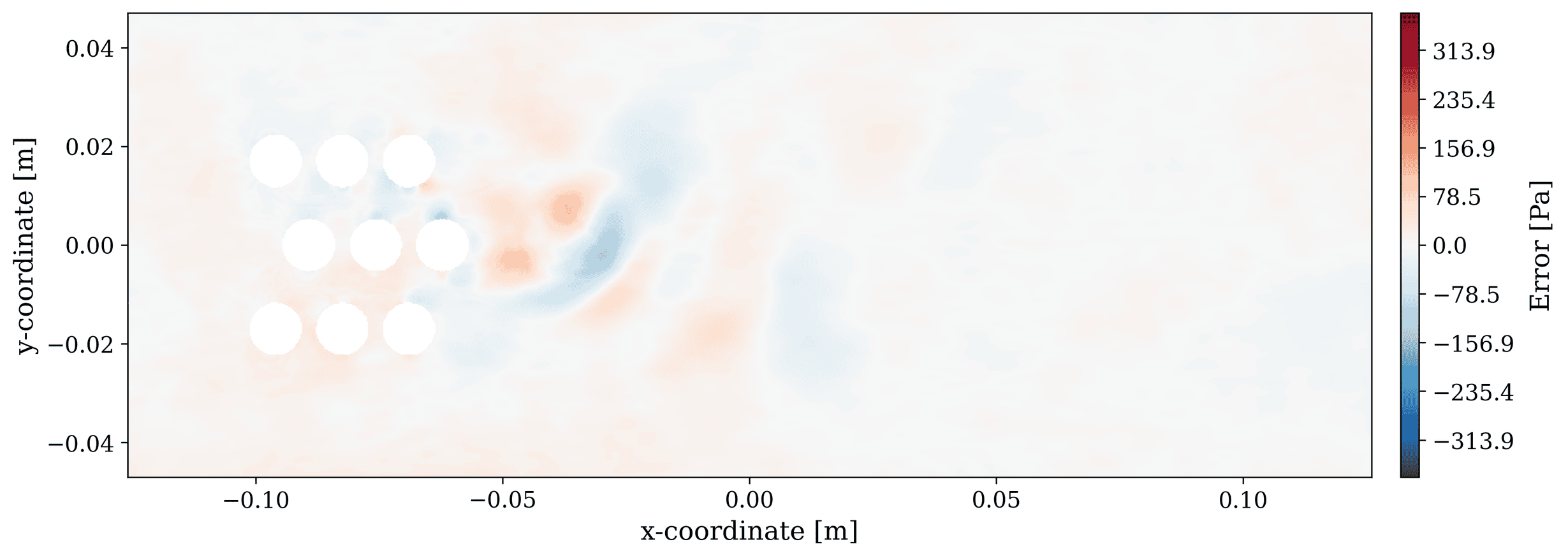} &
        \includegraphics[width=0.45\textwidth,valign=c]{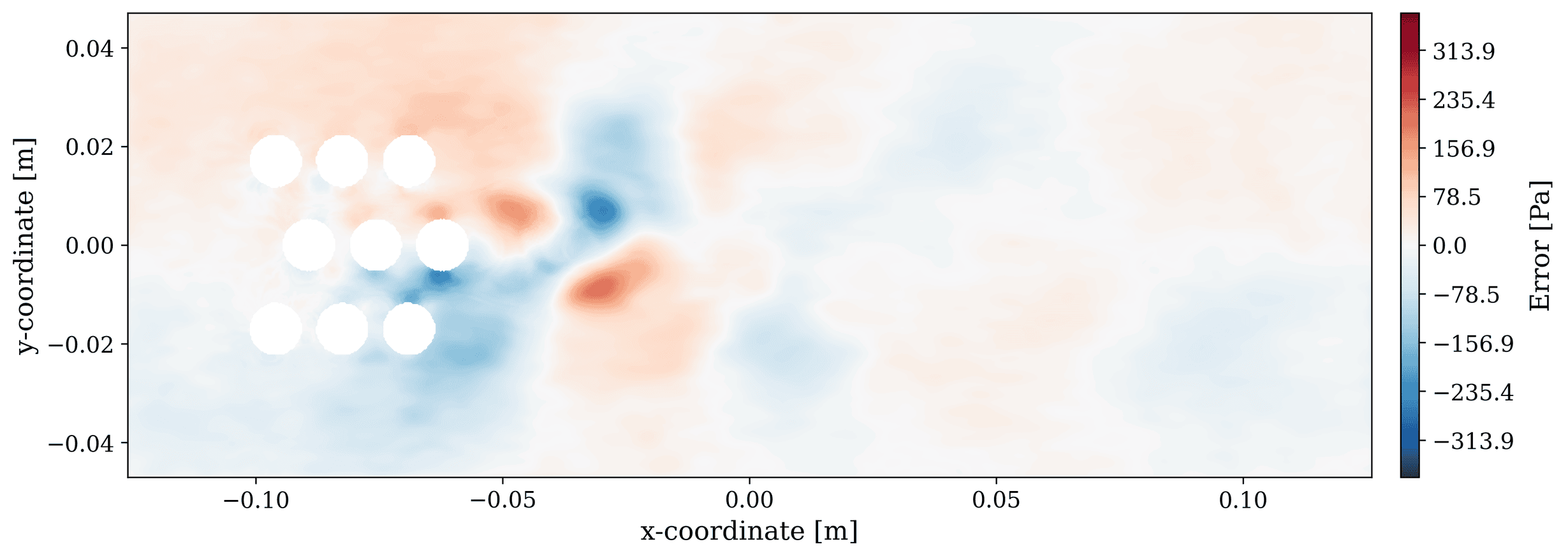} \\
    \end{tabular}
    }%
    \caption{MscaleFNO: pressure field prediction for inlet velocity 0.7 m/s. Top: reference CFD; middle: prediction; bottom: error.}
    \label{fig:mscalefno_pressure_inlet070}
\end{figure}

\bibliographystyle{unsrt}     
\bibliography{reference}
\end{document}